\pdfoutput=1

\documentclass[11pt]{article}

\usepackage[preprint]{acl}

\usepackage{times}
\usepackage{latexsym}
\usepackage{amsmath}
\usepackage{booktabs}
\usepackage{appendix}

\usepackage[T1]{fontenc}

\usepackage[utf8]{inputenc}

\usepackage{microtype}

\usepackage{inconsolata}

\usepackage{graphicx}

\usepackage{array}
\usepackage{caption}
\usepackage{rotating}
\usepackage{adjustbox}

\newlength{\plotfigurewidth}
\title{Temporal Dependencies in In-Context Learning: The Role of Induction Heads}

\author{%
  \textbf{Anooshka Bajaj\thanks{Equal contribution. Authors ordered alphabetically.}}, \textbf{Deven Mahesh Mistry}\footnotemark[1], \textbf{Sahaj Singh Maini}\footnotemark[1], \textbf{Yash Aggarwal}\\
  \textbf{Billy Dickson}, \textbf{Zoran Tiganj}\\
  Department of Computer Science, Indiana University Bloomington
}

\begin{document}
\maketitle
\begin{abstract}
Large language models (LLMs) exhibit strong in-context learning capabilities, but how they track and retrieve information from context remains underexplored. Drawing on the free recall paradigm in cognitive science (where participants recall list items in any order), we show that several open-source LLMs consistently display a serial-recall-like pattern, assigning peak probability to tokens that immediately follow a repeated token in the input sequence. Through systematic ablation experiments, we show that induction heads, specialized attention heads that attend to the token following a previous occurrence of the current token, play an important role in this phenomenon. Removing heads with a high induction score substantially reduces the +1 lag bias, whereas ablating random heads does not reproduce the same reduction. We also show that removing heads with high induction scores impairs the performance of models prompted to do serial recall using few-shot learning to a larger extent than removing random heads. Our findings highlight a mechanistically specific connection between induction heads and temporal context processing in transformers, suggesting that these heads are especially important for ordered retrieval and serial-recall-like behavior during in-context learning.
\end{abstract}

\section{Introduction}

Large language models (LLMs) have demonstrated an extraordinary capacity for in-context learning (ICL), adapting their behavior solely based on the prompt context without explicit parameter updates \cite{brown2020language,olsson2022context,chan2022data,singh2024transient,chan2024toward,xie2021explanation,dong2022survey}. While extensive work has explored their few-shot and emergent reasoning capabilities \cite{brown2020language,wei2022emergent}, comparatively little attention has been devoted to temporal effects: how the serial position of a token within the context window affects the probability that it is retrieved by the model.

We adopt a cognitive-science-inspired paradigm to study LLMs. In cognitive science, serial-position effects have been systematically studied for decades \cite{ebbinghaus1913memory}. Paradigms such as free recall, where participants are asked to recall a list of sequentially presented items in any order, reveal a tendency to better recall items from the beginning or the end of the list (primacy and recency effects). By contrast, serial recall requires reproducing list items in the order in which they were originally presented. Furthermore, after recalling an item from the list, participants tend to recall next an item that was presented near the just recalled item. This effect is called temporal contiguity \cite{kahana1996associative}. The contiguity effect was commonly modeled through associations that bind a gradually decaying memory trace of the recent past with the present item \cite{howard2002contextual,polyn2009context}. This effect is considered critical for recall of episodic memory \cite{b2010temporal,folkerts2018human}, and inability to reinstate temporal context through the contiguity effect in clinical cases such as medial temporal lobe amnesia is linked with deficits in human episodic memory \cite{palombo2019medial}.

Recent studies have begun to bridge human episodic memory concepts with ICL in transformers \cite{pink2024assessing,ji2024linking,mistry2025emergence,guo2024serial,bajaj2026beyond}. A particularly promising direction is the investigation of induction heads. These are attention heads that selectively attend to the token that followed the previous presentation of the current token, thus facilitating pattern matching and the reproduction of observed token sequences \cite{olsson2022context,elhage2021mathematical,singh2024needs}. Prior work has demonstrated that such heads are important for ICL \cite{crosbie2024induction,olsson2022context,yinattention,bansal2022rethinking}, and they have been linked to emergent memory-like processes in LLMs \cite{fountas2024human}. Previous work also demonstrated the presence of a contiguity effect in attention heads with high induction scores \cite{guo2024serial}: when presented with a repeated sequence of tokens, attention heads tend to attend to the temporal neighborhood of the original token presentation. At the same time, recent mechanistic work suggests that broader few-shot ICL can also rely on complementary mechanisms such as task vectors and function vectors, especially in more abstract settings \cite{hendel2023context,todd2023function,yinattention,csahincontext}.

In the present study, we build on previous approaches and examine the temporal properties of ICL and the importance of induction heads using an experimental paradigm motivated by the free recall task. Our goal is intentionally specific. We test whether induction heads are important for temporal retrieval and serial recall behavior. We extend previous analyses from relatively small transformer models (such as GPT-2 medium, with less than 400M parameters) to larger open-weight base and instruction-tuned models with billions of parameters, compare four model families, and connect the mechanistic analysis to a downstream few-shot serial recall task. Similar to the approach in \citet{mistry2025emergence}, we construct a sequence consisting of 500 randomly ordered tokens followed by the token at index 250 from that sequence (using zero-based indexing). The order of tokens was random to reduce any effects coming from semantic similarity. We then quantified the probability of the next token as a function of lag, defined as the distance from the repeated token. The expectation is that if temporal contiguity holds, tokens temporally adjacent to the repeated token (e.g., at lags +1, -1, +2, -2...) will be assigned higher probabilities than more distant tokens. If temporal contiguity is one-sided and restricted to serial recall, the token immediately adjacent to the repeated token (e.g., at lag +1) will be assigned higher probabilities than other tokens. In contrast, primacy and recency effects would manifest as elevated probabilities at the beginning and end of the sequence, respectively \cite{b2010temporal,wang2023primacy}.

Our analysis, performed on a set of LLMs including Llama, Mistral, Qwen, and Gemma, reveals a consistent trend with some model-dependent differences. Instruction-tuned Mistral, Qwen, and Gemma had an overall tendency to assign peak probability to the token immediately following the repeated token (lag +1), a pattern reminiscent of serial recall. Llama, however, exhibited different behavior with relatively flat probabilities across lags.

We conducted targeted ablation experiments to better understand the role of induction heads in shaping these temporal effects. By selectively ablating attention heads with high induction scores, we observed a substantial reduction and, in some models, near elimination of the +1 lag preference. Ablating random heads produced a different pattern and often increased the +1 lag preference, underscoring the specific contribution of induction heads to this serial-recall-like behavior. This observation aligns with recent findings that highlight the mechanistic importance of induction heads for ICL \cite{crosbie2024induction}.

Finally, to link this mechanistic evidence to functional performance, we directly assessed the impact of these ablations on a task requiring ordered recall. Specifically, we performed an additional experiment focused on ICL serial recall. In this experiment, a few-shot example shows a short list of 14 tokens immediately followed by its verbatim reproduction, and the model must do the same for a novel list at test time. The results show that ablating induction heads leads to a larger degradation in serial recall performance than ablating random heads. These findings further underscore the important role of induction heads in tasks demanding ordered sequence recall and support their importance for the observed serial recall bias in LLMs.

\section{Methods}

\subsection{Models}

We used popular open-source LLMs with a similar parameter count (between 7B and 9B parameters): \textit{Meta-Llama/Llama-3.1-8B} \cite{dubey2024llama}, \textit{Qwen/Qwen2.5-7B} \cite{yang2024qwen2}, \textit{Mistralai/Mistral-7B-v0.1} \cite{jiang2023mistral}, and \textit{Google/gemma-2-9b} \cite{team2024gemma}. For each model, we conducted our experiments on both the base and instruction-tuned versions. We also used the open-source library TransformerLens \cite{nanda2022transformerlens} for the ablation of induction heads.

\subsection{Quantifying temporal dependencies in ICL}
\label{sec:temporal}
To quantify how temporal position in the sequence shapes recall probability, we conducted an experiment described in \citet{mistry2025emergence} and inspired by the free recall task. Models were prompted with a sequence of 501 tokens. The first 500 tokens were randomly ordered tokens (we chose 500 common English words such that each word corresponded to a single token), and the last token repeated the token at index 250 in that sequence (using zero-based indexing; e.g., \textit{[G R D B T H M B]}, assuming that each character is a single token). Both base and instruction-tuned models received the same prompt. We then quantified the probability of the next token as a function of lag (distance from the repeated token; in our example, the repeated token would be the first presentation of token \textit{B}). For example, the probability of token \textit{D} would be considered the probability of lag -1, and the probability of token \textit{M} would be considered the probability of lag +3. With sequences of 500 tokens, our lags ranged from -250 to 249. A temporal contiguity effect would cause high recall probability for tokens temporally close to the repeated token (low absolute lags). Human results are characterized by a gradual falloff to the left and right of zero lag, with a slight forward asymmetry (a higher preference for forward than backward recalls \cite{kahana1996associative,howard2002contextual}). Primacy and recency would predict higher recall probability for tokens at the beginning and end of the sequence, respectively.

Naturally, the semantic properties of the tokens would have a dominant impact on the output probabilities. For instance, if token \textit{B} is \textit{blue} and token \textit{H} is \textit{sky}, it is very likely that the models would attribute high probability to \textit{H} to follow \textit{B}. To make our temporal analysis meaningful, we had to ensure that these semantic effects are averaged out. To achieve that, we generated 5000 random permutations of the sequence and averaged the final reported probabilities across these permutations. Results to follow will confirm that recall probabilities were at a baseline level common across all lags, except for the lag +1 and, in some cases, lag 0 and lags close to 249 (recency effect). The fact that probabilities were at a baseline level for most lags indicates that our shuffling and averaging procedure was sufficient to remove the semantic similarity effects and isolate temporal effects. We note that a similar procedure is also commonly applied in human free recall experiments where participants are typically presented with a large number of lists or words, and results are then averaged across the lists to minimize the impact from semantic similarities \cite{manning2012interpreting, steyvers2005word}.

\subsection{Attention heads ablation}

To ablate the induction heads, we followed the approach described in \citet{crosbie2024induction}. Specifically, we set the attention scores for ablated heads to $-\infty$ for \textit{zero ablation} and to their mean value for \textit{mean ablation}. When ablating induction heads, we progressively increased the number of ablated heads starting from heads with high induction scores. For random head ablation, we ensured that none of the selected heads ranked among the top 300 based on the induction score.

\subsection{Computation of induction scores}

To compute the induction score, we followed the procedure described in previous work \cite{nanda2022transformerlens,olsson2022context,elhage2021mathematical}. The models were prompted with a sequence of 1,000 commonly used tokens repeated twice, yielding a final prompt of the form \textit{[A B C D E A B C D E]}. Let the first copy occupy positions $1,\dots,N$ and the repeated copy positions $N+1,\dots,2N$, where $N=1000$ is the length of the unrepeated source sequence. For a given attention head, the induction score is defined as
\begin{equation}
 \mathrm{I} = \frac{1}{N-1}\sum_{i=N+1}^{2N-1} a_{i,\,i-N+1},
\end{equation}
where $a_{i,j}$ is the attention value (after softmax normalization) from token $i$ to token $j$. Intuitively, this quantity measures how strongly the repeated copy attends to the token that immediately followed the previous occurrence of the current token.

\subsection{Serial Recall in an ICL Task}
\label{sec:serial_recall}
To test the mechanistic link between induction heads and ordered recall, we devised a \emph{serial-recall} ICL task, where serial recall refers to reproducing list items in their original presentation order. We evaluated Llama and Qwen (base and instruct); Mistral and Gemma did not reliably learn the task.
Each \textit{study list} \(S^{(k)} = \bigl(s^{(k)}_1,\dots,s^{(k)}_{14}\bigr)\) comprised \(14\) distinct tokens drawn without replacement from the 25 uppercase English letters. Ten such lists were packed into a single \emph{10-shot} prompt that alternates ``study'' and ``recall'' segments:
\begin{center}
\texttt{\textless{}bos\textgreater{} study }$S^{(1)}$\texttt{ \textless{}bos\textgreater{} recall }$S^{(1)}$\texttt{ \textless{}bos\textgreater{} }\(\dots\)\texttt{ study }$S^{(10)}$\texttt{ \textless{}bos\textgreater{}}
\end{center}
The final \texttt{\textless{}bos\textgreater{}} token signals the model to output the reproduction of the last study list, $S^{(10)}$. We computed aggregate statistics over 50 distinct prompts while systematically ablating induction and random (control) heads.

\section{Results}

\subsection{Induction scores before and after instruction tuning: High degree of heterogeneity across models}

\begin{figure*}[h!]
    \centering
    \begin{tabular}{lllll}
    \textbf{A} &
    \textbf{B} &
    \textbf{C} &
    \textbf{D} \\
        {\includegraphics[width=0.22\textwidth]{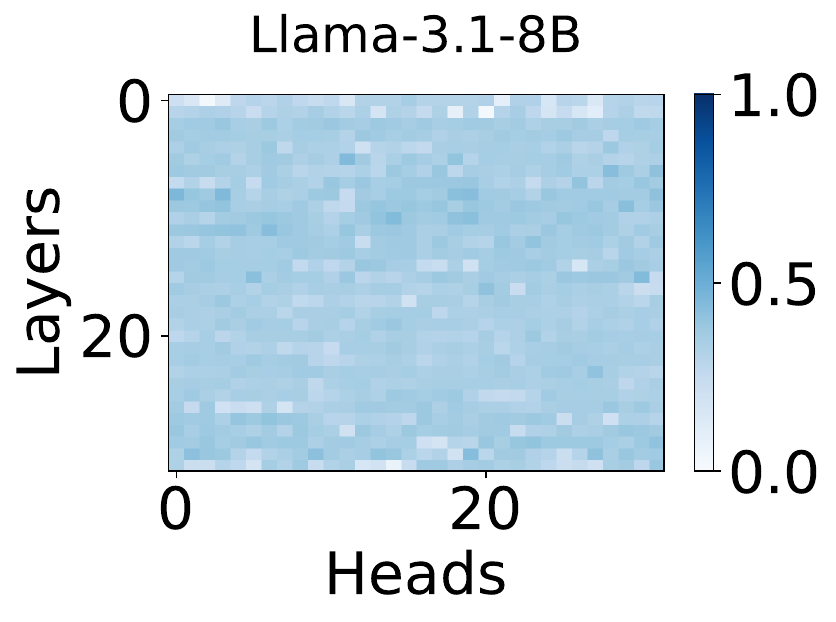}} &
        {\includegraphics[width=0.22\textwidth]{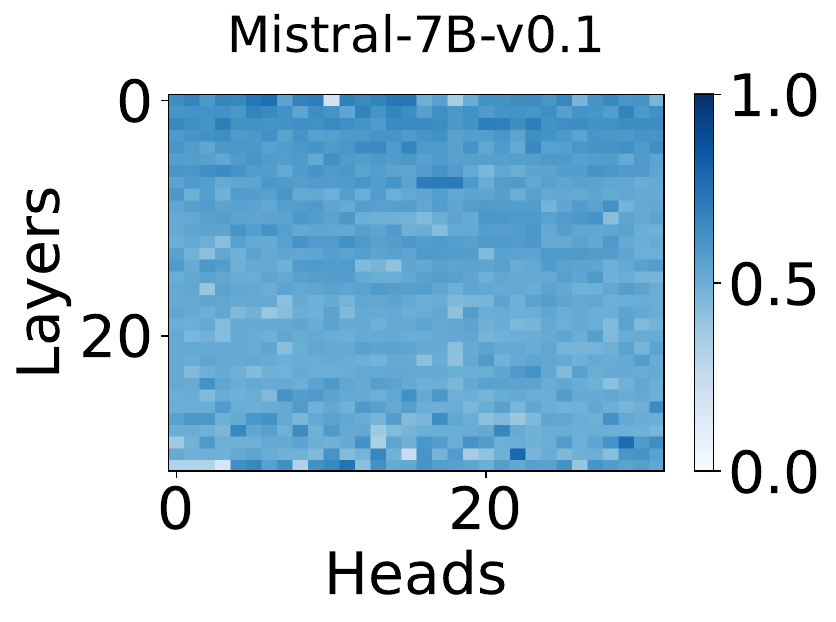}} &
        {\includegraphics[width=0.22\textwidth]{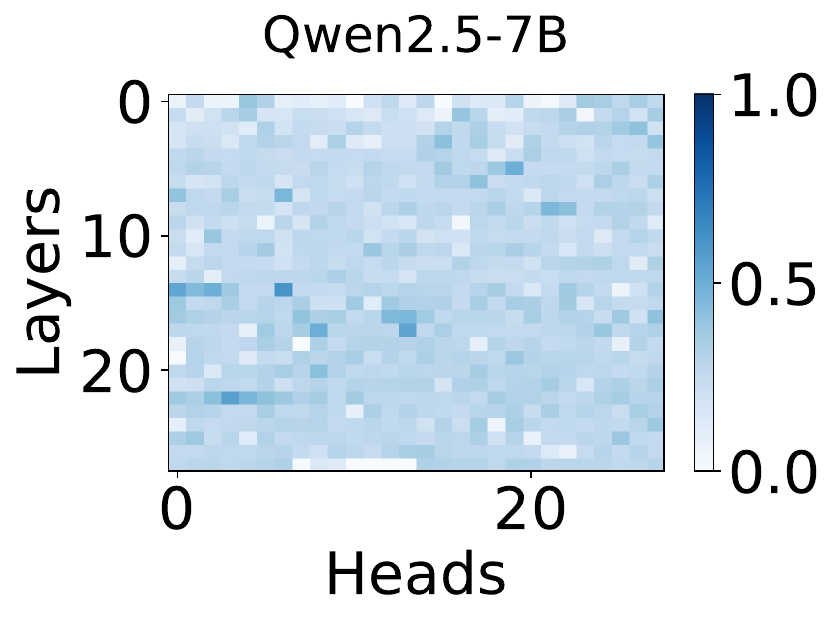}} &
        {\includegraphics[width=0.22\textwidth]{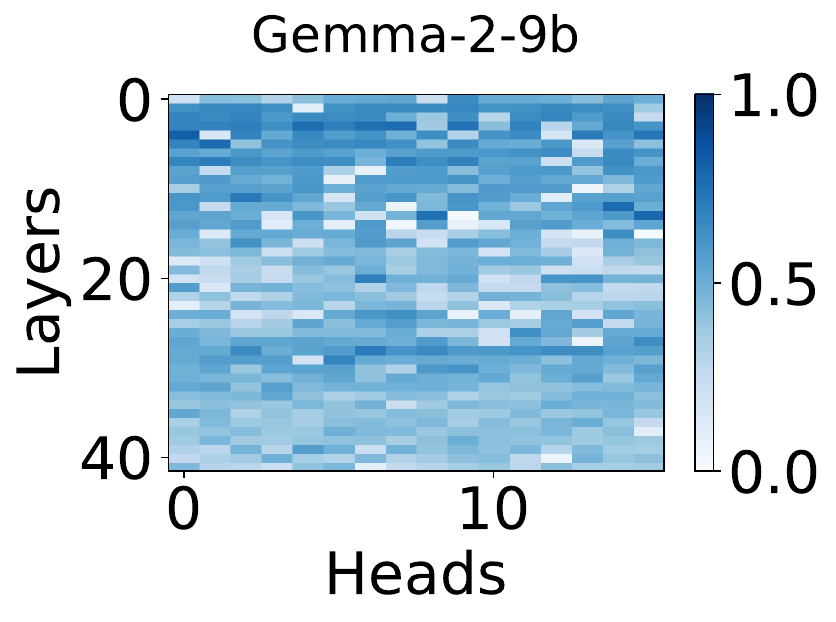}} \\
    \end{tabular}
      \begin{tabular}{lllll}
    \textbf{E} &
    \textbf{F} &
    \textbf{G} &
    \textbf{H} \\
        {\includegraphics[width=0.22\textwidth]{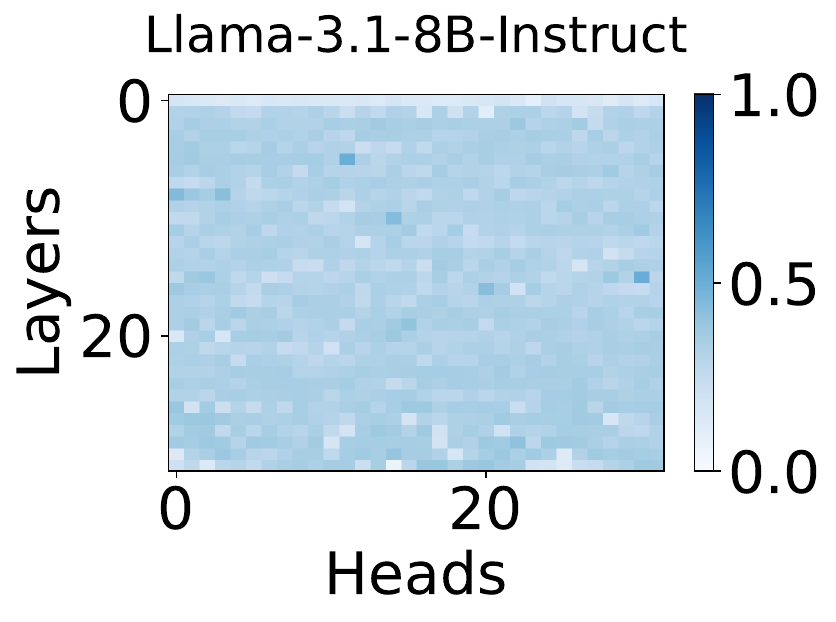}} &
        {\includegraphics[width=0.22\textwidth]{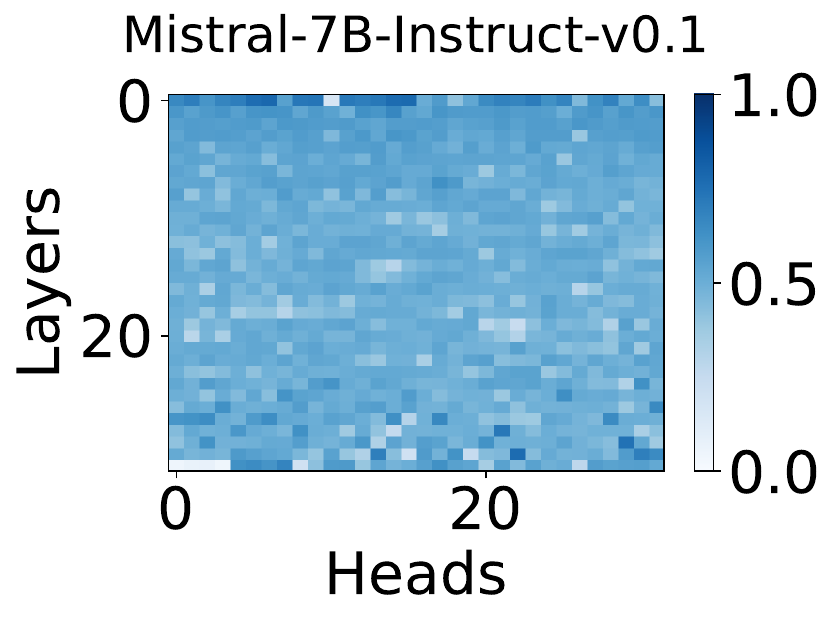}} &
        {\includegraphics[width=0.22\textwidth]{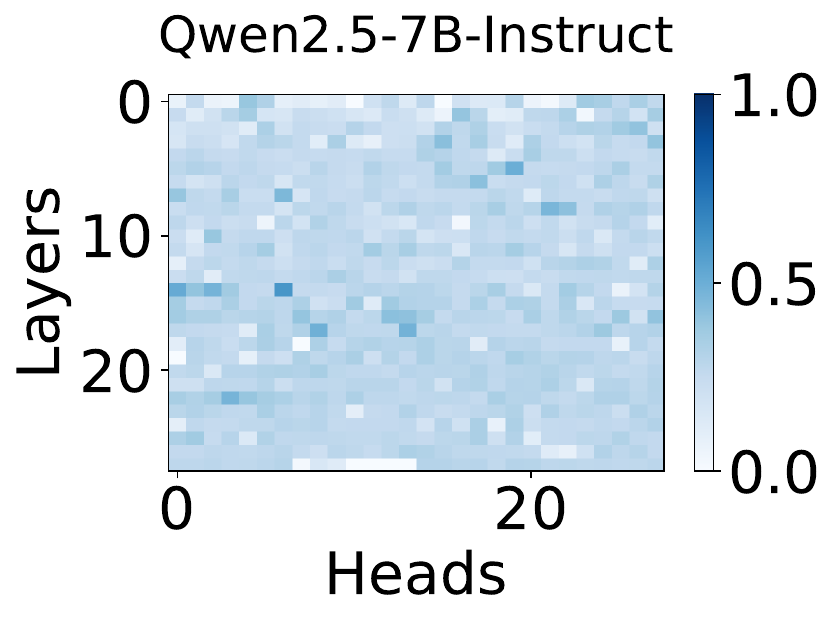}} &
        {\includegraphics[width=0.22\textwidth]{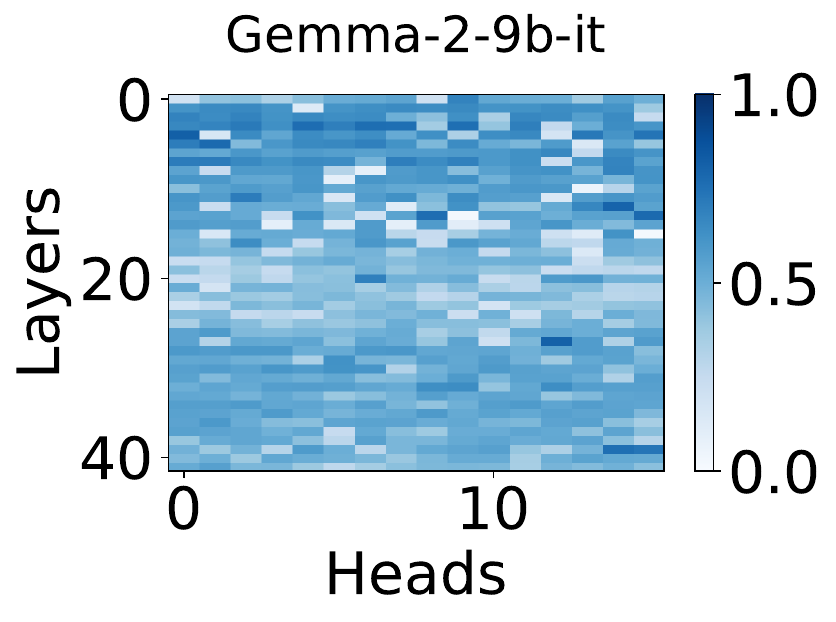}} \\
    \end{tabular}
    \begin{tabular}{lllll}
    \textbf{I} &
    \textbf{J} &
    \textbf{K} &
    \textbf{L} \\
        {\includegraphics[width=0.22\textwidth]{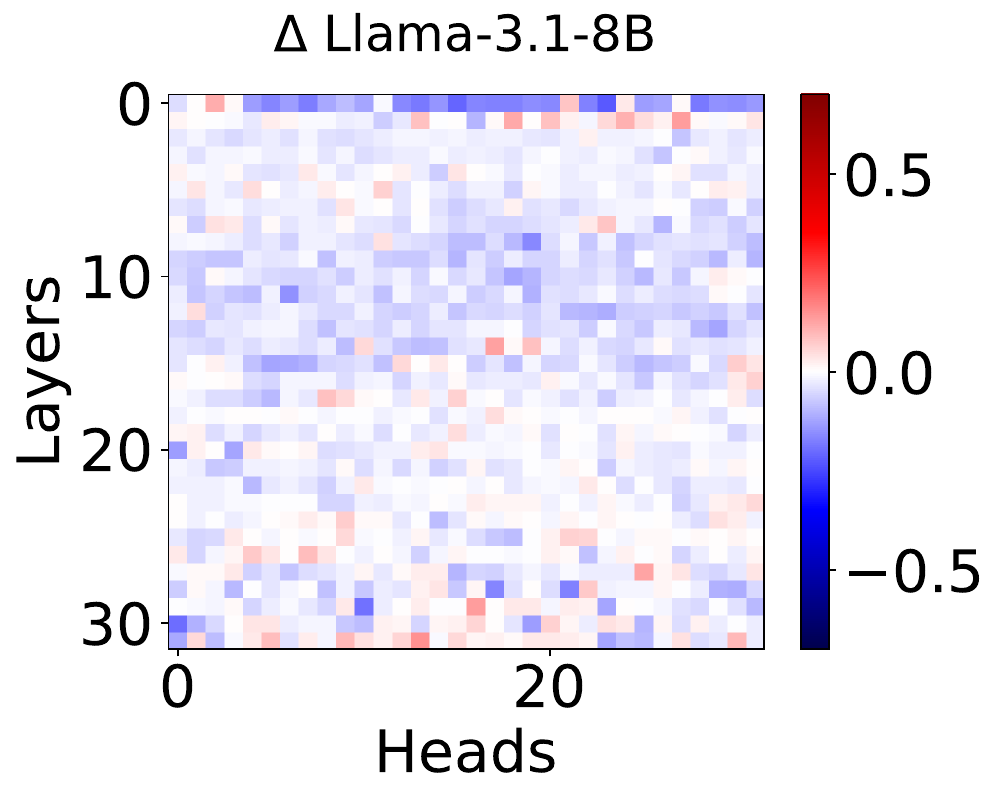}} &
        {\includegraphics[width=0.22\textwidth]{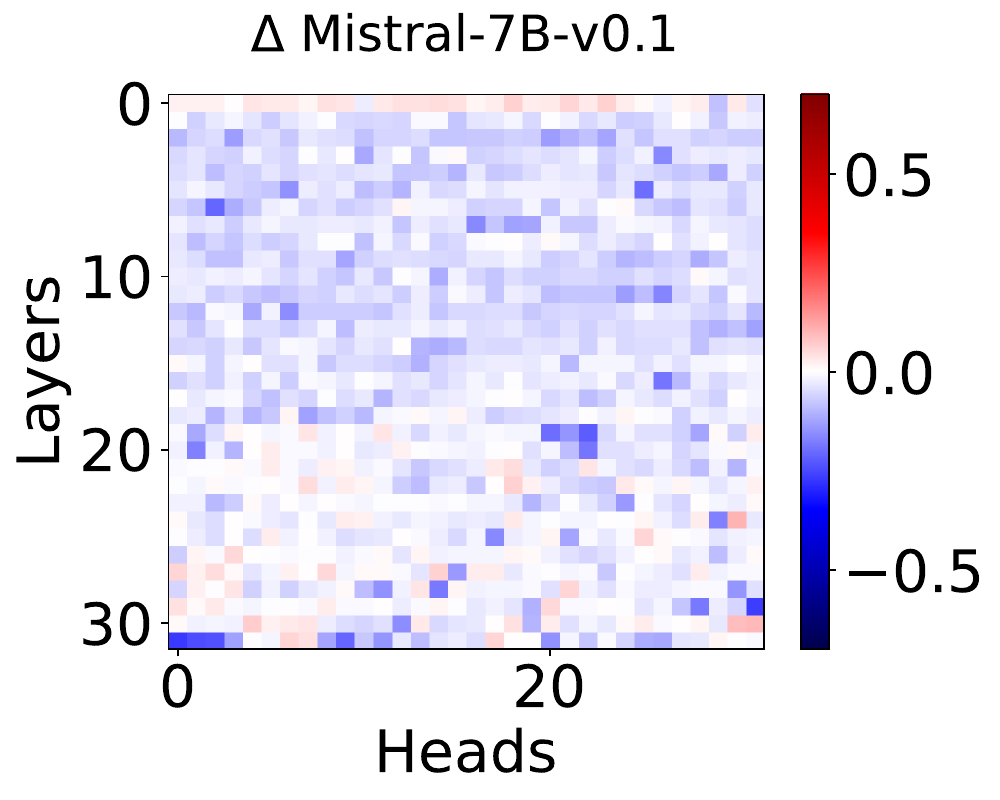}} &
        {\includegraphics[width=0.22\textwidth]{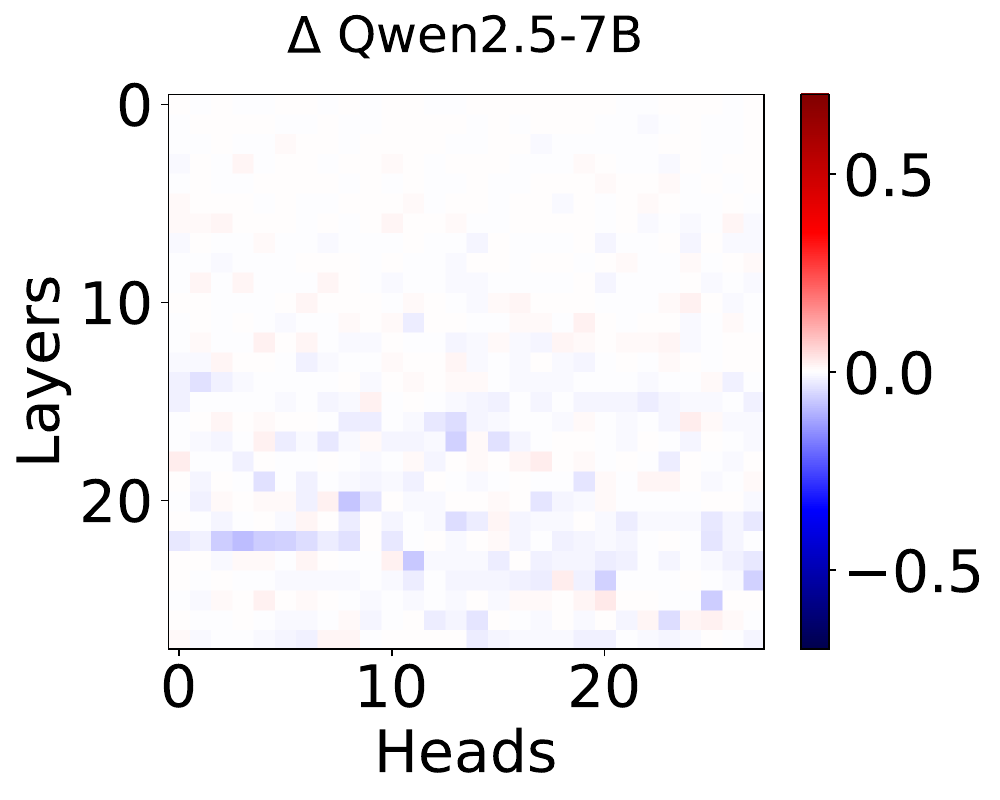}} &
        {\includegraphics[width=0.22\textwidth]{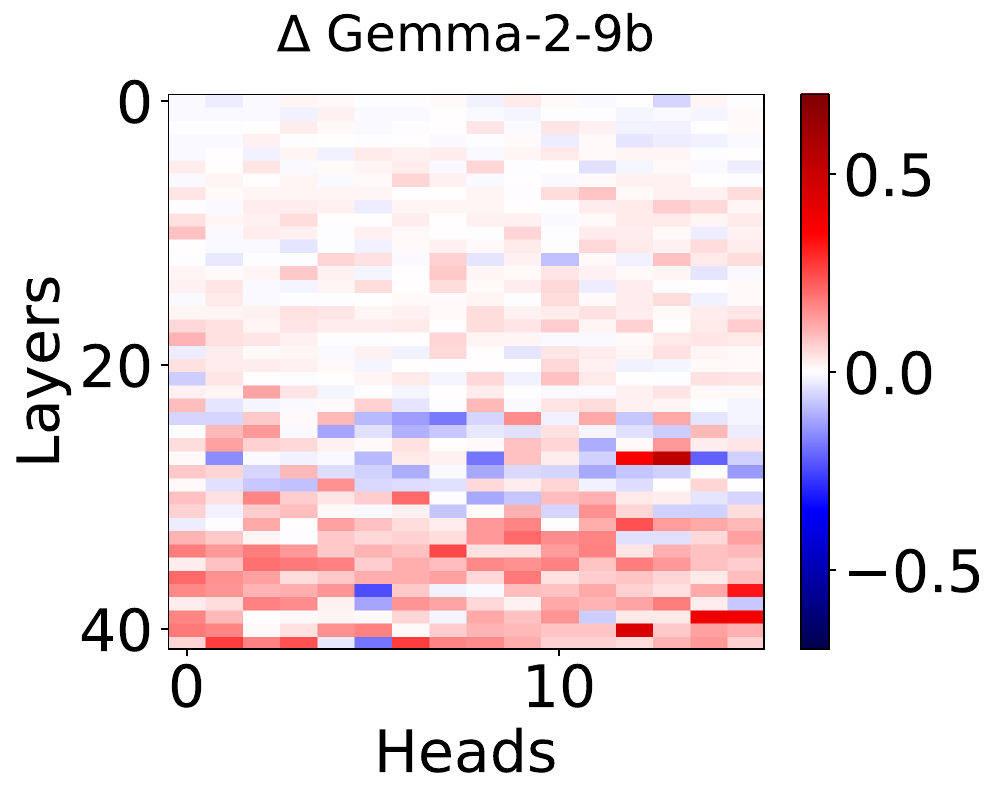}} \\
    \end{tabular}
    \caption{Induction scores for the four models across layers and heads. Top row: Base models. Middle row: Instruction-tuned models. Bottom row: difference in induction scores for each layer and head between instruction-tuned and base models.
    \label{fig:induction_heatmaps}}
\end{figure*}

We visualized induction scores across layers and heads for each of the four examined models: Llama, Mistral, Qwen and Gemma (Fig.~\ref{fig:induction_heatmaps} A-D for base models and  Fig.~\ref{fig:induction_heatmaps} E-H for instruction-tuned models). Induction scores varied across models, with Mistral having the highest average and Qwen the lowest (Table~\ref{tab:mean_std_induction_scores}). Across models, Gemma had relatively high variance in scores, followed by Mistral, Qwen and Llama. To better understand the differences between base and instruction-tuned models, we subtracted attention scores of the base models from the attention scores of the instruction-tuned models (Fig.~\ref{fig:induction_heatmaps} I-L). The largest difference was observed in Gemma, especially in the high layers, where induction scores were higher for the instruction-tuned model. Conversely, Mistral and Llama had higher scores for the base models. Qwen showed very little difference between the base and instruction-tuned versions. Overall, this indicates a high degree of heterogeneity across the models, suggesting multiple possible routes to convergence. Observed heterogeneity contrasts with the strong mid-/deep-layer clustering reported for OPT-66B \citep{bansal2022rethinking}, indicating that model scale and architectural family substantially modulate where induction heads emerge.

\begin{table}[h]
    \caption{Mean and standard deviation of the induction scores}
    \begin{tabular}{ccc}
        \toprule
        &  \textbf{Instruct} & \textbf{Base} \\
        \midrule
            Llama-3.1-8B & 0.31 ± 0.05 & 0.34 ± 0.05 \\
            Mistral-7B-v0.1 & 0.50 ± 0.08 & 0.54 ± 0.06 \\
            Qwen2.5-7B & 0.26 ± 0.07 & 0.26 ± 0.08 \\
            Gemma-2-9b & 0.48 ± 0.13 & 0.45 ± 0.14 \\
        \bottomrule
    \end{tabular}
    \label{tab:mean_std_induction_scores}
\end{table}

The above heterogeneity is further illustrated through visualization of induction scores for each head as a function of layer (Fig.~\ref{fig:ind_scatter} A-D for base models, Fig.~\ref{fig:ind_scatter} E-H for instruction-tuned models, and Fig.~\ref{fig:ind_scatter} I-L for the subtraction). Variability of the induction scores is most prominent in Gemma. In general, most heads exhibit similar induction scores, sometimes with gradual changes across layers.

\begin{figure*}[h!]
    \centering
    \begin{tabular}{lllll}
    \textbf{A} &
    \textbf{B} &
    \textbf{C} &
    \textbf{D} \\
        {\includegraphics[width=0.22\textwidth]{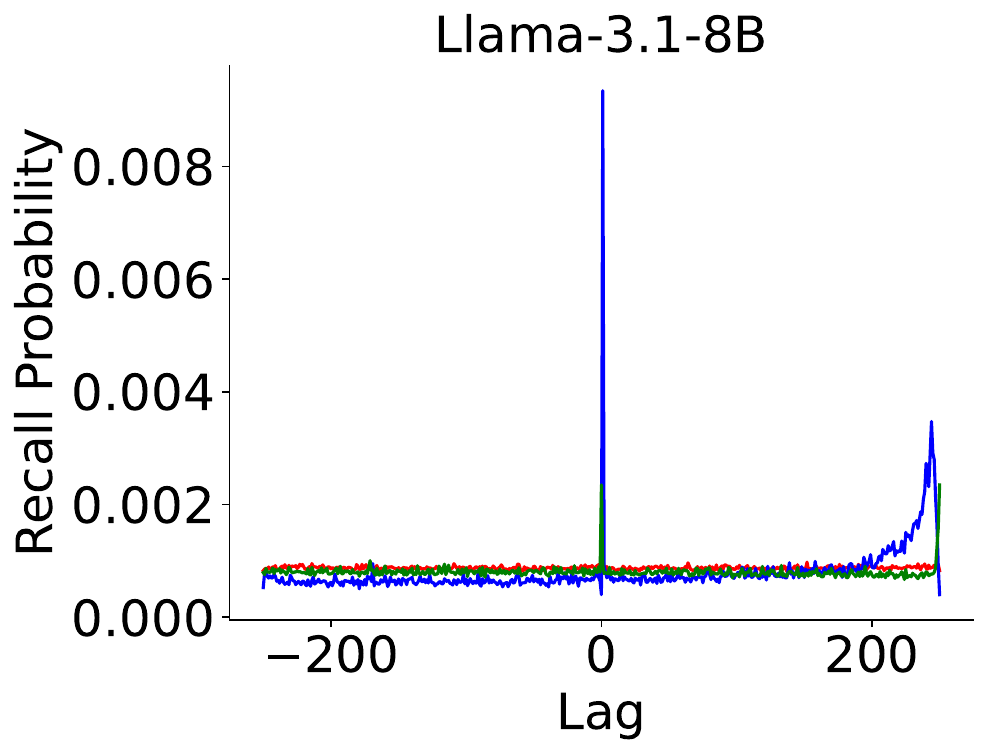
        }} &
        {\includegraphics[width=0.22\textwidth]{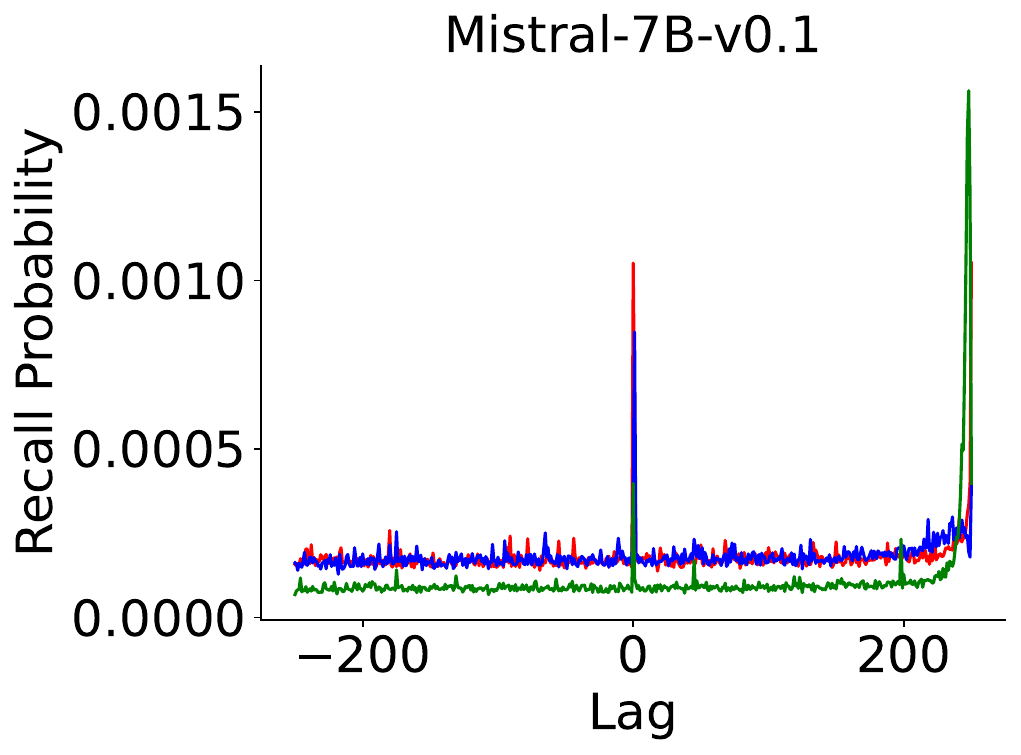}} &
        {\includegraphics[width=0.22\textwidth]{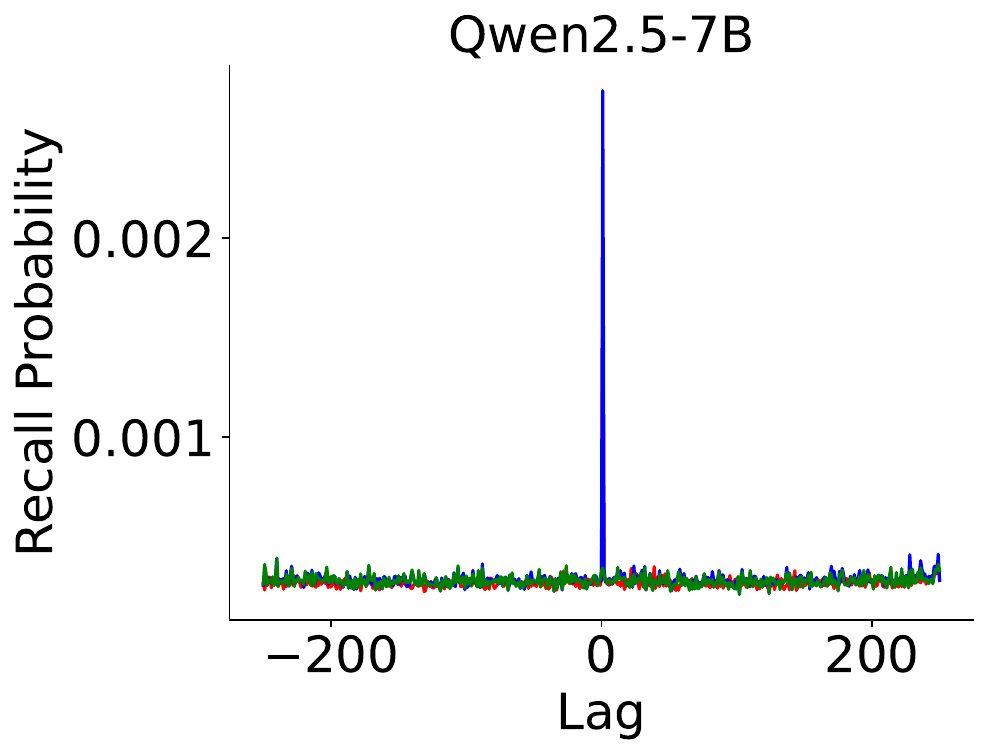}} &
        {\includegraphics[width=0.22\textwidth]{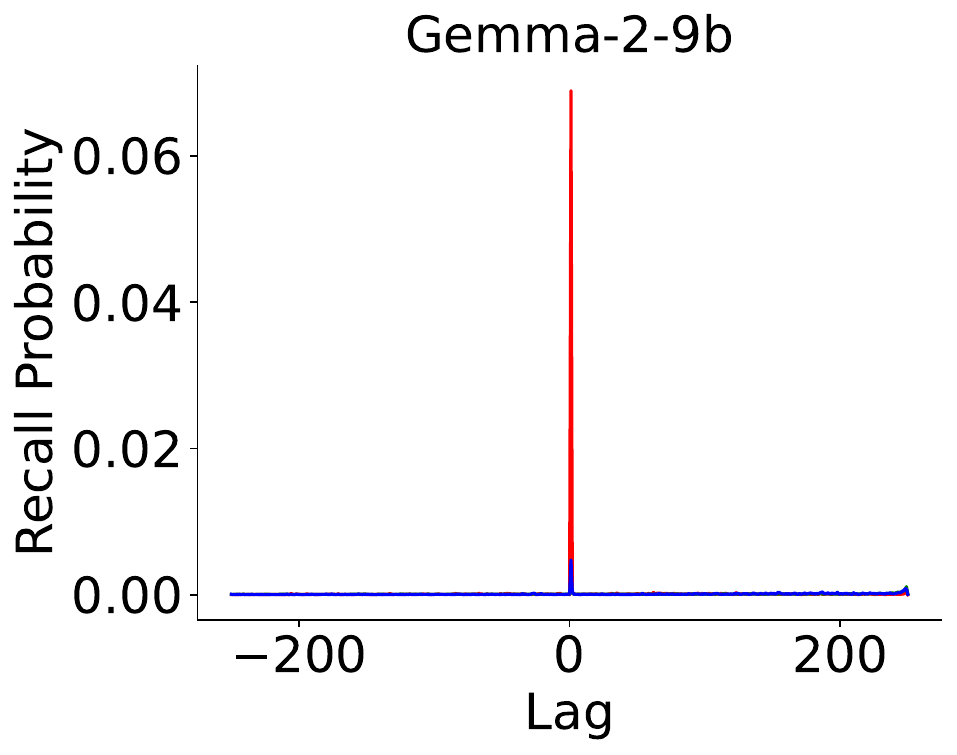}}
        \\
    \end{tabular}
    \begin{tabular}{lllll}
    \textbf{E} &
    \textbf{F} &
    \textbf{G} &
    \textbf{H} \\
        {\includegraphics[width=0.22\textwidth]{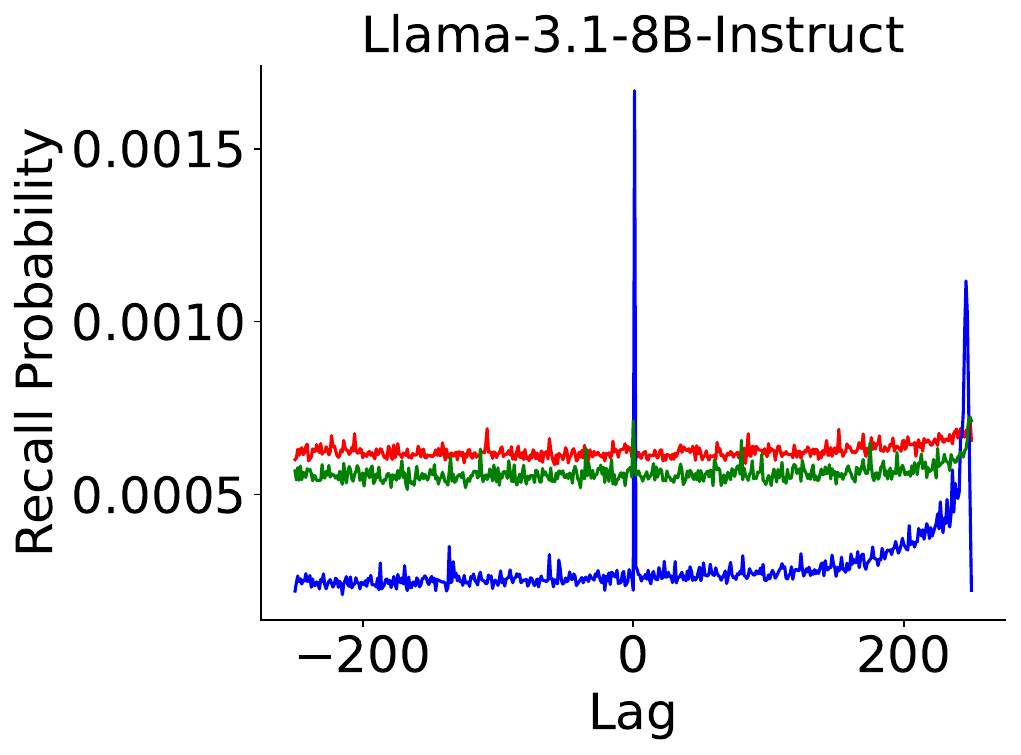
        }} &
        {\includegraphics[width=0.22\textwidth]{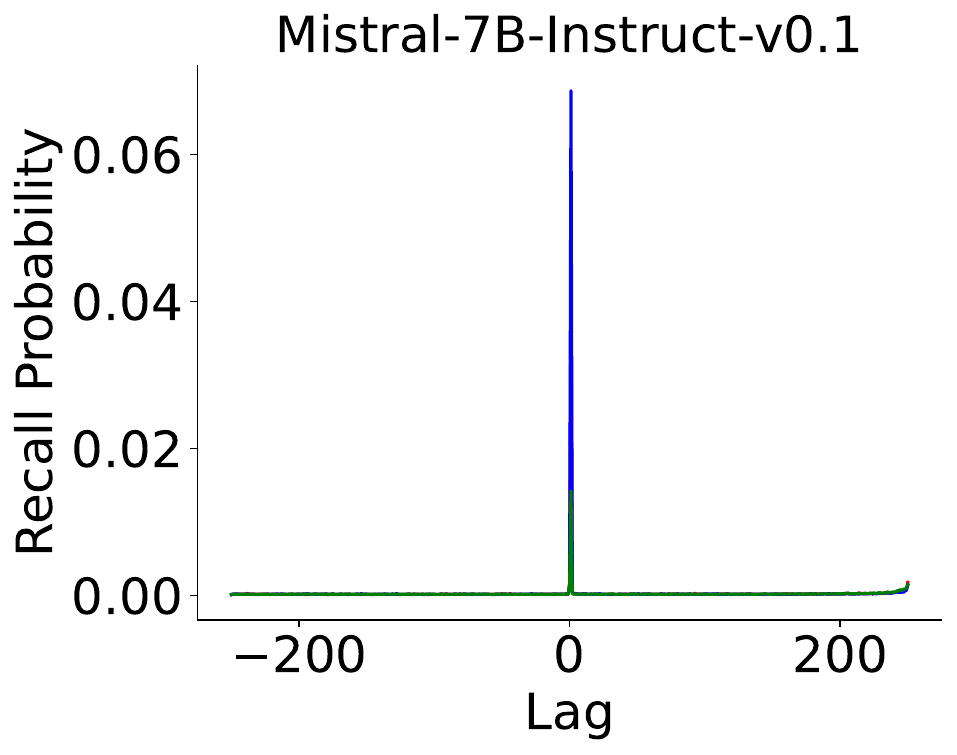}} &
        {\includegraphics[width=0.22\textwidth]{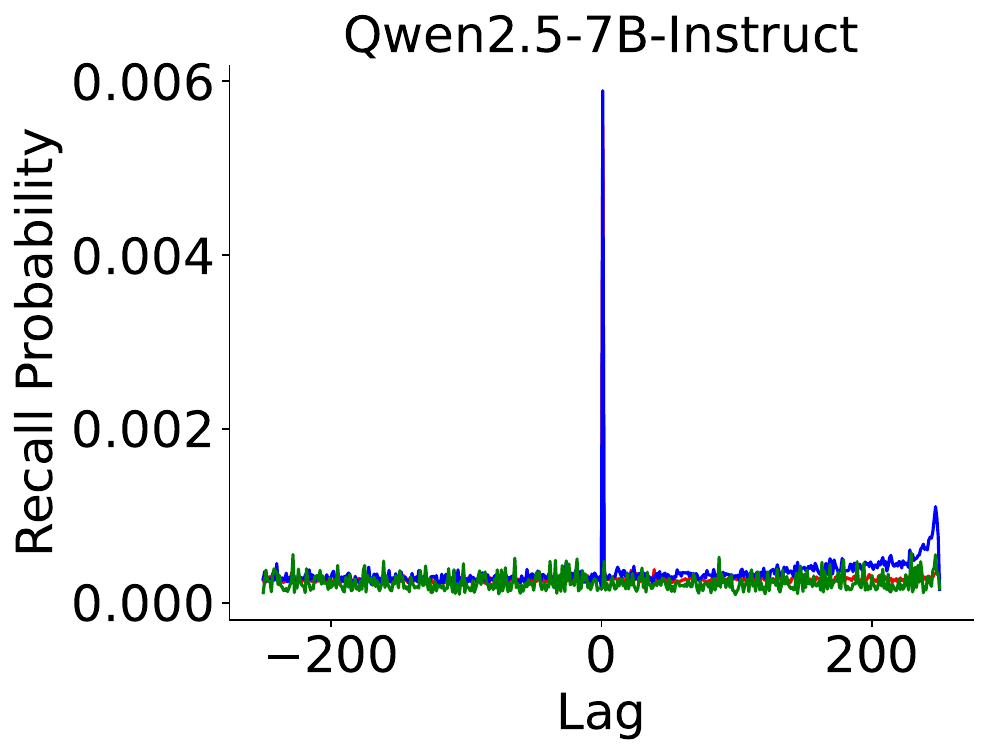}} &
        {\includegraphics[width=0.22\textwidth]{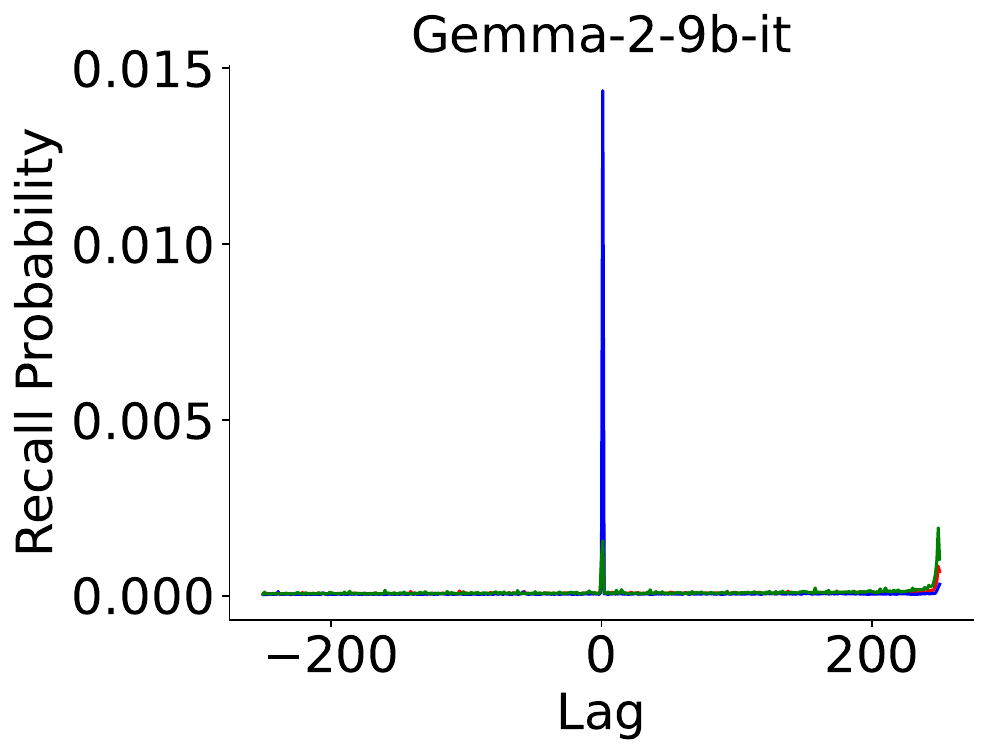}} \\
    \end{tabular}
    \includegraphics[scale=0.5]{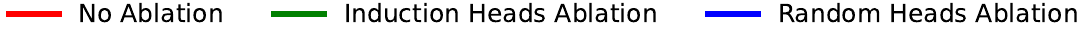}
    \caption{Impact of induction and random head ablation (100 heads in each case) on the model output probability as a function of lag. The models were presented with a sequence of 501 tokens where the last token repeated the token at index 250 and the lag is defined relative to the repeated token (see Methods for more details and Fig.~\ref{fig:CRP_6} for a zoomed version showing lags -6 to 6). The results show averages across 5000 runs with shuffled token sequences.
    Top row: Base models. Bottom row: Instruction-tuned models.
    \label{fig:CRP_250}}
\end{figure*}

\subsection{Temporal dependencies in ICL: Strong tendency for serial recall}

\begin{figure*}[h!]
    \centering
    \begin{tabular}{lllll}
    \textbf{A} &
    \textbf{B} &
    \textbf{C} &
    \textbf{D} \\
        {\includegraphics[width=0.22\textwidth]{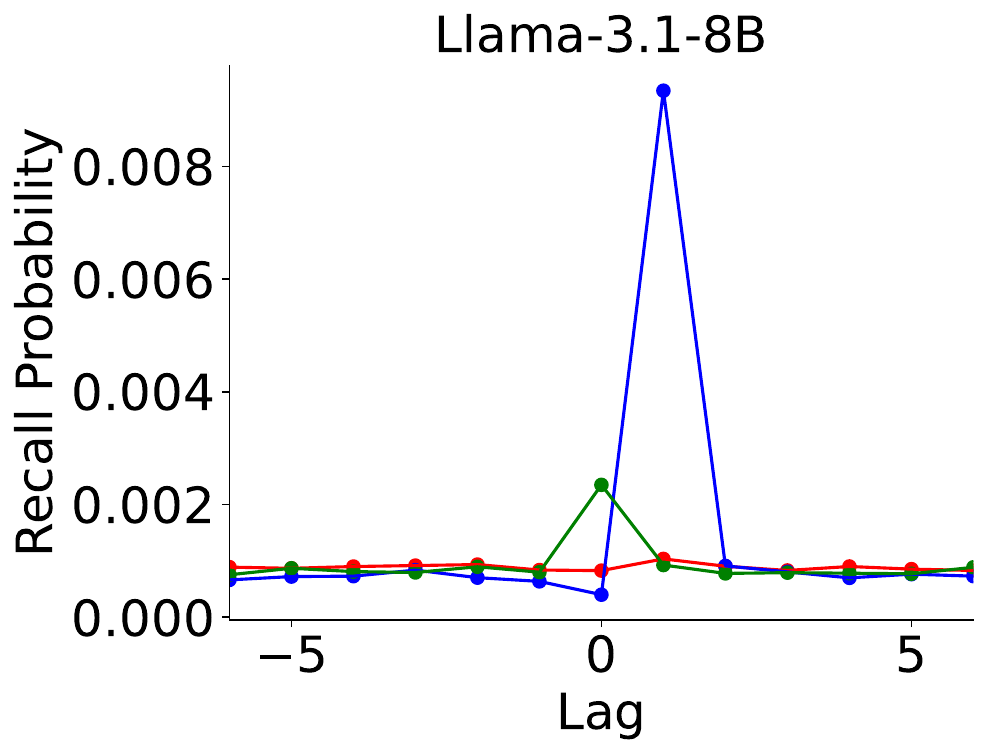
        }} &
        {\includegraphics[width=0.22\textwidth]{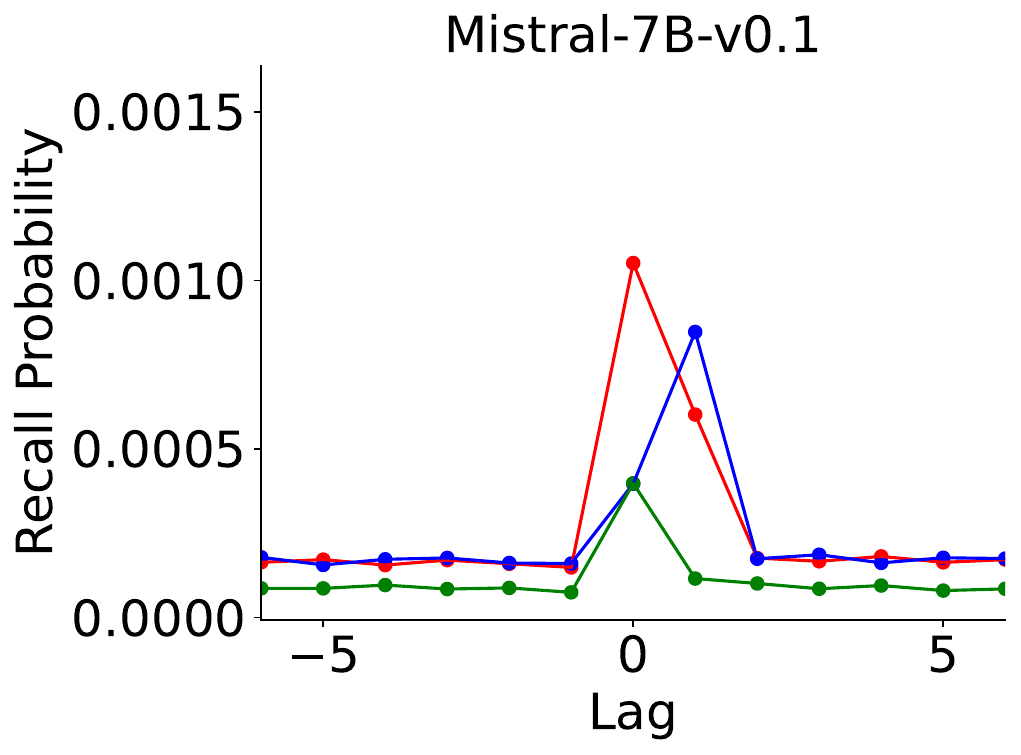}} &
        {\includegraphics[width=0.22\textwidth]{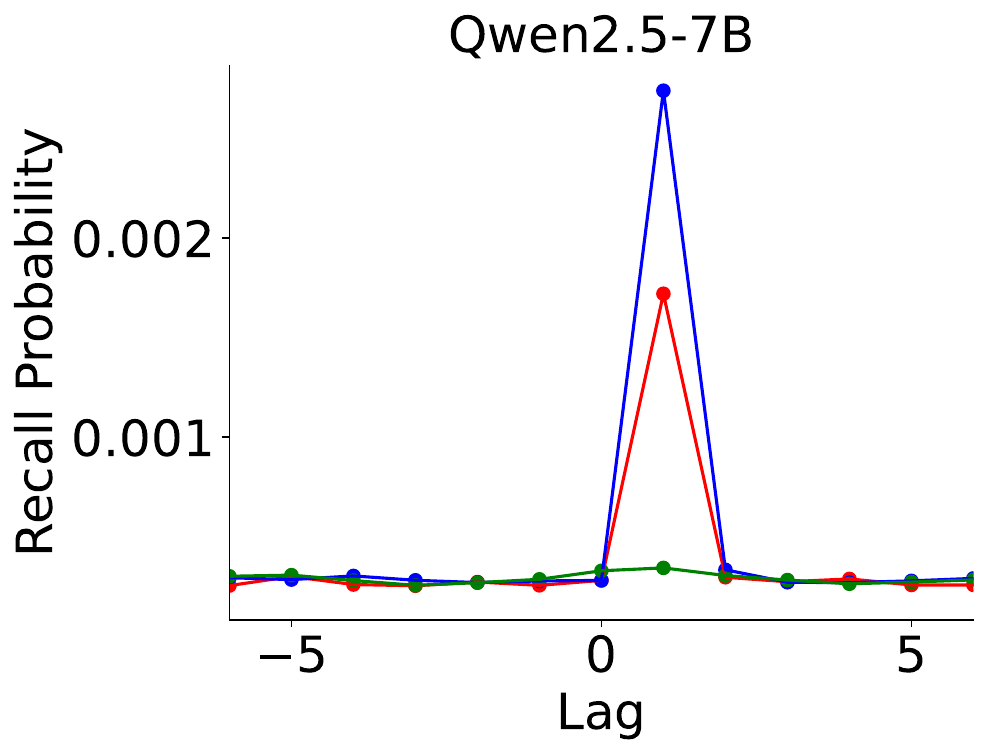}} &
        {\includegraphics[width=0.22\textwidth]{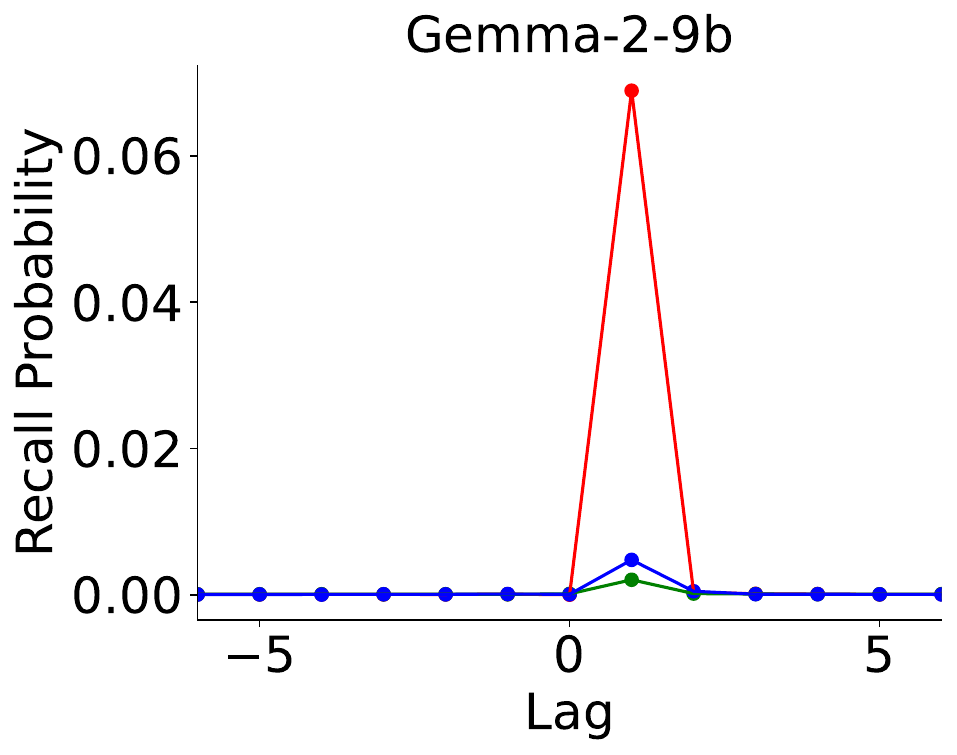}}
        \\
    \end{tabular}
    \begin{tabular}{lllll}
    \textbf{E} &
    \textbf{F} &
    \textbf{G} &
    \textbf{H} \\
        {\includegraphics[width=0.22\textwidth]{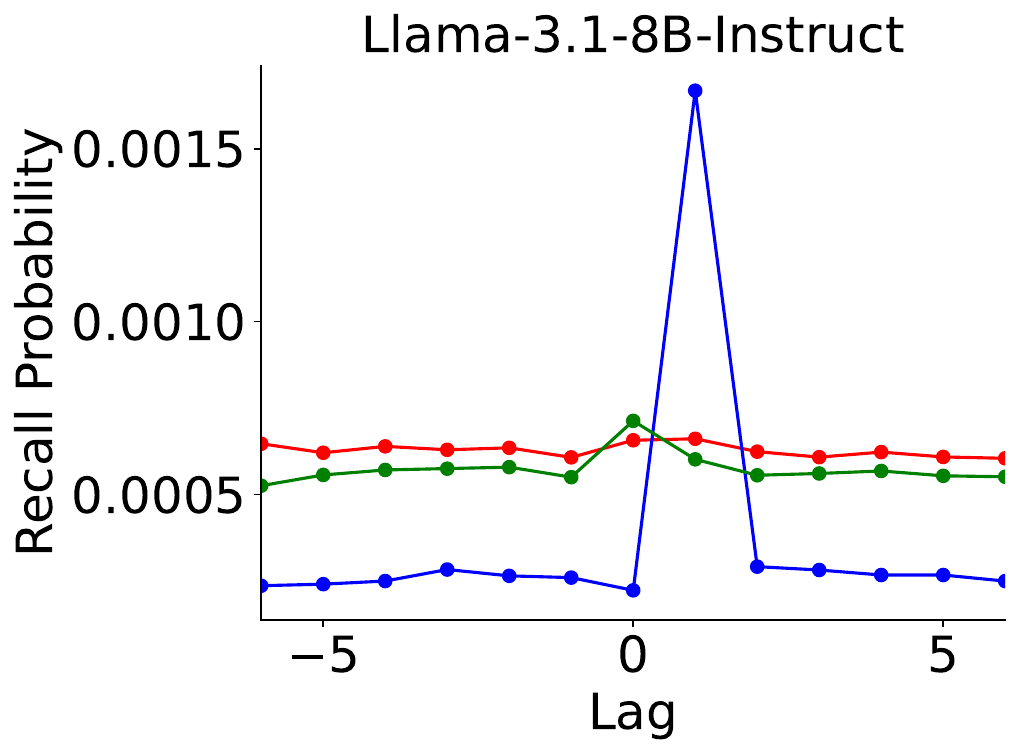
        }} &
        {\includegraphics[width=0.22\textwidth]{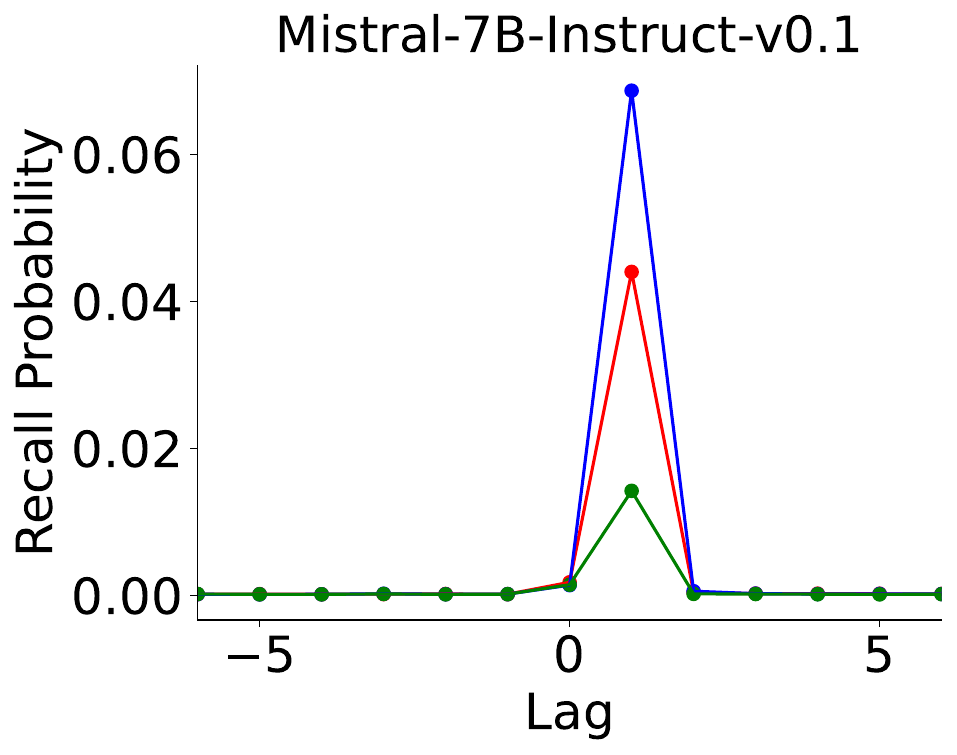}} &
        {\includegraphics[width=0.22\textwidth]{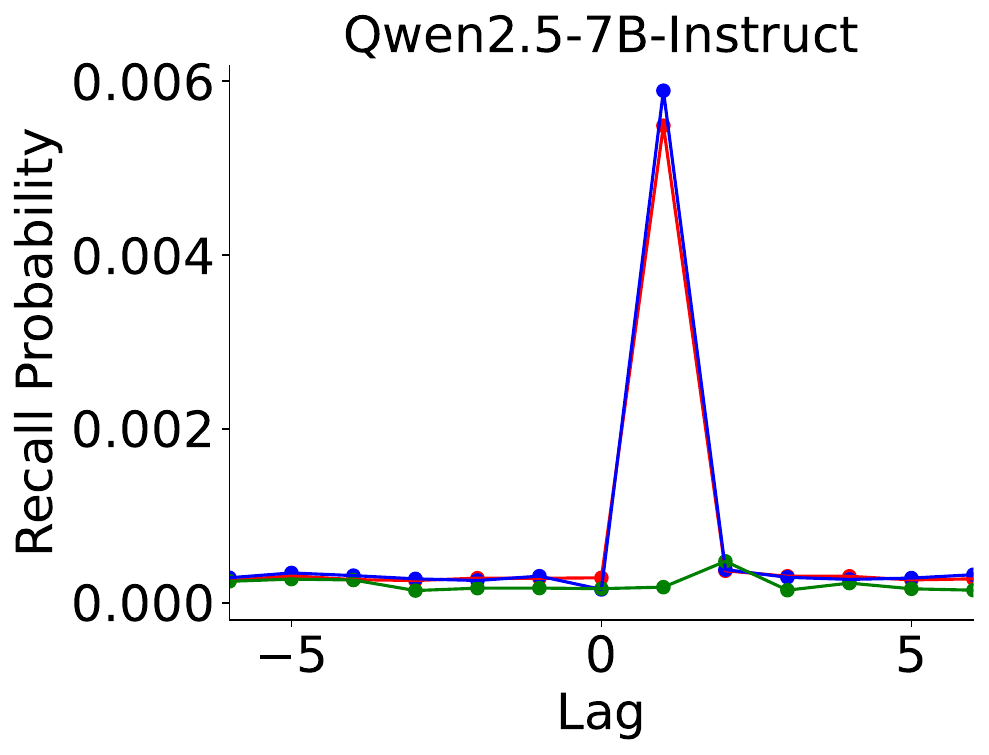}} &
        {\includegraphics[width=0.22\textwidth]{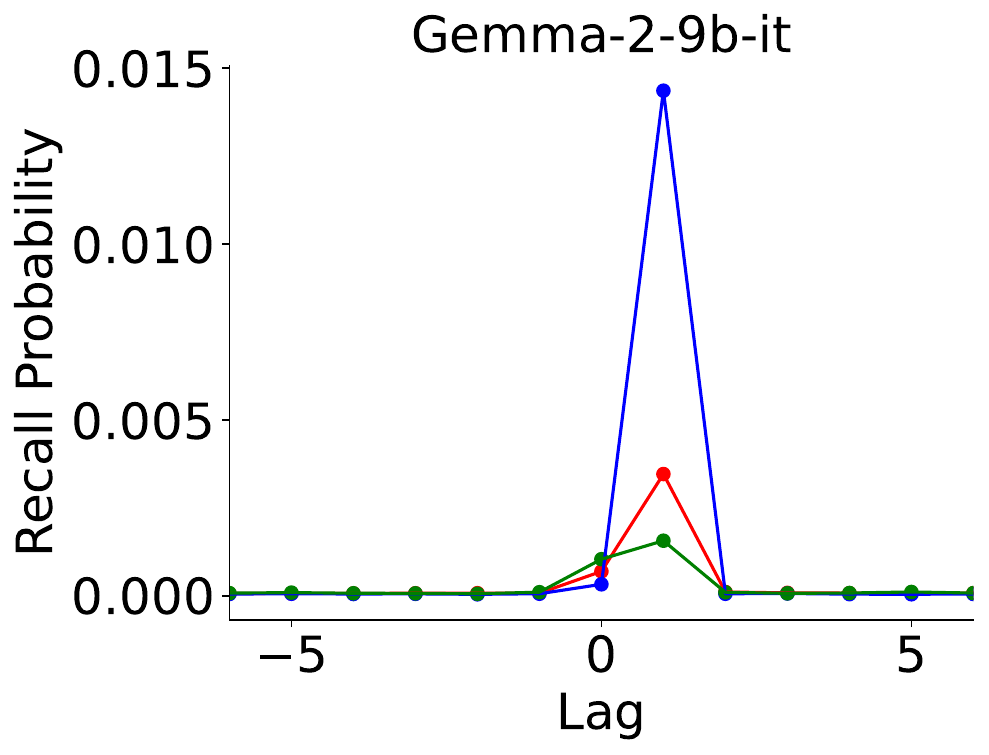}} \\
    \end{tabular}
    \includegraphics[scale=0.5]{Figures/legend.pdf}
    \caption{Impact of induction and random head ablation (100 heads in each case) on the model output probability as a function of lag: same as Fig.~\ref{fig:CRP_250}, but with zoom on lags -6 to 6 to emphasize that the highest probabilities were at lags 1 or 0. Top row: Base models. Bottom row: Instruction-tuned models.
    \label{fig:CRP_6}}
\end{figure*}

We quantified temporal dependencies in ICL using the protocol described in the Methods section (Sec.~\ref{sec:temporal}). We used a sequence of 501 tokens, with the last (501st) token repeating the token at index 250 in the 500-token list, and quantified the output probabilities for each of the remaining 499 tokens. After averaging across 5000 permutations, we visualized those probabilities as a function of lag (distance) from the repeated token. The results are shown in Fig.~\ref{fig:CRP_250} for all 499 lags, and in Fig.~\ref{fig:CRP_6} we show a zoomed-in version focusing on the central lags from -6 to +6 (red line).

From Fig.~\ref{fig:CRP_6}, we notice that the base versions of Qwen and Gemma, and the instruction-tuned versions of Mistral, Qwen and Gemma, show a prominent peak at lag +1 (serial recall). The base version of Mistral shows a prominent peak at lag 0 (copying of the current token), while Llama had only a slight increase at lag +1. Notably, Mistral shifts from lag 0 in the base model to lag +1 after instruction tuning, suggesting a change in retrieval mode from copying toward successor retrieval. From Fig.~\ref{fig:CRP_250} we observe a slight increase for higher lags (recency effect) in Llama, Mistral, and Gemma. Apart from these slight deviations, the models maintain relatively flat output probabilities across lags, indicating that this approach effectively removed semantic components and isolated temporal dependencies.

\subsection{Impact of induction head ablation on the temporal dependencies: Systematic reduction of the serial recall tendency}

\begin{figure*}[h!]
    \centering
    \begin{tabular}{lllll}
    \textbf{A} &
    \textbf{B} &
    \textbf{C} &
    \textbf{D} \\
        {\includegraphics[width=0.22\textwidth]{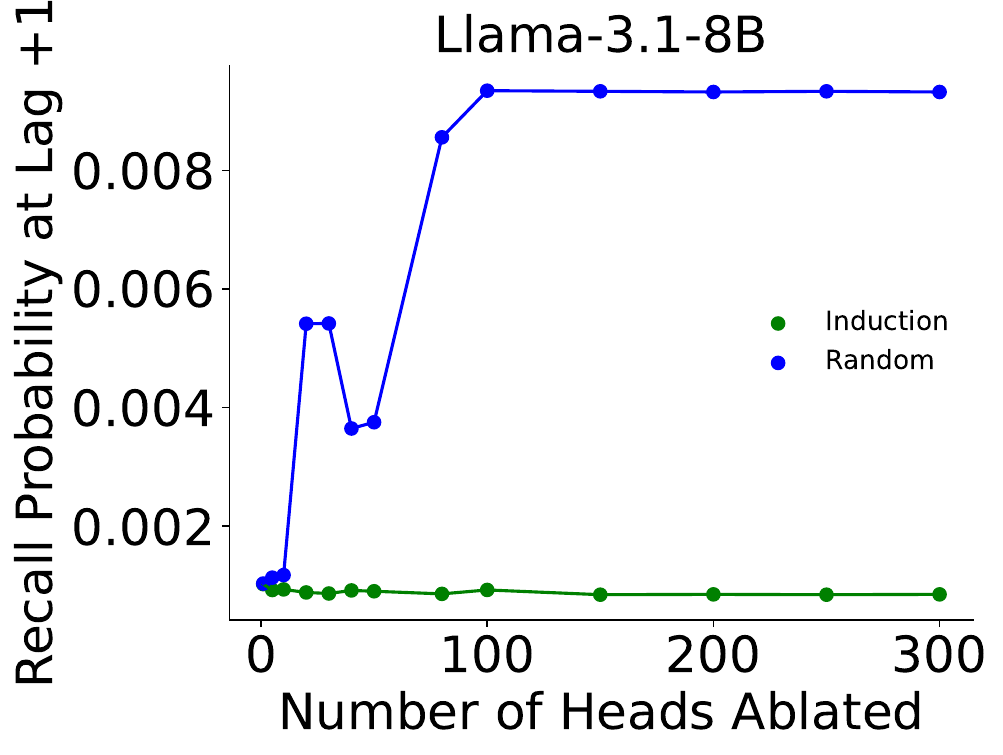
        }} &
        {\includegraphics[width=0.22\textwidth]{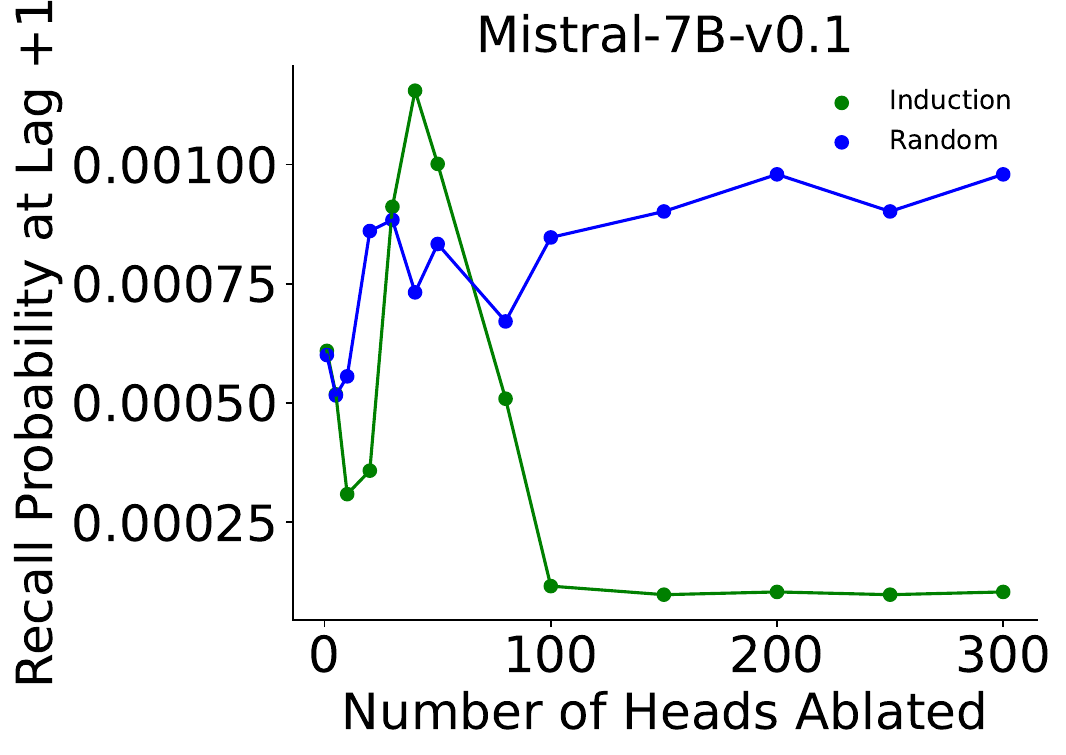}} &
        {\includegraphics[width=0.22\textwidth]{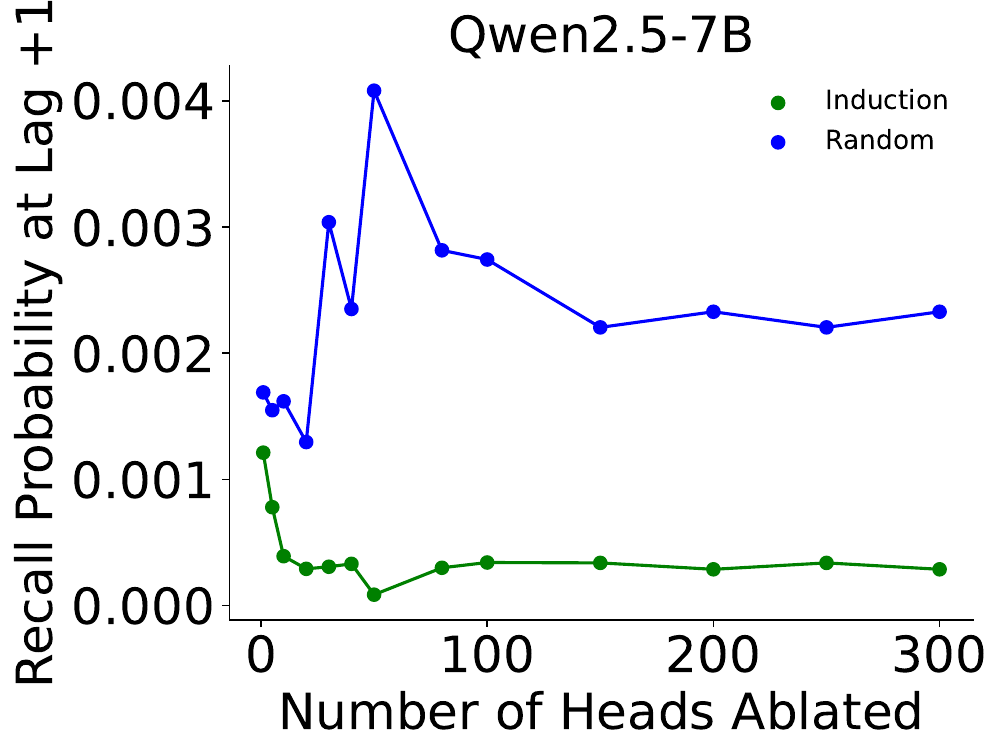}} &
        {\includegraphics[width=0.22\textwidth]{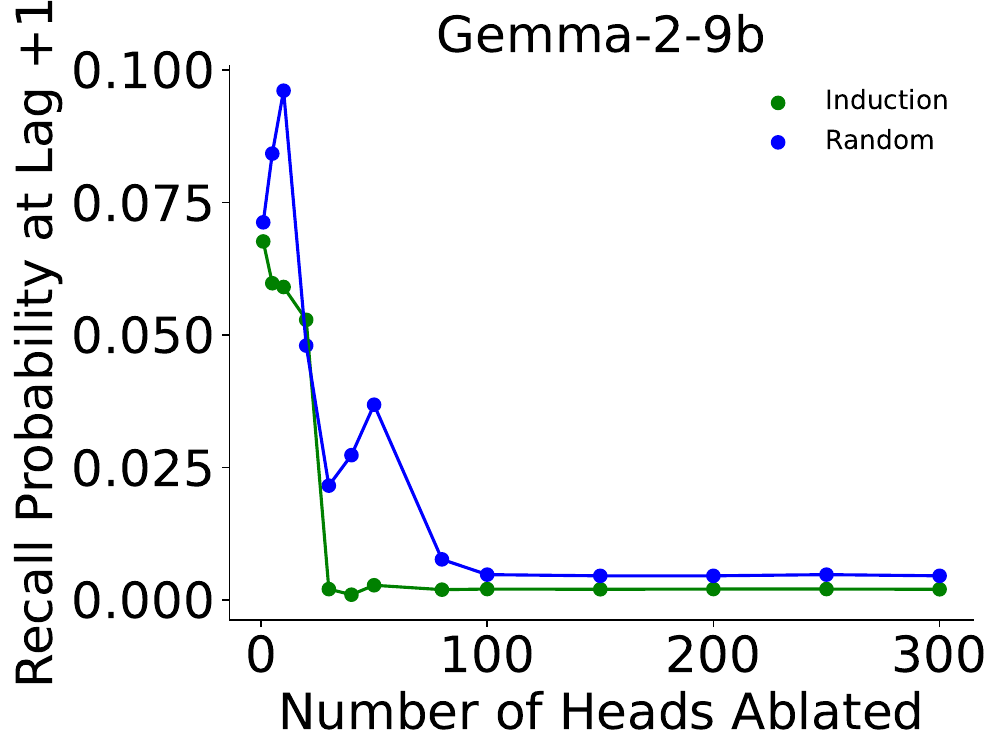}} \\
    \end{tabular}
    \begin{tabular}{lllll}
    \textbf{E} &
    \textbf{F} &
    \textbf{G} &
    \textbf{H} \\
         {\includegraphics[width=0.22\textwidth]{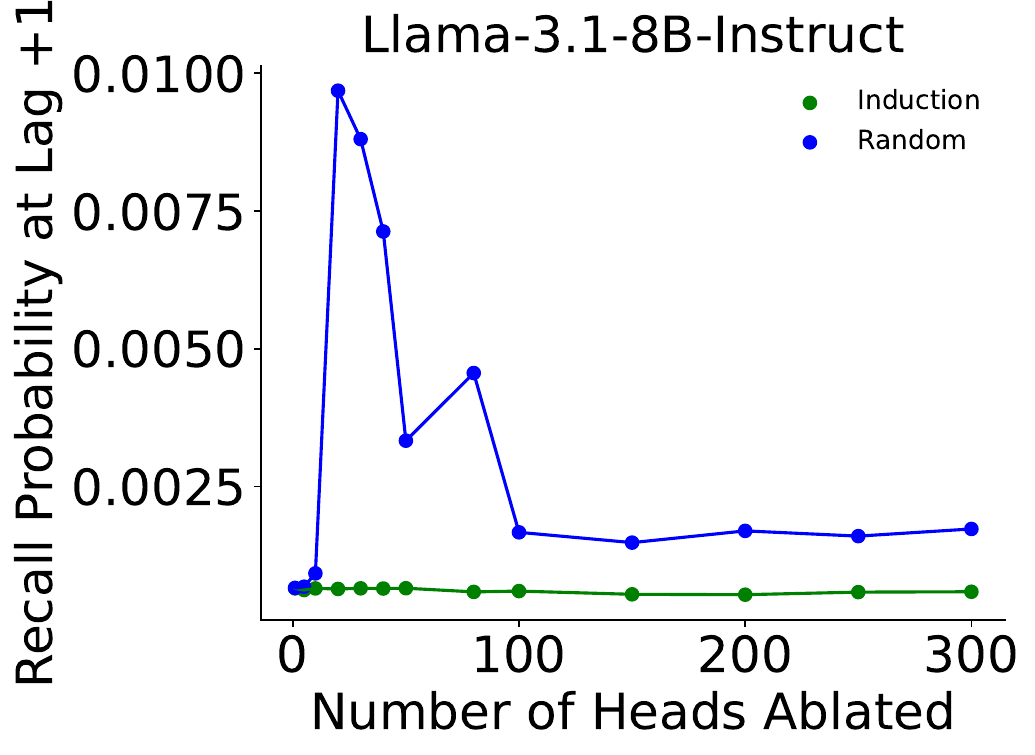
        }} &
        {\includegraphics[width=0.22\textwidth]{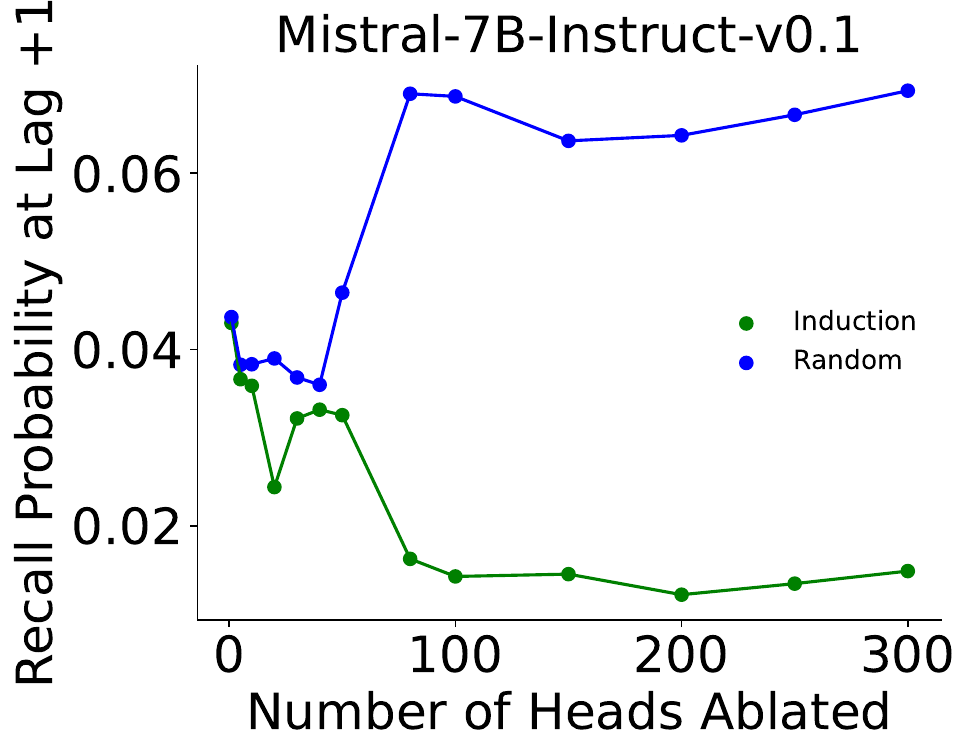}} &
        {\includegraphics[width=0.22\textwidth]{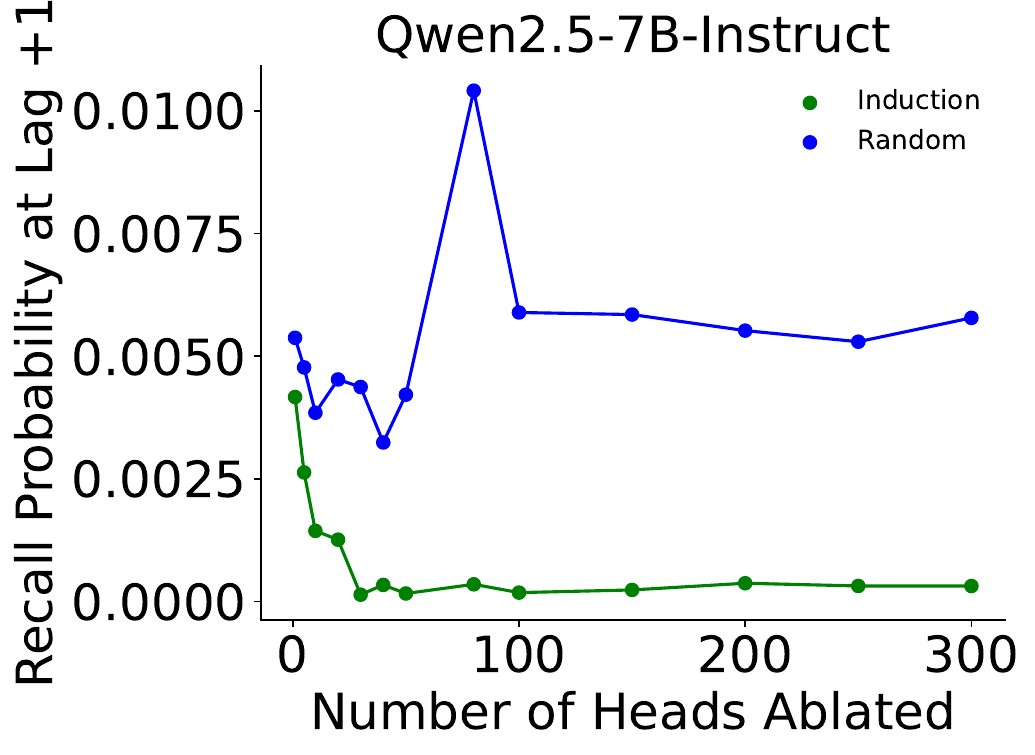}} &
        {\includegraphics[width=0.22\textwidth]{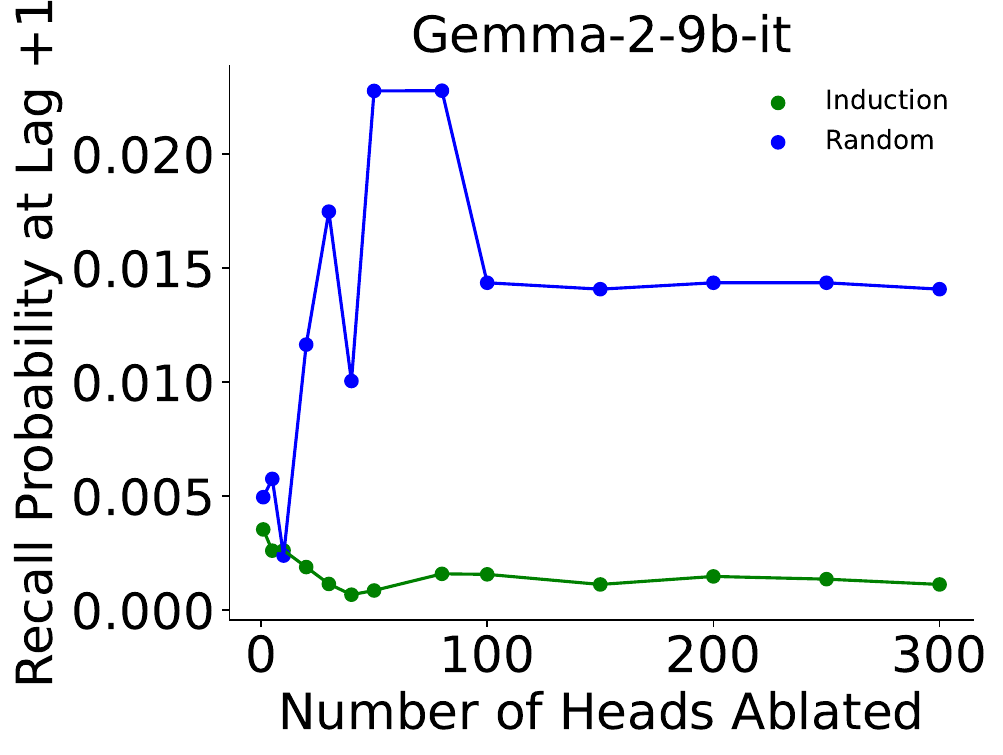}} \\
    \end{tabular}
    \caption{Impact of induction head ablation on the model output probability for the token at lag +1. The models were presented with a sequence of 501 tokens where the last token repeated the token at index 250 and the lag is defined relative to the repeated token, hence the probability at lag +1 is the probability that the model assigns to token 251 (see Methods for more details). The results show averages across 5000 runs with shuffled token sequences.  We ablated the following numbers of induction heads (sorted by the induction scores) and random heads (x-axis): 1, 5, 10, 20, 30, 40, 50, 80, 100, 150, 200, 250 and 300. Top row: Base models. Bottom row: Instruction-tuned models. Exact numerical values for these experiments are shown in Table~\ref{tab:tab_indhead_base} and Table~\ref{tab:tab_indhead_inst} for induction head ablation in base and instruction-tuned models respectively and in Table~\ref{tab:tab_randhead_base} and Table~\ref{tab:tab_randhead_inst} for random head ablation in base and instruction-tuned models respectively.
    \label{fig:lag1_probability}}
\end{figure*}

 \begin{table*}[h]
    \centering
    \caption{Impact of \textit{induction} head ablation on the model output probability for the token at lag +1 for different models. \textit{The results are for instruction-tuned models.} These results are visualized in Fig.~\ref{fig:lag1_probability}. Corresponding results for ablation of random heads are shown in Table~\ref{tab:tab_randhead_inst}.}
    {\setlength{\tabcolsep}{4pt}
    \resizebox{\textwidth}{!}{%
    \begin{tabular}{ccccc}
        \toprule
        \textbf{\# Induction Heads Ablated} & \textbf{Llama-3.1-8B-Instruct} & \textbf{Mistral-7B-Instruct-v0.1} & \textbf{Qwen2.5-7B-Instruct} & \textbf{Gemma-2-9b-it} \\
        \midrule
        1   & 0.00065  & 0.04300  & 0.00417  & 0.00354 \\
        5   & 0.00062  & 0.03662  & 0.00263  & 0.00261 \\
        10  & 0.00065  & 0.03587  & 0.00144  & 0.00262 \\
        20  & 0.00064  & 0.02438  & 0.00126  & 0.00189 \\
        30  & 0.00065  & 0.03216  & 0.00014  & 0.00115 \\
        40  & 0.00065  & 0.03316  & 0.00034  & 0.00068 \\
        50  & 0.00065  & 0.03253  & 0.00016  & 0.00086 \\
        80  & 0.00059  & 0.01623  & 0.00035  & 0.00159 \\
        100 & 0.00060  & 0.01424  & 0.00018  & 0.00157 \\
        150 & 0.00054  & 0.01450  & 0.00023  & 0.00113 \\
        200 & 0.00054  & 0.01217  & 0.00037  & 0.00148 \\
        250 & 0.00058  & 0.01343  & 0.00032  & 0.00136 \\
        300 & 0.00059  & 0.01485  & 0.00031  & 0.00113 \\
        \bottomrule
    \end{tabular}}}
    \label{tab:tab_indhead_inst}
\end{table*}

To better understand the key components behind the temporal dependencies in LLMs, we ablated induction heads through a procedure described in the Methods section. We ablated heads gradually, starting with the one that had the highest induction score. The results after ablating 100 heads are again shown in Fig.~\ref{fig:CRP_250} for all 499 lags and in Fig.~\ref{fig:CRP_6} for -6 to +6 central lags (green line for induction head ablation and blue line for ablation of random heads).

We observed a clear reduction of the lag +1 probability. For the base versions of Qwen and Gemma and for the instruction-tuned versions of Mistral, Qwen, and Gemma, the prominent increase at lag +1 prior to ablation was substantially reduced and, in some cases, nearly eliminated. For models with low lag +1 probability in the non-ablated condition, induction-head ablation kept lag +1 low. For base Mistral and Llama, ablation caused an increase at lag 0, suggesting a switch from serial-recall behavior to repetition. Random-head ablation showed a different pattern: in all cases except base Gemma, it increased the lag +1 probability, indicating an increased tendency toward serial recall when fewer non-induction heads were present. This pattern is consistent with the possibility of competing circuitry, in which some non-induction heads counterbalance or dilute successor-style retrieval in the intact model. For base Mistral, random-head ablation shifted the peak from lag 0 to lag +1. On a larger scale (Fig.~\ref{fig:CRP_250}), ablations also tended to amplify the recency effect.

More detailed examination indicates a relatively gradual impact of induction head ablation on the lag +1  probability. This is shown in Fig~\ref{fig:lag1_probability} where lag +1 probability is shown as a function of the number of ablated induction and random heads (for detailed numerical results see Table~\ref{tab:tab_indhead_base}, Table~\ref{tab:tab_indhead_inst}, Table~\ref{tab:tab_randhead_base}, and Table~\ref{tab:tab_randhead_inst}). Although the reduction in lag +1 probability was not smooth, it had an overall downward trend as more induction heads were ablated, particularly for Mistral, Qwen, and Gemma. This trend was not present for random-head ablation, except for base Gemma. In general, gradual ablation of random heads had a more variable impact across models and often increased lag +1 probability.

To determine whether the serial‐recall bias mediated by induction heads is localized to particular parts of the network, we performed ablations restricted to either the top 50\% of layers or the bottom 50\% of layers (see top‐half ablation in Fig.~\ref{fig:CRP_250_top} and Fig.~\ref{fig:lag1_probability_top} and bottom‐half ablation in Fig.~\ref{fig:CRP_250_bottom} and Fig.~\ref{fig:lag1_probability_bottom}). Ablating induction heads from only the top or bottom layers resulted in a less substantial reduction of the +1 lag probability compared to ablating an equivalent number of induction heads across all layers (see Fig.~\ref{fig:CRP_6_top} and Fig.~\ref{fig:CRP_6_bottom} for zoomed views emphasizing lags -6 to 6). This trend was consistent across the models tested, including Llama, Mistral, Qwen, and Gemma, in both base and instruction-tuned versions. Together, these results point to a distributed circuit: the induction heads driving serial-recall-like behavior are spread throughout model depth rather than being confined to only early or late layers.

Finally, we tested a mean ablation, an alternative to zero ablation, where the attention scores of the ablated heads were replaced with their mean value rather than set to $-\infty$. This method enforces uniform attention across the sequence, neutralizing the specific biases of the targeted heads. Compared to zero ablation, mean ablation of induction heads produced qualitatively similar results, but with a less pronounced reduction of the +1 lag probability (Fig.~\ref{fig:CRP_250_mean}-Fig.~\ref{fig:lag1_probability_mean}), indicating that while the serial recall effect was reduced, completely eliminating the heads' influence had a stronger effect.

\subsection{Induction Head Ablation Impairs Serial Recall in ICL}
\label{sec:serial_recall_ICL_results}

\begin{figure}[h!]
    \centering
        {\includegraphics[width=1\columnwidth]{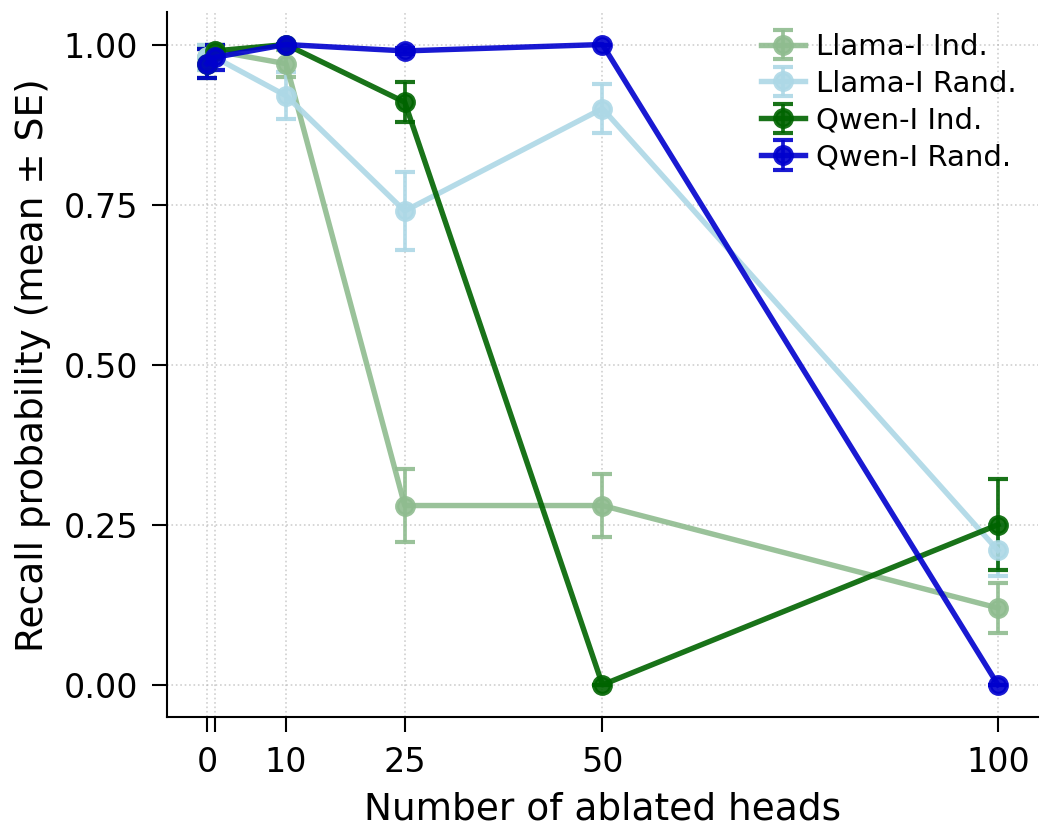}}
    \caption{Model performance in ICL is more sensitive to ablation of induction than random heads. Conditional recall probability at lag +1 for Llama-3.1-8B-Instruct and Qwen2.5-7B-Instruct as a function of the number of ablated induction and random attention heads. Exact values are shown in Table~\ref{tab:ICL_serial_recall}.
    \label{fig:ICL_serial_recall}}
\end{figure}

The results of the ICL serial recall experiment described in Methods (Sec.~\ref{sec:serial_recall}) are presented in Table~\ref{tab:ICL_serial_recall} and Fig.~\ref{fig:ICL_serial_recall} (see also Fig.~\ref{fig:CRP_ablation_grid} for CRP plots with different numbers of ablated induction and random heads, as well as Fig.~\ref{fig:SPC_ablation_grid} for recall probability at each serial position in the same ablation conditions). We evaluated Llama and Qwen in both base and instruction-tuned variants; Mistral and Gemma did not reliably learn the task. Figure~\ref{fig:ICL_serial_recall} visualizes the instruction-tuned models, whereas Table~\ref{tab:ICL_serial_recall} reports all four model variants. For both instruction-tuned models, ablating induction heads generally led to a larger degradation in serial recall performance than ablating random heads or applying no ablation. Without any ablation, both models demonstrated high lag +1 probability (Llama-I: $0.98$; Qwen-I: $0.97$). While ablating a single head had minimal impact, the difference between induction- and random-head ablation became evident starting from 10 ablated heads and increased thereafter. For instance, with 50 induction heads ablated, Llama-I's lag +1 probability dropped to $0.28$, whereas random ablation yielded $0.90$. The same qualitative pattern is present in the base models reported in Table~\ref{tab:ICL_serial_recall}. Overall recall probability was typically more affected by induction-head ablation than by random-head ablation (Fig.~\ref{fig:SPC_ablation_grid}).

\section{Discussion}

Our findings indicate that in-context temporal effects in LLMs can be linked to underlying attention mechanisms, particularly induction heads. The observation that models such as Qwen, Gemma, and Mistral consistently show an increased probability for the token immediately following the repeated token (i.e., lag +1) indicates reliance on temporal context during ICL. Unlike humans, who tend to show elevated recall probability for a temporal neighborhood surrounding the repeated token (highest recall probability for lag +1, followed by lag -1, lag +2, lag -2, lag +3, etc. \cite{kahana1996associative,b2010temporal}), the LLMs we studied show a stronger tendency toward serial recall, with elevated probabilities concentrated at +1 and only occasional increases at lag 0 and higher lags that indicate recency effects. This sharper, +1-centered profile suggests a transformer-specific retrieval phenomenon rather than simply a reproduction of human-like memory.

Our ablation studies revealed an important role of induction heads in these temporal effects. The substantial decrease and, in some models, near loss of the +1 lag bias upon ablating heads with high induction scores aligns with prior mechanistic studies \cite{crosbie2024induction,singh2024needs}. This mechanistic relationship was further supported by the few-shot serial recall task, where induction-head ablation impaired performance more strongly than ablating the same number of random heads. Taken together, these results point to mechanistic specificity: induction heads appear especially important for temporal retrieval and ordered recall. The top-half and bottom-half ablations likewise suggest a distributed circuit rather than a sharply localized module. Recent work has also highlighted complementary task-vector and function-vector mechanisms, especially for more abstractive settings \cite{hendel2023context,todd2023function,yinattention,csahincontext}. Even in light of those broader accounts, our results point to a mechanistically specific role for induction heads in temporal retrieval and ordered recall.

We found that instruction tuning had a relatively modest impact on induction scores and downstream in-context temporal properties, with Mistral standing out as the clearest exception. Overall, we observed a relatively high degree of heterogeneity across models. This observation resonates with recent investigations into how different training and fine-tuning regimes affect memory-like properties in LLMs \cite{liu2024lost,wang2023primacy}. Future work should systematically explore how various fine-tuning strategies influence the interplay between architectural components and temporal processing.

Ablation of random heads had a tendency to increase the probability of the +1 lag. This is not surprising since random heads were always chosen from the heads that were not in the top 300 heads in terms of the induction score. Hence, having fewer heads with low induction scores caused the increase in lag +1 probability, consistent with the hypothesis that the heads with high induction scores shaped this effect. The fact that removing these lower-induction heads often sharpened the +1 effect is also compatible with competing circuitry, where some heads oppose or dilute successor-style retrieval in the intact model.

Ablation of induction and random heads commonly caused an increase in recency. We leave for future work to explore what specific properties in attention heads might support the recency effect, commonly observed in humans \cite{kahana1996associative} and LLMs (manifested through high importance of more recent context \cite{liu2024lost}). After the ablation of induction heads, some models (Llama and the instruction-tuned version of Gemma) showed an increase at lag 0 probability, which indicates a copying tendency (i.e., repeating the same token). This is consistent with copying attention heads, which show properties similar to induction heads but at lag 0 \cite{elhage2021mathematical}.

This study demonstrates that induction heads are important for the temporal dynamics of ICL, mediating serial-recall-like behavior in LLMs. These findings, situated within a framework informed by cognitive science, offer new insights into how transformer architectures process sequential information and provide a promising direction for future research aimed at advancing LLMs and bridging artificial and human cognition.

\section{Limitations}
Our objective was to understand temporal dependencies in ICL and the role of induction heads in those dependencies. Thus, we focused on quantifying token probabilities in a task inspired by the human free recall and serial recall tasks. We did not evaluate the models on other ICL tasks. For this, we refer to recent literature. For instance, \citet{crosbie2024induction}, who examined the effects of ablating induction and random heads on various ICL tasks. The authors found that induction heads play an important role in these tasks, making a much larger impact on performance than random heads. Future work could consider other types of heads and complementary mechanisms \citep{bansal2022rethinking,hendel2023context,todd2023function,yinattention,park2025does,csahincontext} to understand further how attention heads shape the temporal structure of model outputs.

Our analysis was restricted to 500 frequent words from the English language. While conducting experiments with frequent words is common for free and serial recall studies with human participants, we do not know whether using different sets of tokens might lead to different results. Finally, while our models were relatively large (7B to 9B parameters), they are not among the largest currently available models. The results might differ if the same approach is applied to those models.

\section*{Acknowledgments}
This research was supported in part by Lilly Endowment, Inc., through its support for the Indiana University Pervasive Technology Institute.

\bibliography{references}

\clearpage
\appendix

\begin{appendices}

\section{Additional figures and tables}

\renewcommand{\thetable}{A\arabic{table}}
\renewcommand{\thefigure}{A\arabic{figure}}
\setcounter{table}{0}
\setcounter{figure}{0}


\begin{figure}[!htbp]
    \centering
    \begin{tabular}{lllll}
    \textbf{A} &
    \textbf{B} &
    \textbf{C} &
    \textbf{D} \\
        {\includegraphics[width=0.215\textwidth]{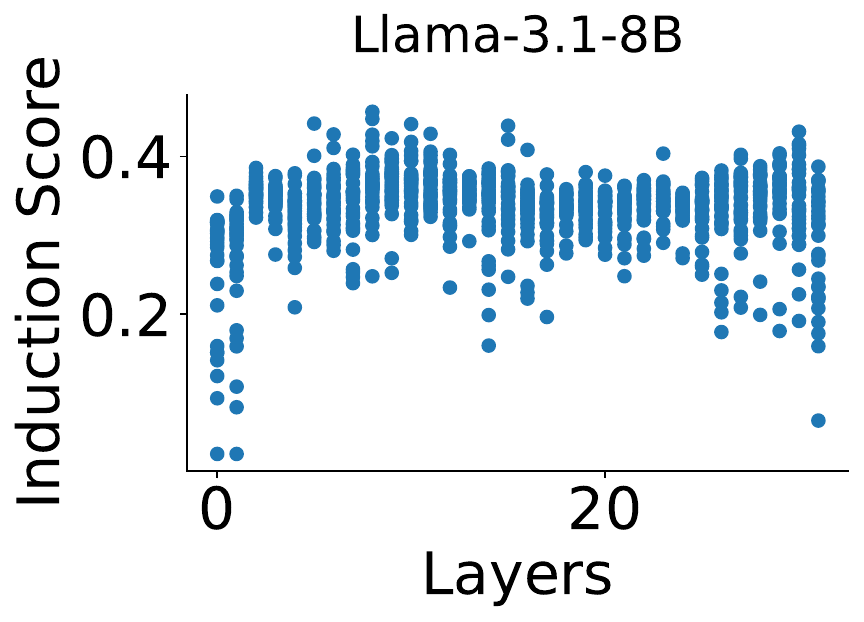}} &
        {\includegraphics[width=0.215\textwidth]{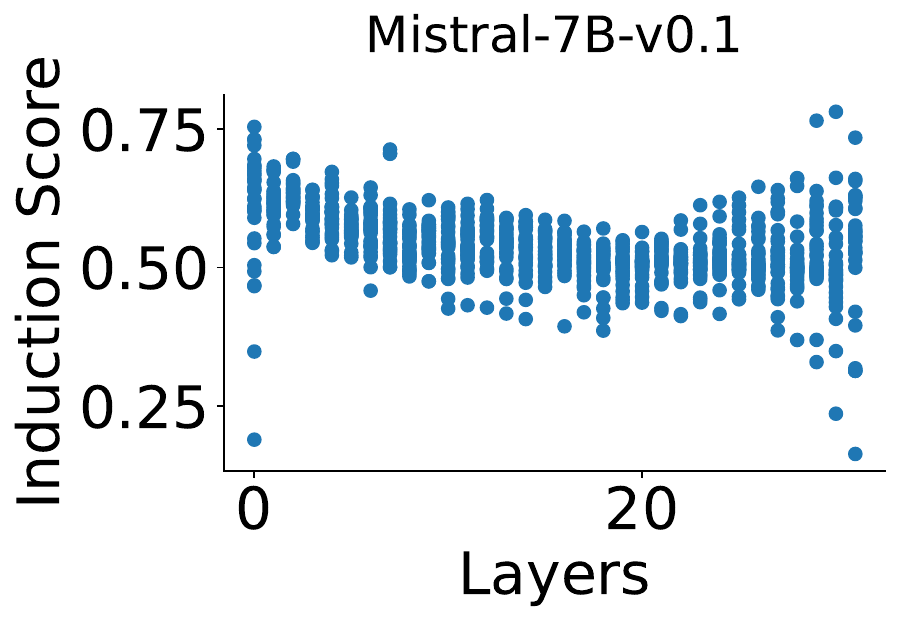}} &
        {\includegraphics[width=0.215\textwidth]{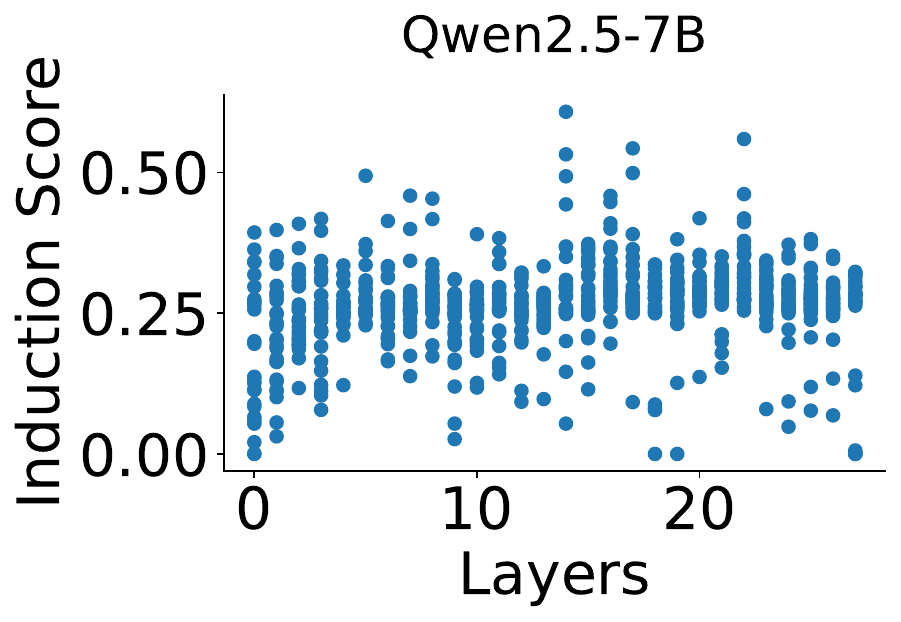}} &
        {\includegraphics[width=0.215\textwidth]{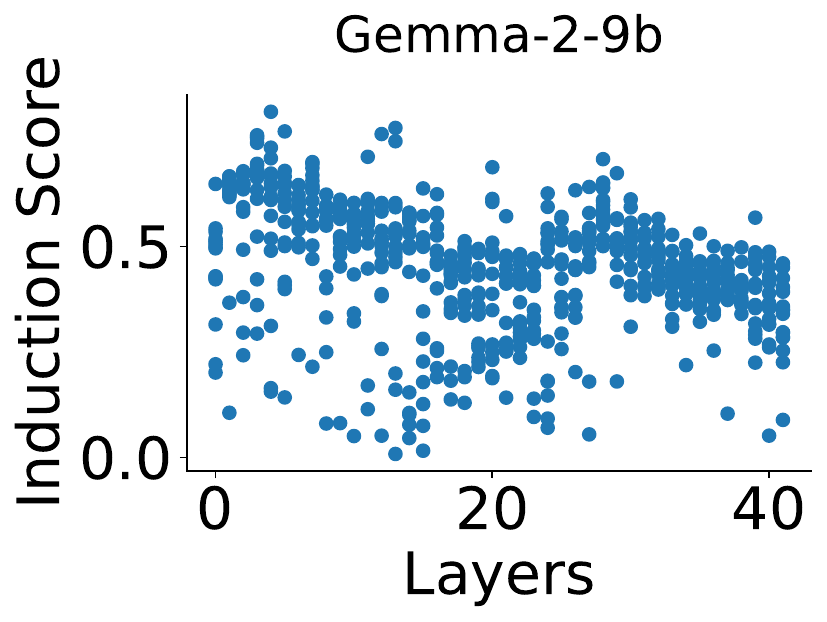}} \\
    \end{tabular}
    \begin{tabular}{lllll}
    \textbf{E} &
    \textbf{F} &
    \textbf{G} &
    \textbf{H} \\
        {\includegraphics[width=0.215\textwidth]{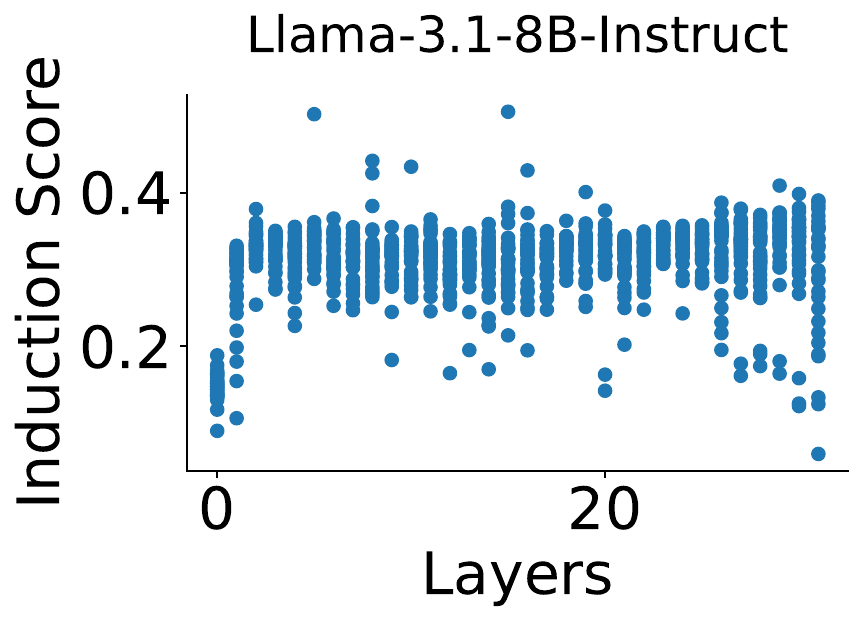}} &
        {\includegraphics[width=0.215\textwidth]{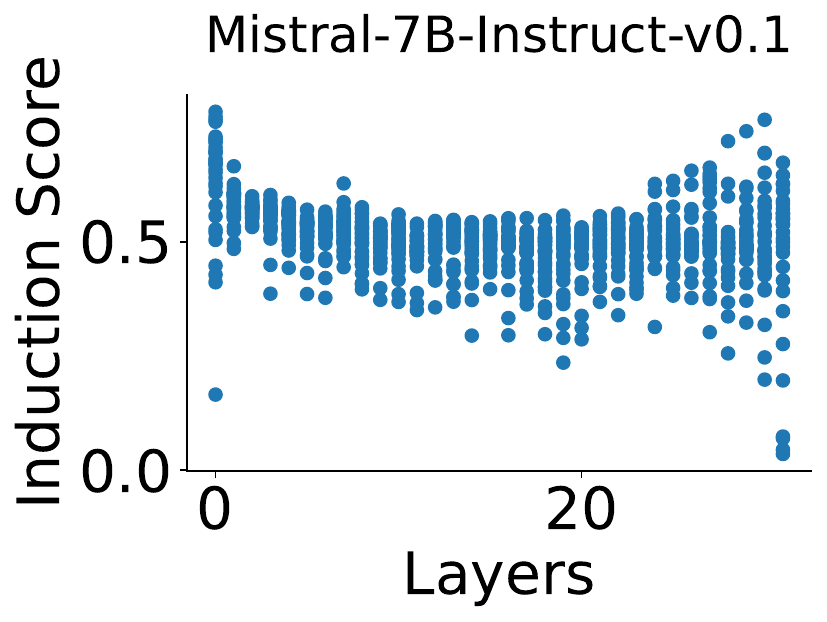}} &
        {\includegraphics[width=0.215\textwidth]{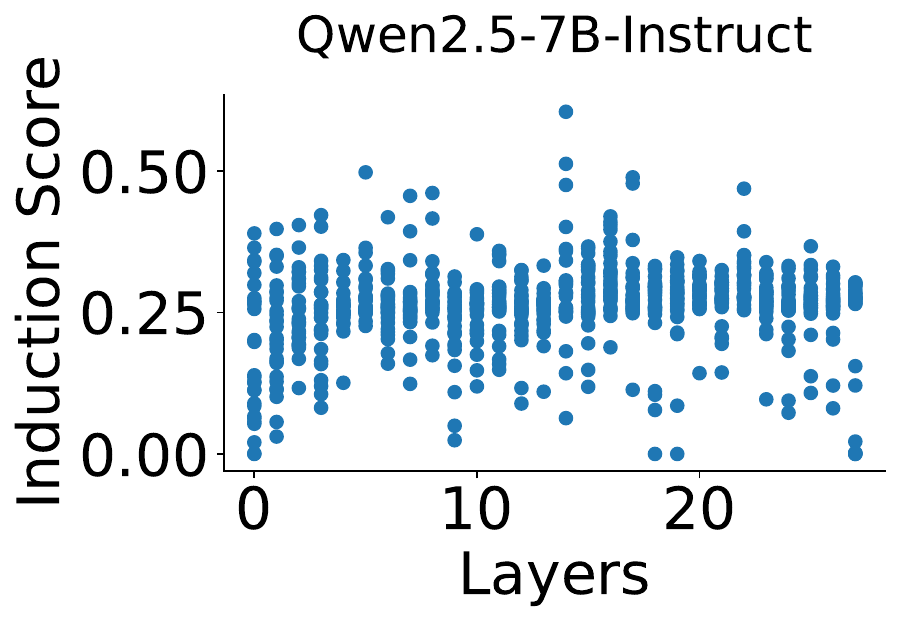}} &
        {\includegraphics[width=0.215\textwidth]{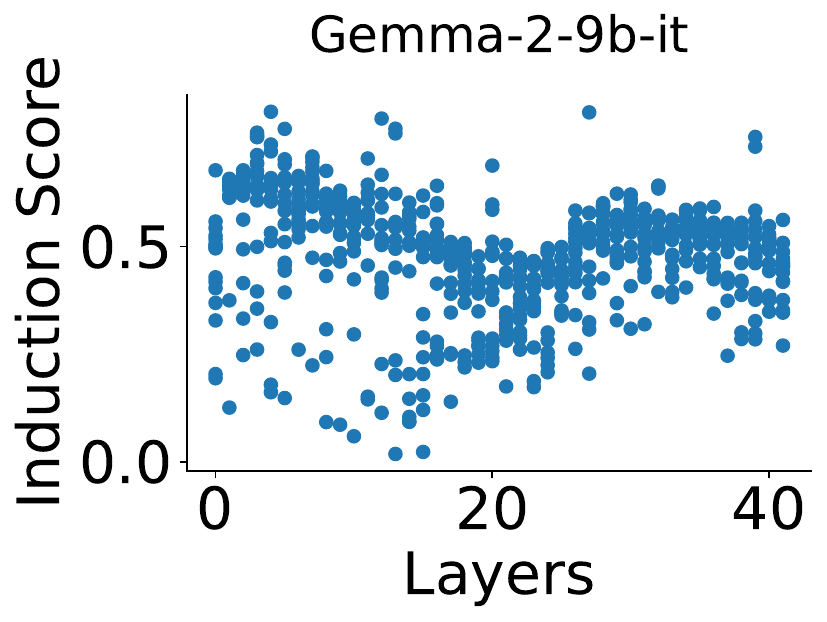}} \\
    \end{tabular}
    \begin{tabular}{lllll}
    \textbf{I} &
    \textbf{J} &
    \textbf{K} &
    \textbf{L} \\
        {\includegraphics[width=0.215\textwidth]{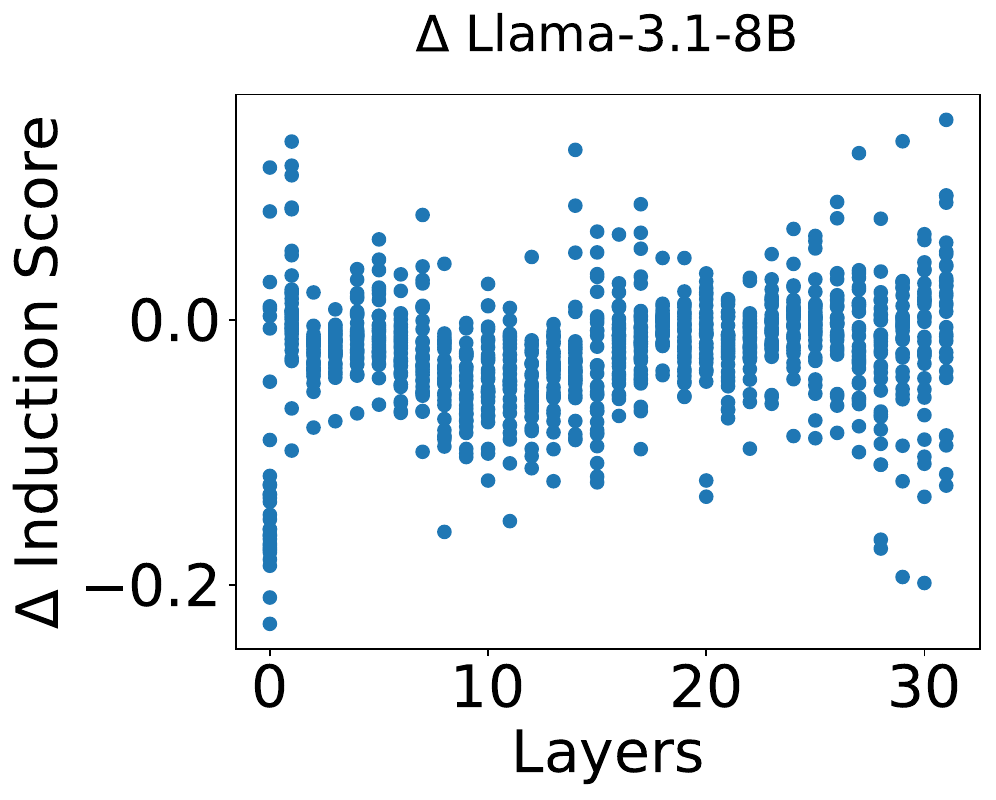}} &
        {\includegraphics[width=0.215\textwidth]{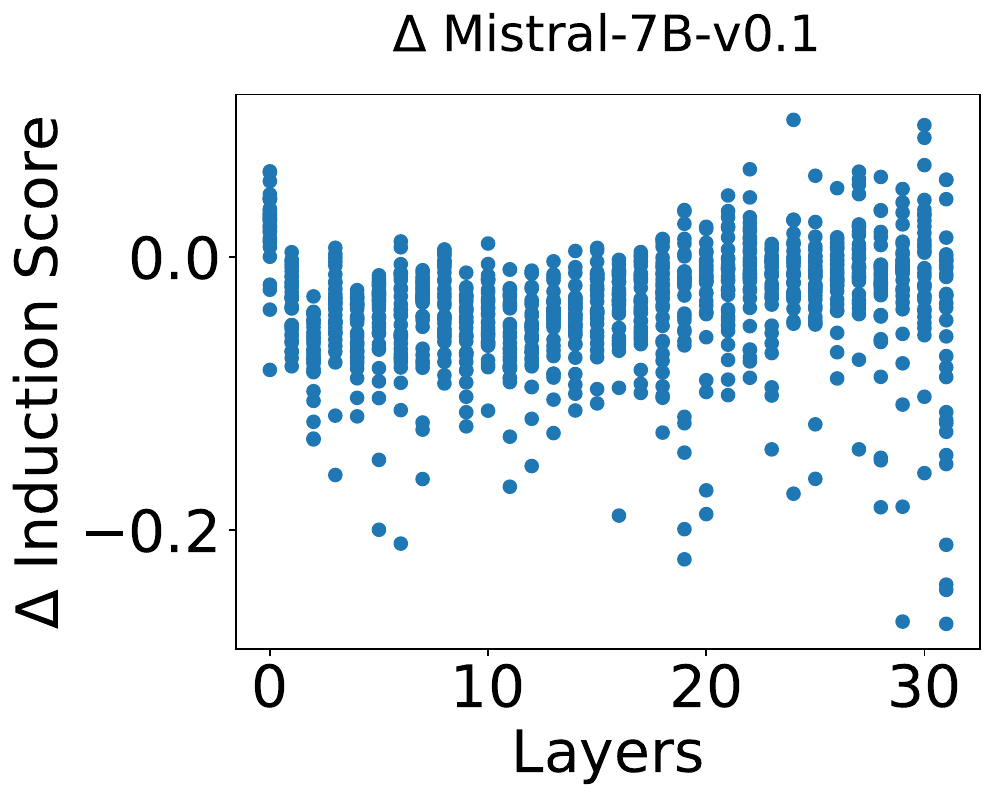}} &
        {\includegraphics[width=0.215\textwidth]{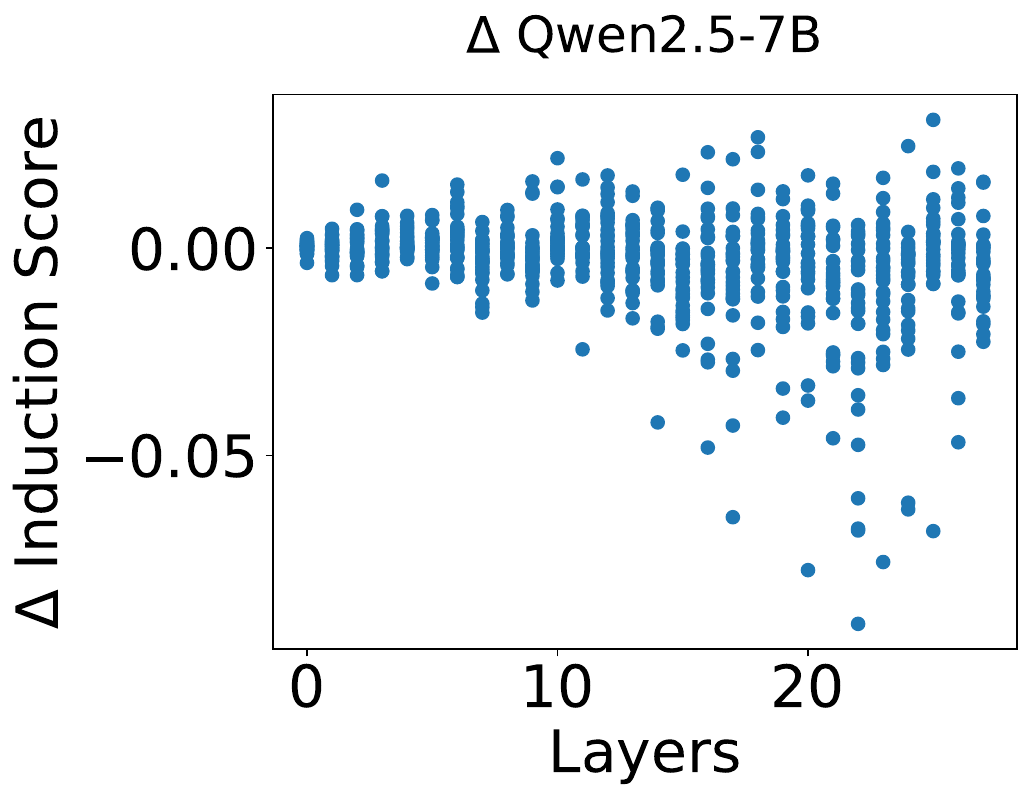}} &
        {\includegraphics[width=0.215\textwidth]{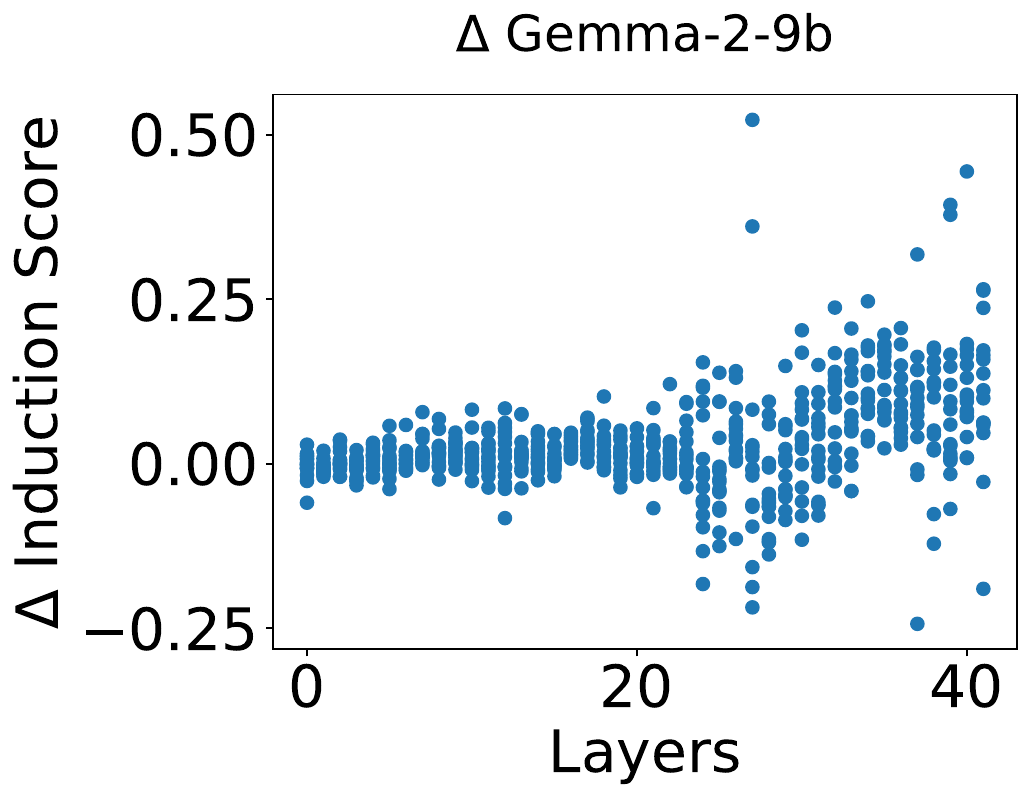}} \\
    \end{tabular}
    \caption{Induction scores across layers for the four  models. Each dot represents induction score at a particular attention head. Top row: Base models. Middle row: Instruction-tuned models. Bottom row: difference in induction scores for each layer and head between instruction-tuned and base models.
    \label{fig:ind_scatter}}
\end{figure}

 \begin{table*}[h]
    \centering
    \caption{Impact of \textit{induction} head ablation on the model output probability for the token at lag +1 for different models. \textit{The results are for base models.} These results are visualized in Fig.~\ref{fig:lag1_probability}. Corresponding results for ablation of random heads are shown in Table~\ref{tab:tab_randhead_base}.}
    \begin{tabular}{ccccc}
        \toprule
        \textbf{\# Induction Heads Ablated} & \textbf{Llama-3.1-8B} & \textbf{Mistral-7B-v0.1} & \textbf{Qwen2.5-7B} & \textbf{Gemma-2-9b} \\
        \midrule
        1   & 0.00102  & 0.00061  & 0.00121  & 0.06764 \\
        5   & 0.00092  & 0.00052  & 0.00078  & 0.05976 \\
        10  & 0.00093  & 0.00031  & 0.00039  & 0.05905 \\
        20  & 0.00088  & 0.00036  & 0.00029  & 0.05285 \\
        30  & 0.00086  & 0.00091  & 0.00031  & 0.00205 \\
        40  & 0.00091  & 0.00116  & 0.00033  & 0.00100 \\
        50  & 0.00090  & 0.00100  & 0.00009  & 0.00278 \\
        80  & 0.00086  & 0.00051  & 0.00030  & 0.00194 \\
        100 & 0.00092  & 0.00012  & 0.00034  & 0.00206 \\
        150 & 0.00084  & 0.00010  & 0.00034  & 0.00202 \\
        200 & 0.00085  & 0.00010  & 0.00029  & 0.00206 \\
        250 & 0.00084  & 0.00010  & 0.00034  & 0.00206 \\
        300 & 0.00085  & 0.00010  & 0.00029  & 0.00202 \\
        \bottomrule
    \end{tabular}
    \label{tab:tab_indhead_base}
\end{table*}

\begin{table*}[h]
    \centering
    \caption{Impact of \textit{random} head ablation on the model output probability for the token at lag +1 for different models. \textit{The results are for base models.} These results are visualized in Fig.~\ref{fig:lag1_probability}.}
    \begin{tabular}{ccccc}
        \toprule
        \textbf{\# Random Heads Ablated} & \textbf{Llama-3.1-8B} & \textbf{Mistral-7B-v0.1} & \textbf{Qwen2.5-7B} & \textbf{Gemma-2-9b} \\
        \midrule
        0   & 0.00103  & 0.00060  & 0.00169  & 0.07125 \\
        5   & 0.00113  & 0.00052  & 0.00155  & 0.08422 \\
        10  & 0.00117  & 0.00056  & 0.00162  & 0.09609 \\
        20  & 0.00541  & 0.00086  & 0.00130  & 0.04800 \\
        30  & 0.00542  & 0.00088  & 0.00304  & 0.02157 \\
        40  & 0.00365  & 0.00073  & 0.00235  & 0.02732 \\
        50  & 0.00375  & 0.00083  & 0.00408  & 0.03684 \\
        80  & 0.00856  & 0.00067  & 0.00282  & 0.00766 \\
        100 & 0.00934  & 0.00085  & 0.00274  & 0.00478 \\
        150 & 0.00933  & 0.00090  & 0.00221  & 0.00456 \\
        200 & 0.00932  & 0.00098  & 0.00233  & 0.00456 \\
        250 & 0.00933  & 0.00090  & 0.00221  & 0.00478 \\
        300 & 0.00932  & 0.00098  & 0.00233  & 0.00456 \\
        \bottomrule
    \end{tabular}
    \label{tab:tab_randhead_base}
\end{table*}

 \begin{table*}[h]
    \centering
    \caption{Impact of \textit{random} head ablation on the model output probability for the token at lag +1 for different models. \textit{The results are for instruction-tuned models.} These results are visualized in Fig.~\ref{fig:lag1_probability}.}
    {\setlength{\tabcolsep}{4pt}
    \resizebox{\textwidth}{!}{%
    \begin{tabular}{ccccc}
        \toprule
        \textbf{\# Random Heads Ablated} & \textbf{Llama-3.1-8B-Instruct} & \textbf{Mistral-7B-Instruct-v0.1} & \textbf{Qwen2.5-7B-Instruct} & \textbf{Gemma-2-9b-it} \\
        \midrule
        1   & 0.00066  & 0.04366  & 0.00537  & 0.00496 \\
        5   & 0.00068  & 0.03826  & 0.00477  & 0.00576 \\
        10  & 0.00092  & 0.03831  & 0.00385  & 0.00239 \\
        20  & 0.00968  & 0.03897  & 0.00452  & 0.01165 \\
        30  & 0.00881  & 0.03681  & 0.00437  & 0.01748 \\
        40  & 0.00713  & 0.03598  & 0.00324  & 0.01005 \\
        50  & 0.00333  & 0.04644  & 0.00421  & 0.02278 \\
        80  & 0.00456  & 0.06900  & 0.01041  & 0.02279 \\
        100 & 0.00167  & 0.06869  & 0.00589  & 0.01436 \\
        150 & 0.00148  & 0.06365  & 0.00585  & 0.01408 \\
        200 & 0.00169  & 0.06428  & 0.00552  & 0.01436 \\
        250 & 0.00160  & 0.06661  & 0.00529  & 0.01436 \\
        300 & 0.00173  & 0.06934  & 0.00578  & 0.01408 \\
        \bottomrule
    \end{tabular}}}
    \label{tab:tab_randhead_inst}
\end{table*}


\newpage

\begin{figure*}[h!]
    \centering
    \footnotesize

    \setlength{\tabcolsep}{1.5pt}

    \begin{tabular}{c c c c c c c}
    & \parbox{\plotfigurewidth}{\centering\scriptsize\textbf{No Ablation}}
    & \parbox{\plotfigurewidth}{\centering\scriptsize\textbf{1 Head Ablated}}
    & \parbox{\plotfigurewidth}{\centering\scriptsize\textbf{10 Heads Ablated}}
    & \parbox{\plotfigurewidth}{\centering\scriptsize\textbf{25 Heads Ablated}}
    & \parbox{\plotfigurewidth}{\centering\scriptsize\textbf{50 Heads Ablated}}
    & \parbox{\plotfigurewidth}{\centering\scriptsize\textbf{100 Heads Ablated}} \\

    \rotatebox{90}{\parbox{2.2cm}{\centering\scriptsize\textbf{Llama-8B}\\Ind. Abl.}} &
    \includegraphics[width=\plotfigurewidth]{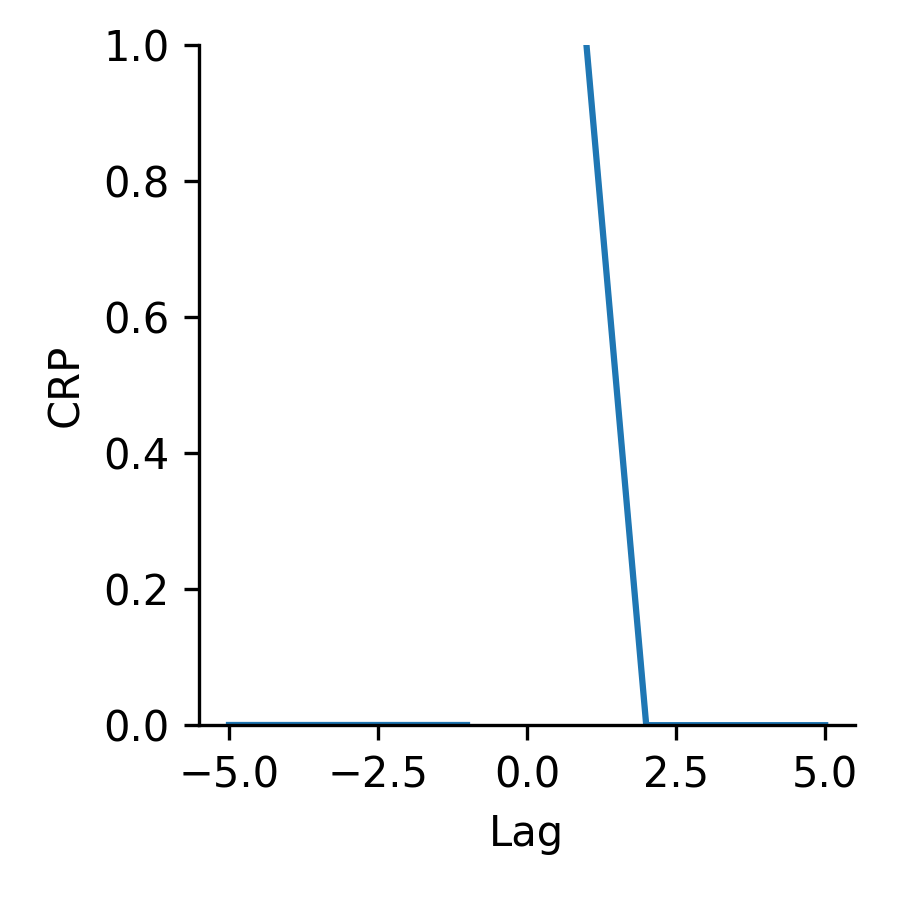} &
    \includegraphics[width=\plotfigurewidth]{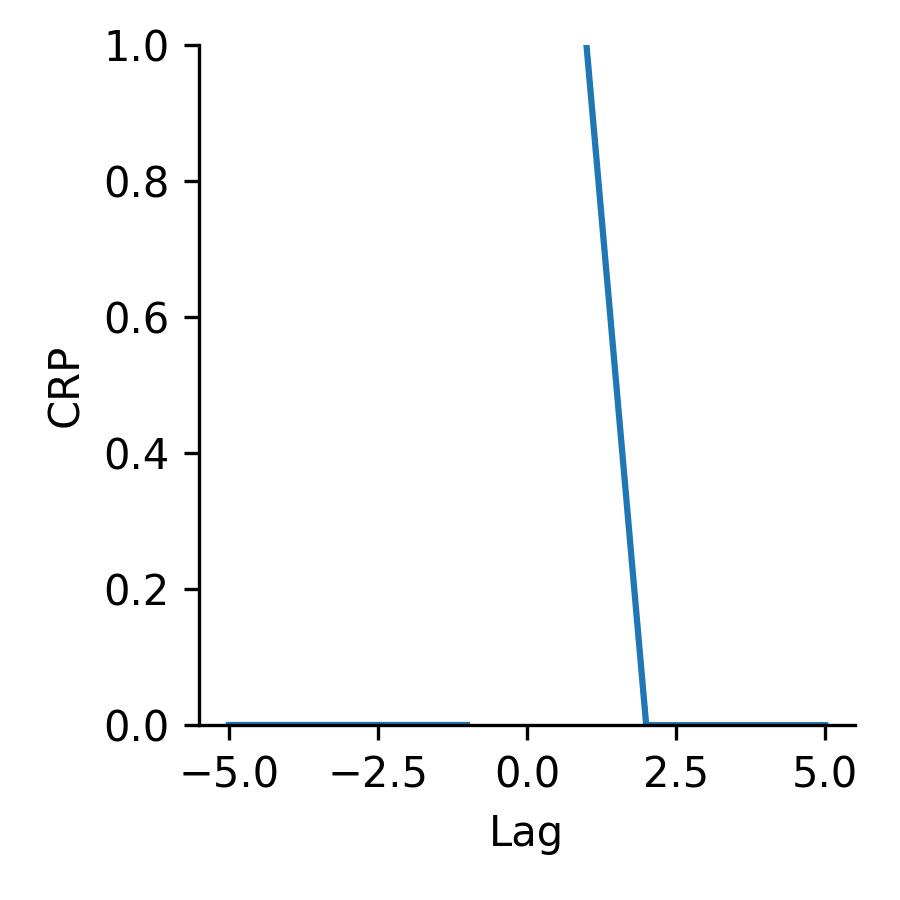} &
    \includegraphics[width=\plotfigurewidth]{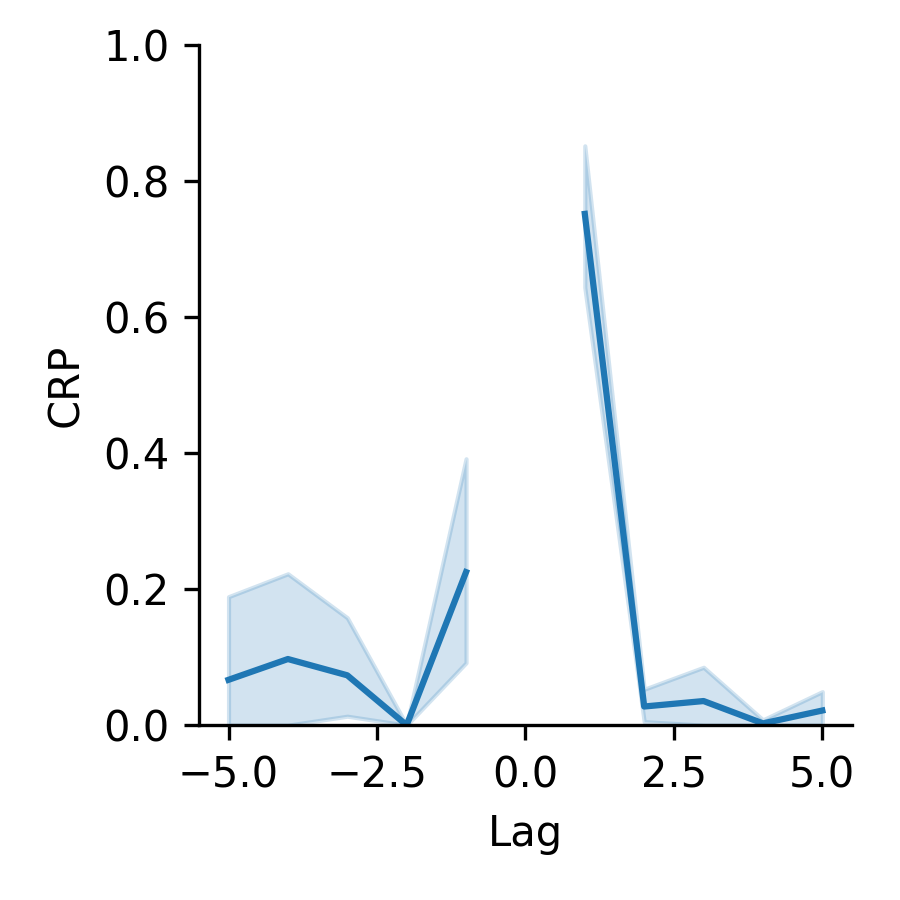} &
    \includegraphics[width=\plotfigurewidth]{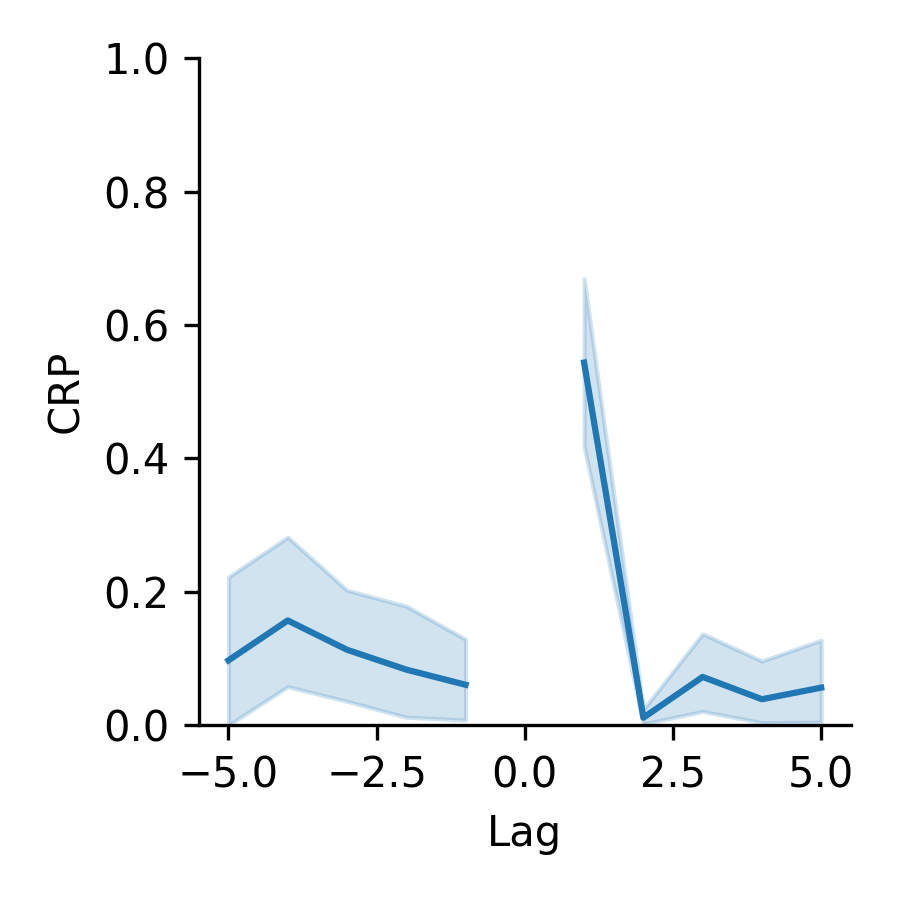} &
    \includegraphics[width=\plotfigurewidth]{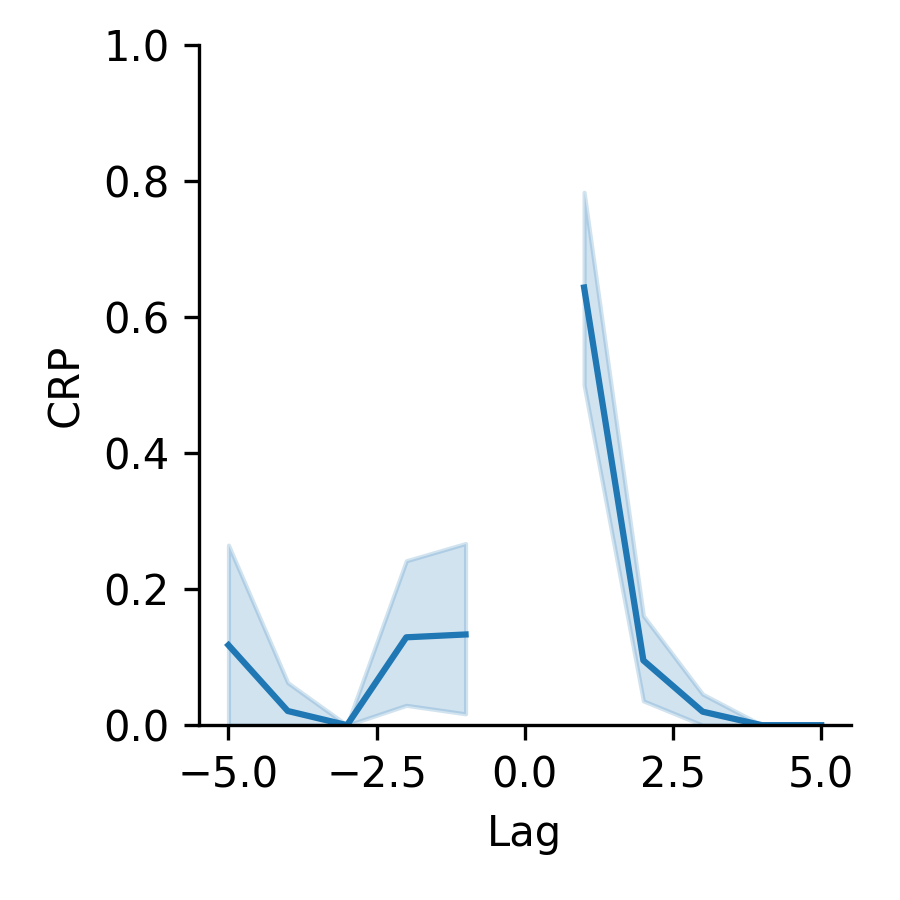} &
    \includegraphics[width=\plotfigurewidth]{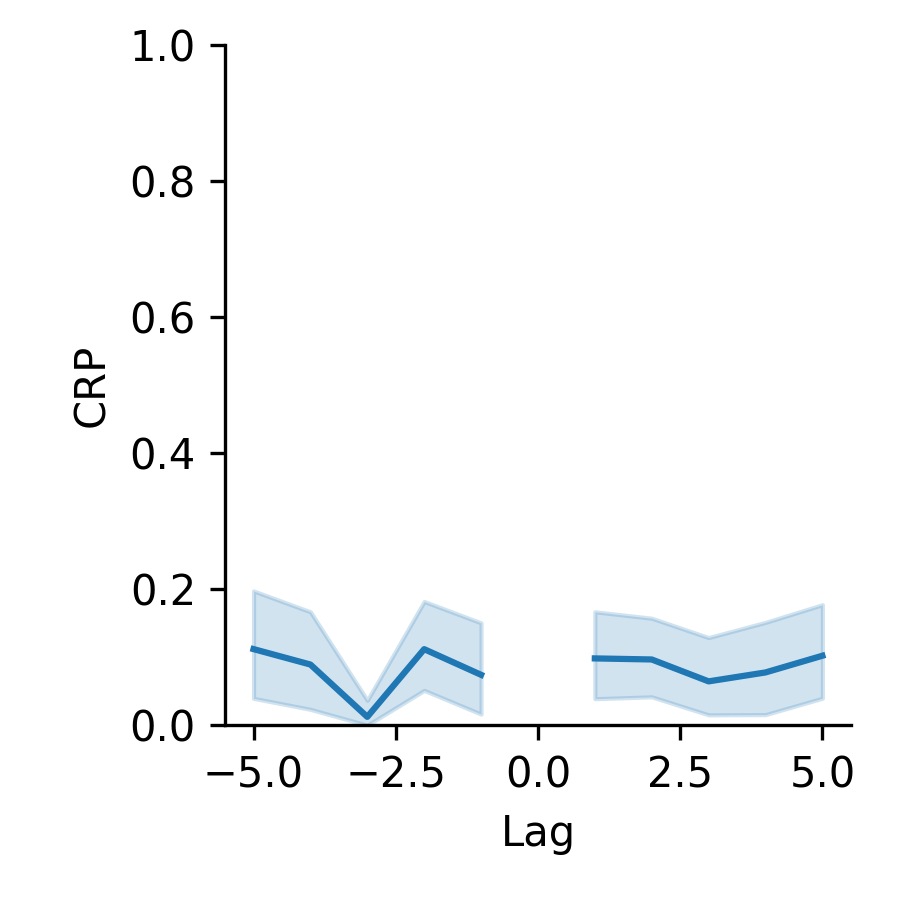} \\

    \rotatebox{90}{\parbox{2.2cm}{\centering\scriptsize\textbf{Llama-8B}\\Rand. Abl.}} &
    \includegraphics[width=\plotfigurewidth]{Figures/CRP/Llama-3.1-8B_few_10_shot_no_ablation_14_50_crp.png} &
    \includegraphics[width=\plotfigurewidth]{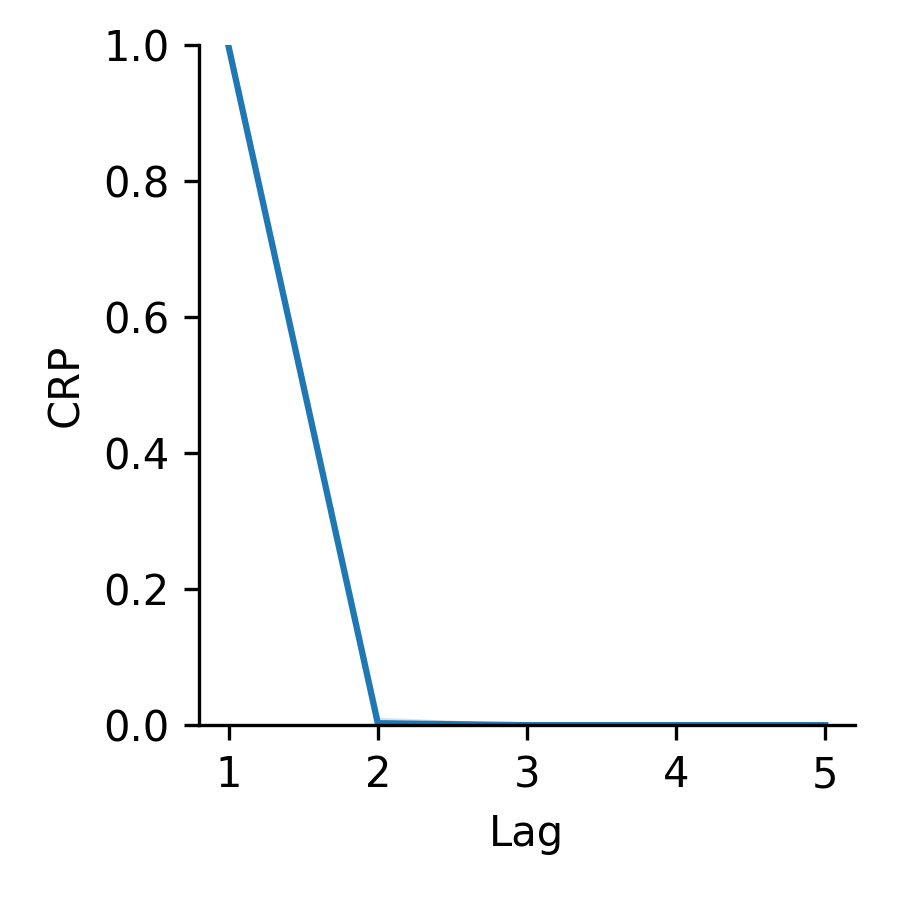} &
    \includegraphics[width=\plotfigurewidth]{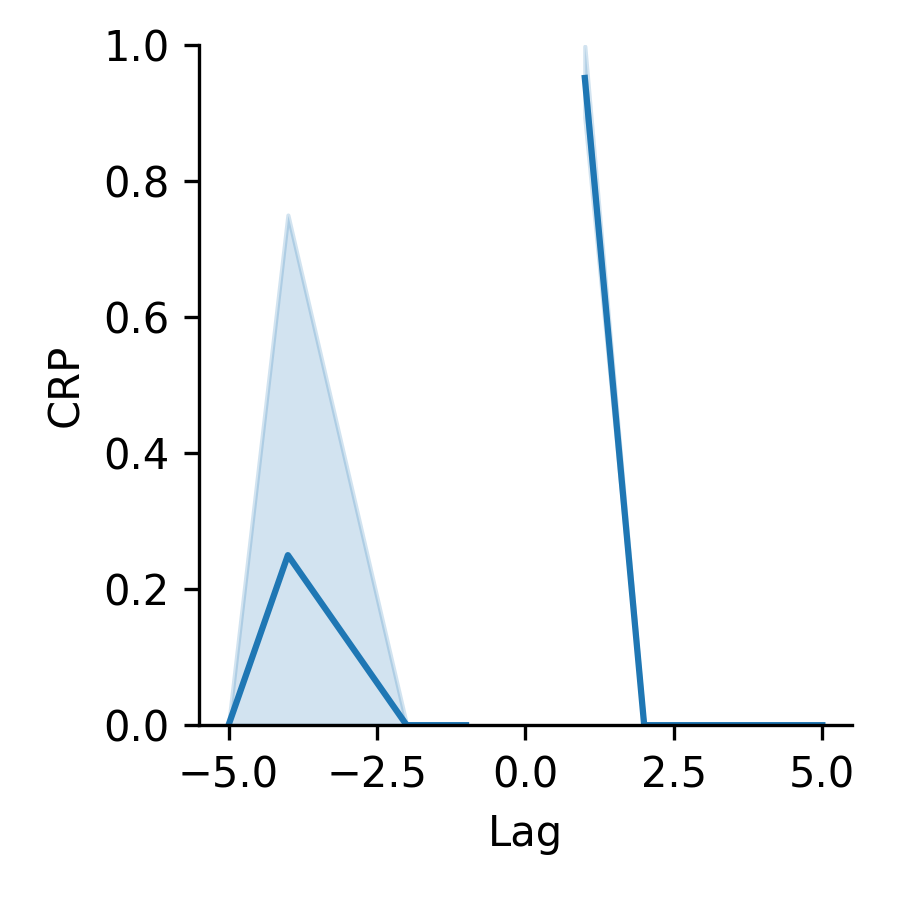} &
    \includegraphics[width=\plotfigurewidth]{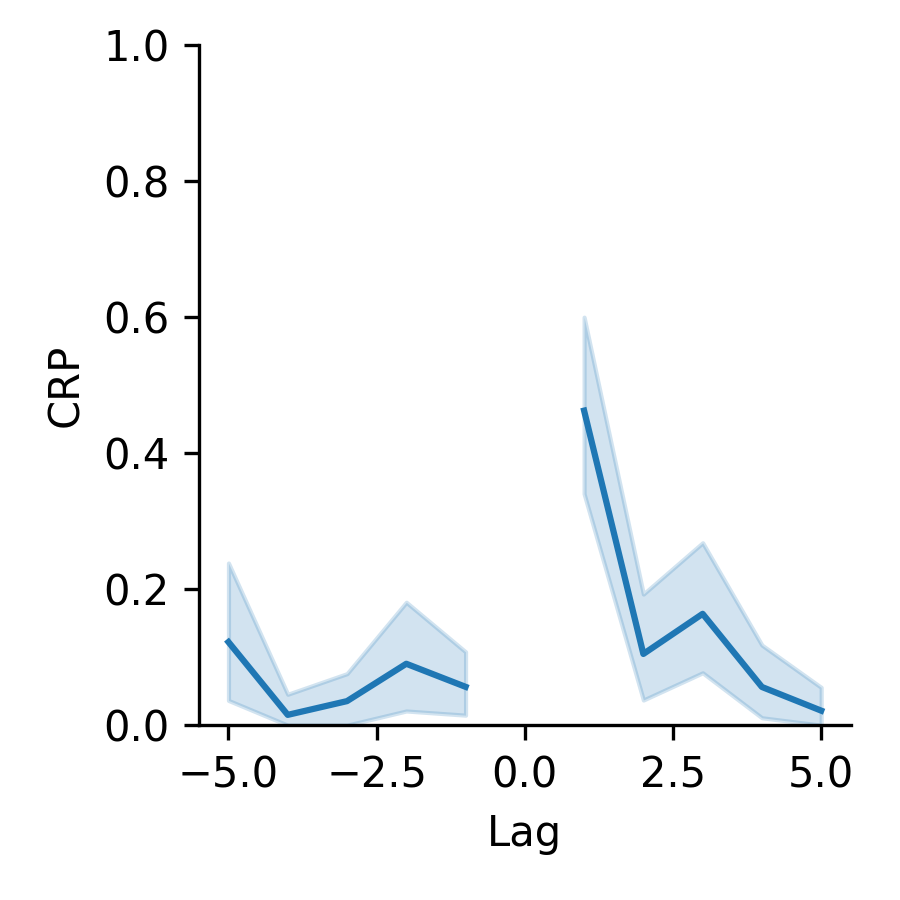} &
    \includegraphics[width=\plotfigurewidth]{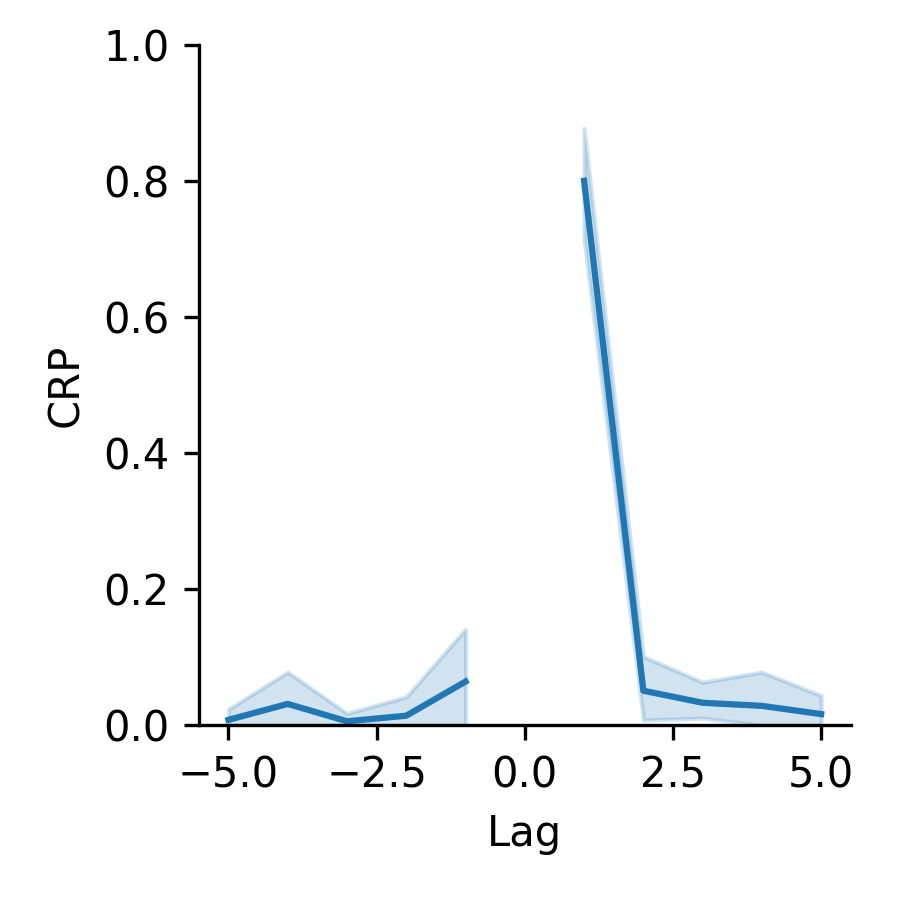} &
    \includegraphics[width=\plotfigurewidth]{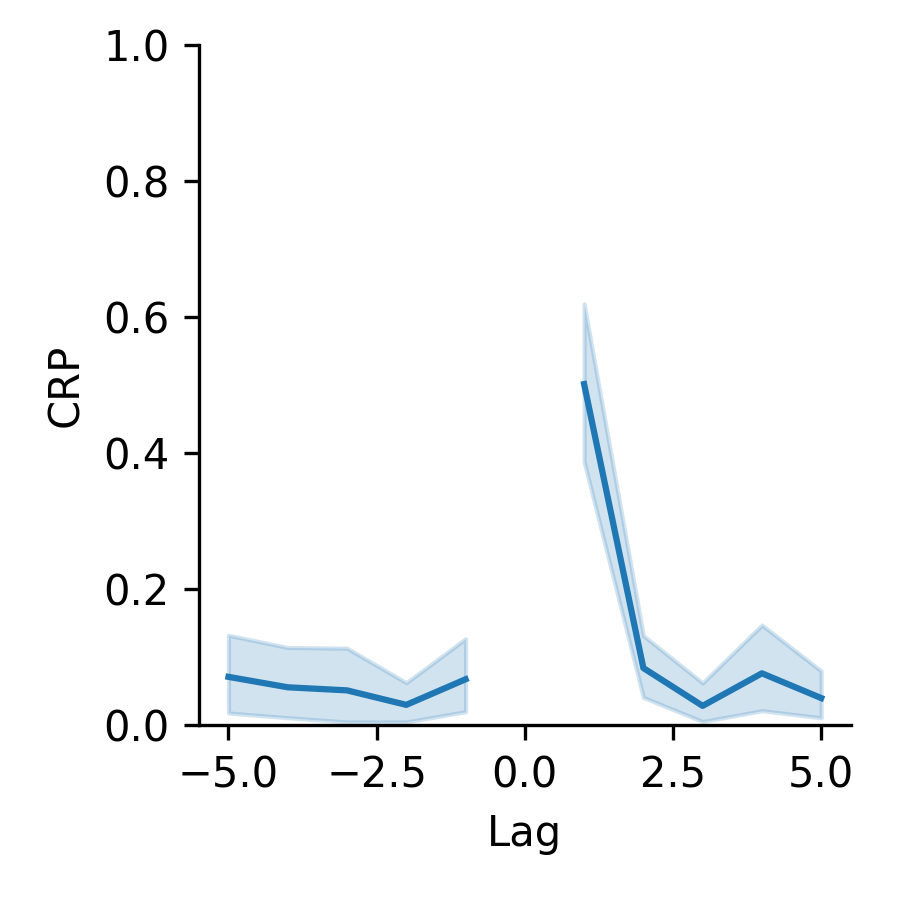} \\

    \rotatebox{90}{\parbox{2.2cm}{\centering\scriptsize\textbf{Llama-8B-I}\\Ind. Abl.}} &
    \includegraphics[width=\plotfigurewidth]{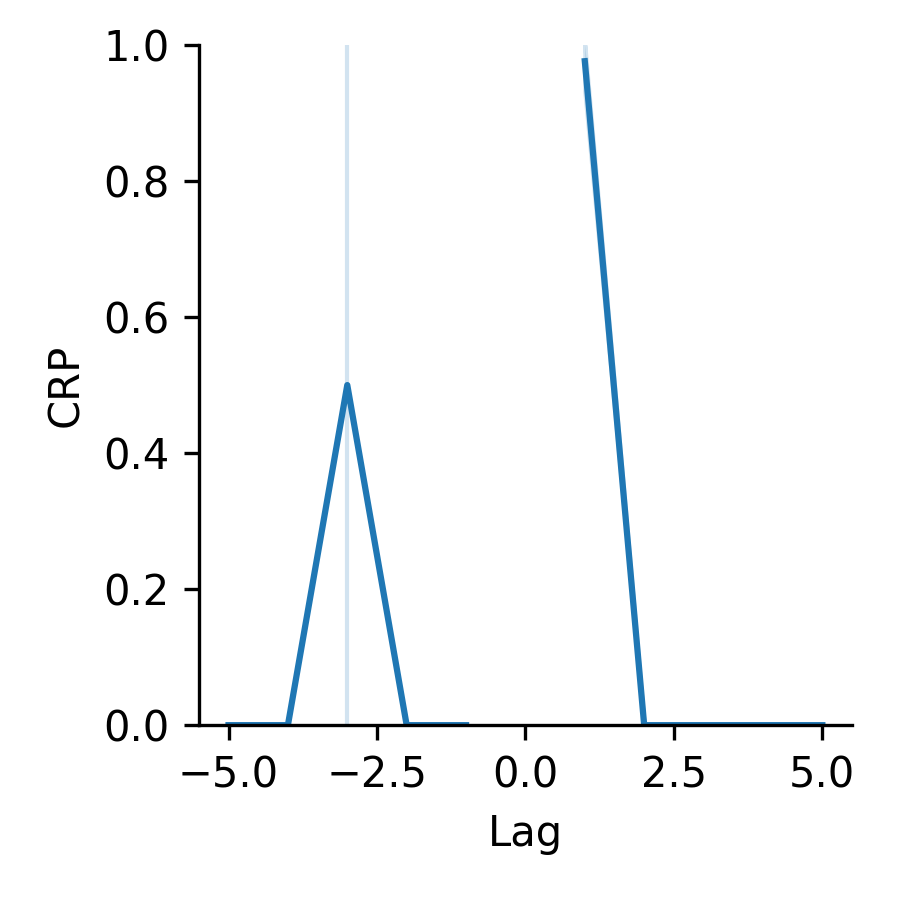} &
    \includegraphics[width=\plotfigurewidth]{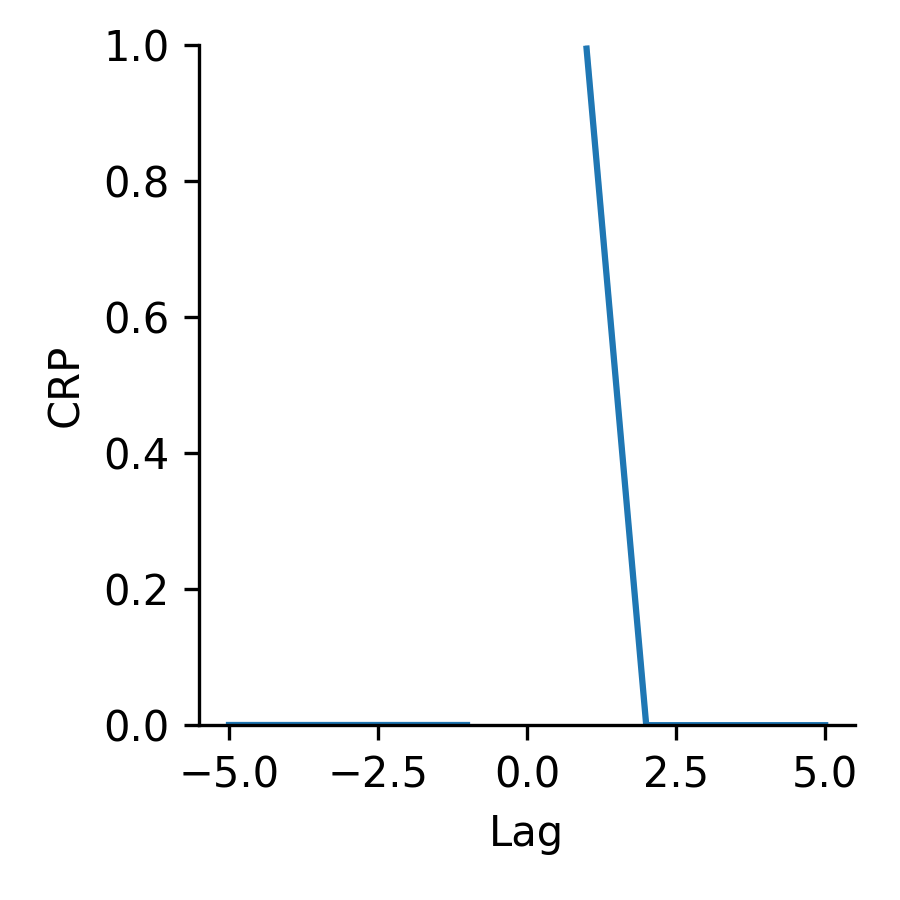} &
    \includegraphics[width=\plotfigurewidth]{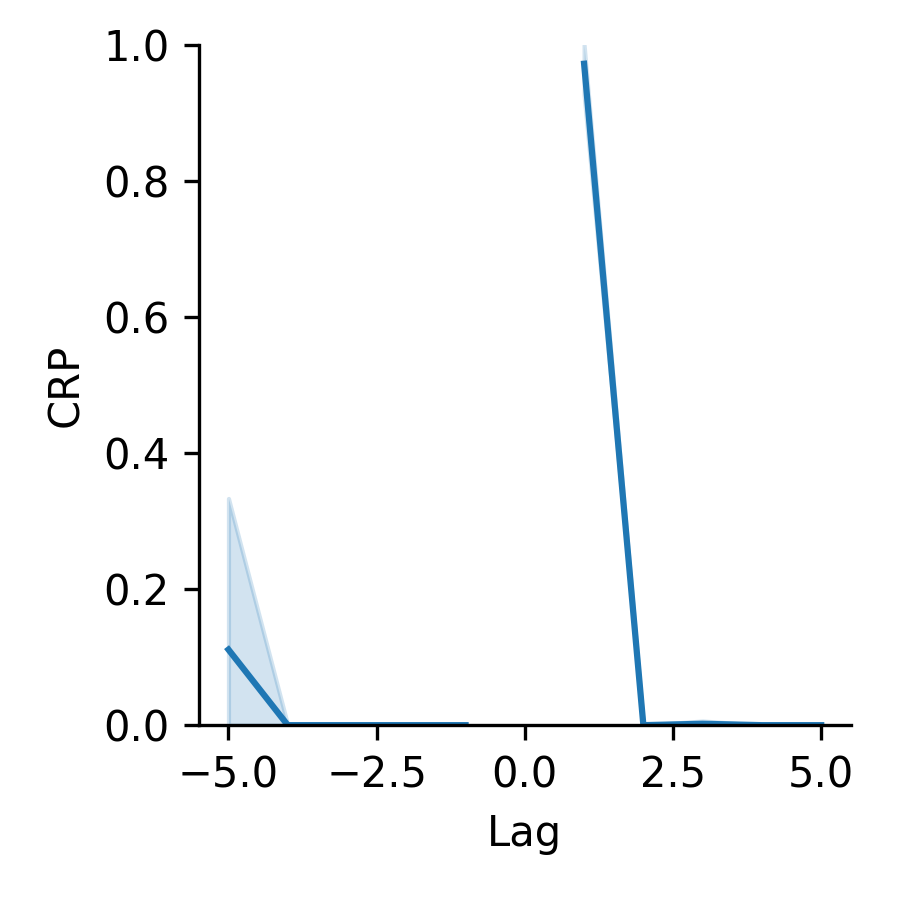} &
    \includegraphics[width=\plotfigurewidth]{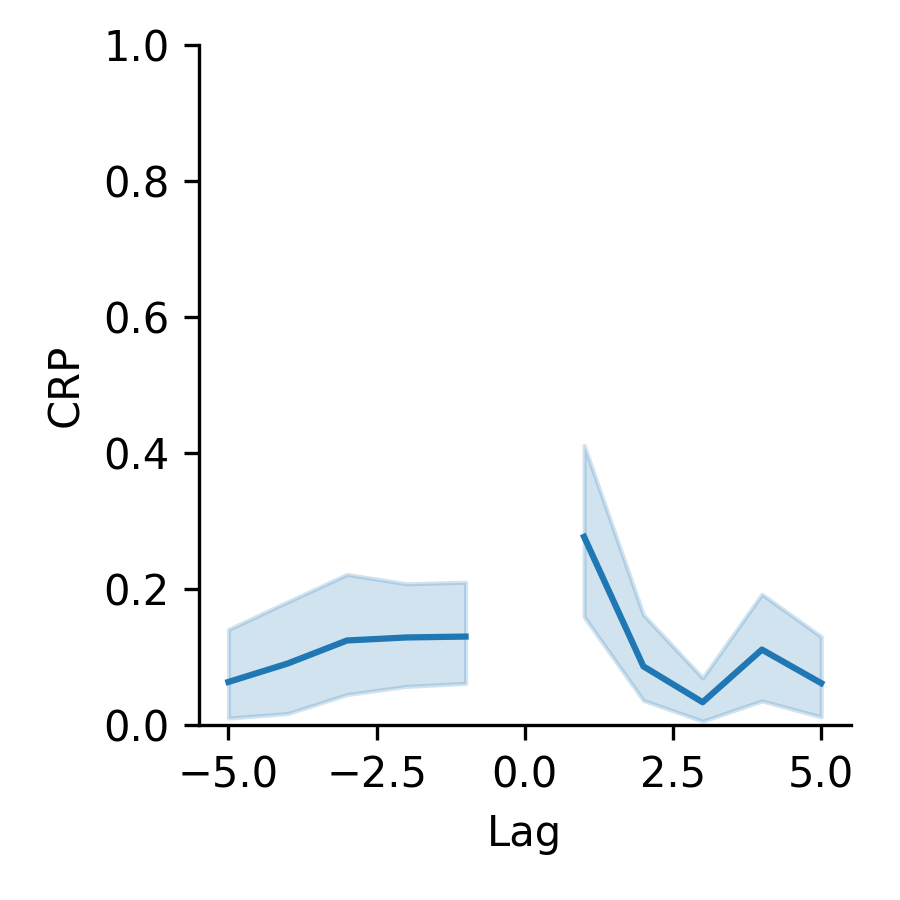} &
    \includegraphics[width=\plotfigurewidth]{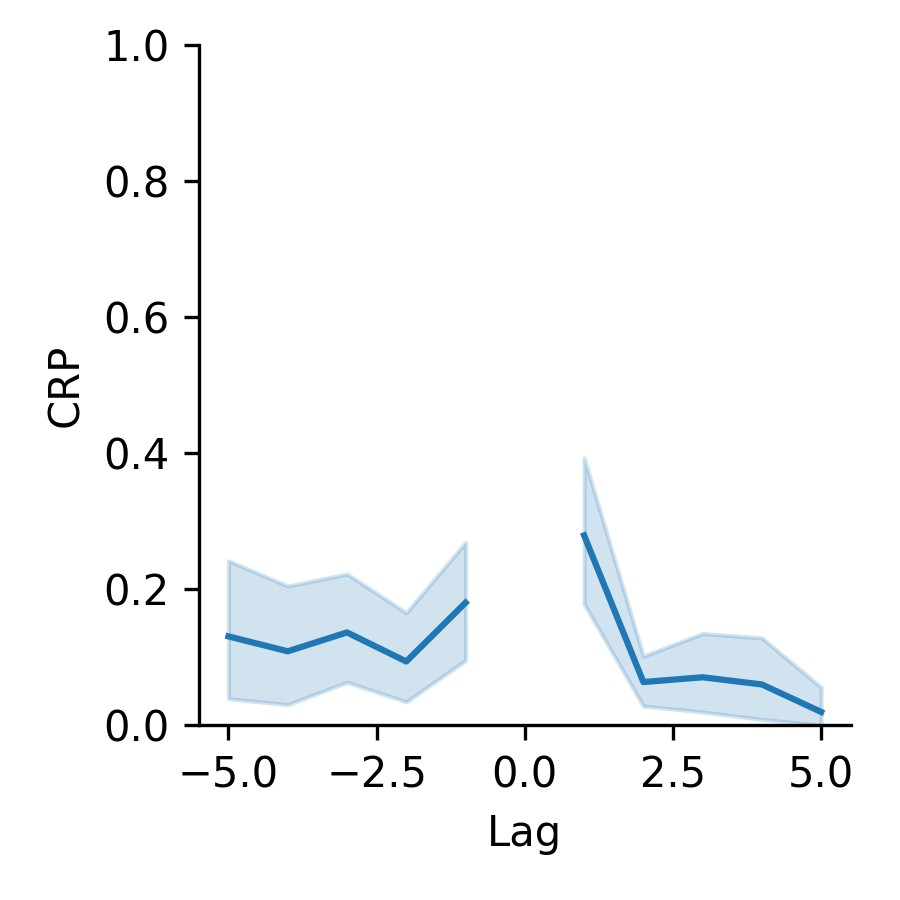} &
    \includegraphics[width=\plotfigurewidth]{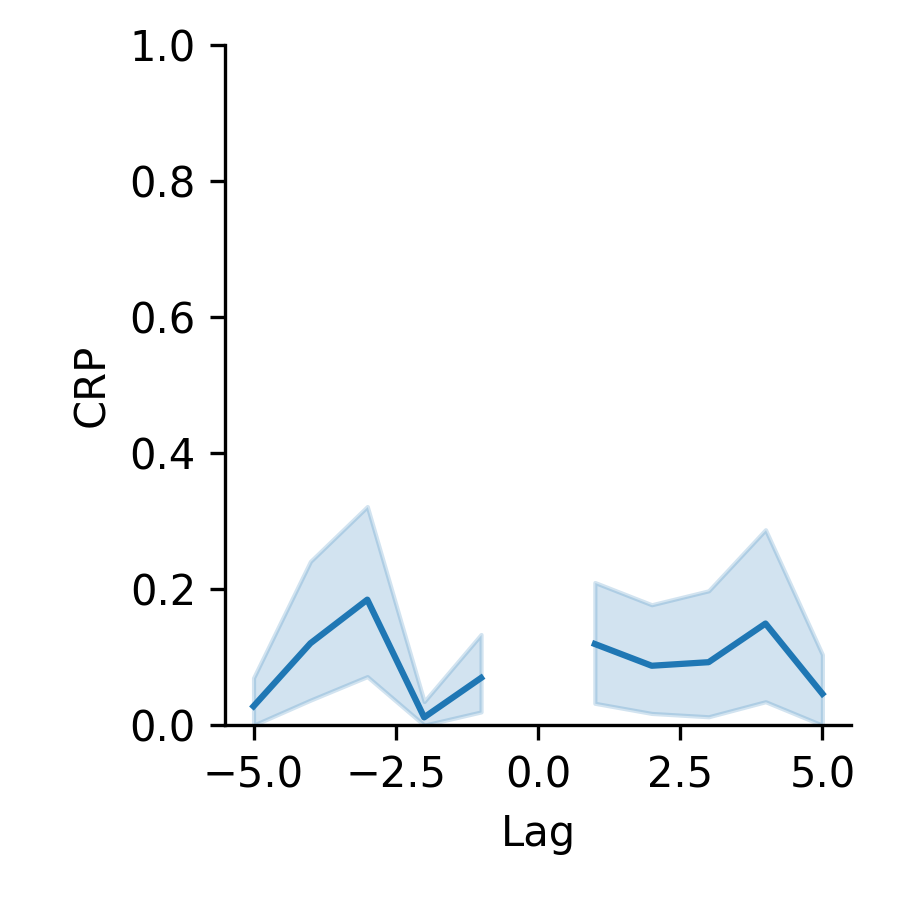} \\

    \rotatebox{90}{\parbox{2.2cm}{\centering\scriptsize\textbf{Llama-8B-I}\\Rand. Abl.}} &
    \includegraphics[width=\plotfigurewidth]{Figures/CRP/Llama-3.1-8B-Instruct_few_10_shot_no_ablation_14_50_crp.png} &
    \includegraphics[width=\plotfigurewidth]{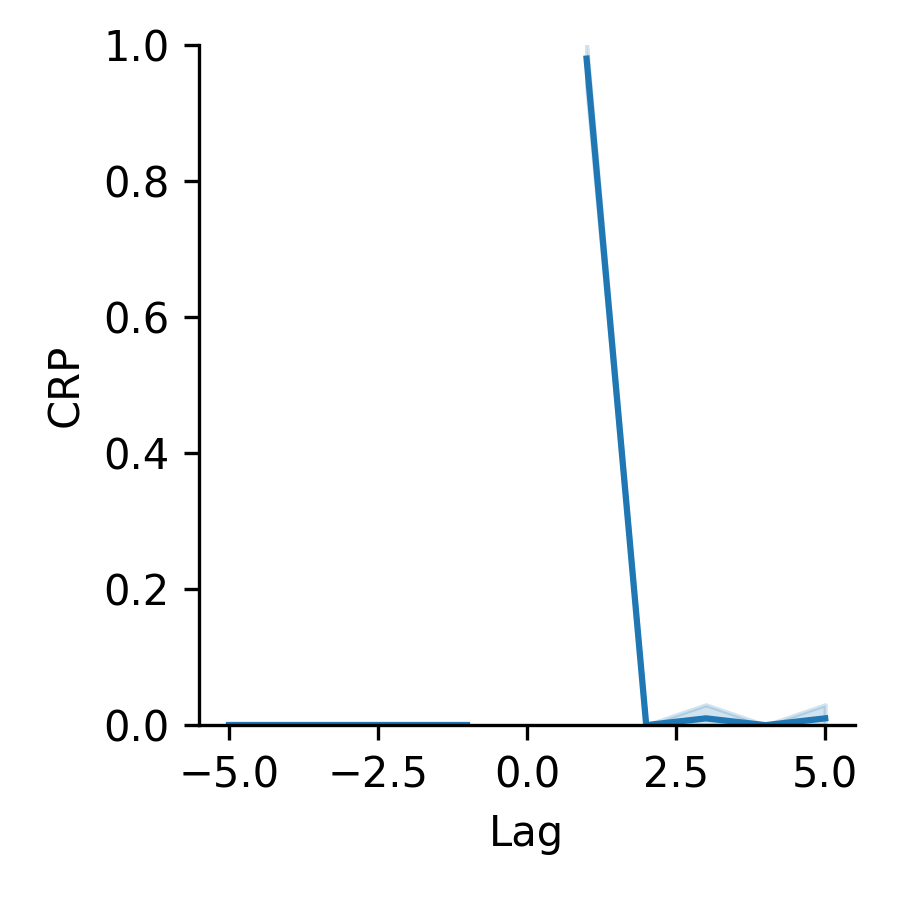} &
    \includegraphics[width=\plotfigurewidth]{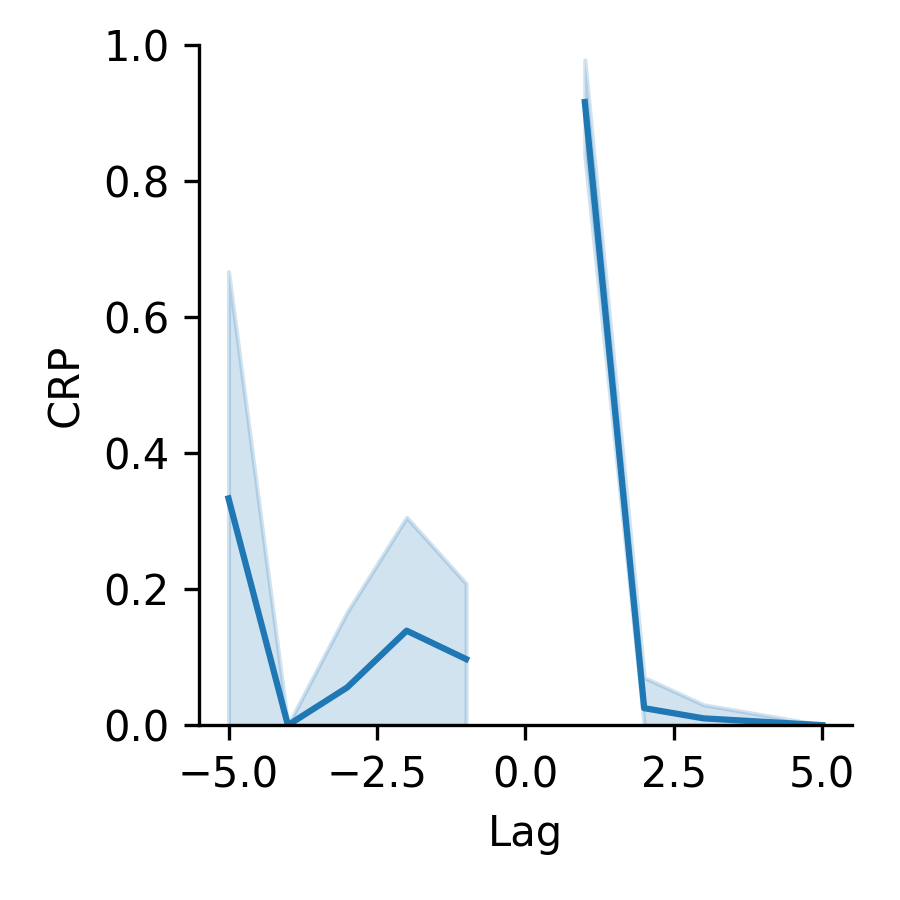} &
    \includegraphics[width=\plotfigurewidth]{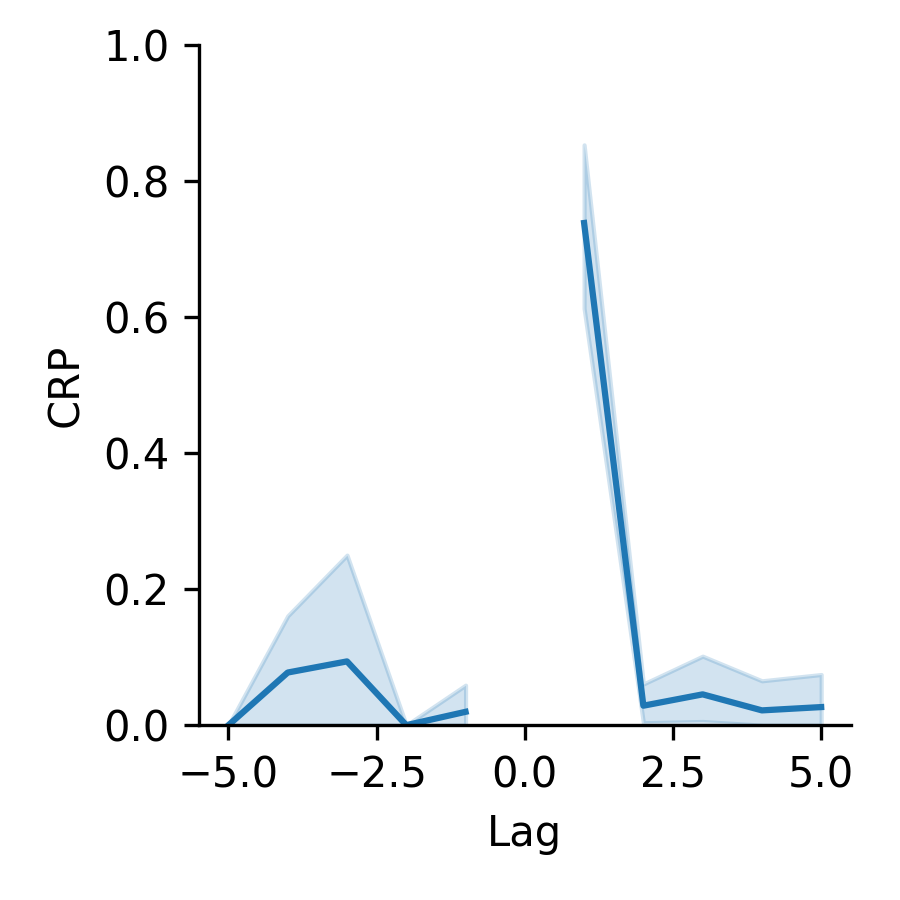} &
    \includegraphics[width=\plotfigurewidth]{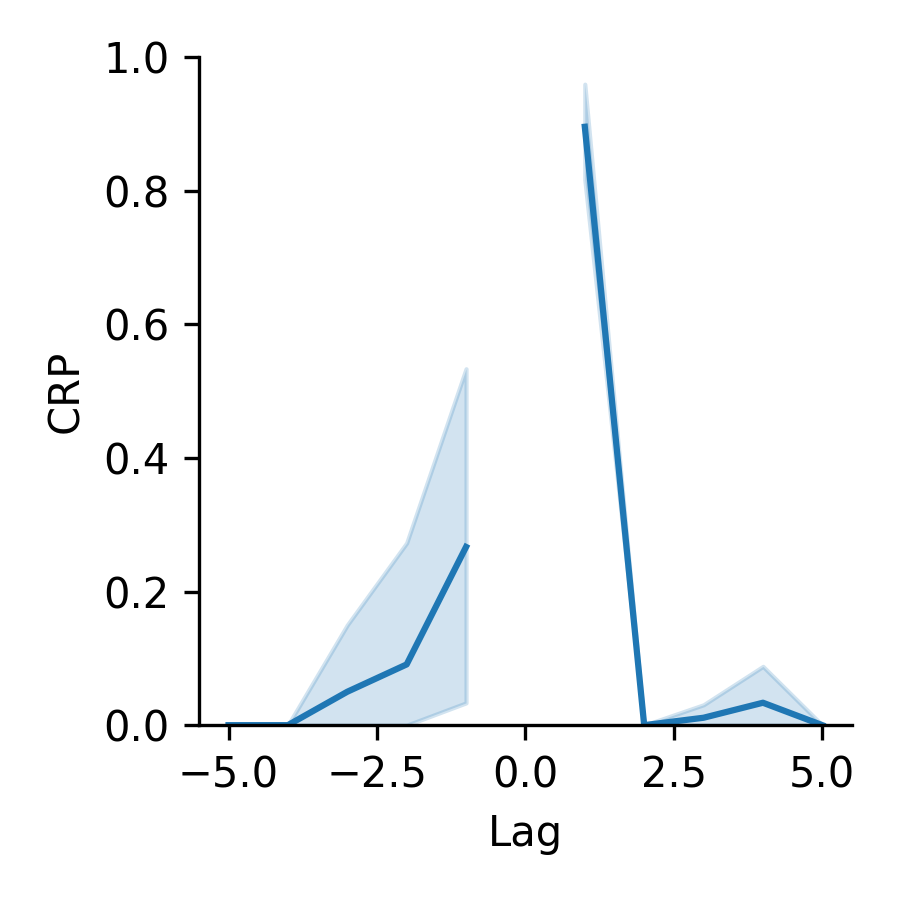} &
    \includegraphics[width=\plotfigurewidth]{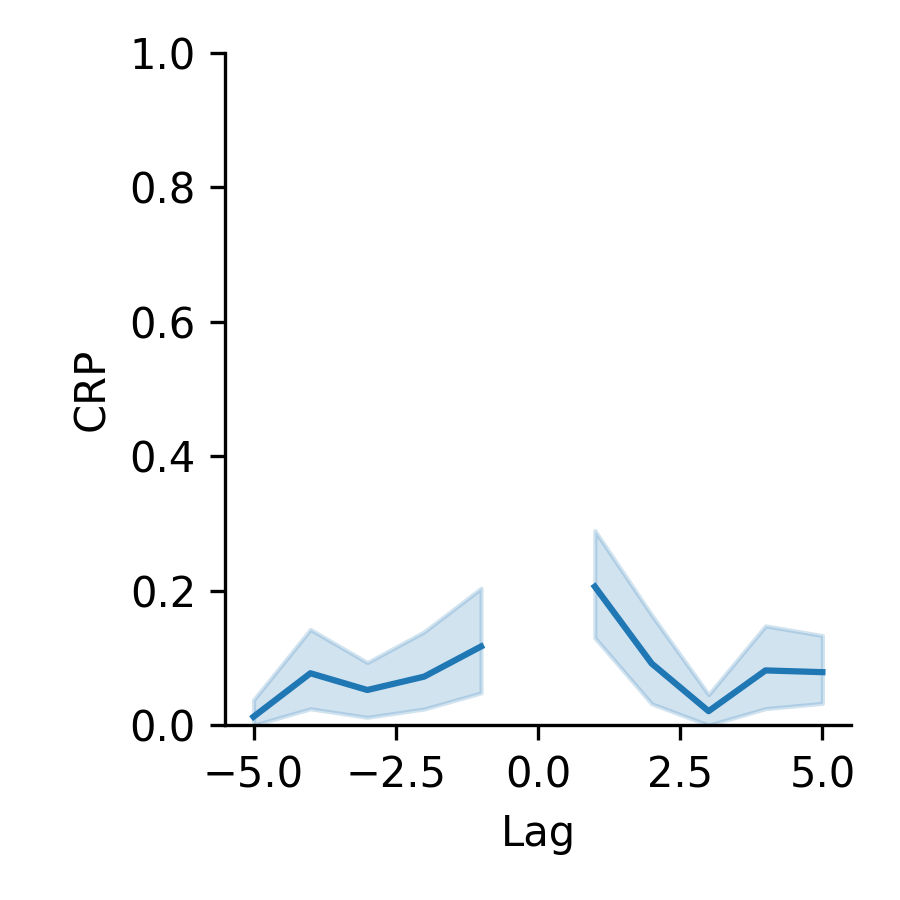} \\

    \rotatebox{90}{\parbox{2.2cm}{\centering\scriptsize\textbf{Qwen-7B}\\Ind. Abl.}} &
    \includegraphics[width=\plotfigurewidth]{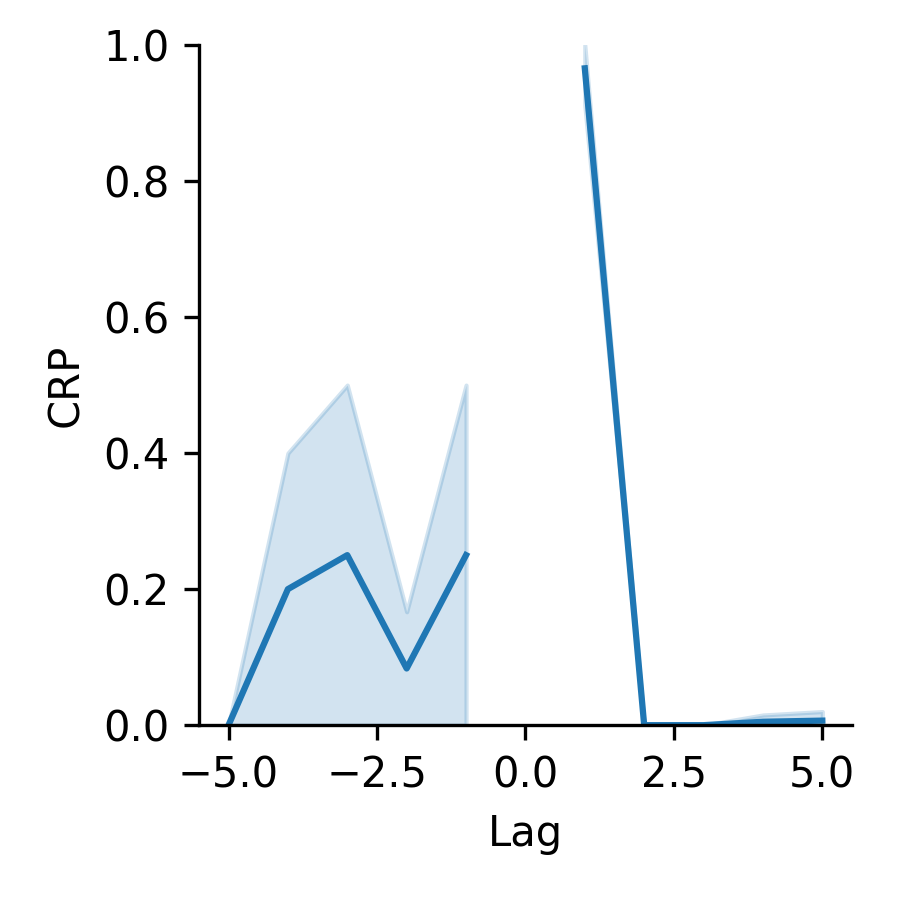} &
    \includegraphics[width=\plotfigurewidth]{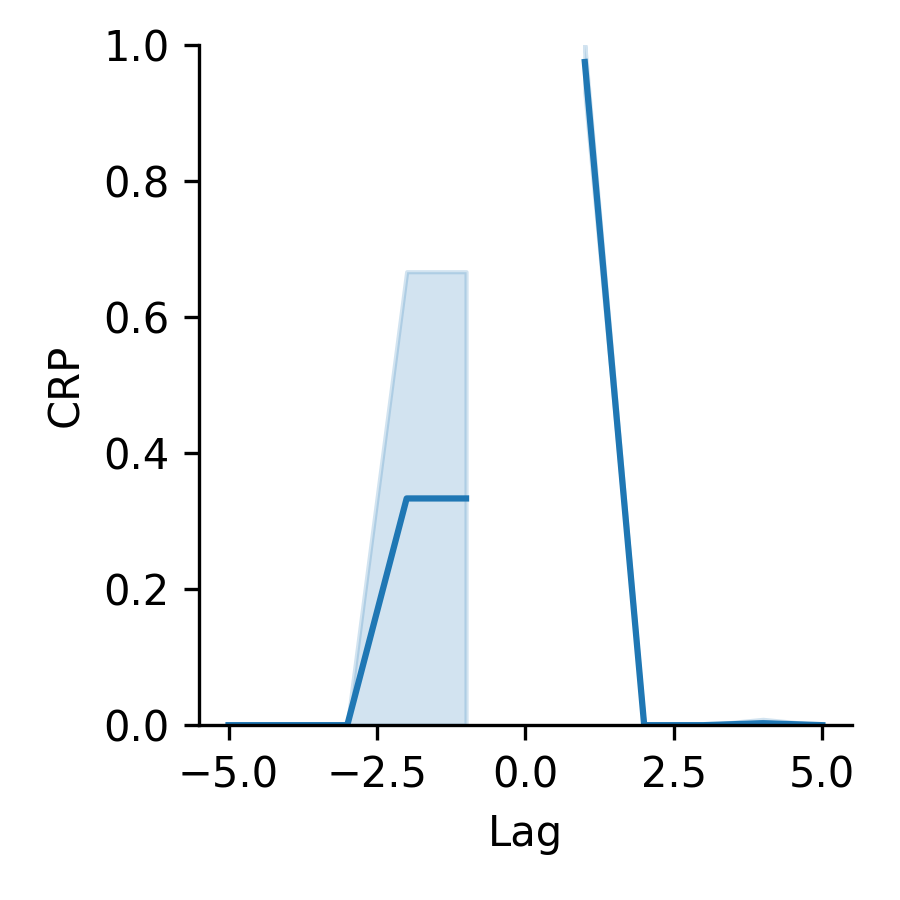} &
    \includegraphics[width=\plotfigurewidth]{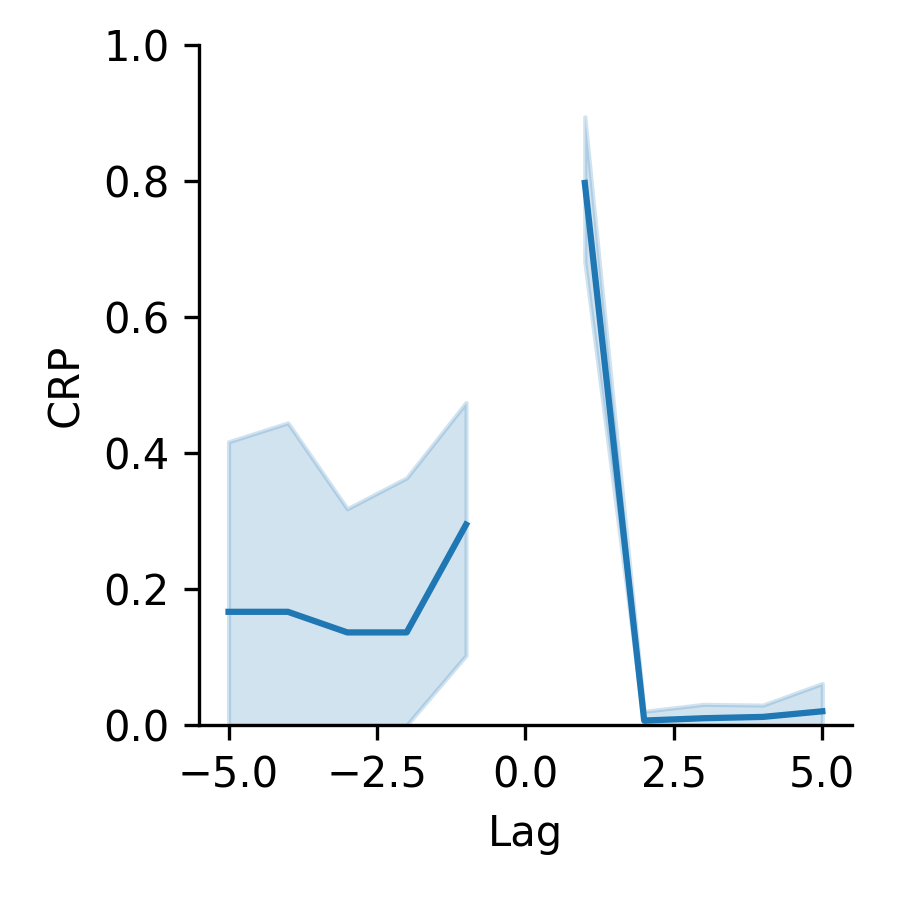} &
    \includegraphics[width=\plotfigurewidth]{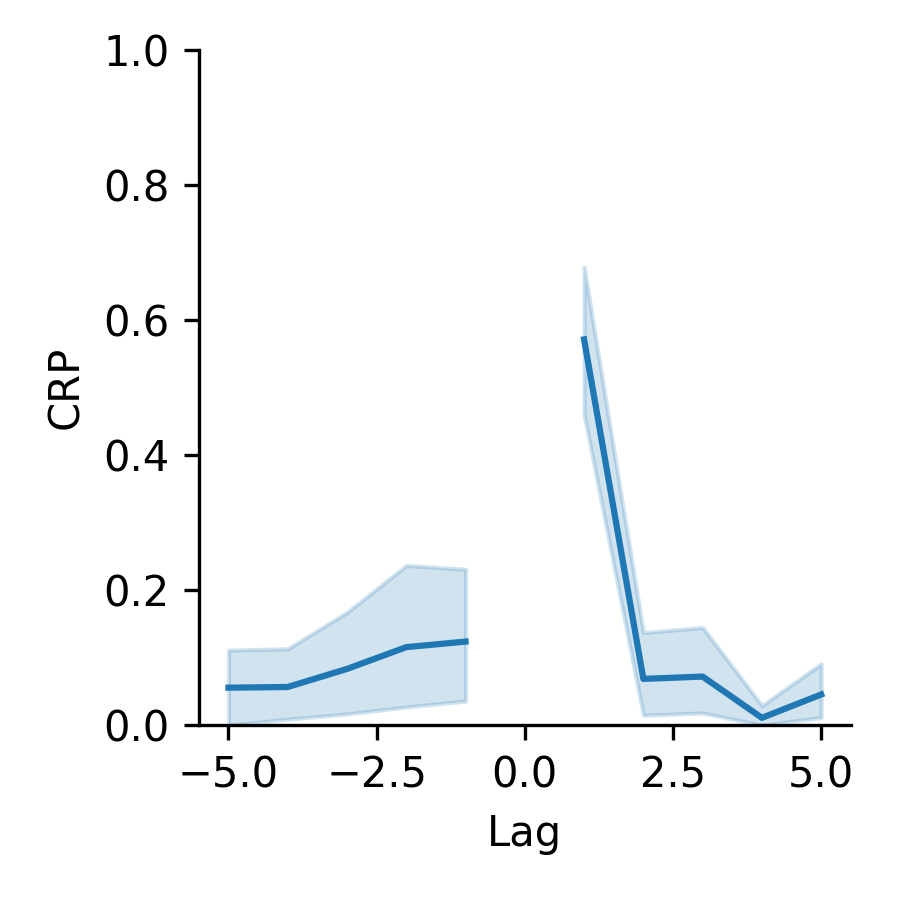} &
    \includegraphics[width=\plotfigurewidth]{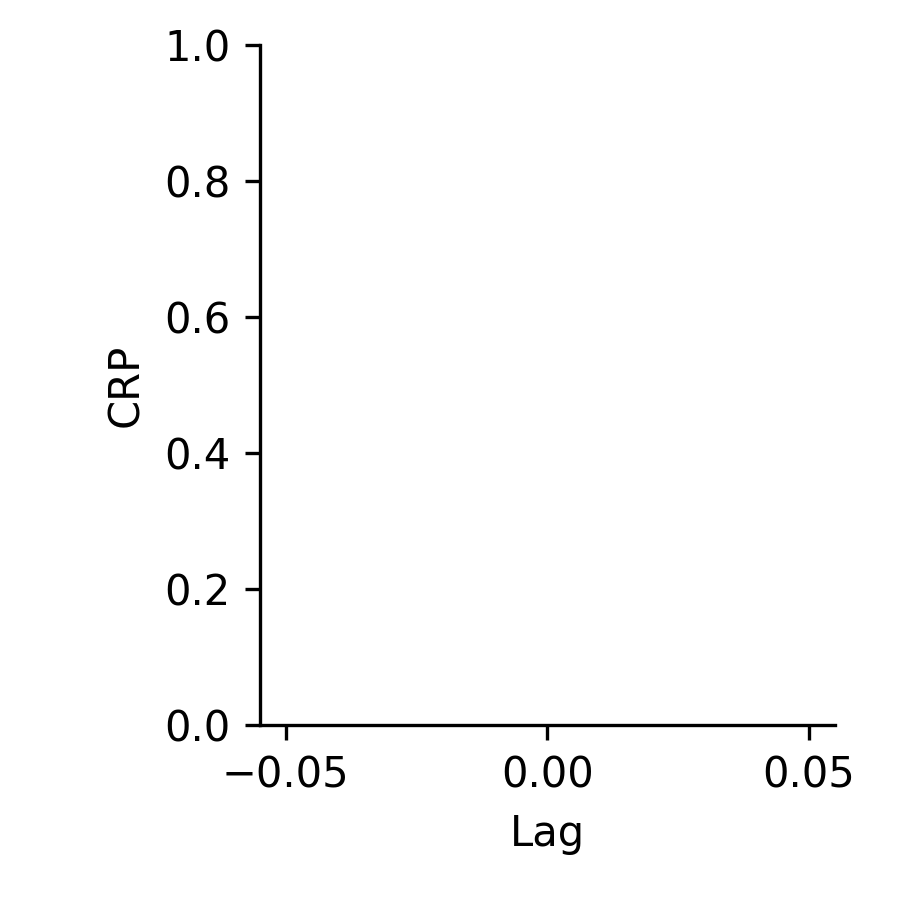} &
    \includegraphics[width=\plotfigurewidth]{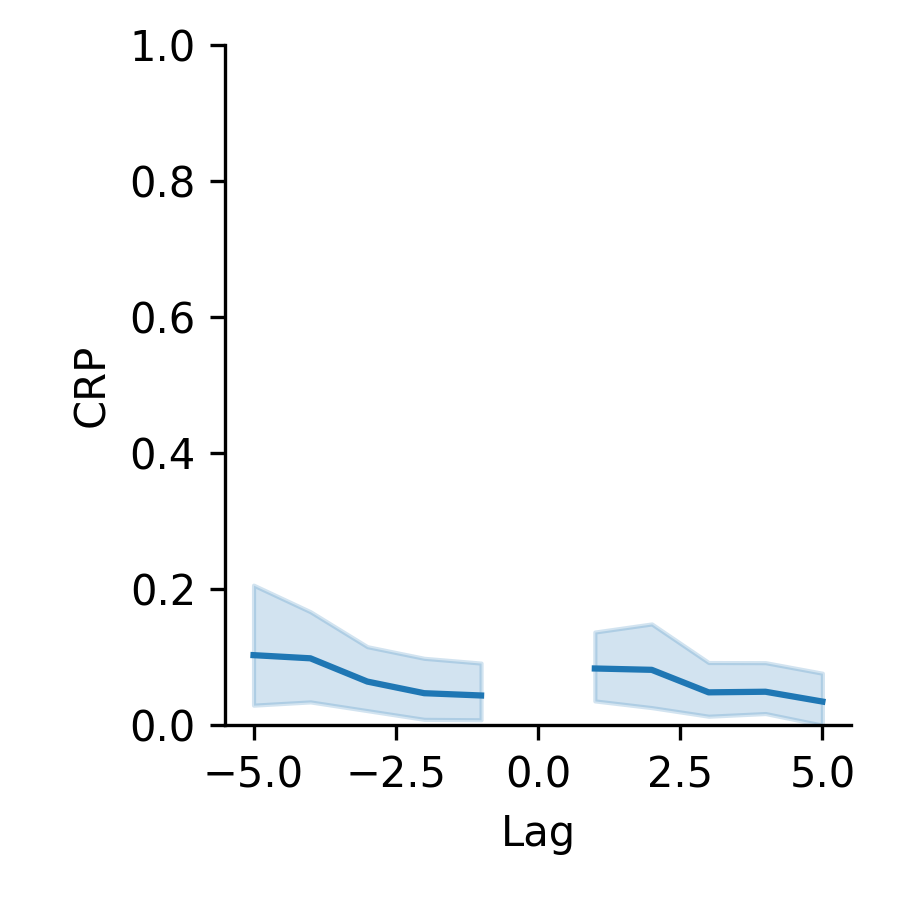} \\

    \rotatebox{90}{\parbox{2.2cm}{\centering\scriptsize\textbf{Qwen-7B}\\Rand. Abl.}} &
    \includegraphics[width=\plotfigurewidth]{Figures/CRP/Qwen2.5-7B_few_10_shot_no_ablation_14_50_crp.png} &
    \includegraphics[width=\plotfigurewidth]{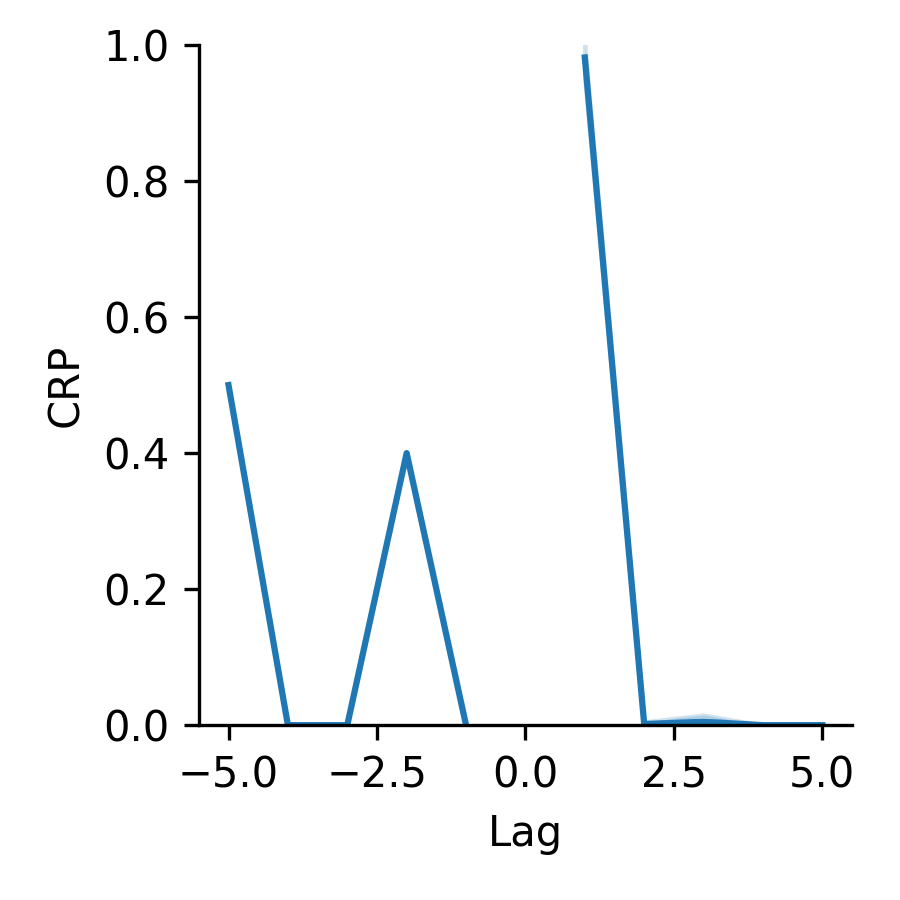} &
    \includegraphics[width=\plotfigurewidth]{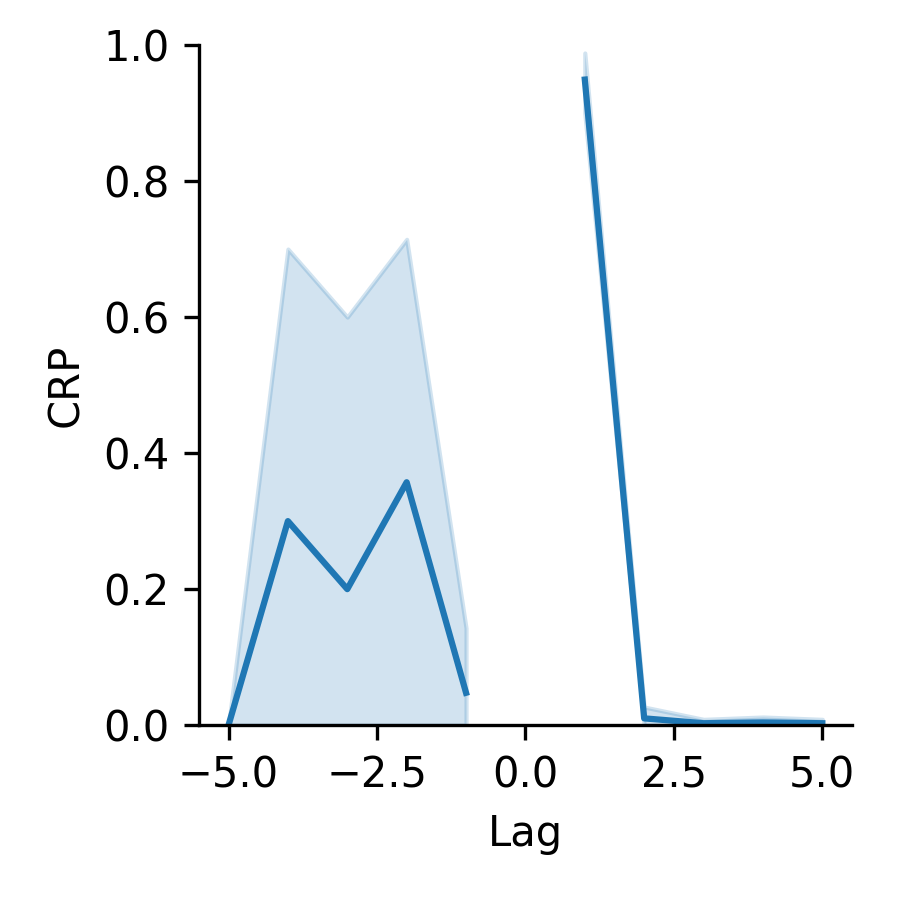} &
    \includegraphics[width=\plotfigurewidth]{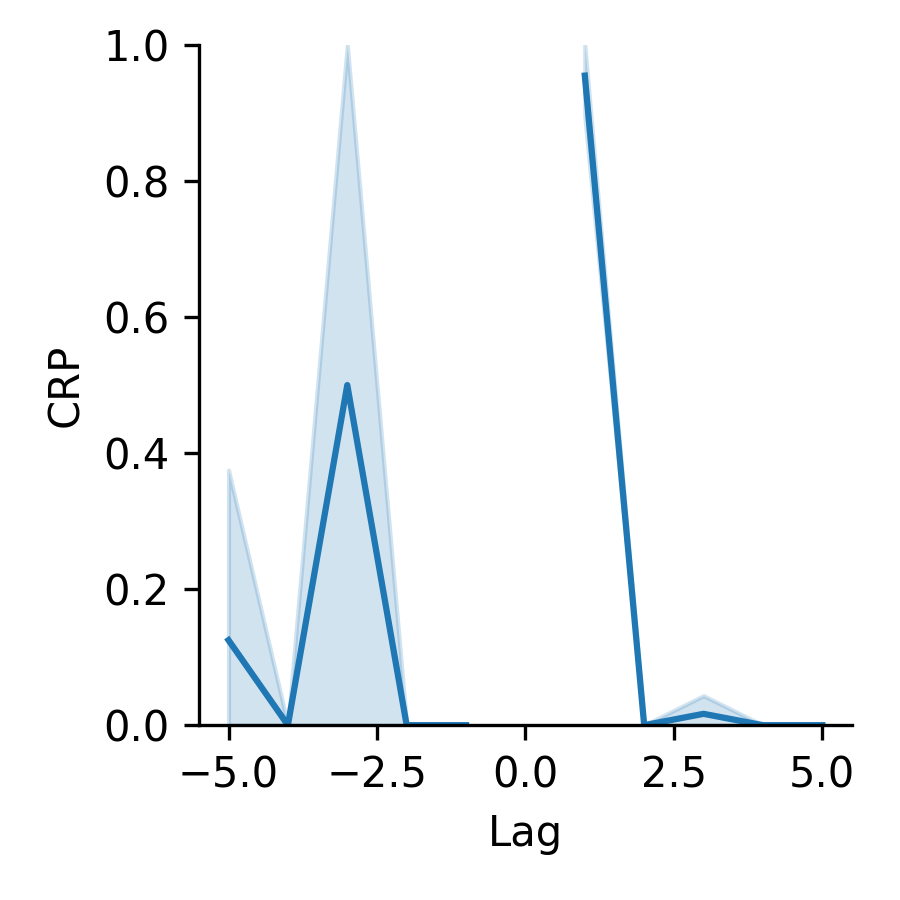} &
    \includegraphics[width=\plotfigurewidth]{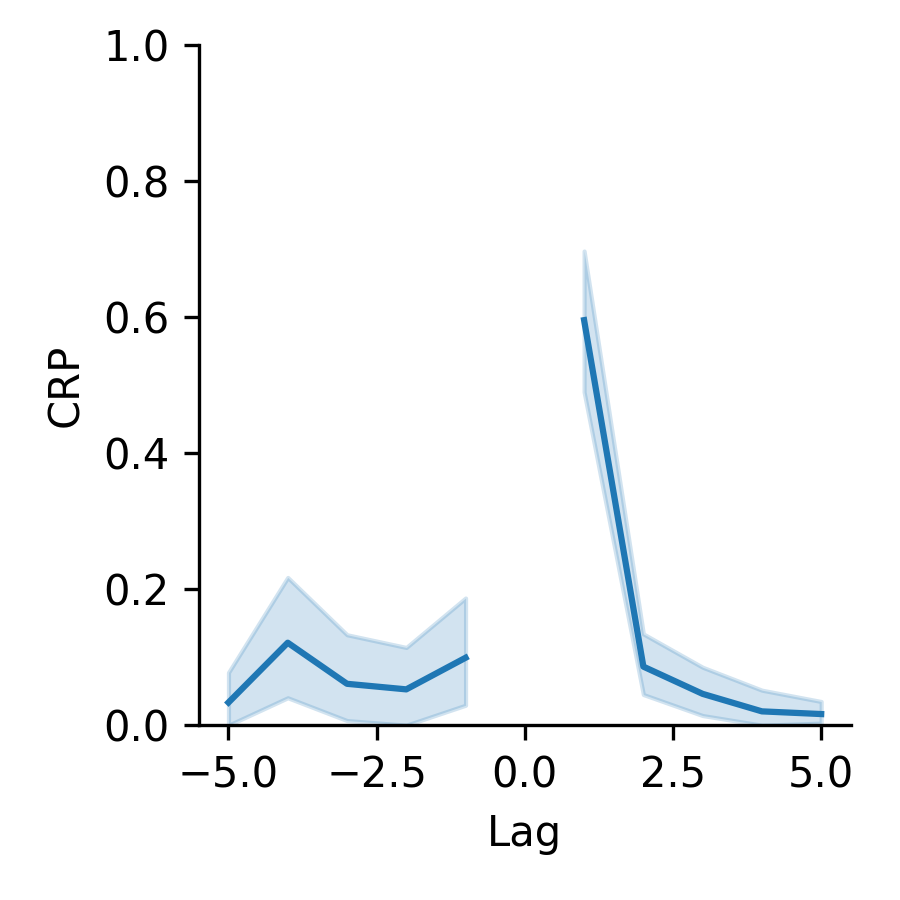} &
    \includegraphics[width=\plotfigurewidth]{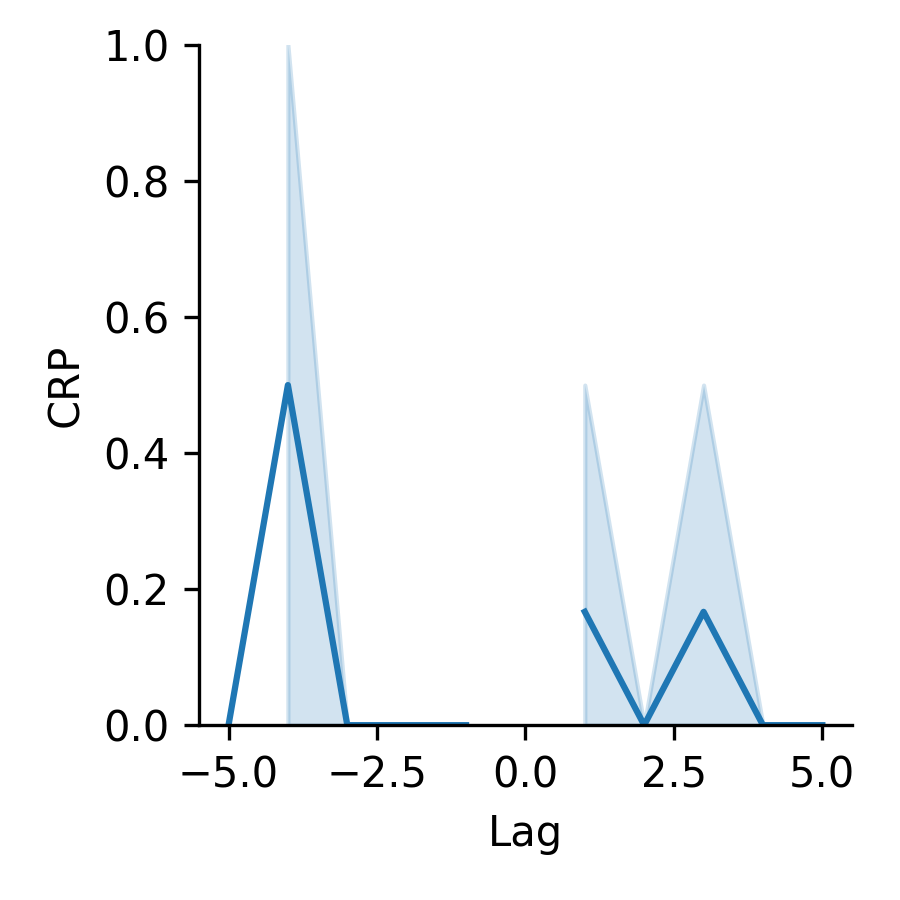} \\

    \rotatebox{90}{\parbox{2.2cm}{\centering\scriptsize\textbf{Qwen-7B-I}\\Ind. Abl.}} &
    \includegraphics[width=\plotfigurewidth]{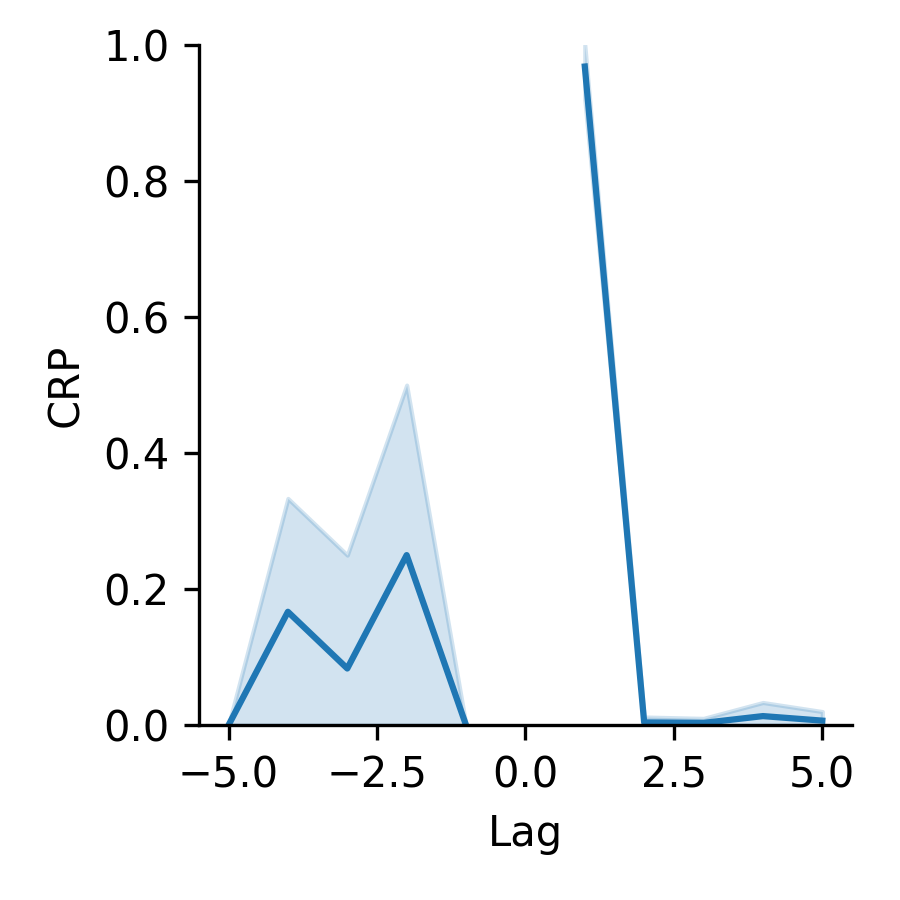} &
    \includegraphics[width=\plotfigurewidth]{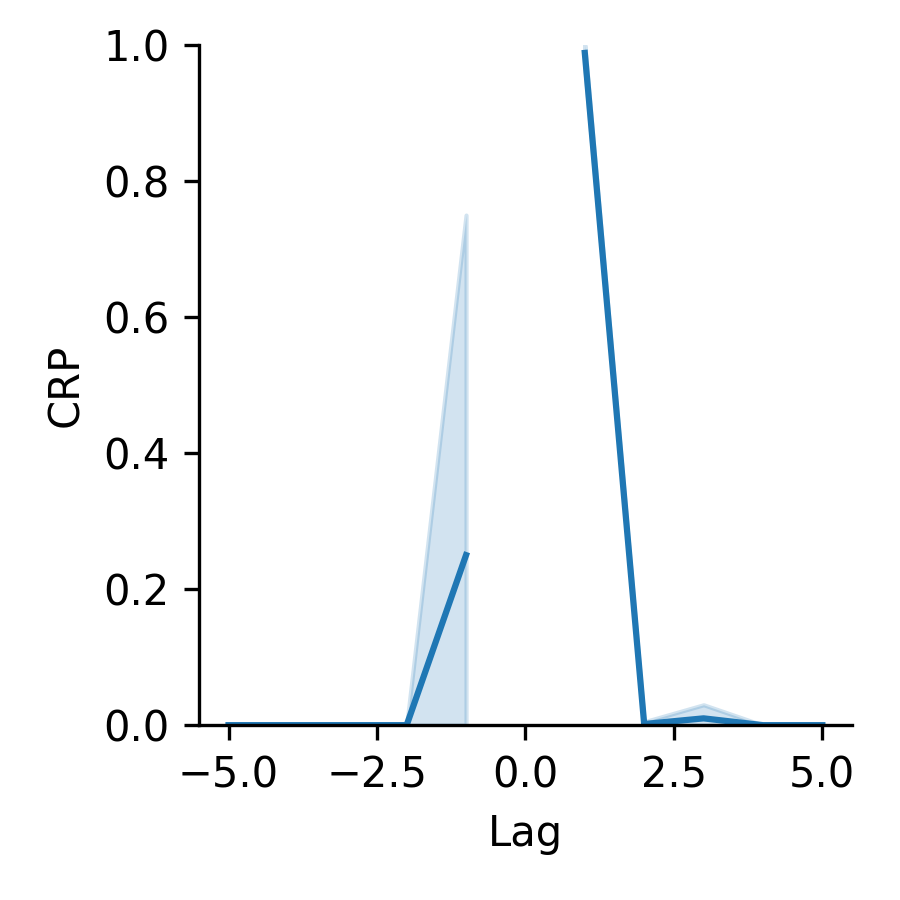} &
    \includegraphics[width=\plotfigurewidth]{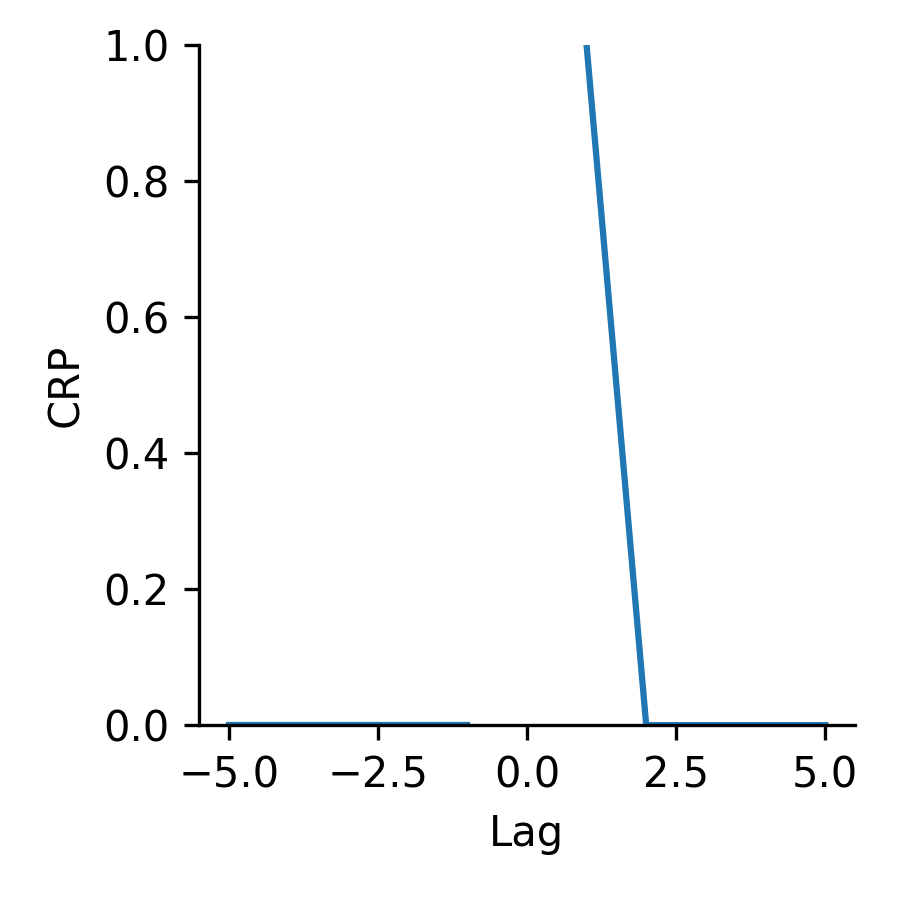} &
    \includegraphics[width=\plotfigurewidth]{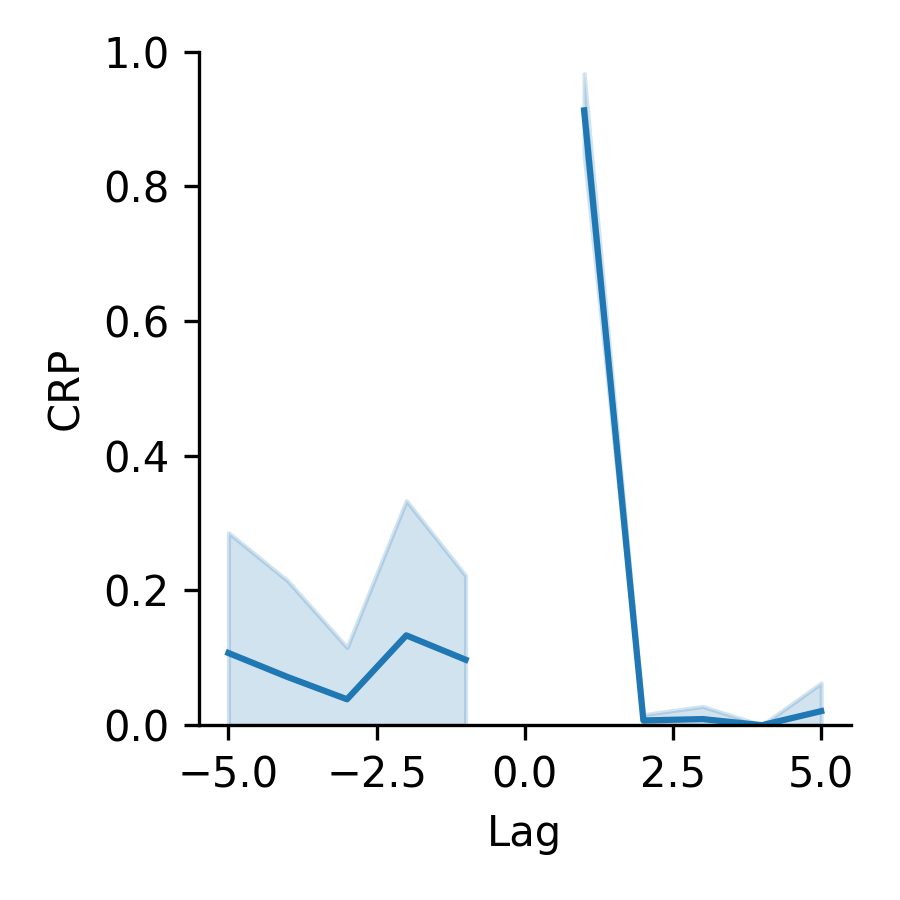} &
    \includegraphics[width=\plotfigurewidth]{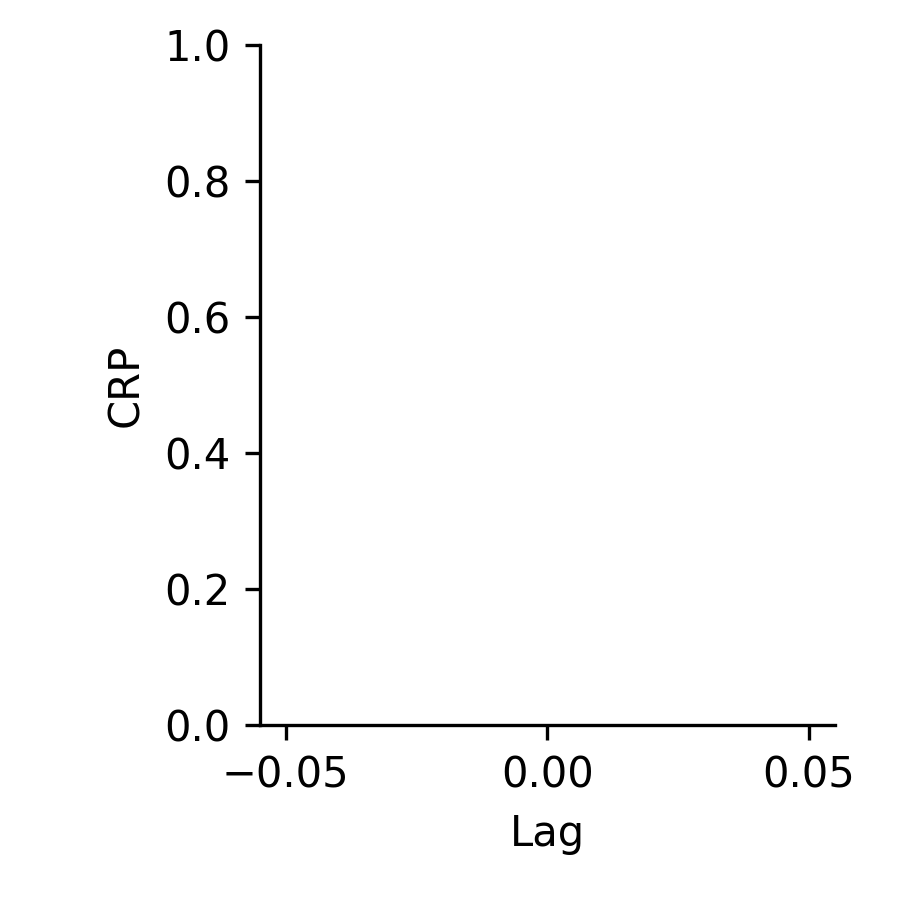} &
    \includegraphics[width=\plotfigurewidth]{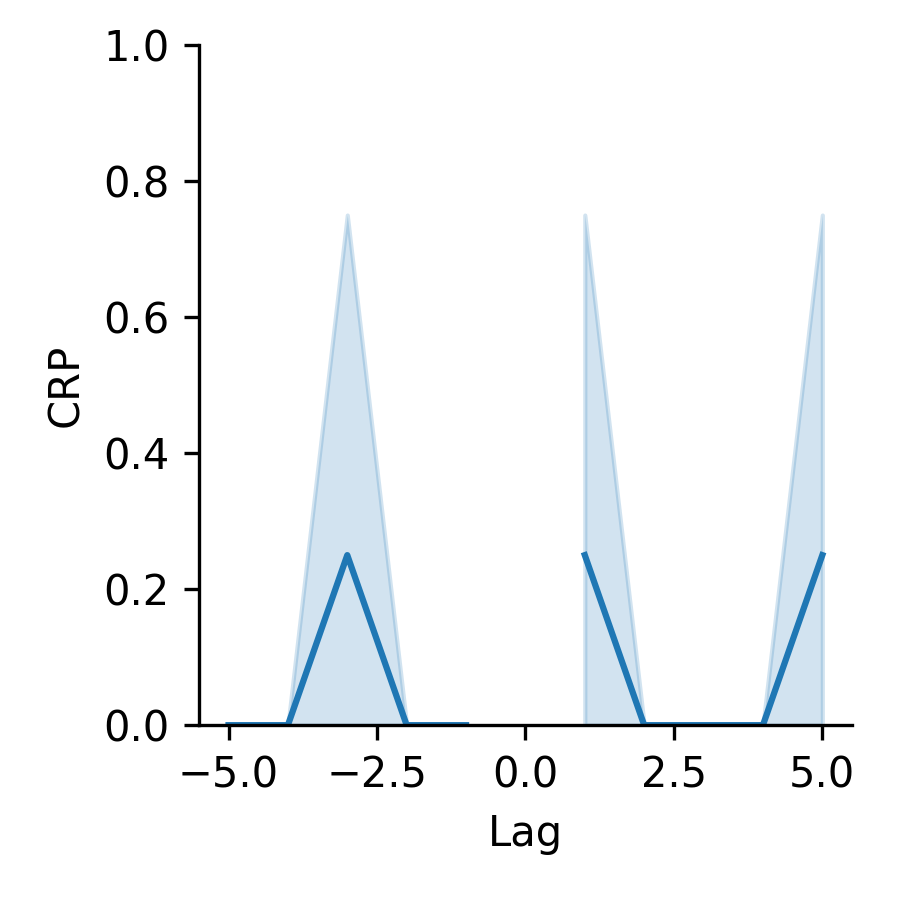} \\

    \rotatebox{90}{\parbox{2.2cm}{\centering\scriptsize\textbf{Qwen-7B-I}\\Rand. Abl.}} &
    \includegraphics[width=\plotfigurewidth]{Figures/CRP/Qwen2.5-7B-Instruct_few_10_shot_no_ablation_14_50_crp.png} &
    \includegraphics[width=\plotfigurewidth]{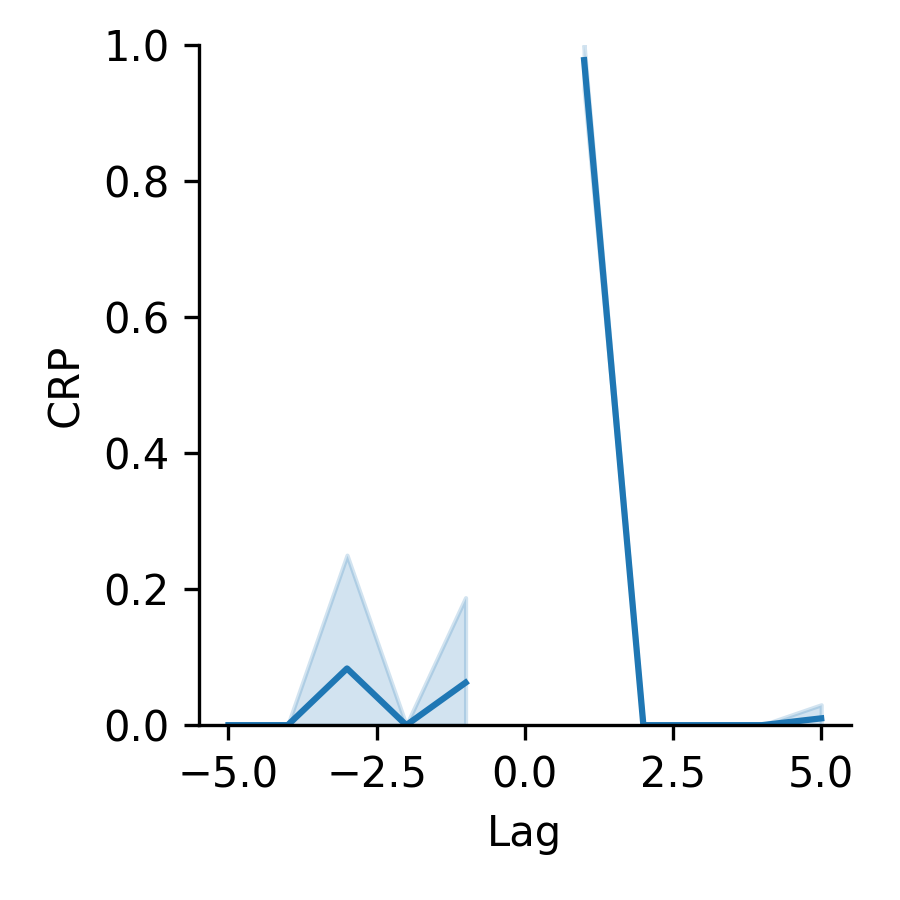} &
    \includegraphics[width=\plotfigurewidth]{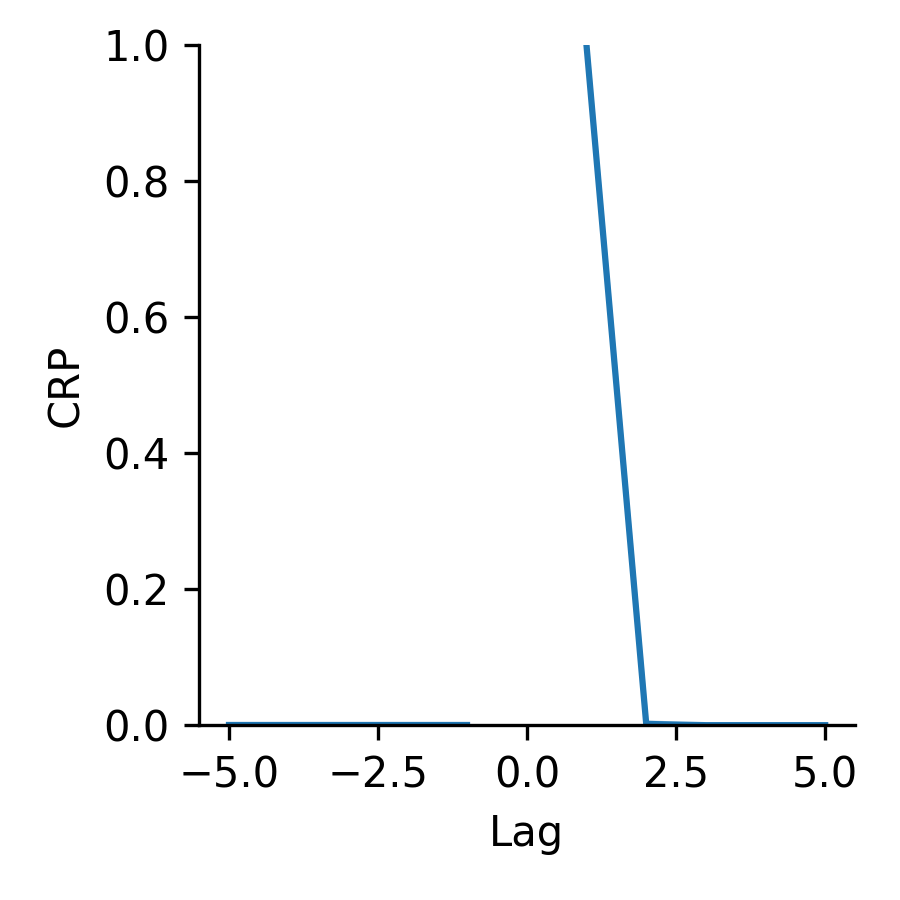} &
    \includegraphics[width=\plotfigurewidth]{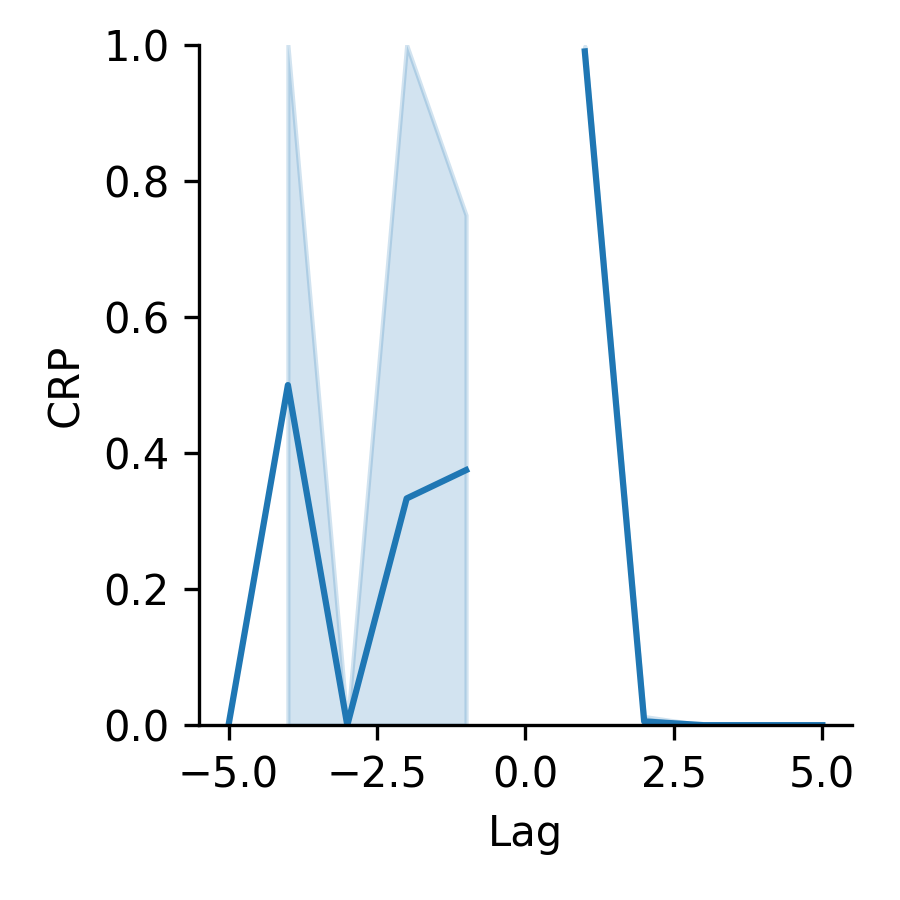} &
    \includegraphics[width=\plotfigurewidth]{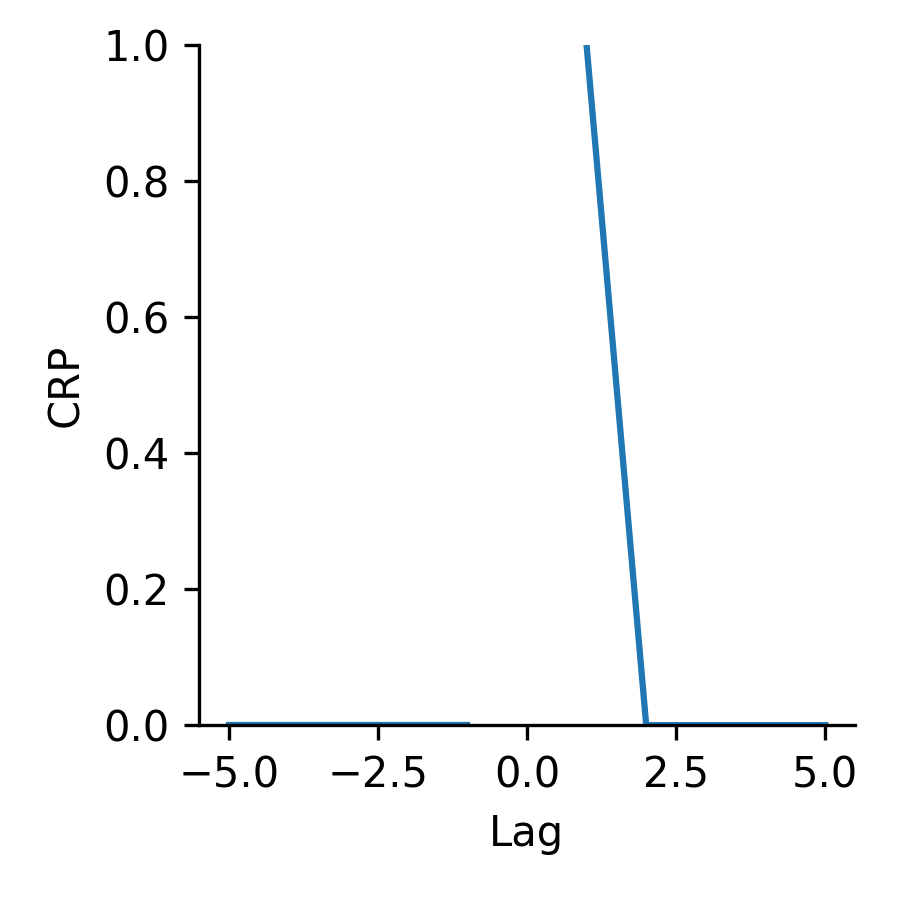} &
    \includegraphics[width=\plotfigurewidth]{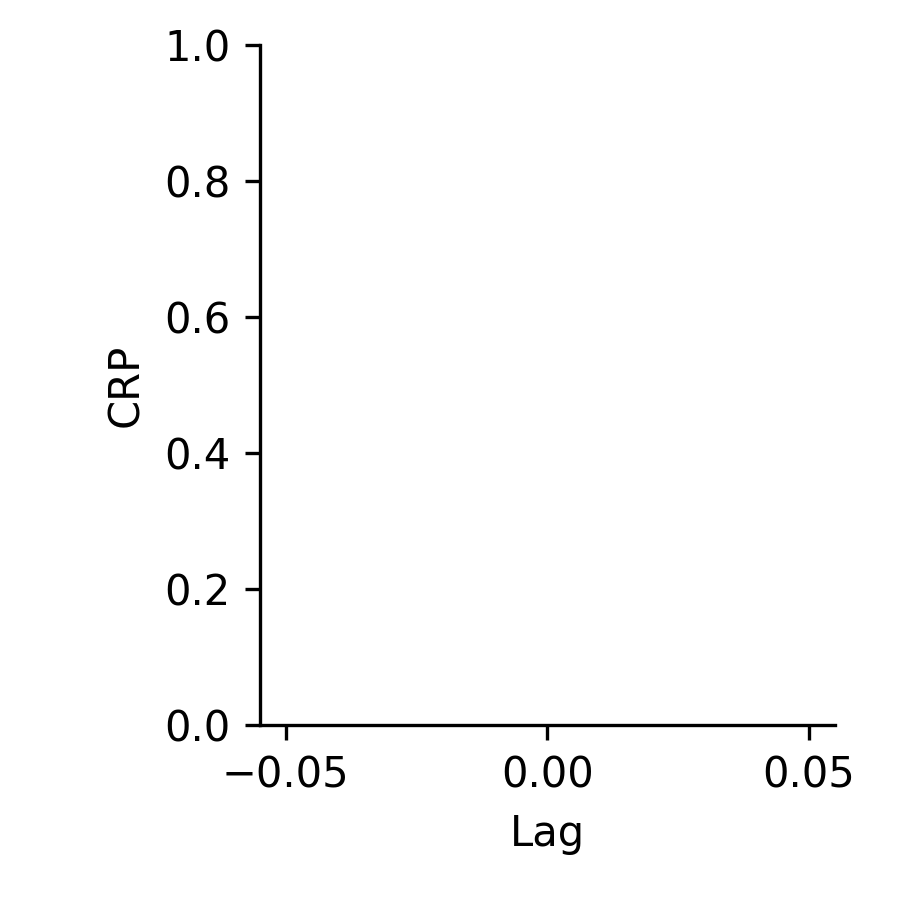} \\
    \end{tabular}

    \caption{CRP in serial recall ICL task for different models and ablation conditions.}
    \label{fig:CRP_ablation_grid}
\end{figure*}

\begin{figure*}[h!]
    \centering
    \footnotesize

    \setlength{\tabcolsep}{1.5pt}

    \begin{tabular}{c c c c c c c}
    & \parbox{\plotfigurewidth}{\centering\scriptsize\textbf{No Ablation}}
    & \parbox{\plotfigurewidth}{\centering\scriptsize\textbf{1 Head Ablated}}
    & \parbox{\plotfigurewidth}{\centering\scriptsize\textbf{10 Heads Ablated}}
    & \parbox{\plotfigurewidth}{\centering\scriptsize\textbf{25 Heads Ablated}}
    & \parbox{\plotfigurewidth}{\centering\scriptsize\textbf{50 Heads Ablated}}
    & \parbox{\plotfigurewidth}{\centering\scriptsize\textbf{100 Heads Ablated}} \\

    \rotatebox{90}{\parbox{2.2cm}{\centering\scriptsize\textbf{Llama-8B}\\Ind. Abl.}} &
    \includegraphics[width=\plotfigurewidth]{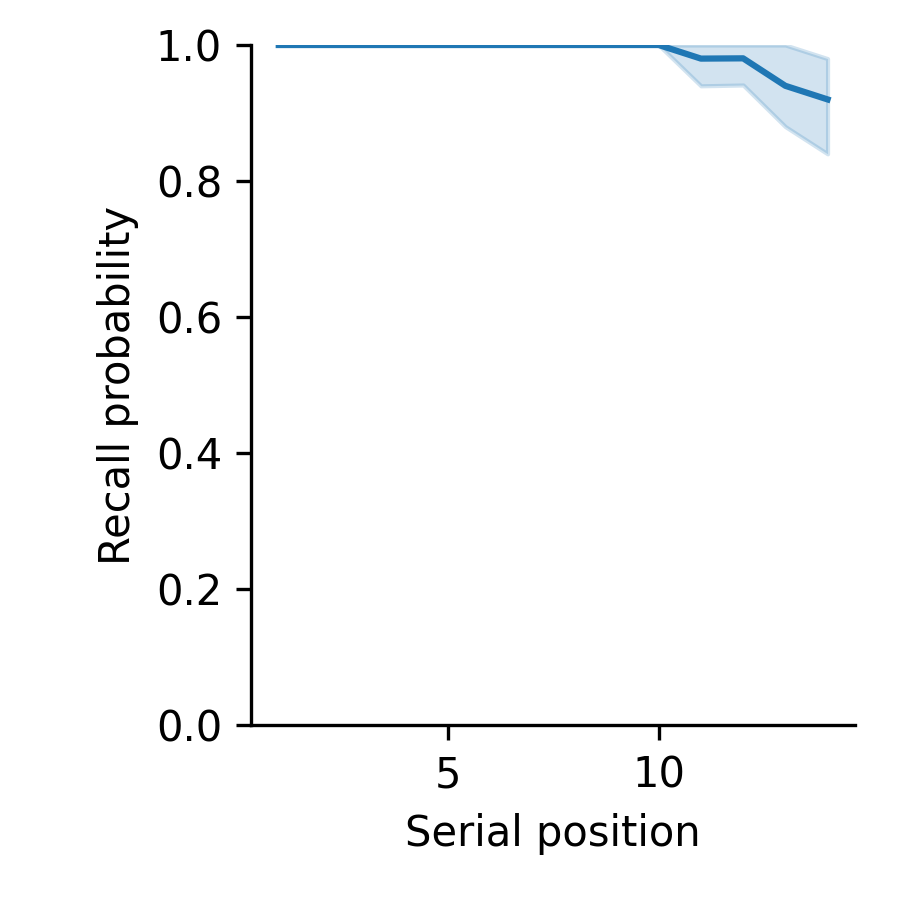} &
    \includegraphics[width=\plotfigurewidth]{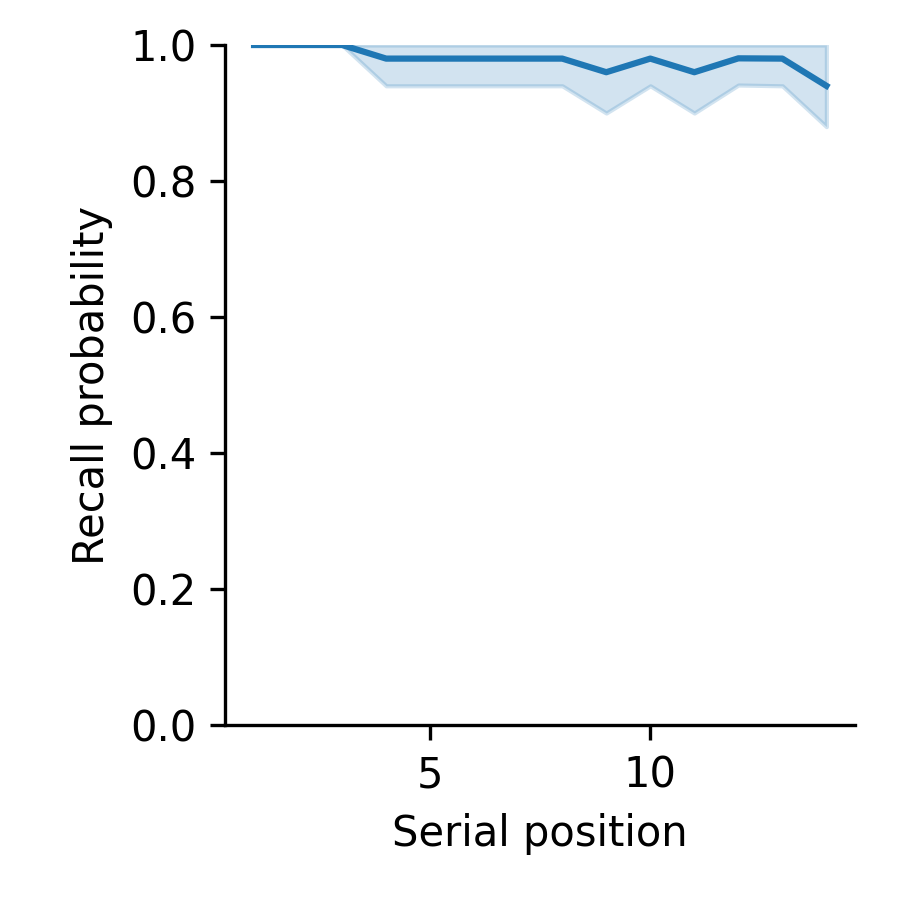} &
    \includegraphics[width=\plotfigurewidth]{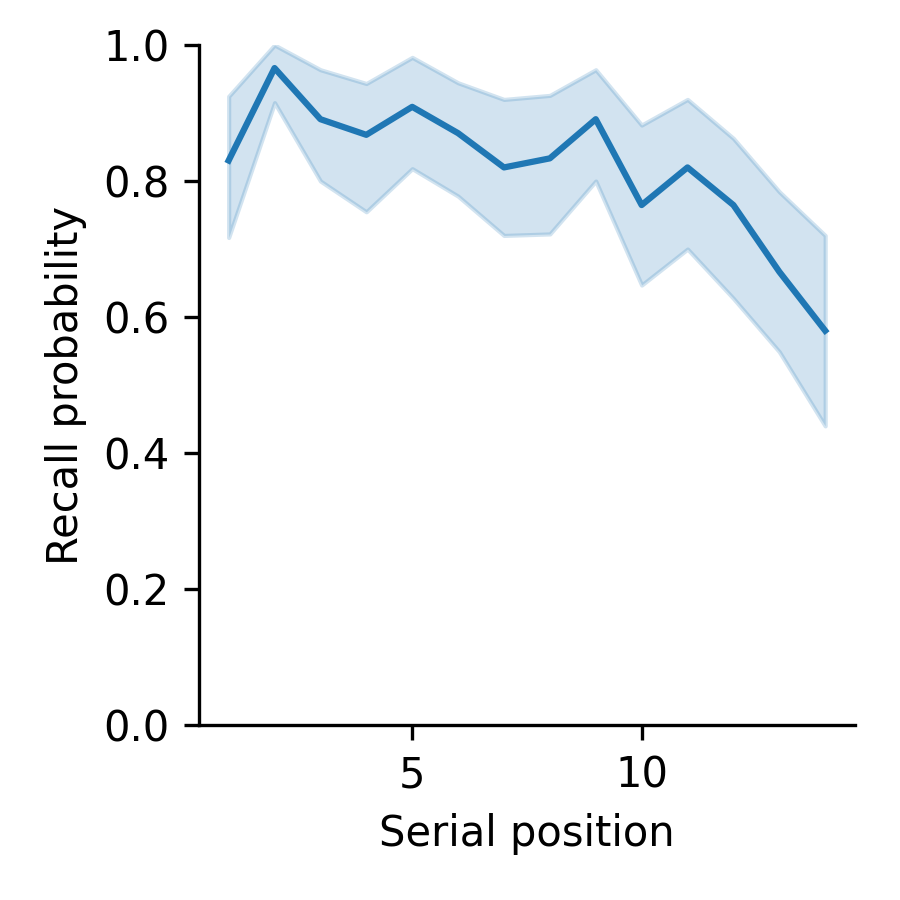} &
    \includegraphics[width=\plotfigurewidth]{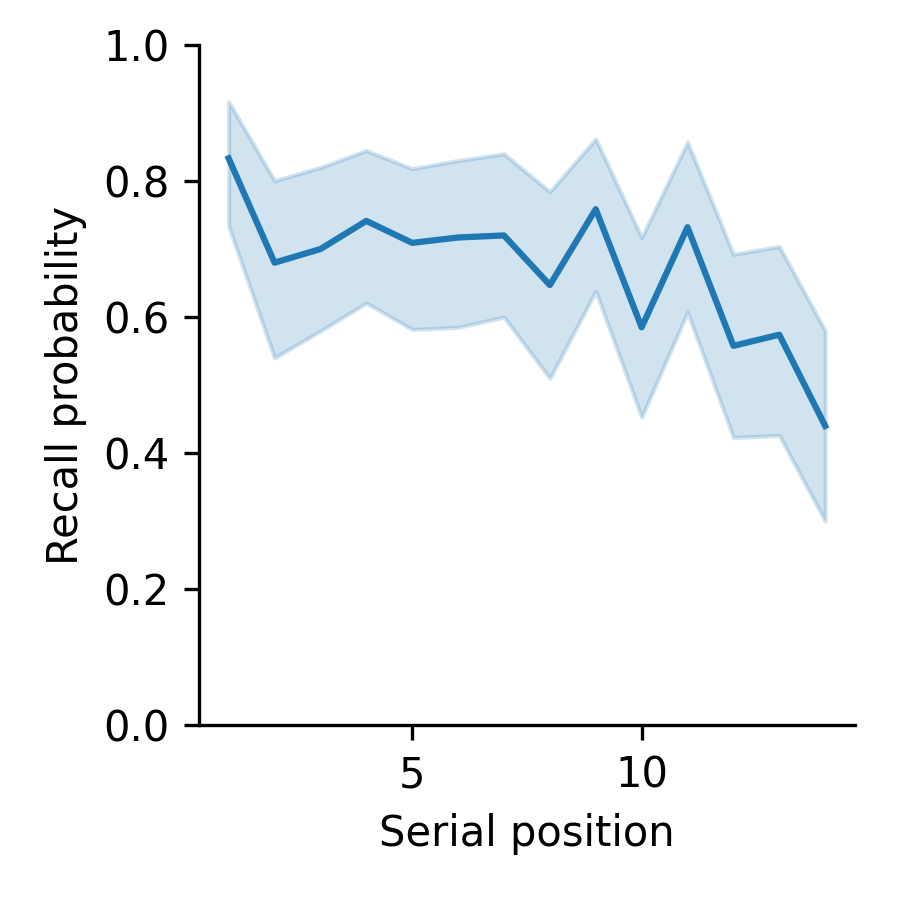} &
    \includegraphics[width=\plotfigurewidth]{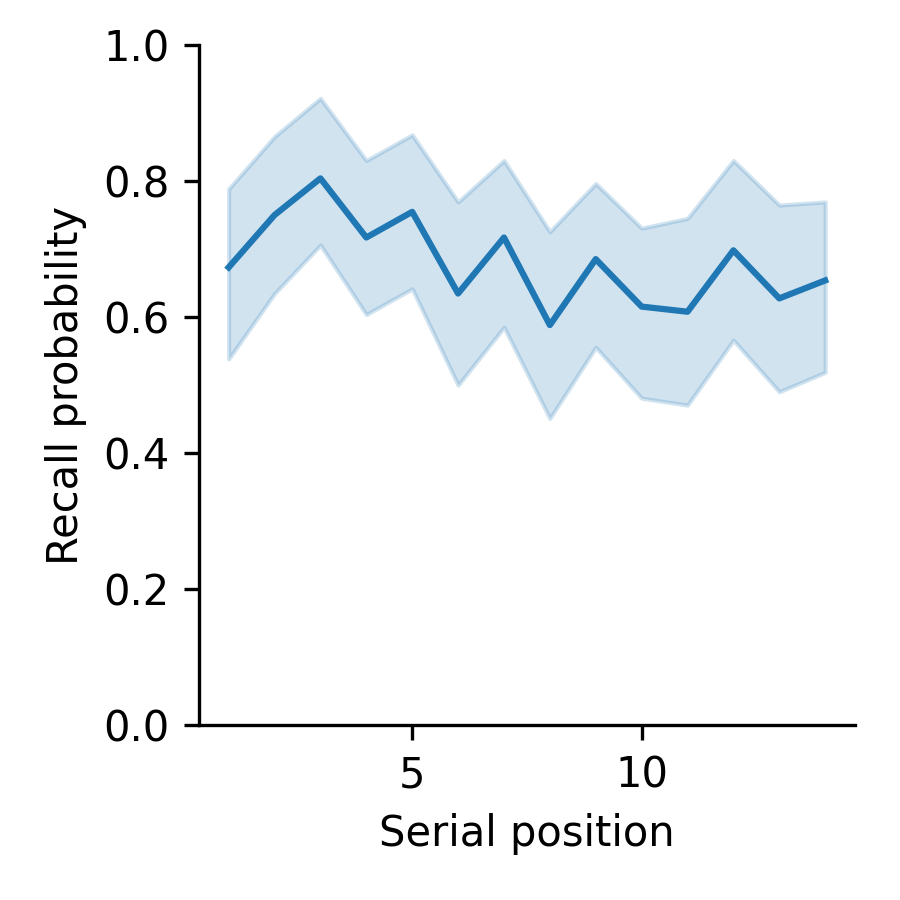} &
    \includegraphics[width=\plotfigurewidth]{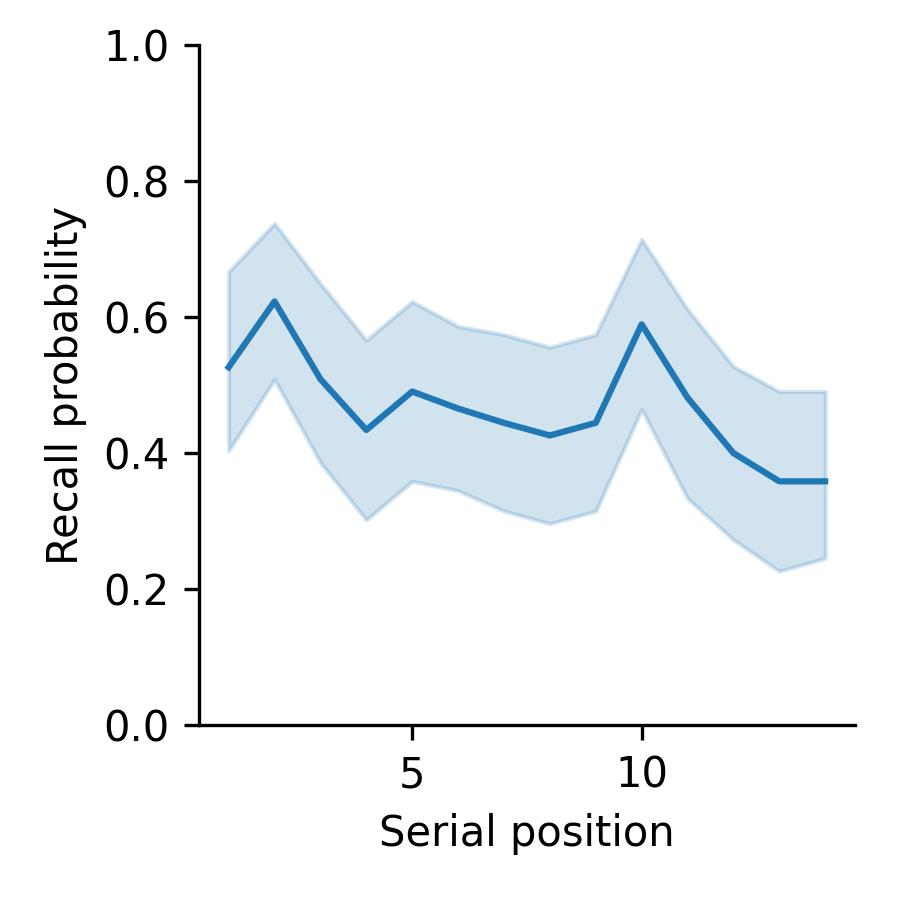} \\

    \rotatebox{90}{\parbox{2.2cm}{\centering\scriptsize\textbf{Llama-8B}\\Rand. Abl.}} &
    \includegraphics[width=\plotfigurewidth]{Figures/CRP/Llama-3.1-8B_few_10_shot_no_ablation_14_50_spc.png} &
    \includegraphics[width=\plotfigurewidth]{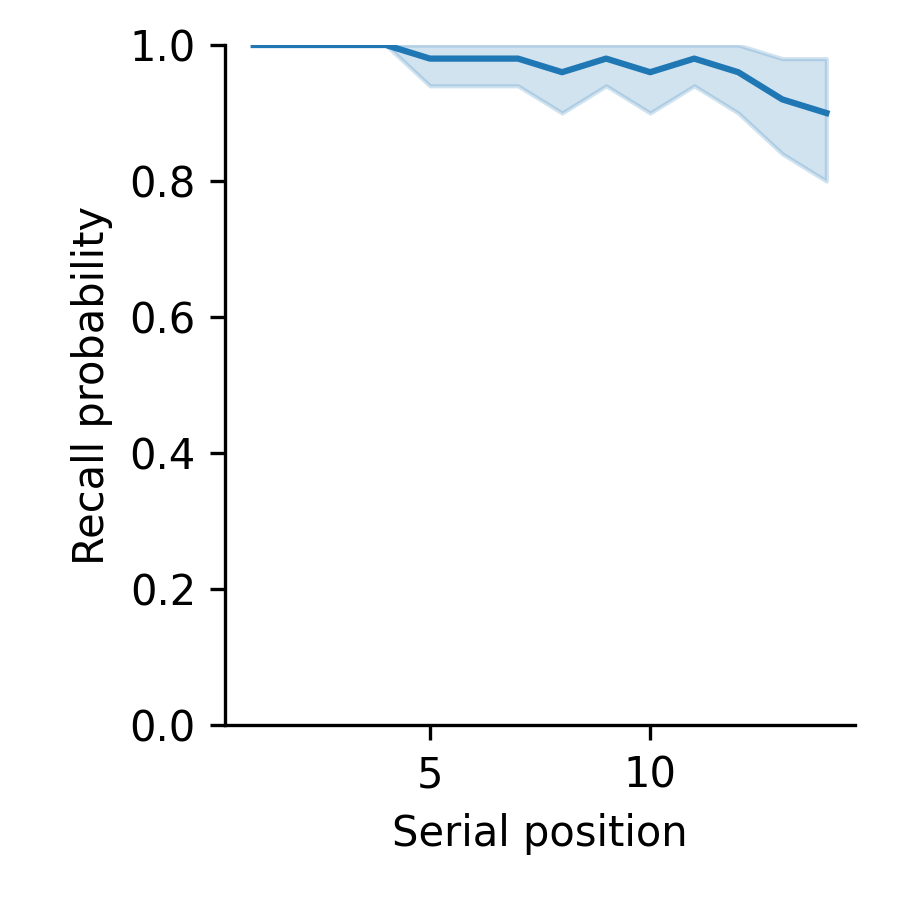} &
    \includegraphics[width=\plotfigurewidth]{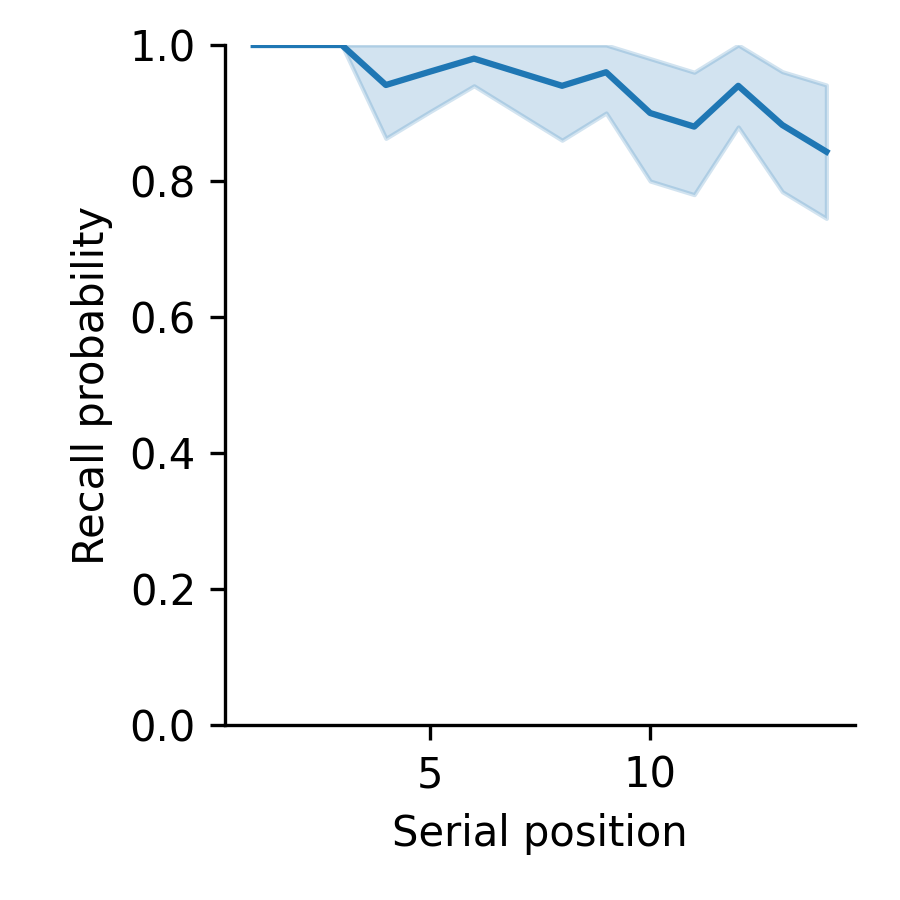} &
    \includegraphics[width=\plotfigurewidth]{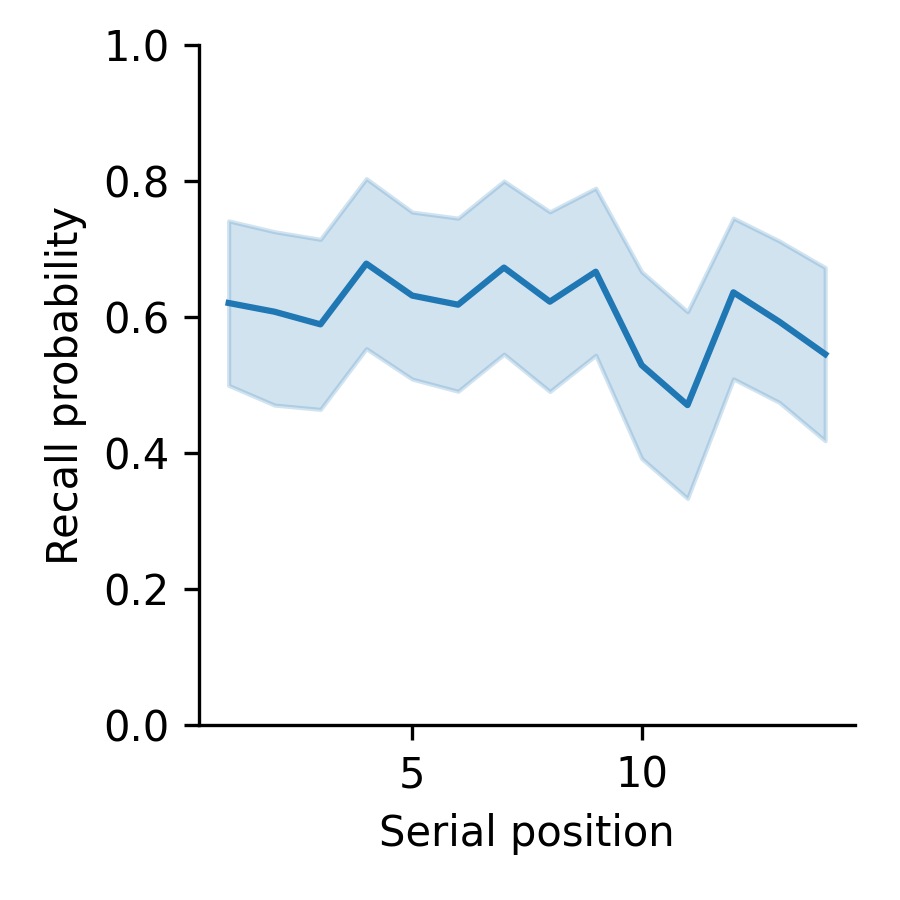} &
    \includegraphics[width=\plotfigurewidth]{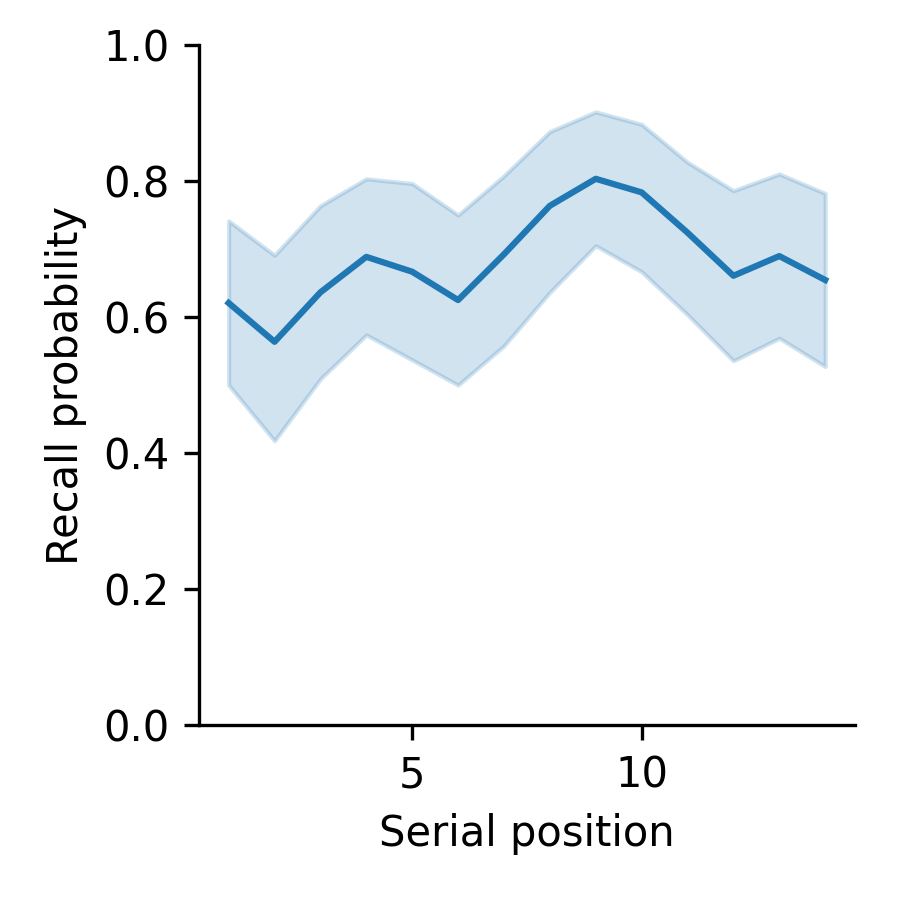} &
    \includegraphics[width=\plotfigurewidth]{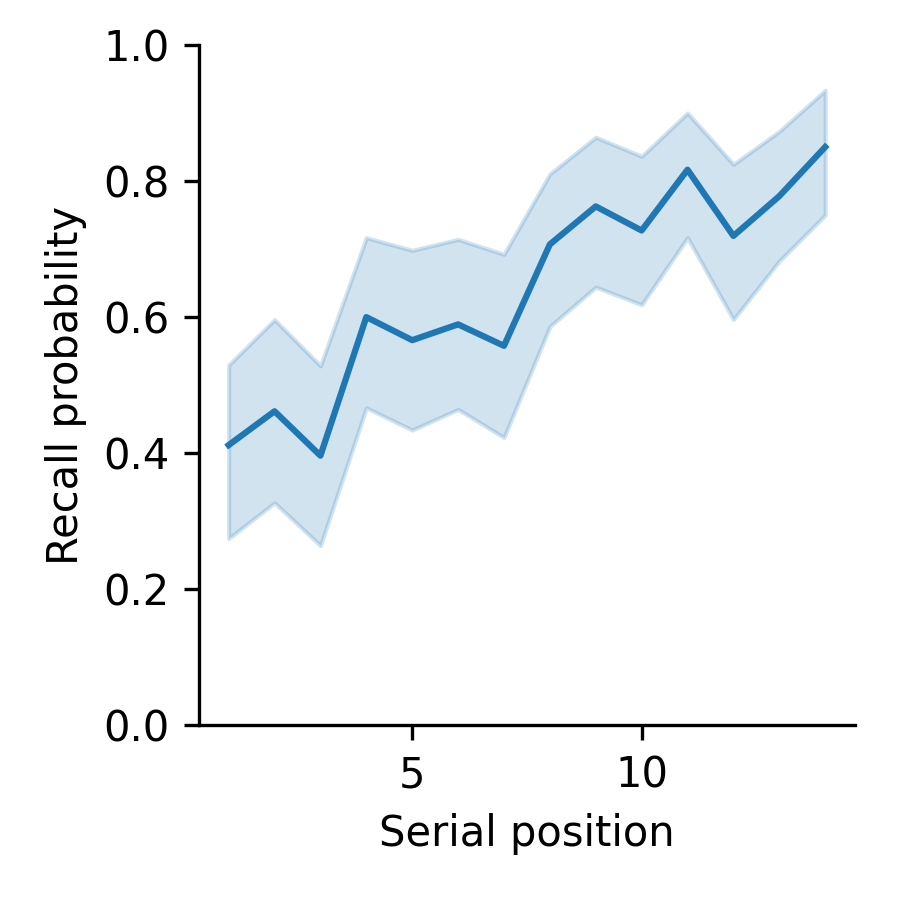} \\

    \rotatebox{90}{\parbox{2.2cm}{\centering\scriptsize\textbf{Llama-8B-I}\\Ind. Abl.}} &
    \includegraphics[width=\plotfigurewidth]{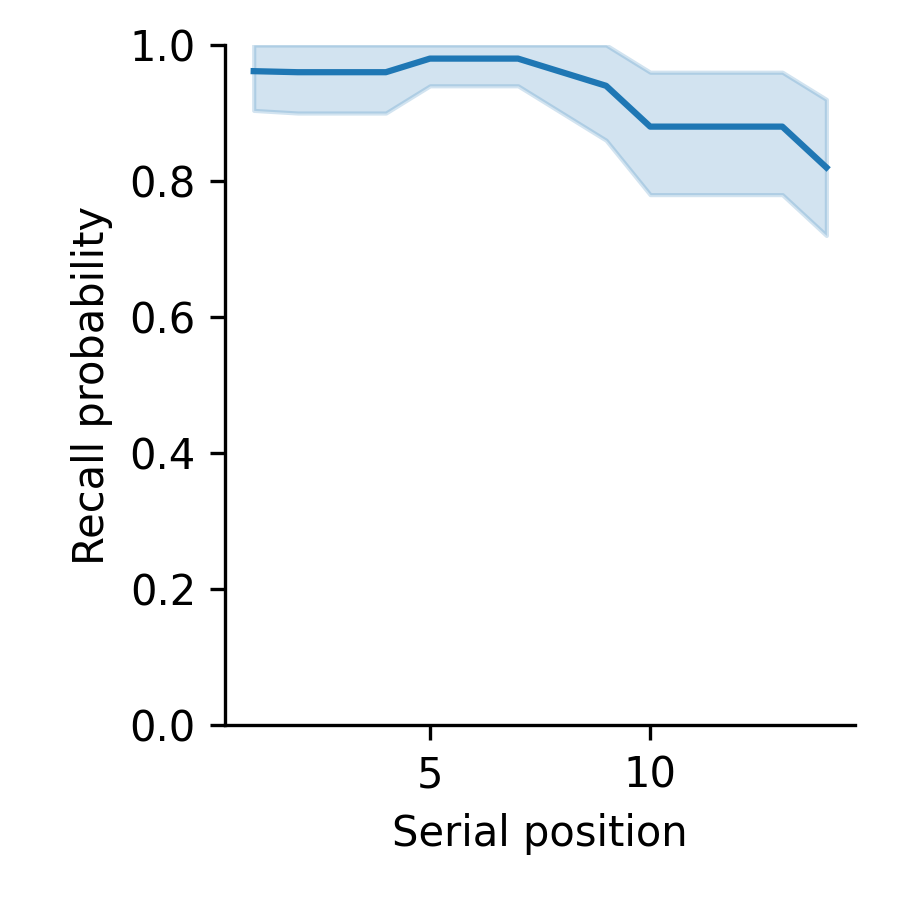} &
    \includegraphics[width=\plotfigurewidth]{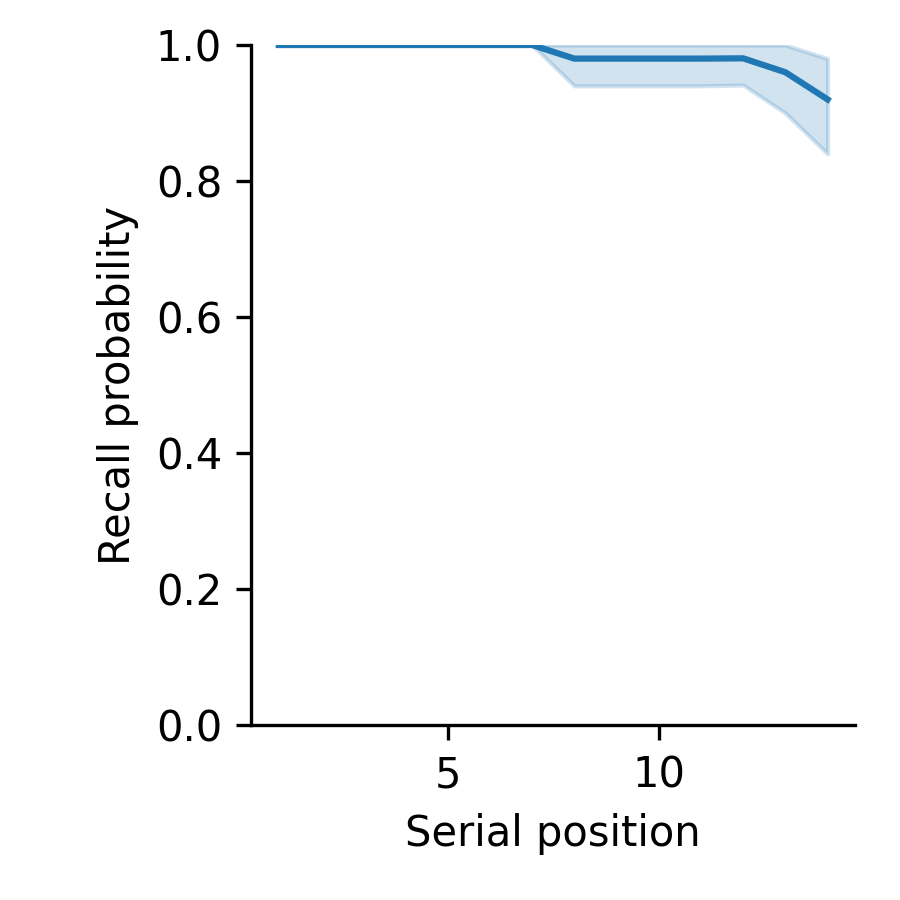} &
    \includegraphics[width=\plotfigurewidth]{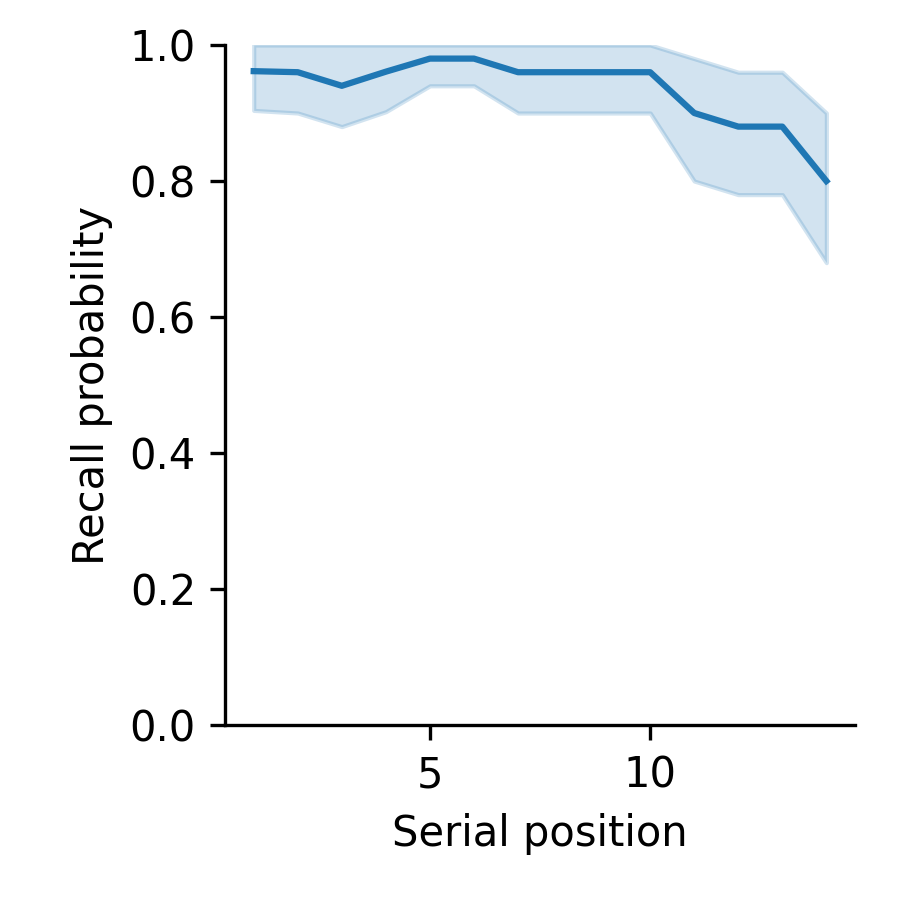} &
    \includegraphics[width=\plotfigurewidth]{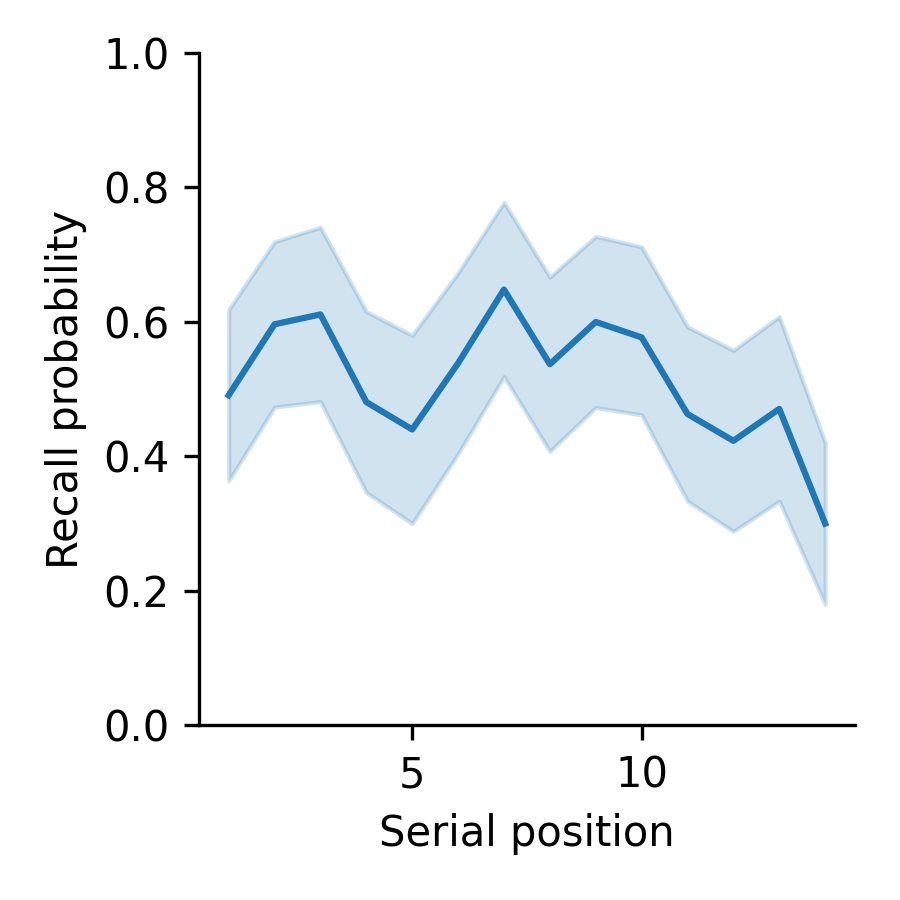} &
    \includegraphics[width=\plotfigurewidth]{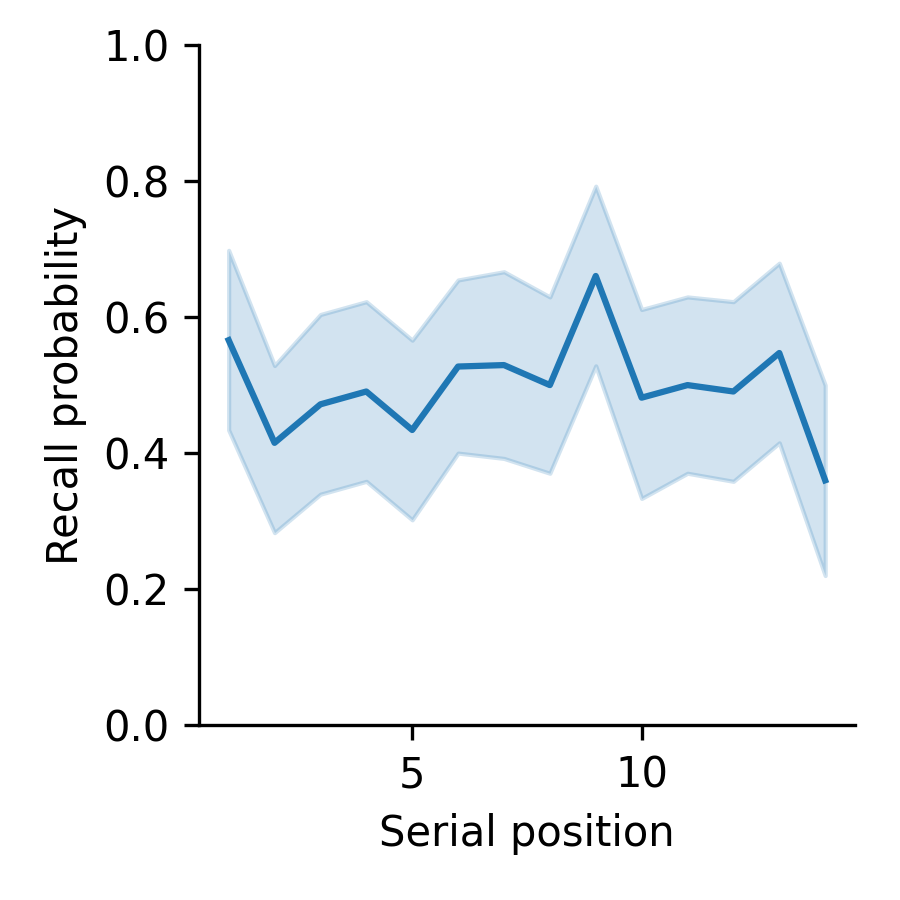} &
    \includegraphics[width=\plotfigurewidth]{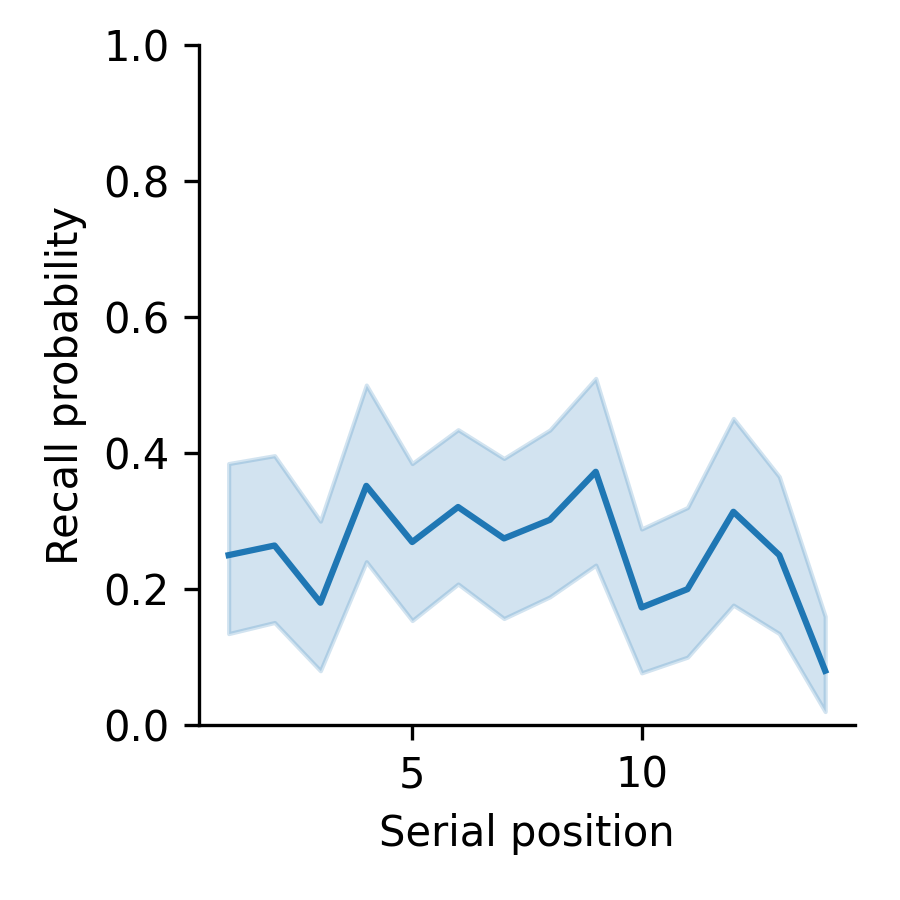} \\

    \rotatebox{90}{\parbox{2.2cm}{\centering\scriptsize\textbf{Llama-8B-I}\\Rand. Abl.}} &
    \includegraphics[width=\plotfigurewidth]{Figures/CRP/Llama-3.1-8B-Instruct_few_10_shot_no_ablation_14_50_spc.png} &
    \includegraphics[width=\plotfigurewidth]{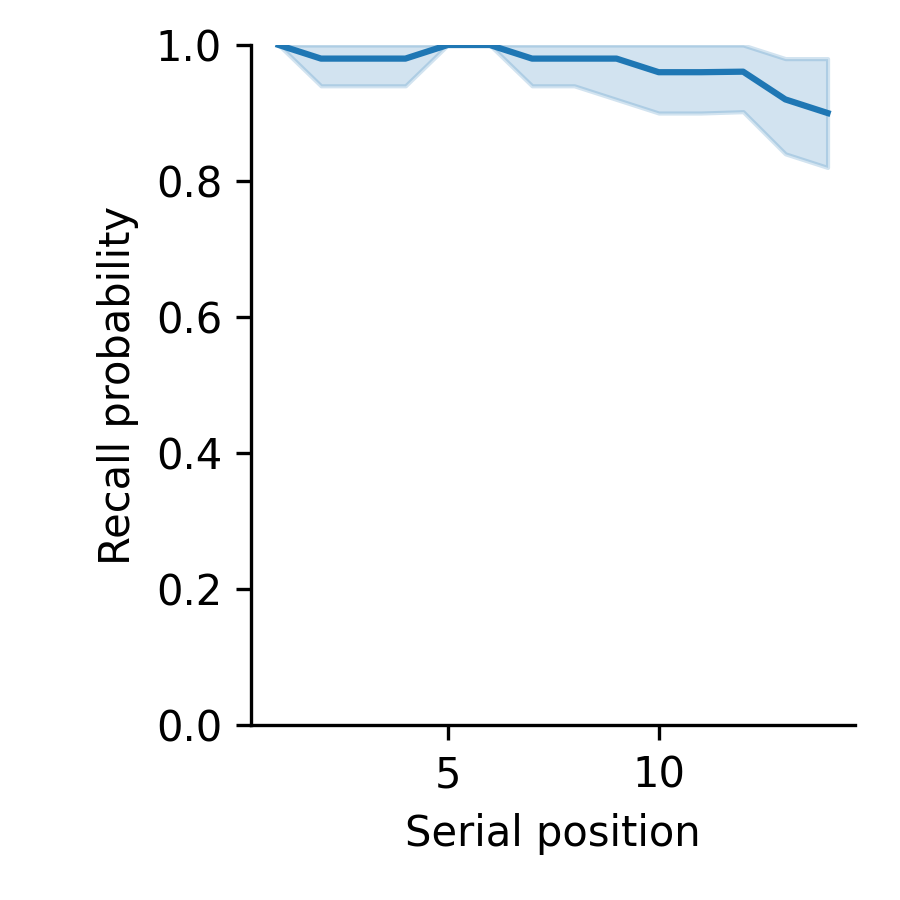} &
    \includegraphics[width=\plotfigurewidth]{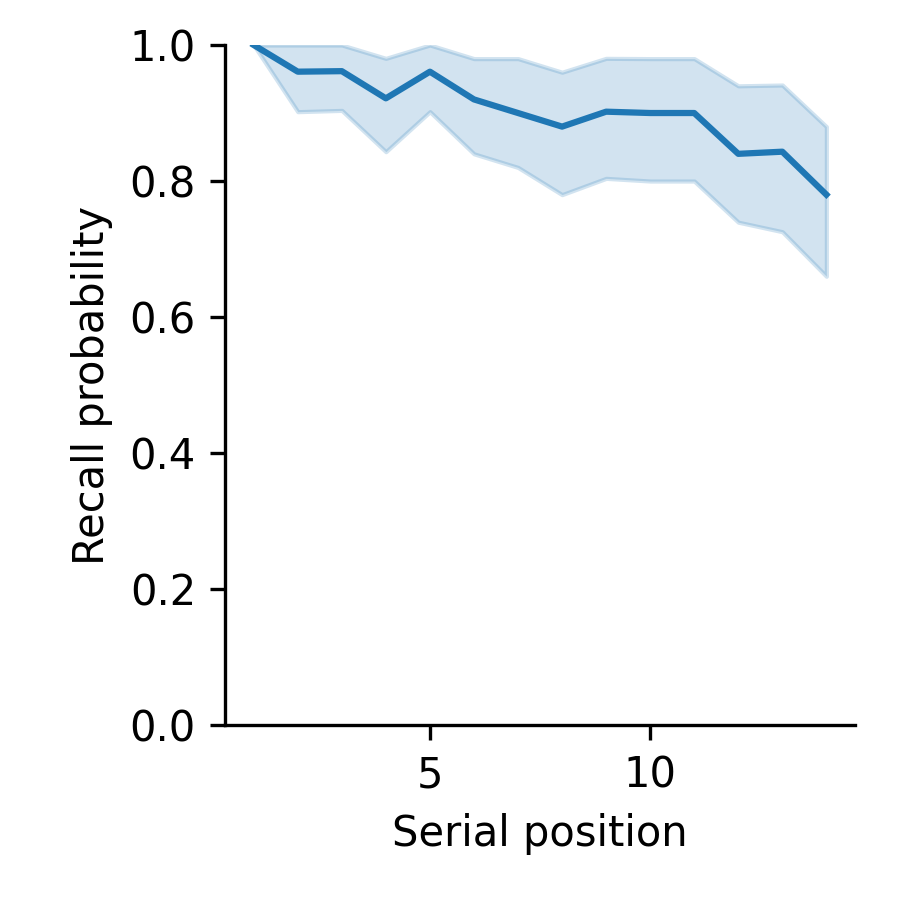} &
    \includegraphics[width=\plotfigurewidth]{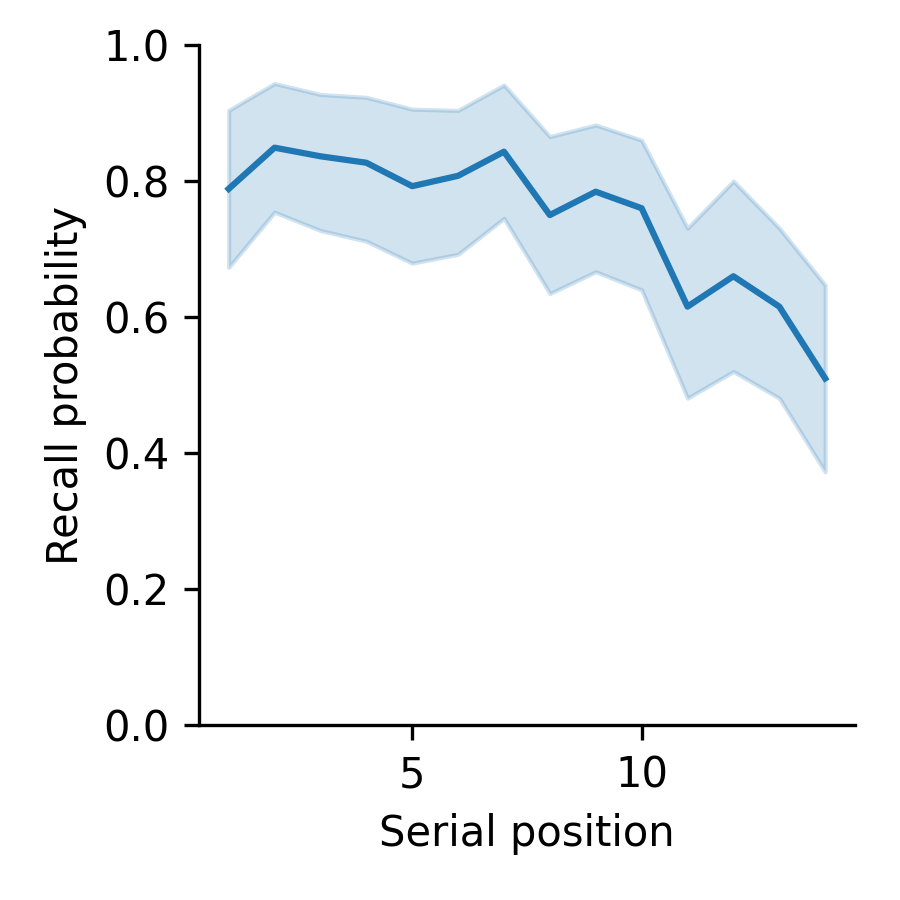} &
    \includegraphics[width=\plotfigurewidth]{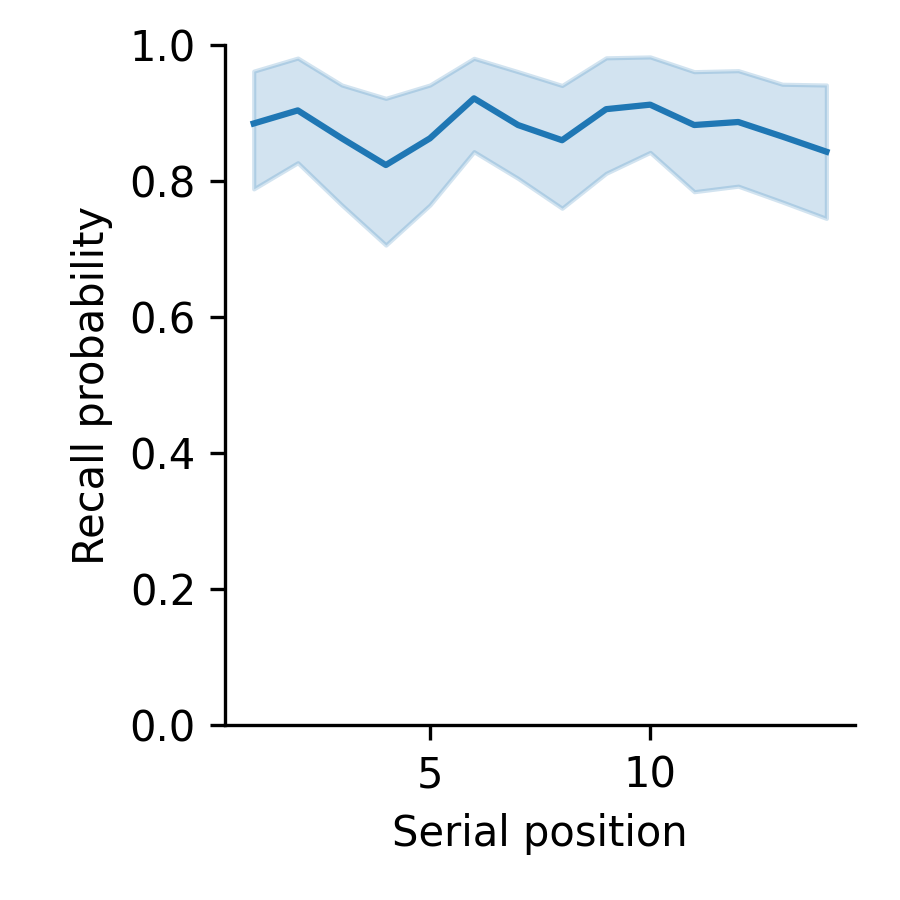} &
    \includegraphics[width=\plotfigurewidth]{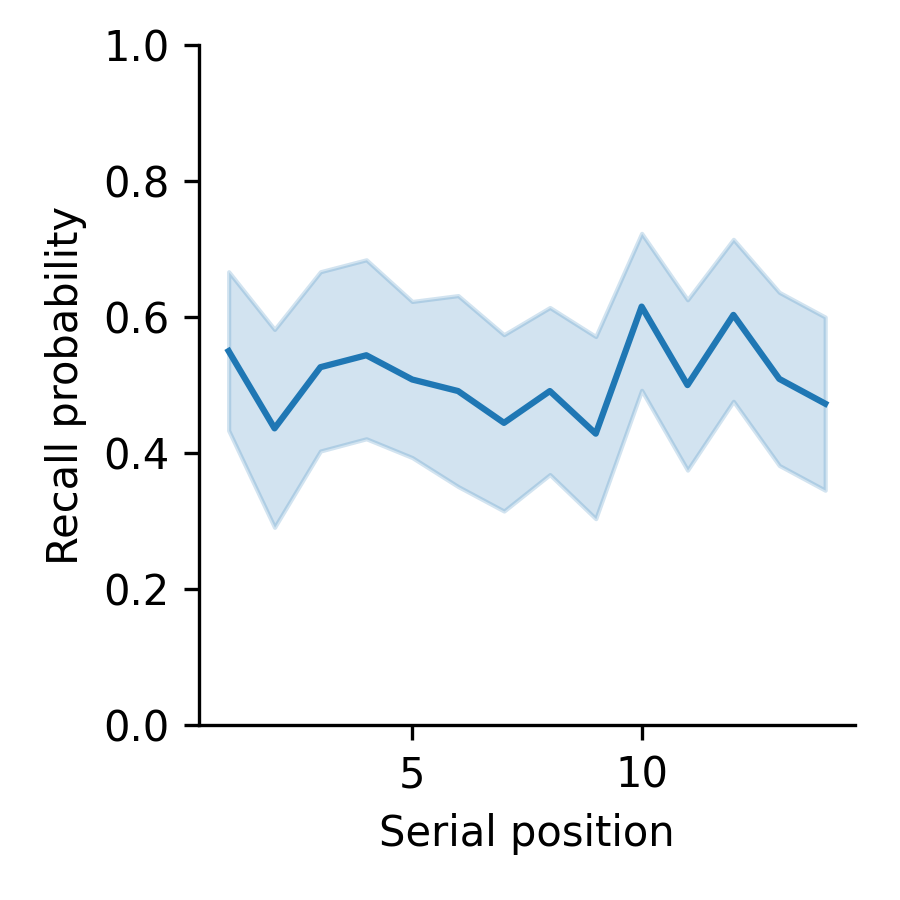} \\

    \rotatebox{90}{\parbox{2.2cm}{\centering\scriptsize\textbf{Qwen-7B}\\Ind. Abl.}} &
    \includegraphics[width=\plotfigurewidth]{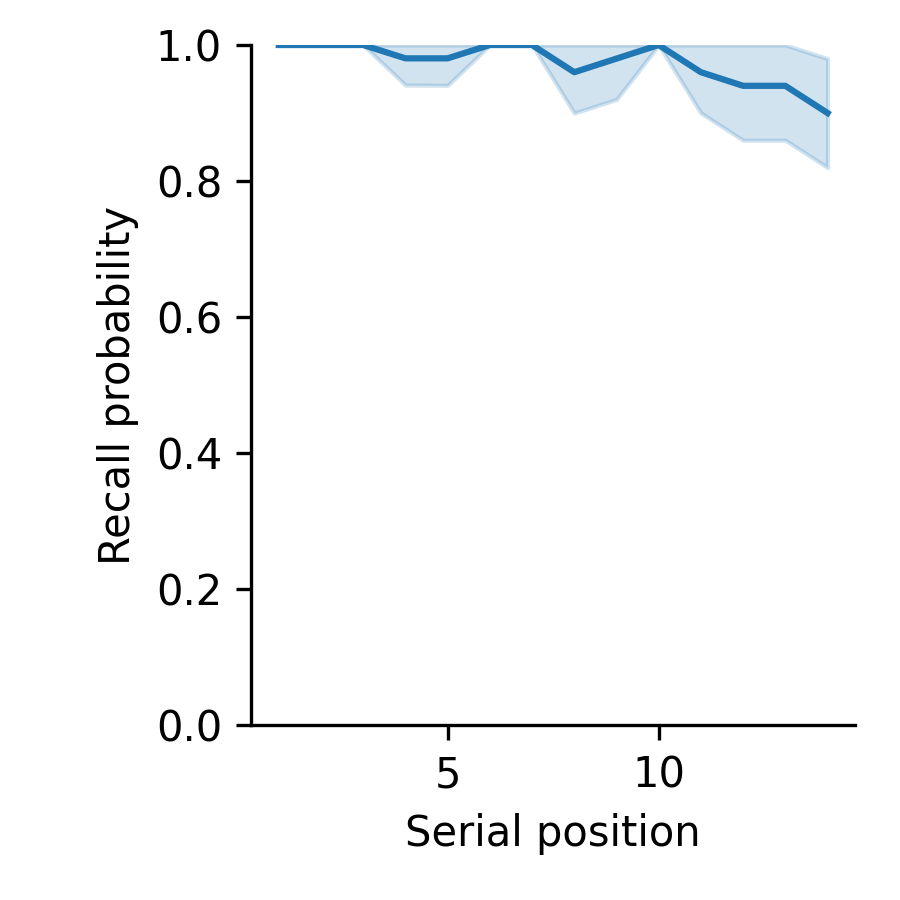} &
    \includegraphics[width=\plotfigurewidth]{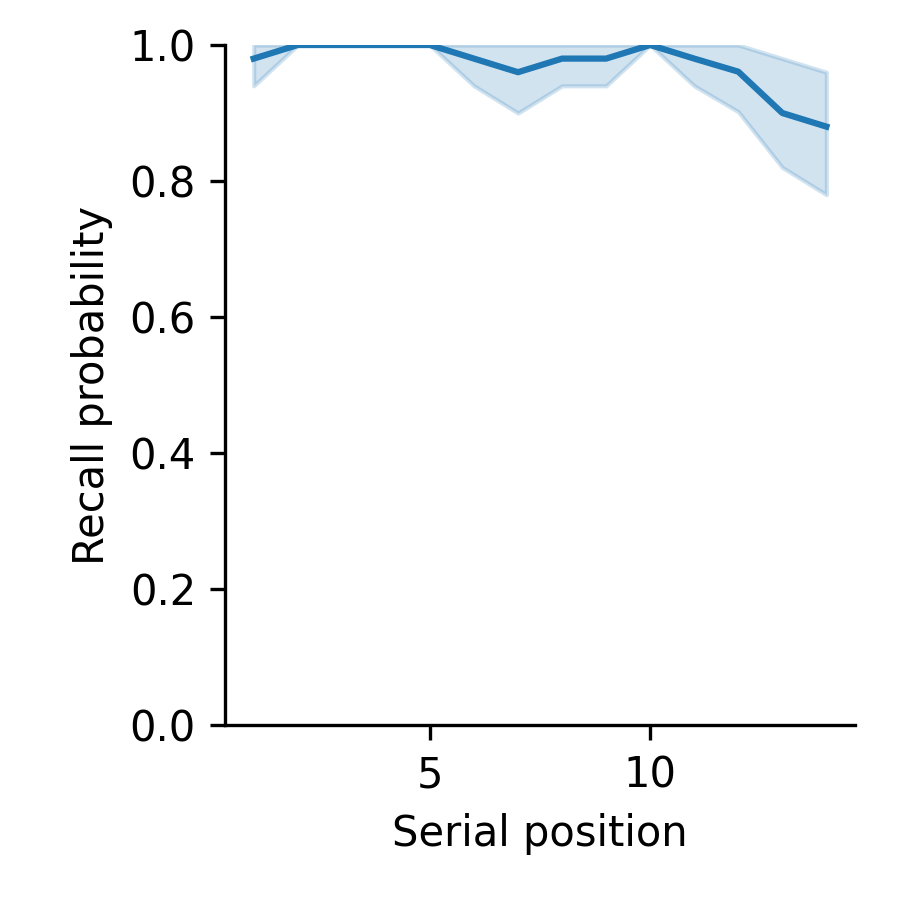} &
    \includegraphics[width=\plotfigurewidth]{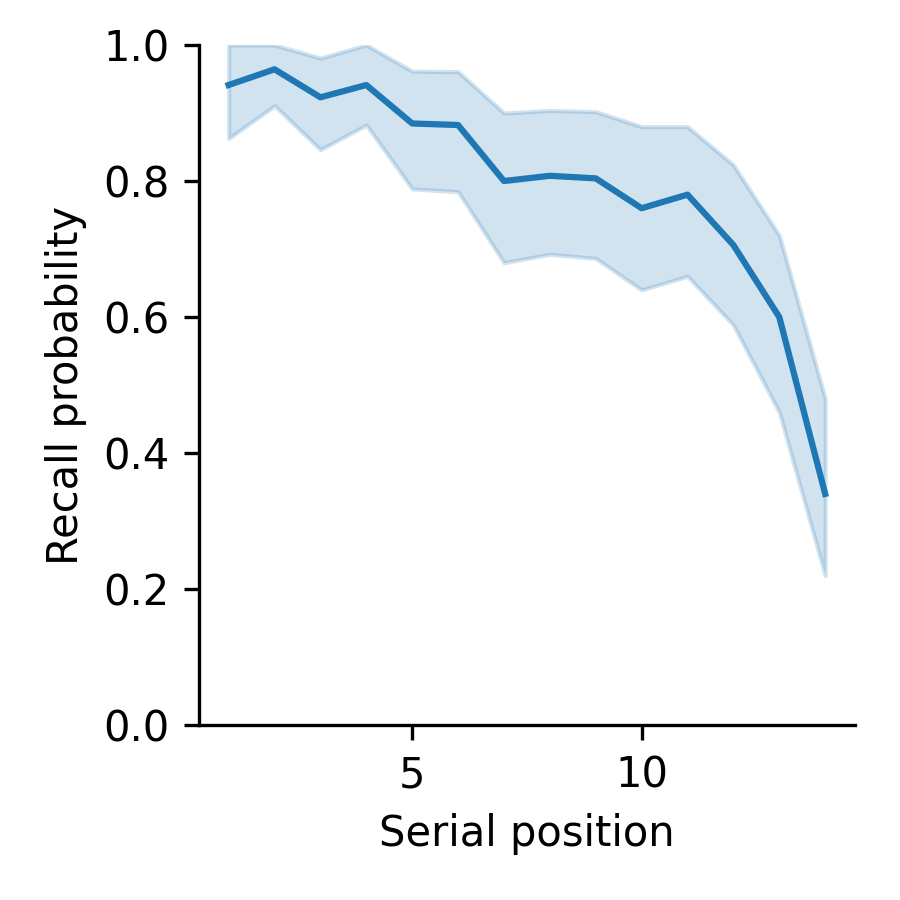} &
    \includegraphics[width=\plotfigurewidth]{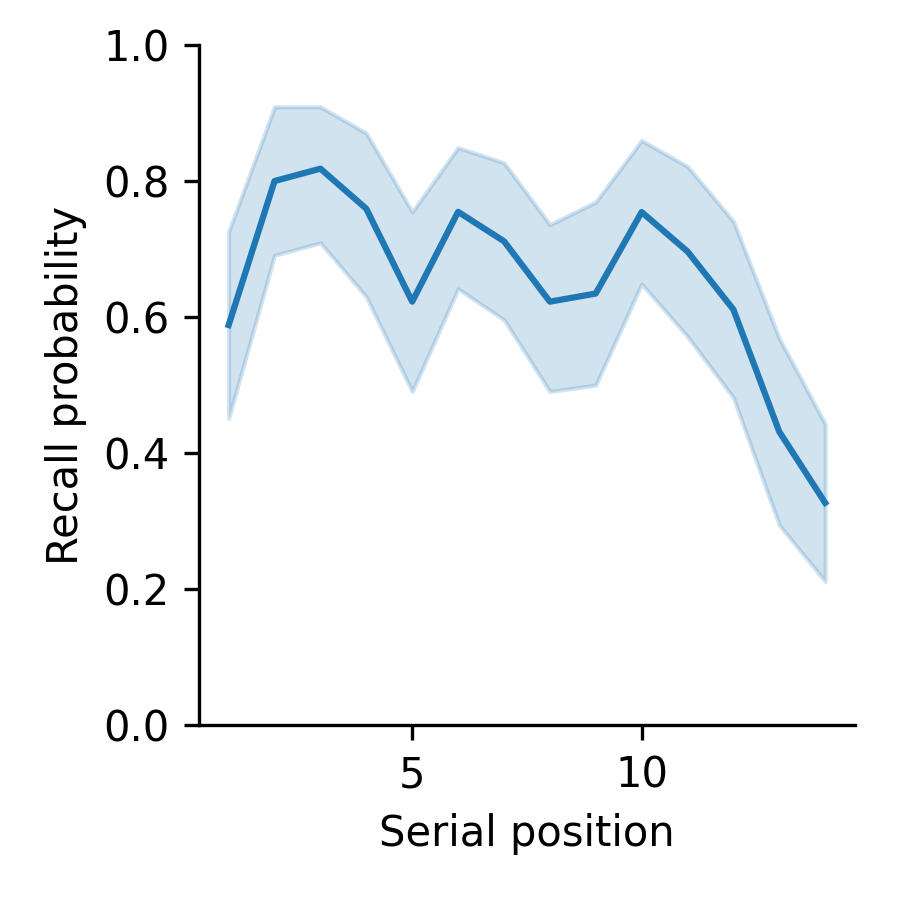} &
    \includegraphics[width=\plotfigurewidth]{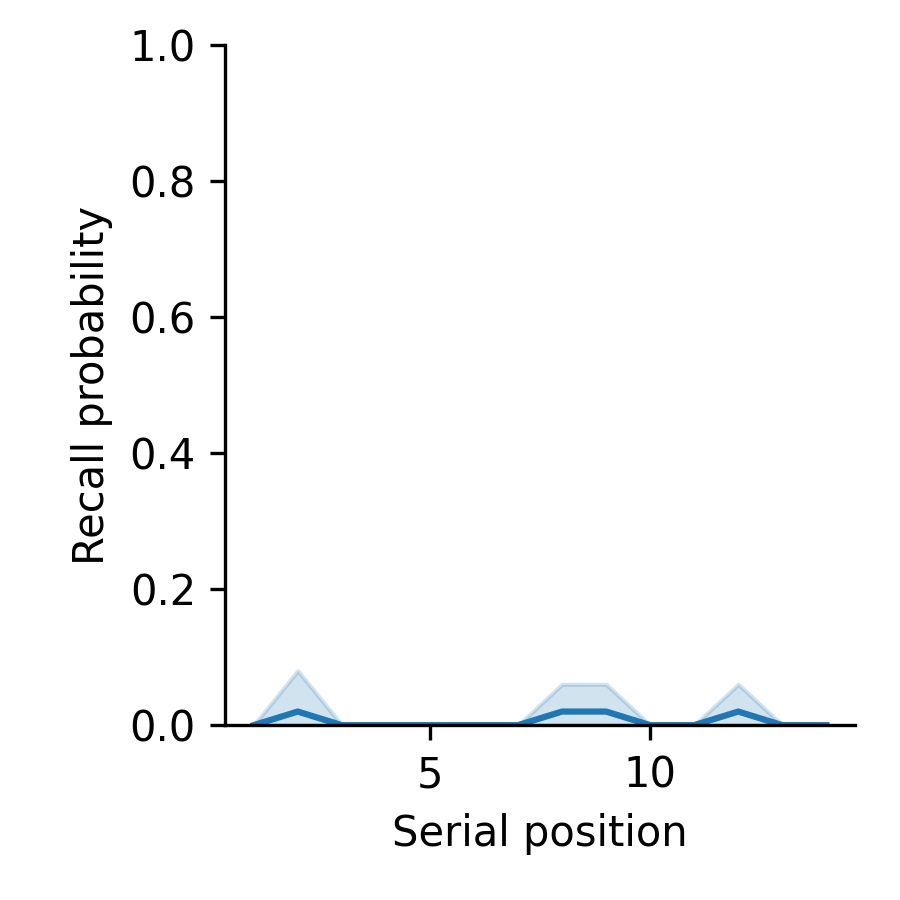} &
    \includegraphics[width=\plotfigurewidth]{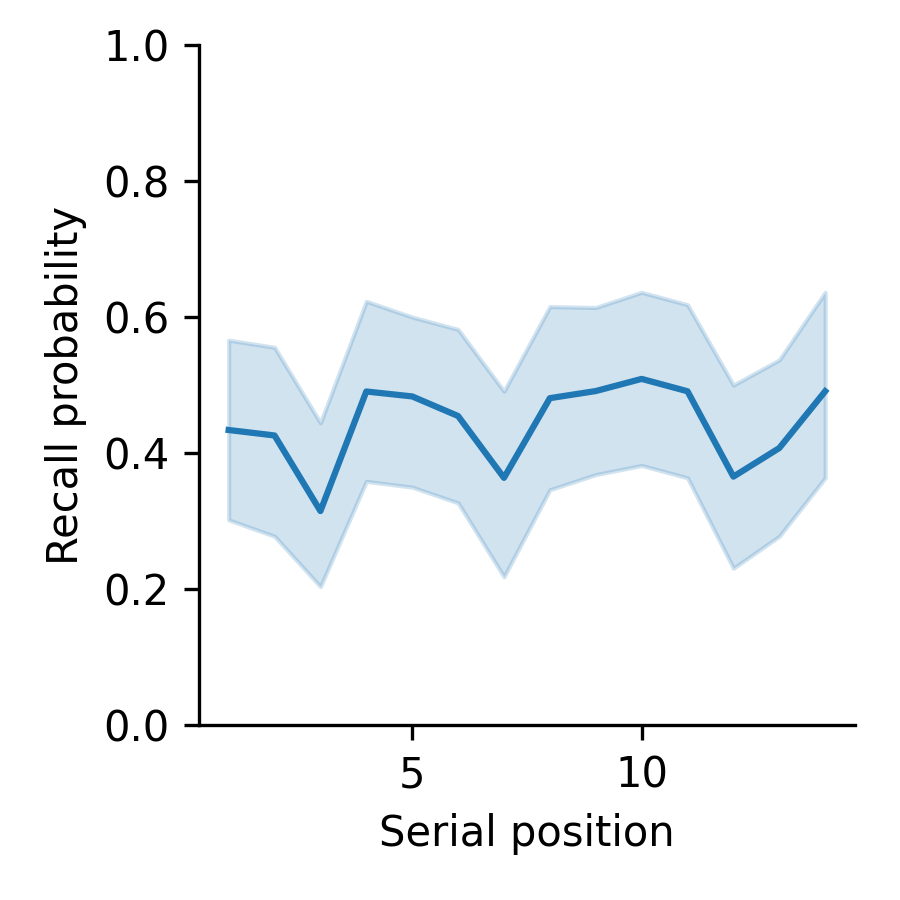} \\

    \rotatebox{90}{\parbox{2.2cm}{\centering\scriptsize\textbf{Qwen-7B}\\Rand. Abl.}} &
    \includegraphics[width=\plotfigurewidth]{Figures/CRP/Qwen2.5-7B_few_10_shot_no_ablation_14_50_spc.png} &
    \includegraphics[width=\plotfigurewidth]{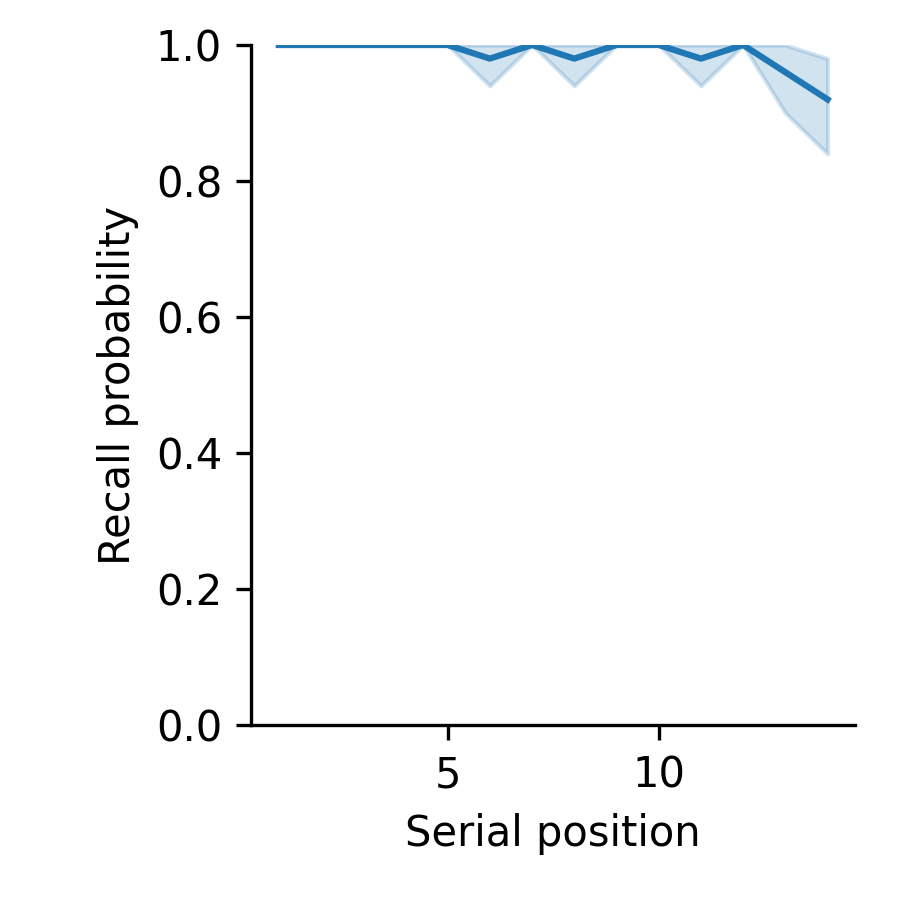} &
    \includegraphics[width=\plotfigurewidth]{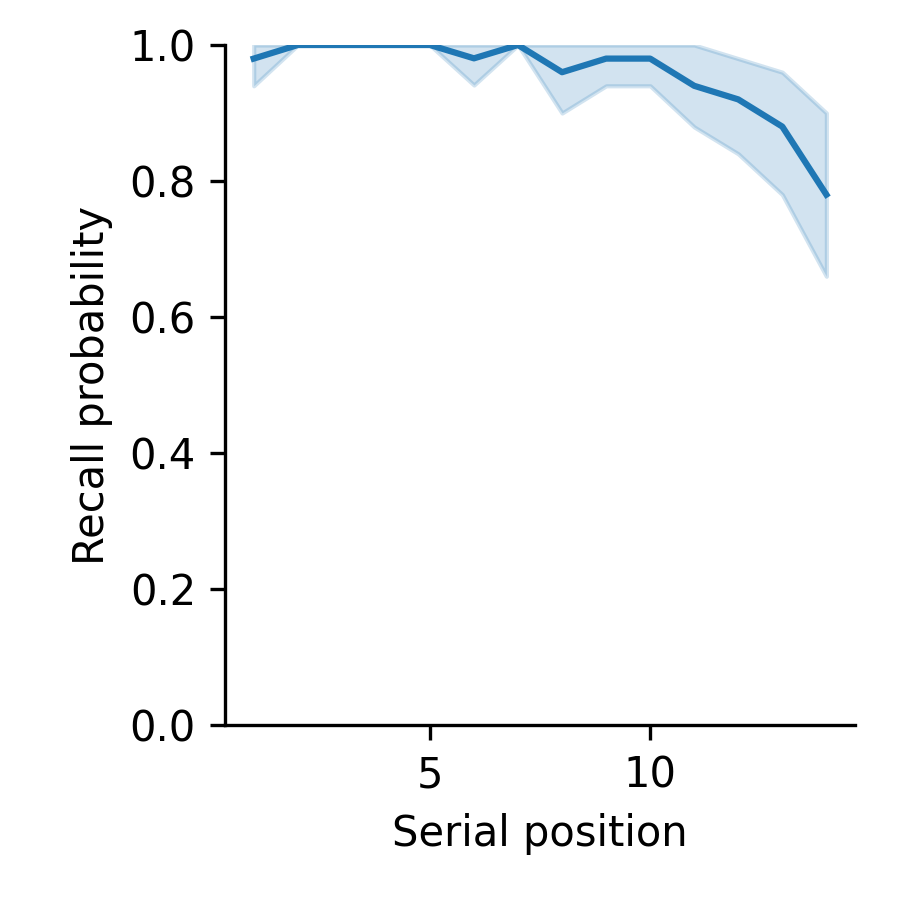} &
    \includegraphics[width=\plotfigurewidth]{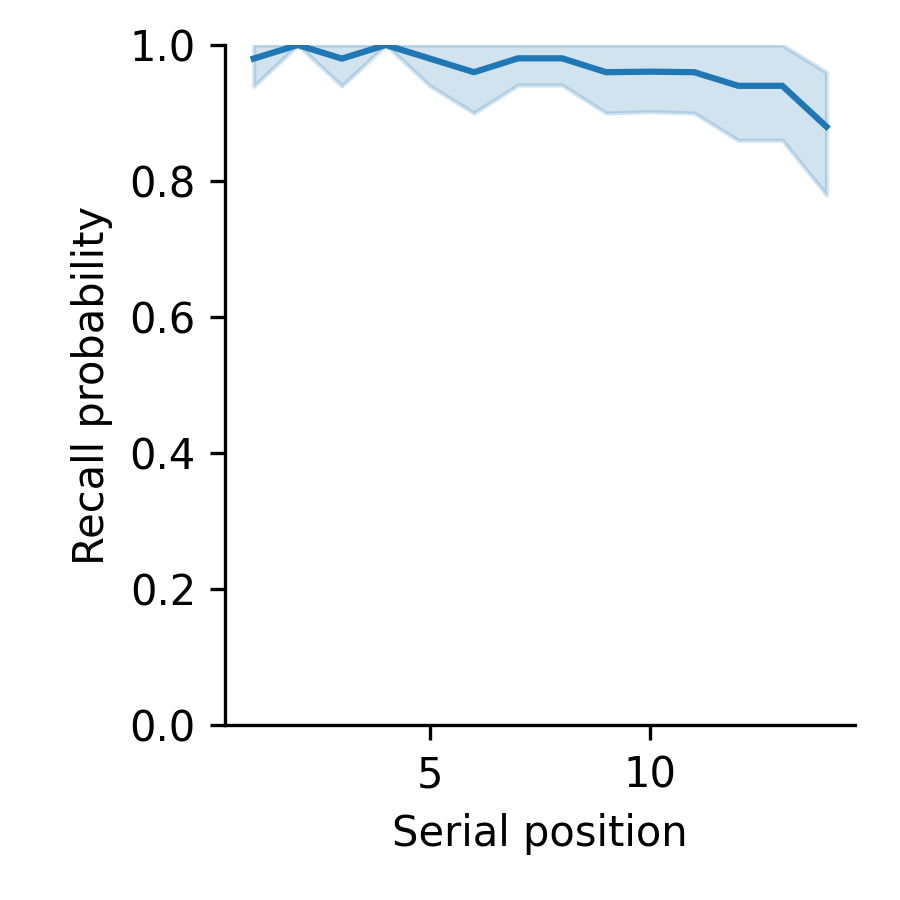} &
    \includegraphics[width=\plotfigurewidth]{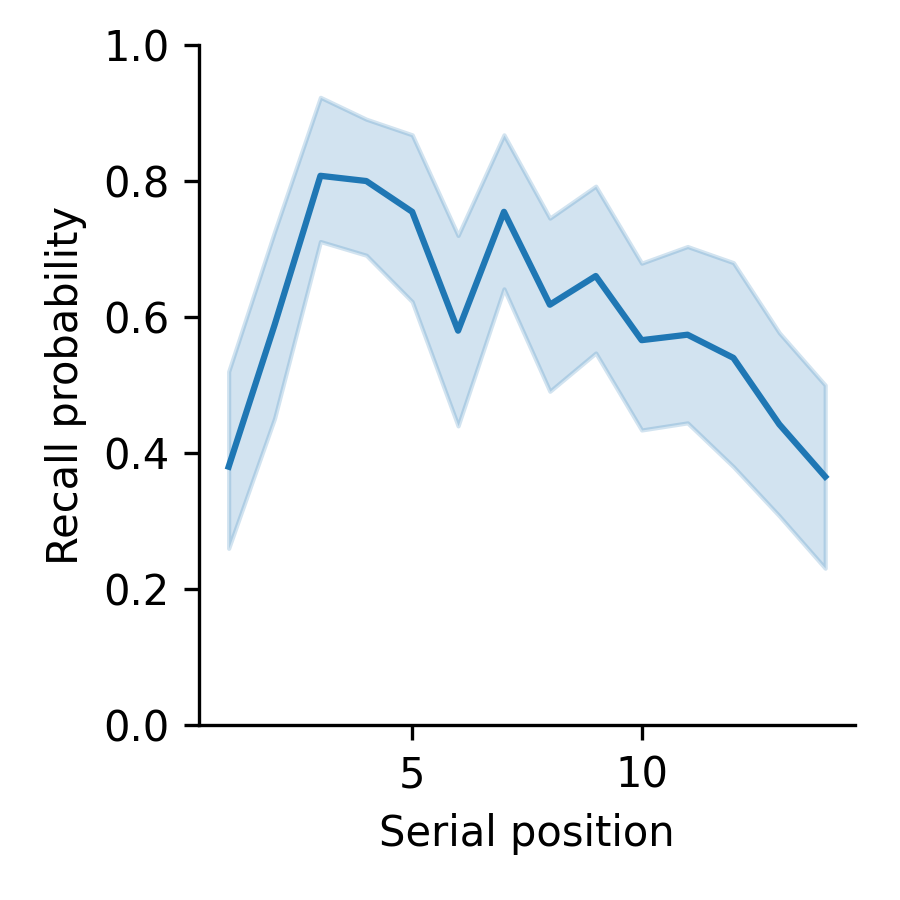} &
    \includegraphics[width=\plotfigurewidth]{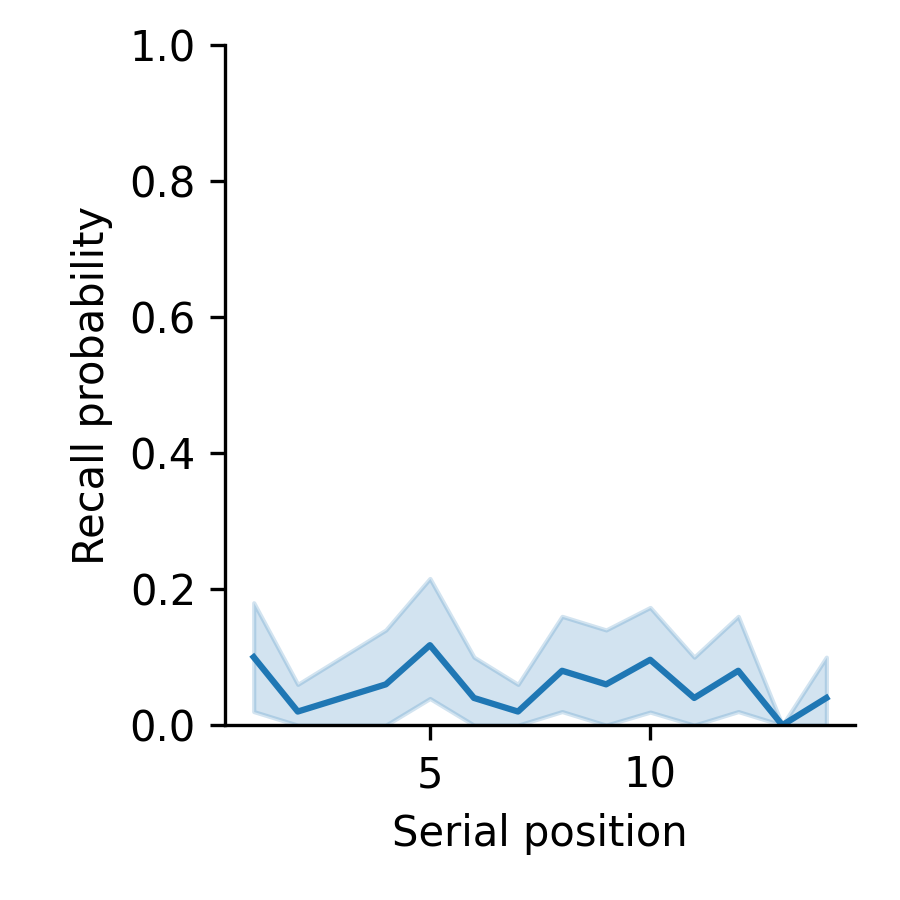} \\

    \rotatebox{90}{\parbox{2.2cm}{\centering\scriptsize\textbf{Qwen-7B-I}\\Ind. Abl.}} &
    \includegraphics[width=\plotfigurewidth]{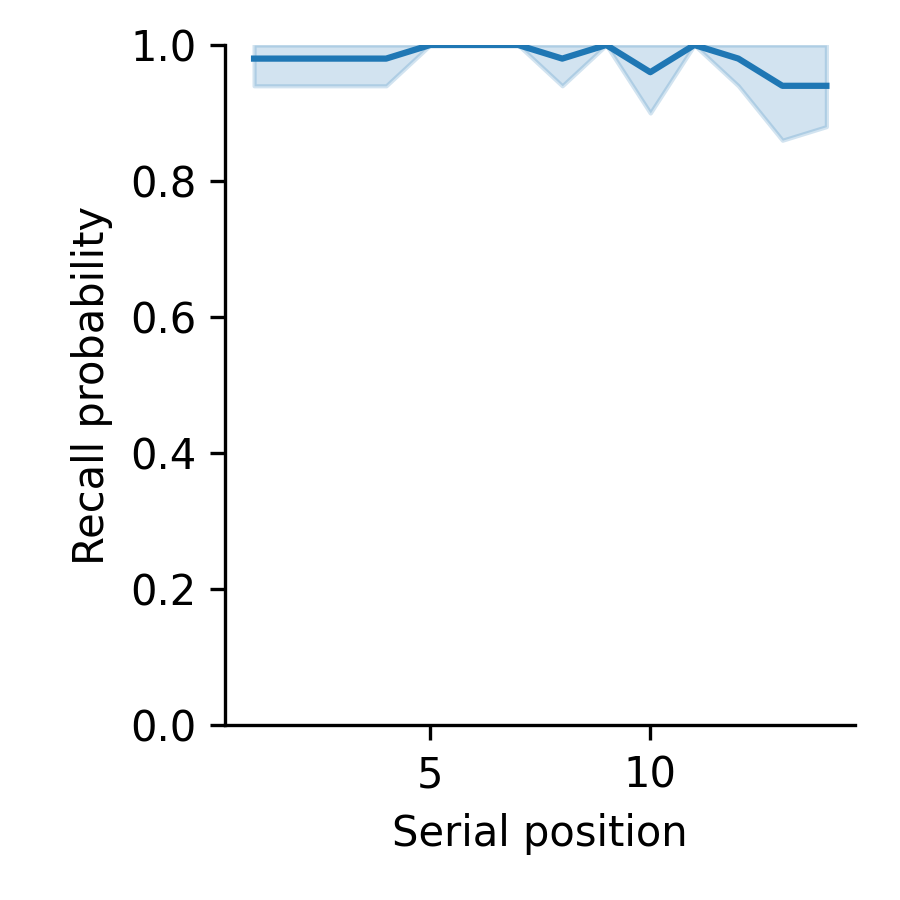} &
    \includegraphics[width=\plotfigurewidth]{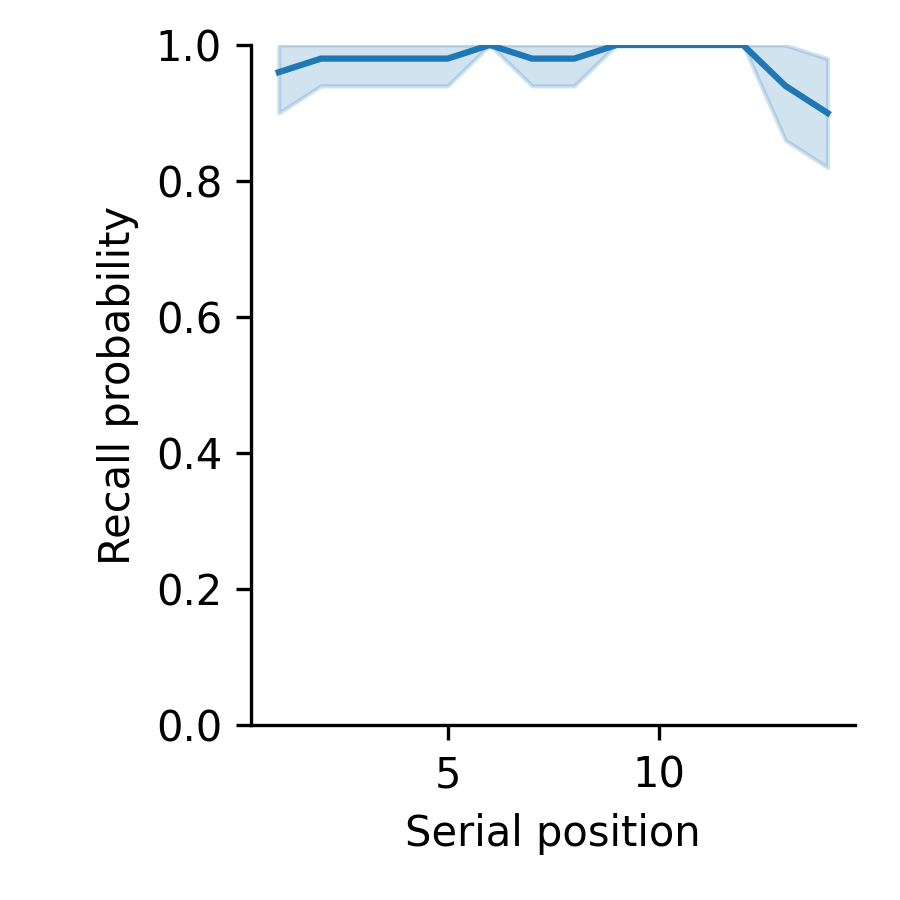} &
    \includegraphics[width=\plotfigurewidth]{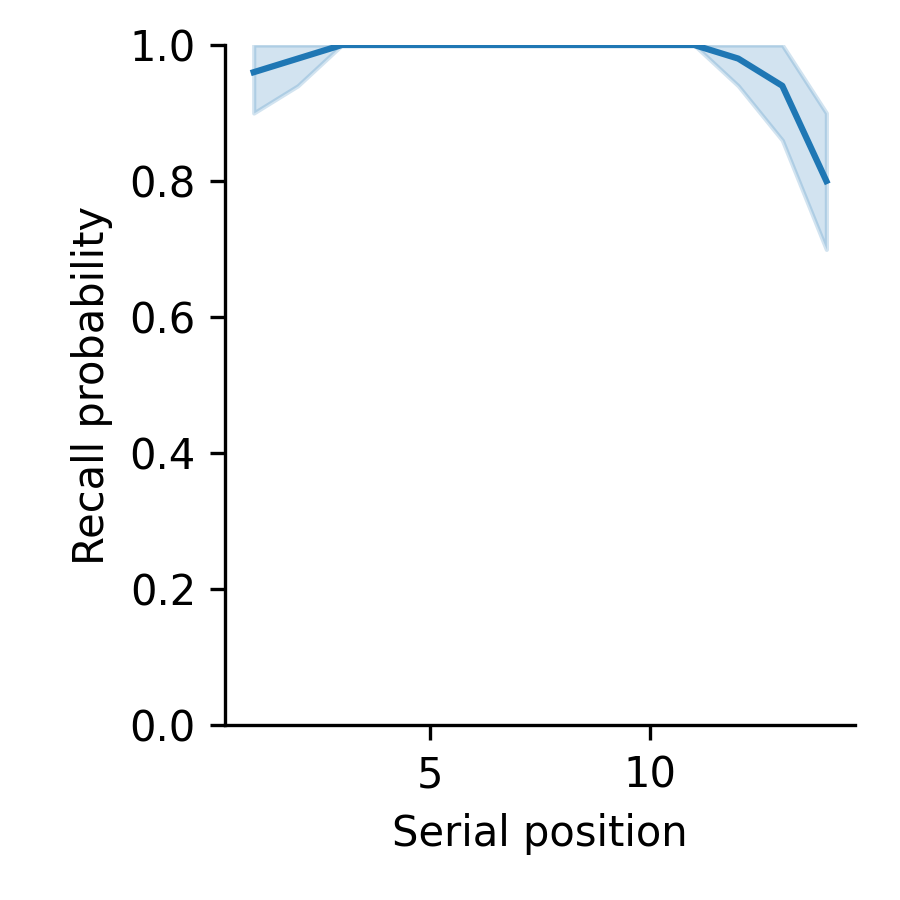} &
    \includegraphics[width=\plotfigurewidth]{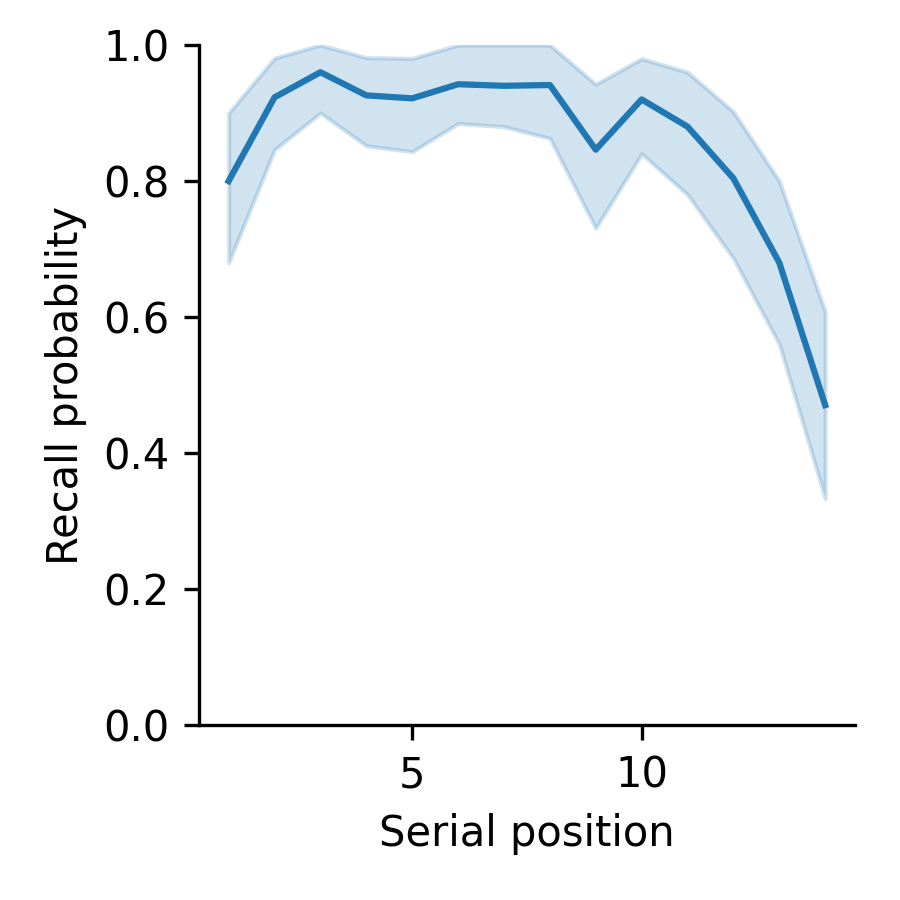} &
    \includegraphics[width=\plotfigurewidth]{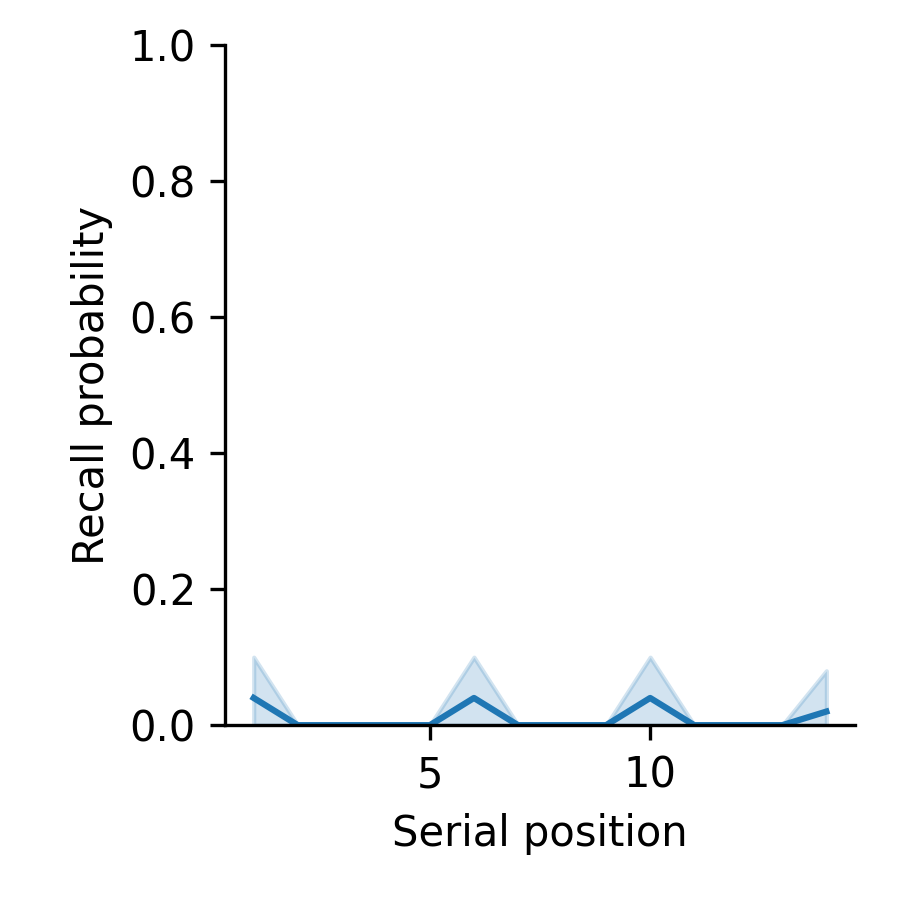} &
    \includegraphics[width=\plotfigurewidth]{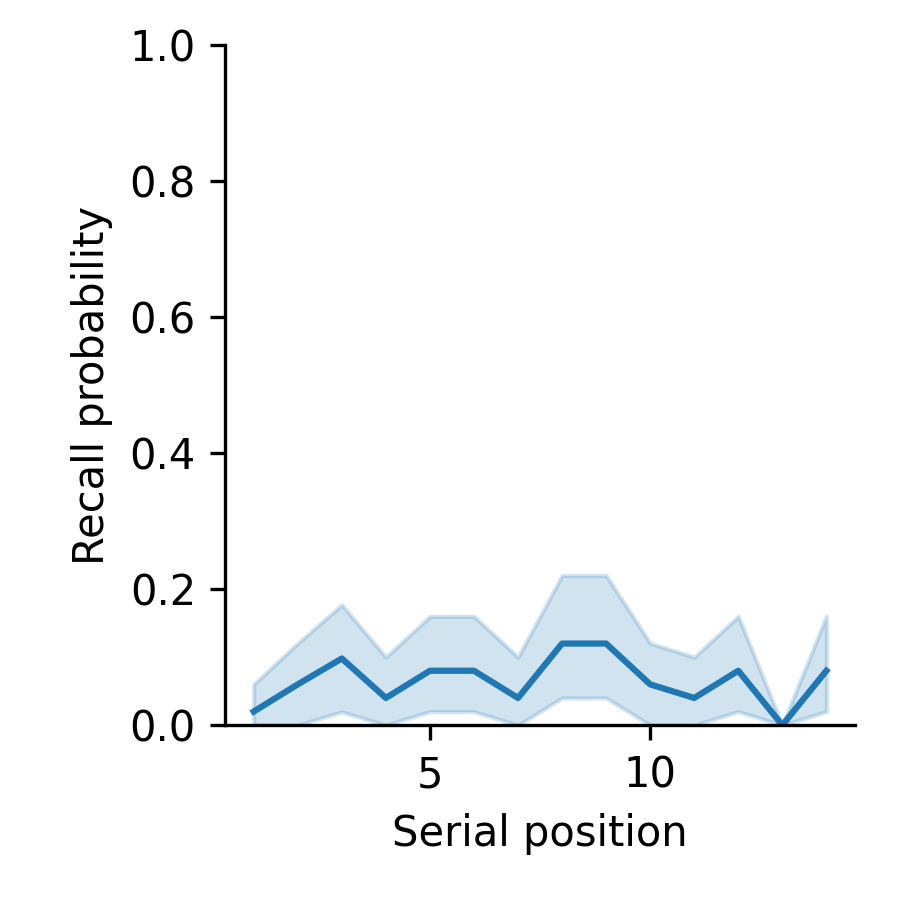} \\

    \rotatebox{90}{\parbox{2.2cm}{\centering\scriptsize\textbf{Qwen-7B-I}\\Rand. Abl.}} &
    \includegraphics[width=\plotfigurewidth]{Figures/CRP/Qwen2.5-7B-Instruct_few_10_shot_no_ablation_14_50_spc.png} &
    \includegraphics[width=\plotfigurewidth]{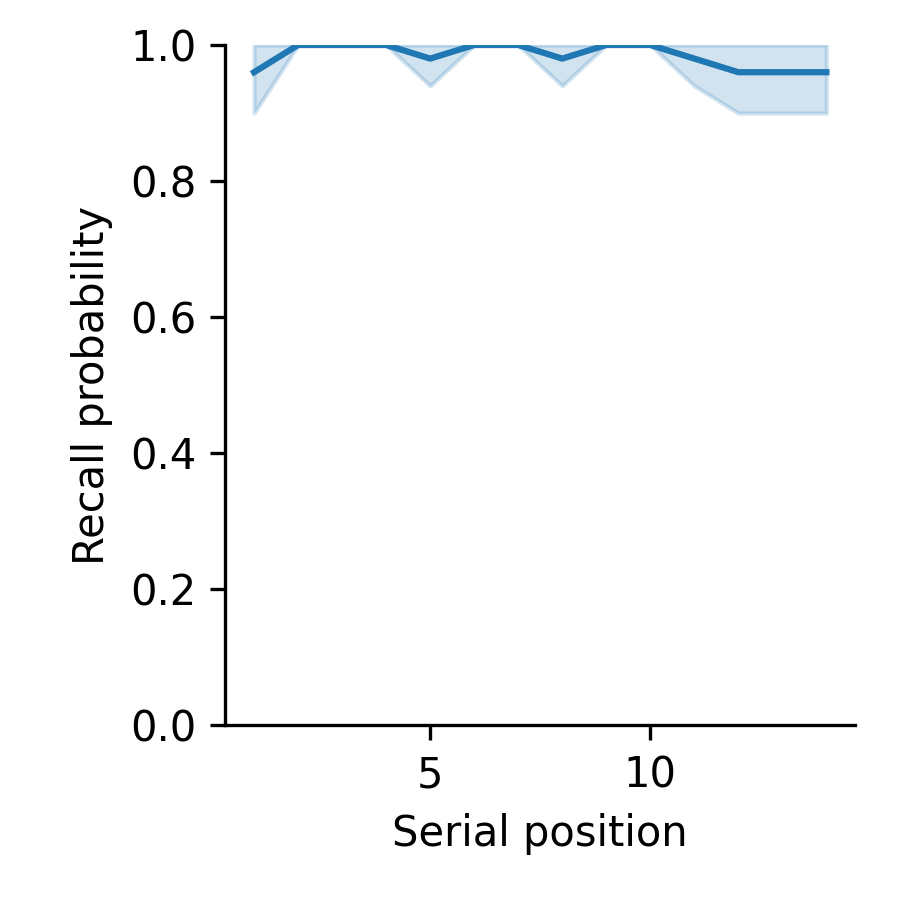} &
    \includegraphics[width=\plotfigurewidth]{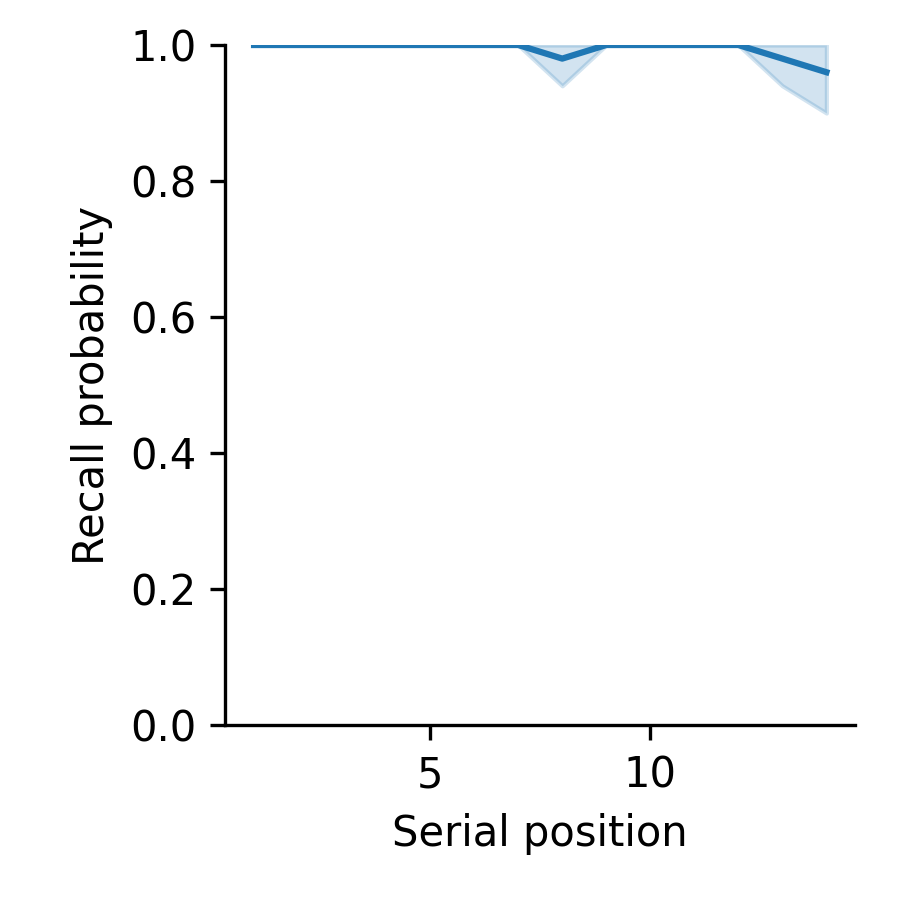} &
    \includegraphics[width=\plotfigurewidth]{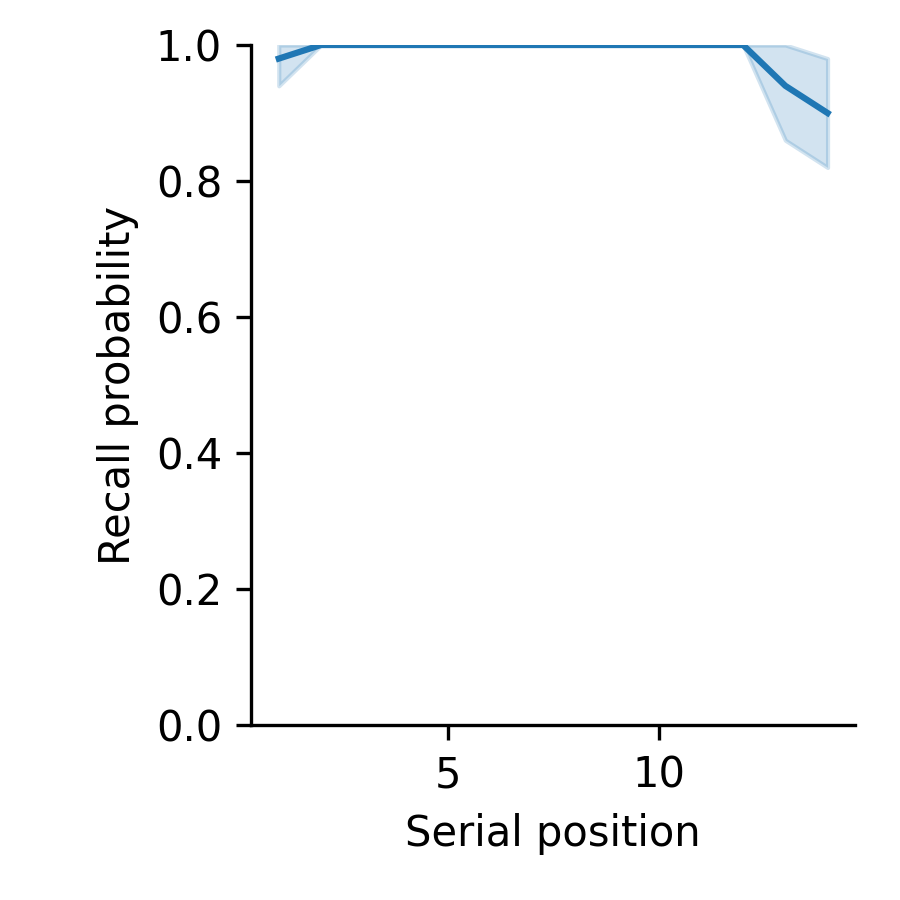} &
    \includegraphics[width=\plotfigurewidth]{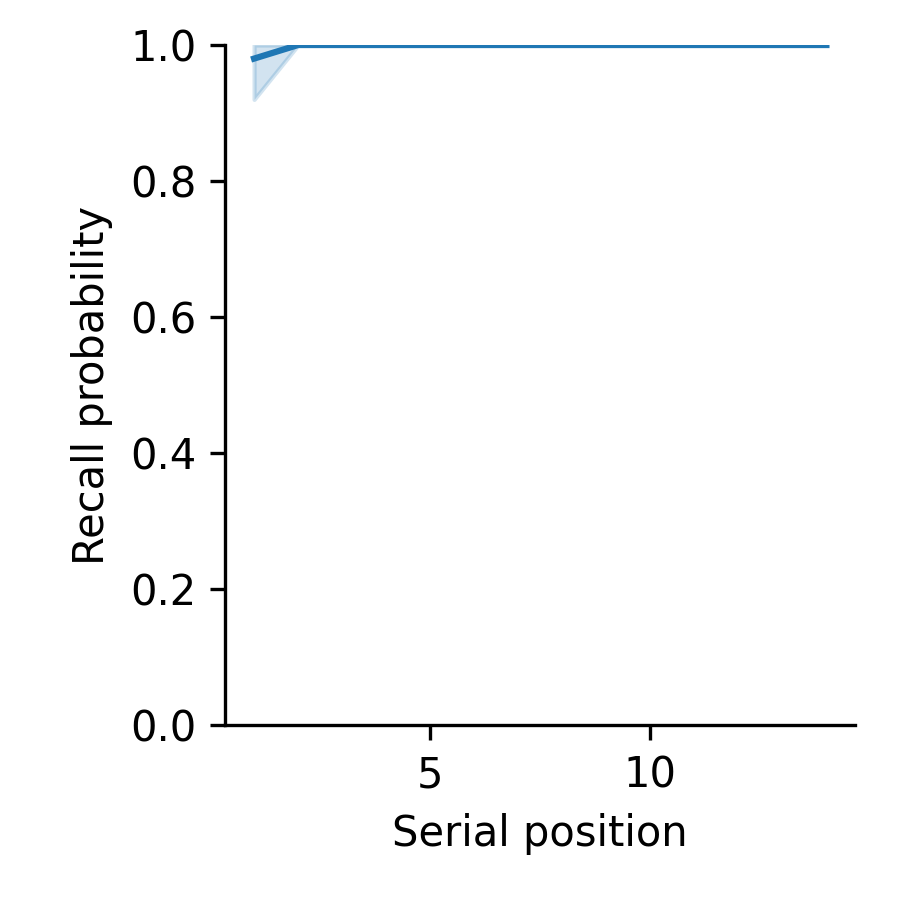} &
    \includegraphics[width=\plotfigurewidth]{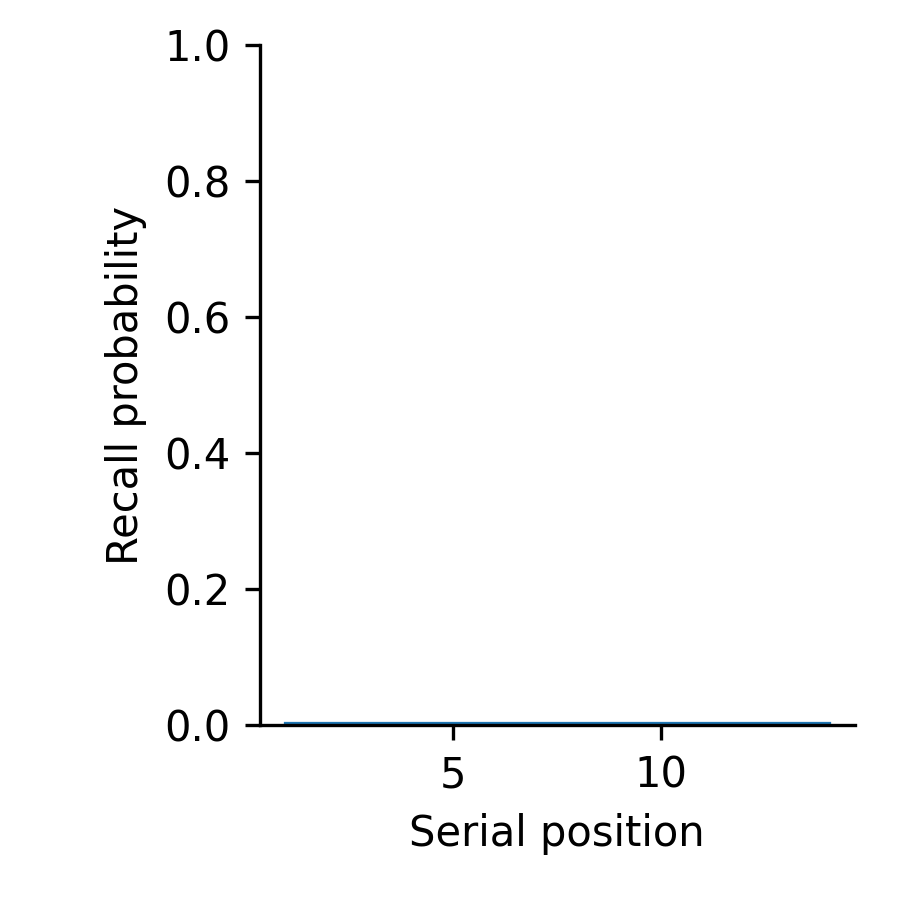} \\
    \end{tabular}

    \caption{Probability of recall for items at different serial positions in serial recall ICL task for different models and ablation conditions (serial position 1 is the first item in the list, serial position 14 is the last item in the list). }
    \label{fig:SPC_ablation_grid}
\end{figure*}

\begin{table*}[h]
    \centering
    \caption{Conditional recall probability at lag +1 illustrating the impact of induction and random head ablation on serial recall performance (mean recall probability $\pm$ standard deviation) in an ICL serial recall task. Results are shown for Llama-3.1-8B and Qwen2.5-7B models, both base and instruction-tuned versions, across varying numbers of ablated heads. These results are visualized in Fig.~\ref{fig:ICL_serial_recall}.}
    {
    \setlength{\tabcolsep}{5pt}
    \renewcommand{\arraystretch}{1.1}
    \begin{tabular}{@{}llcccccc@{}}
        \toprule
        \textbf{Model} & \textbf{Abl.} & \textbf{0 Heads} & \textbf{1 Head} & \textbf{10 Heads} & \textbf{25 Heads} & \textbf{50 Heads} & \textbf{100 Heads} \\
        \midrule
        Llama-I & Ind. & $0.98\pm0.14$ & $0.99\pm0.03$ & $0.97\pm0.15$ & $0.28\pm0.40$ & $0.28\pm0.35$ & $0.12\pm0.28$ \\
                & Rand.& $0.98\pm0.14$ & $0.98\pm0.14$ & $0.92\pm0.26$ & $0.74\pm0.43$ & $0.90\pm0.27$ & $0.21\pm0.28$ \\
        \midrule
        Qwen-I  & Ind. & $0.97\pm0.16$ & $0.99\pm0.07$ & $1.00\pm0.00$ & $0.91\pm0.22$ & $0.00\pm0.00$ & $0.25\pm0.50$ \\
                & Rand.& $0.97\pm0.16$ & $0.98\pm0.14$ & $1.00\pm0.01$ & $0.99\pm0.03$ & $1.00\pm0.00$ & $0.00\pm0.00$ \\
        \midrule
        Llama   & Ind. & $1.00\pm0.01$ & $1.00\pm0.01$ & $0.75\pm0.38$ & $0.54\pm0.46$ & $0.64\pm0.46$ & $0.10\pm0.22$ \\
                & Rand.& $1.00\pm0.01$ & $1.00\pm0.02$ & $0.95\pm0.20$ & $0.46\pm0.44$ & $0.80\pm0.30$ & $0.50\pm0.42$ \\
        \midrule
        Qwen    & Ind. & $0.97\pm0.17$ & $0.97\pm0.15$ & $0.80\pm0.38$ & $0.57\pm0.42$ & $0.00\pm0.00$ & $0.08\pm0.17$ \\
                & Rand.& $0.97\pm0.17$ & $0.98\pm0.12$ & $0.95\pm0.17$ & $0.95\pm0.20$ & $0.60\pm0.37$ & $0.17\pm0.29$ \\
        \bottomrule
    \end{tabular}
    }
    \label{tab:ICL_serial_recall}
\end{table*}

\begin{figure*}[h!]
    \centering
    \begin{tabular}{lllll}
    \textbf{A} &
    \textbf{B} &
    \textbf{C} &
    \textbf{D} \\
        {\includegraphics[width=0.22\textwidth]{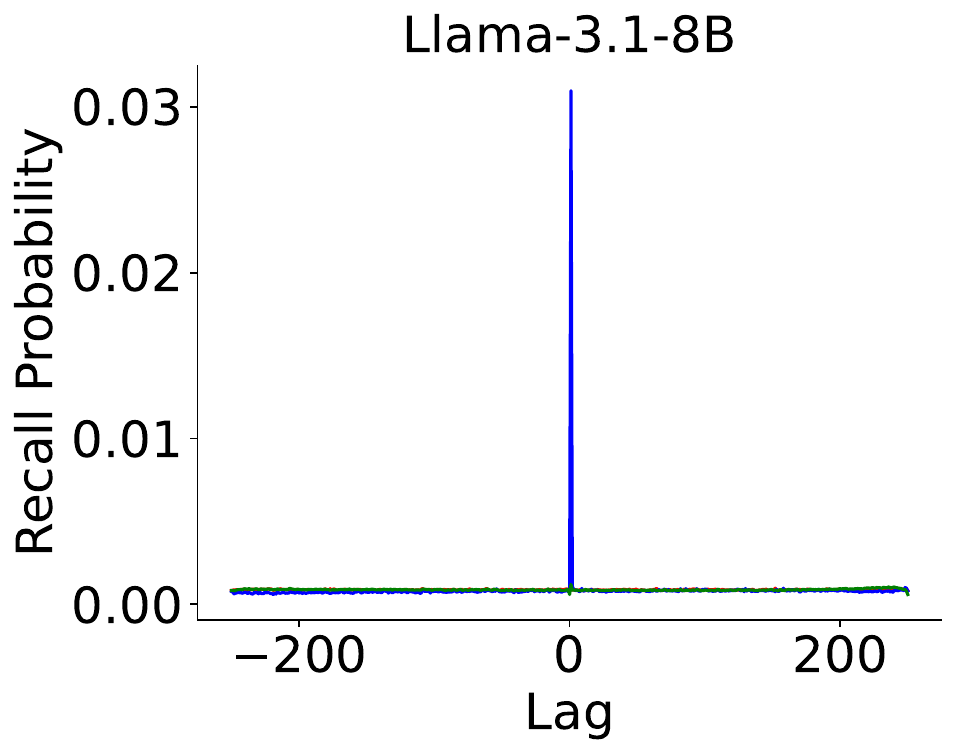
        }} &
        {\includegraphics[width=0.22\textwidth]{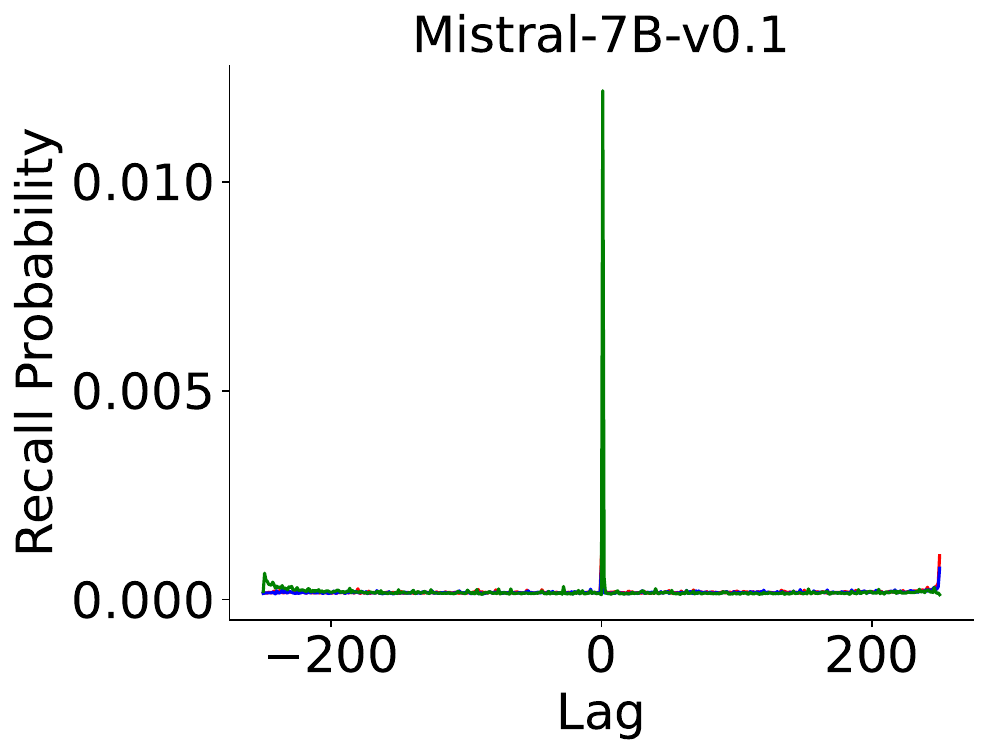}} &
        {\includegraphics[width=0.22\textwidth]{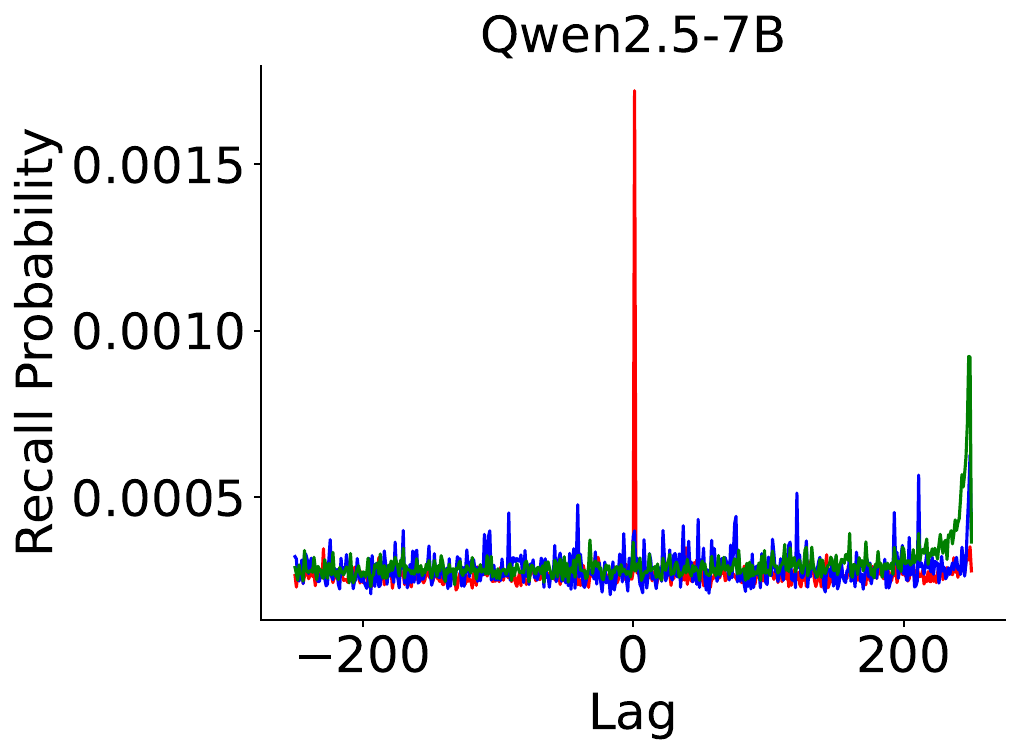}} &
        {\includegraphics[width=0.22\textwidth]{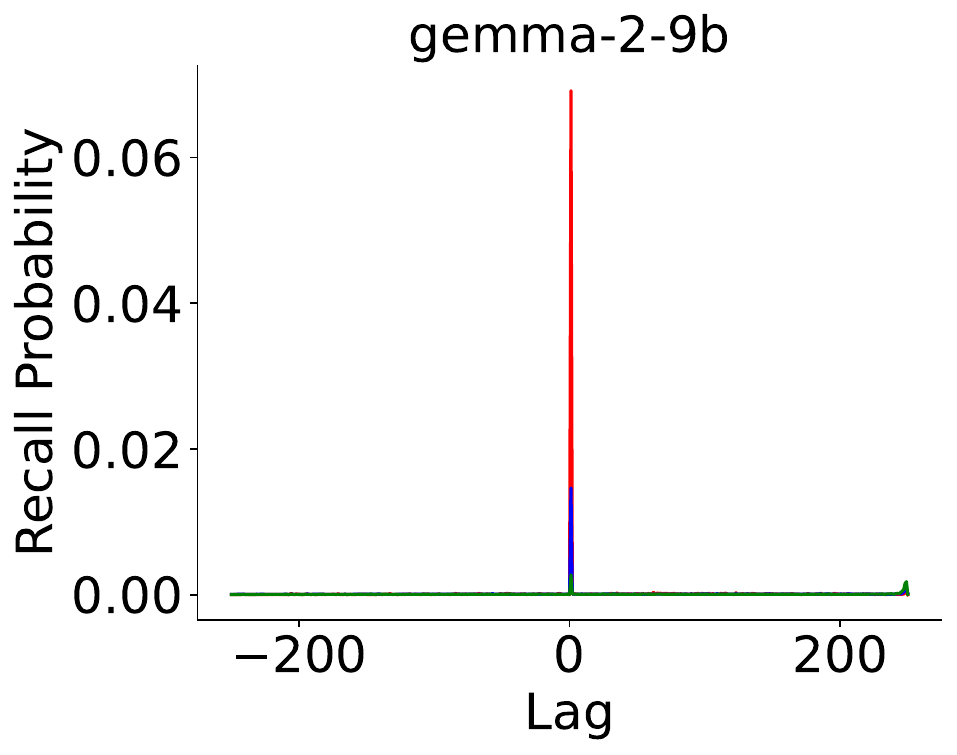}}
        \\
    \end{tabular}
    \begin{tabular}{lllll}
    \textbf{E} &
    \textbf{F} &
    \textbf{G} &
    \textbf{H} \\
        {\includegraphics[width=0.22\textwidth]{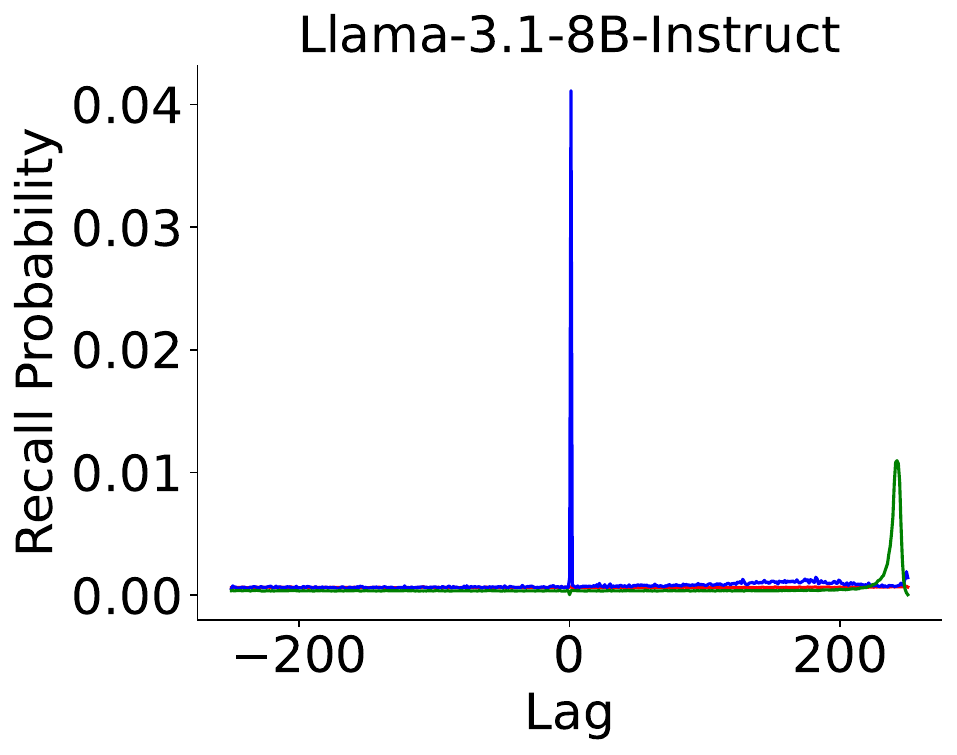
        }} &
        {\includegraphics[width=0.22\textwidth]{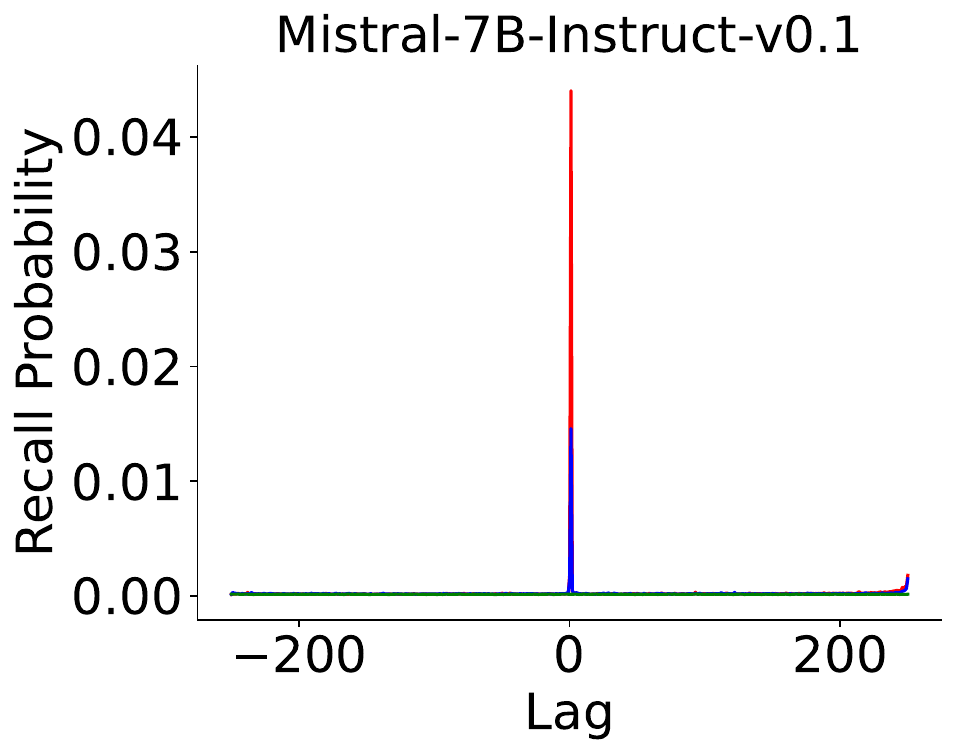}} &
        {\includegraphics[width=0.22\textwidth]{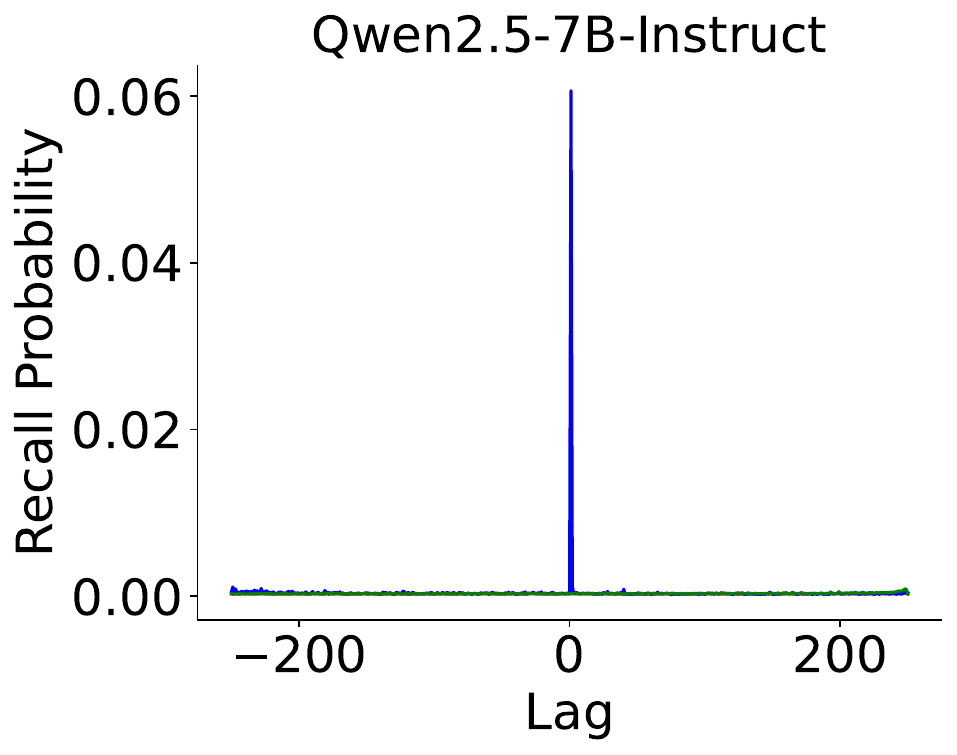}} &
        {\includegraphics[width=0.22\textwidth]{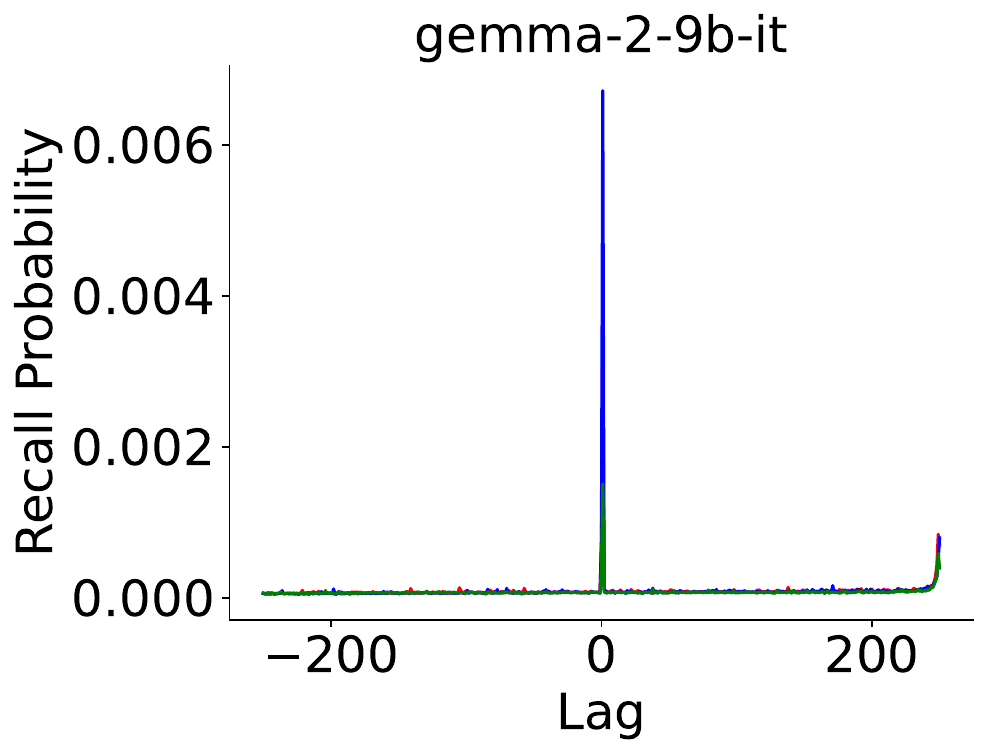}} \\
    \end{tabular}
    \includegraphics[scale=0.5]{Figures/legend.pdf}
    \caption{Impact of induction and random head mean ablation (100 heads in each case) on the model output probability as a function of lag. The models were presented with a sequence of 501 tokens where the last token repeated the token at index 250 and the lag is defined relative to the repeated token (see Methods for more details and Fig.~\ref{fig:CRP_6_mean} for a zoomed version showing lags -6 to 6). The results show averages across 5000 runs with shuffled token sequences.
    Top row: Base models. Bottom row: Instruction-tuned models.
    \label{fig:CRP_250_mean}}
\end{figure*}
\begin{figure*}[h!]
    \centering
    \begin{tabular}{lllll}
    \textbf{A} &
    \textbf{B} &
    \textbf{C} &
    \textbf{D} \\
        {\includegraphics[width=0.22\textwidth]{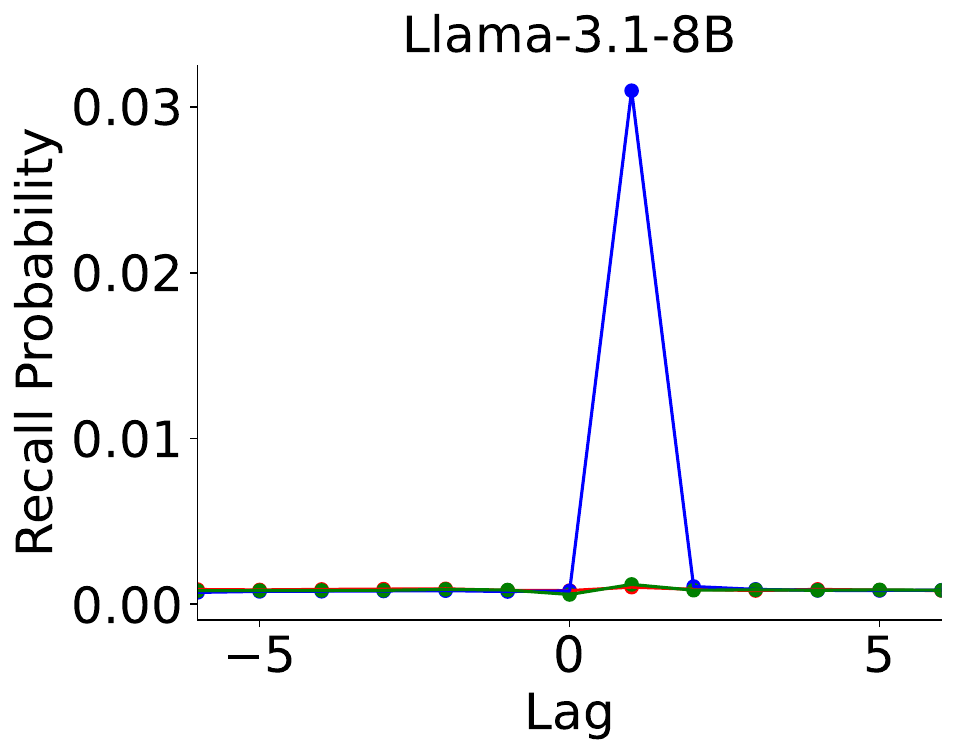
        }} &
        {\includegraphics[width=0.22\textwidth]{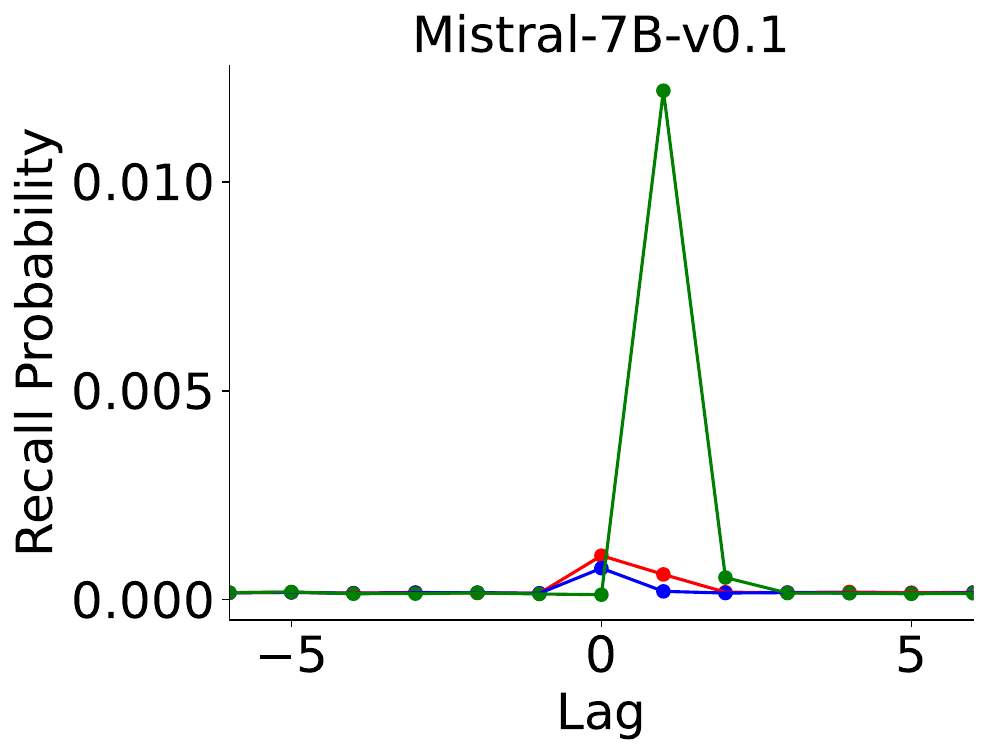}} &
        {\includegraphics[width=0.22\textwidth]{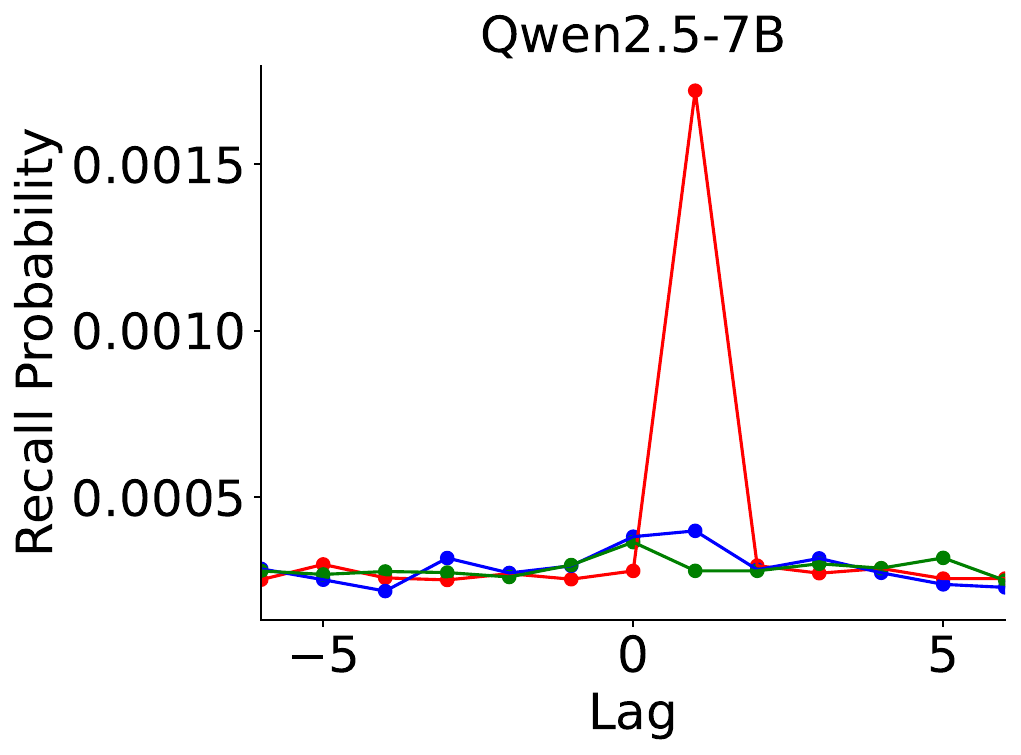}} &
        {\includegraphics[width=0.22\textwidth]{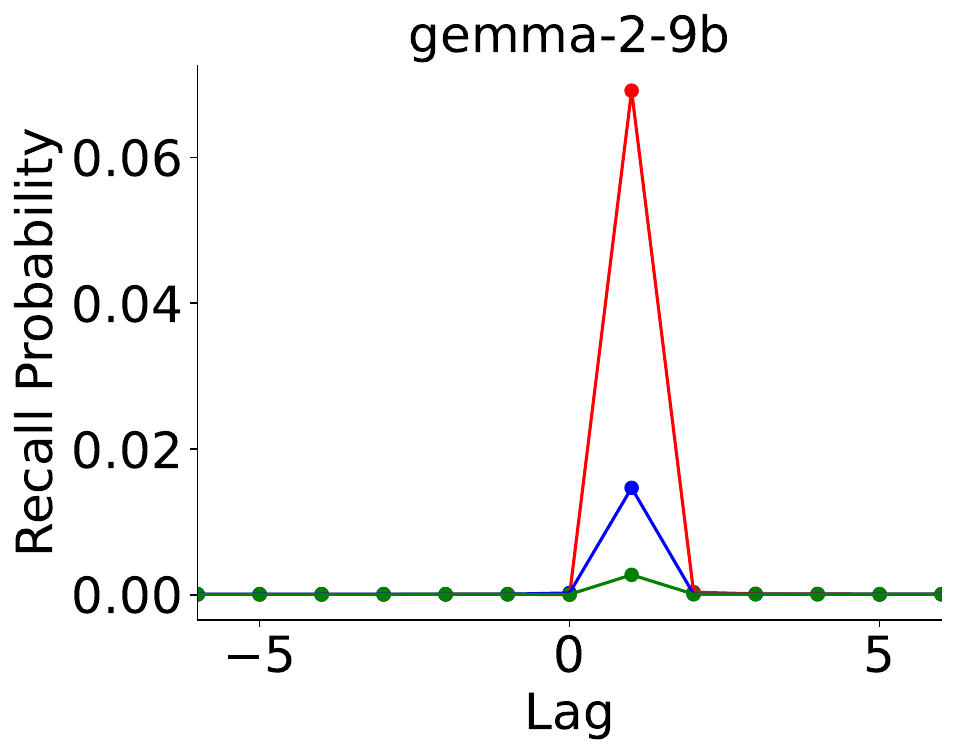}}
        \\
    \end{tabular}
    \begin{tabular}{lllll}
    \textbf{E} &
    \textbf{F} &
    \textbf{G} &
    \textbf{H} \\
        {\includegraphics[width=0.22\textwidth]{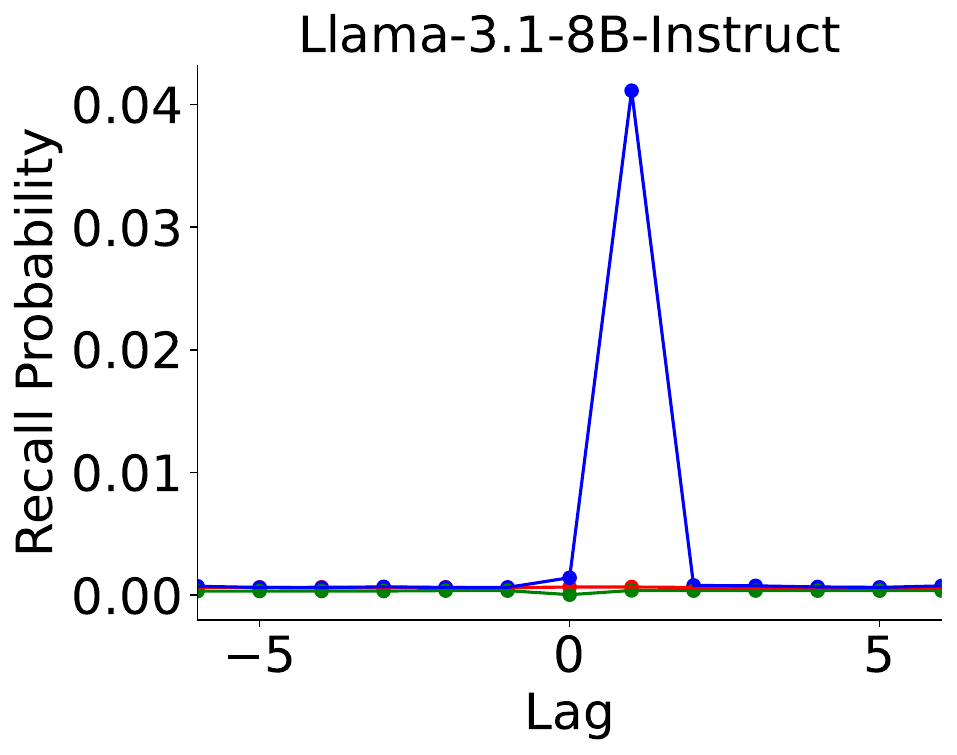
        }} &
        {\includegraphics[width=0.22\textwidth]{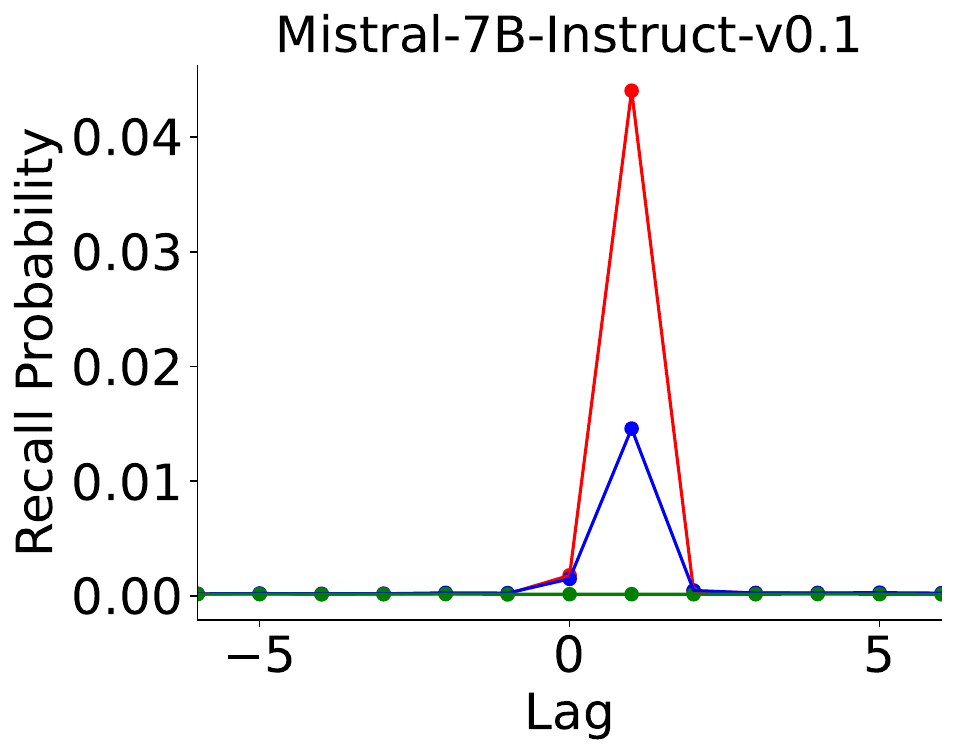}} &
        {\includegraphics[width=0.22\textwidth]{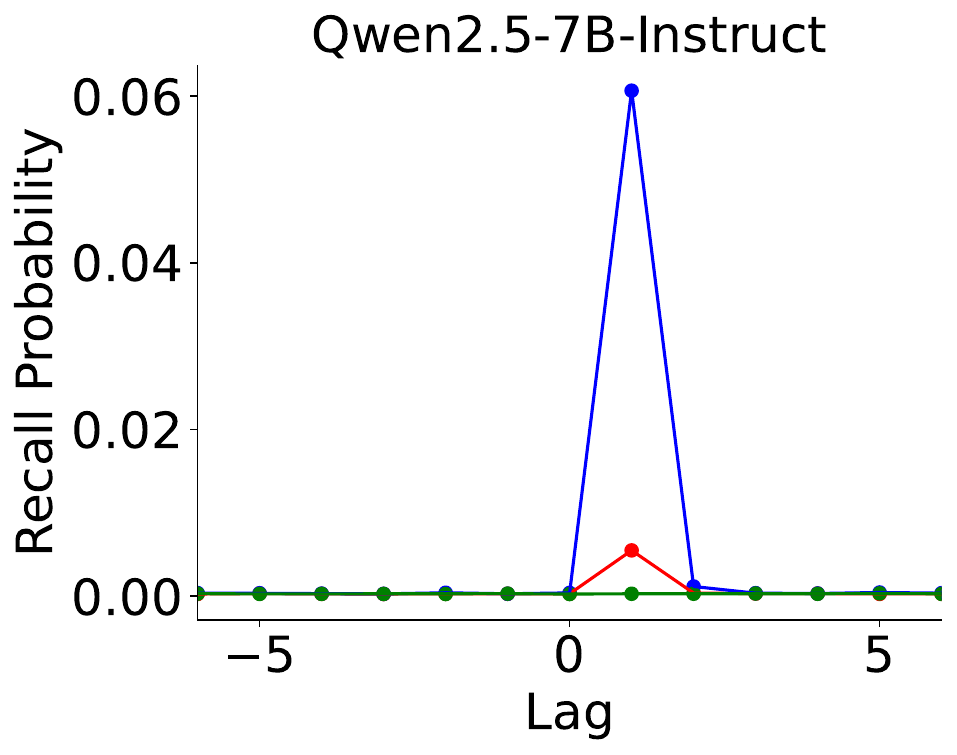}} &
        {\includegraphics[width=0.22\textwidth]{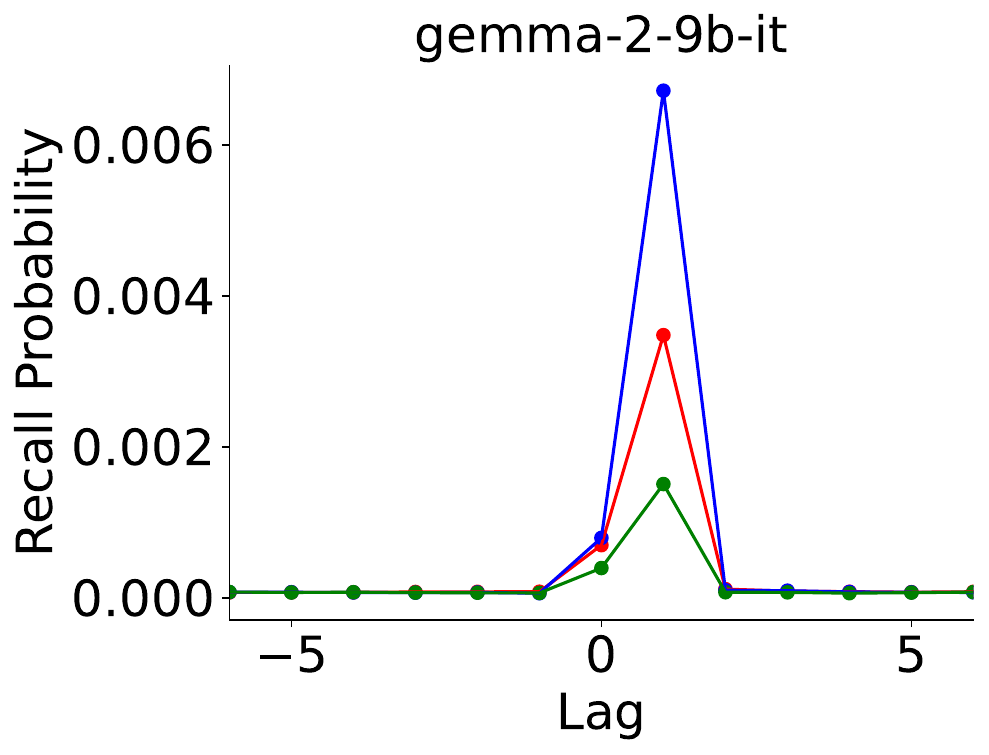}} \\
    \end{tabular}
    \includegraphics[scale=0.5]{Figures/legend.pdf}
    \caption{Impact of induction and random head mean ablation (100 heads in each case) on the model output probability as a function of lag: same as Fig.~\ref{fig:CRP_250_mean}, but with zoom on lags -6 to 6 to emphasize that the highest probabilities were at lags 1 or 0. Top row: Base models. Bottom row: Instruction-tuned models.
    \label{fig:CRP_6_mean}}
\end{figure*}

\begin{figure*}[h!]
    \centering
    \begin{tabular}{lllll}
    \textbf{A} &
    \textbf{B} &
    \textbf{C} &
    \textbf{D} \\
        {\includegraphics[width=0.22\textwidth]{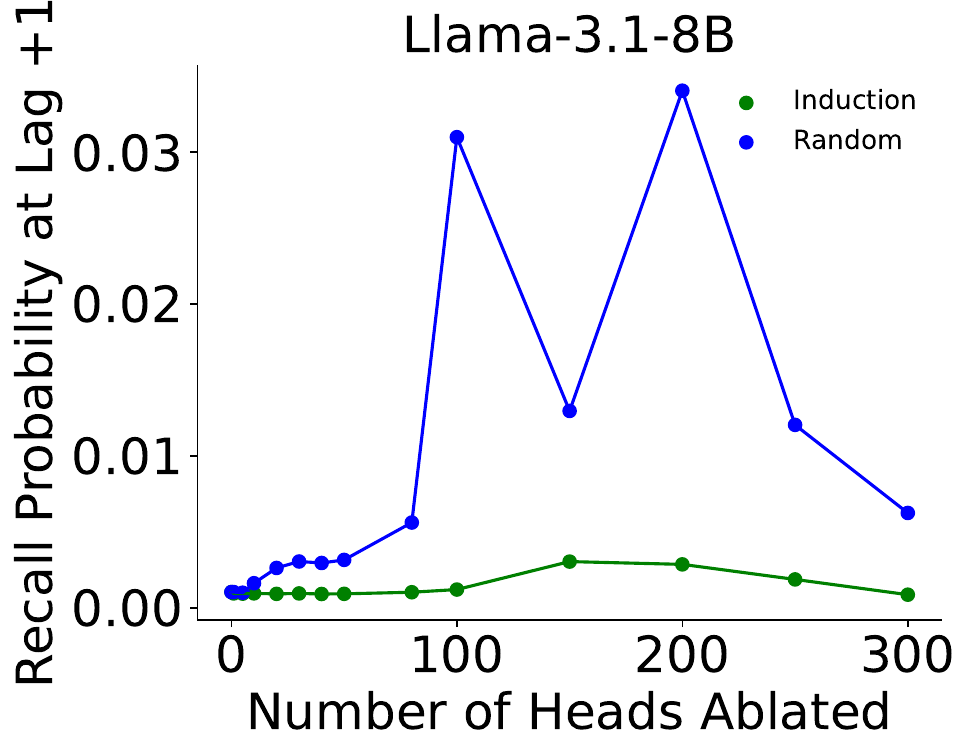
        }} &
        {\includegraphics[width=0.22\textwidth]{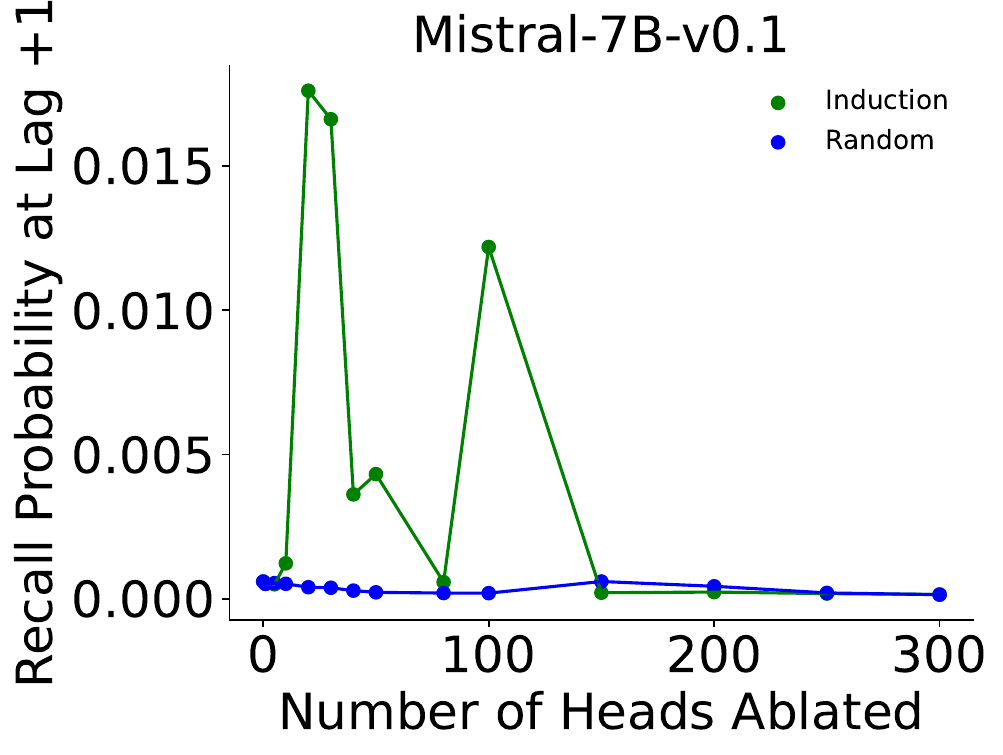}} &
        {\includegraphics[width=0.22\textwidth]{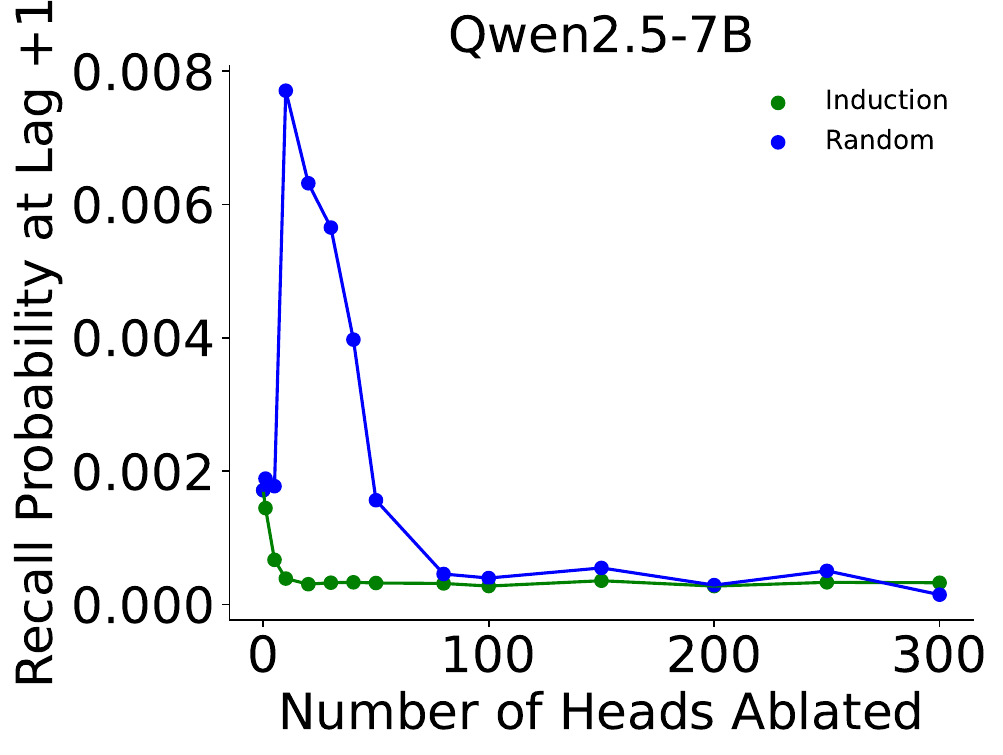}} &
        {\includegraphics[width=0.22\textwidth]{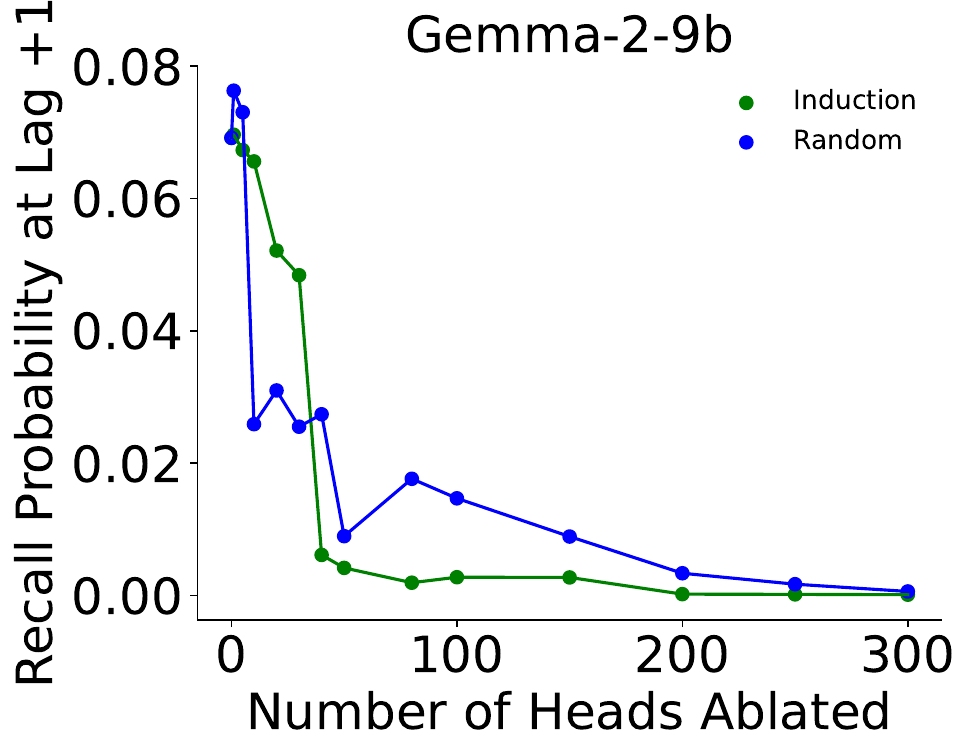}} \\
    \end{tabular}
    \begin{tabular}{lllll}
    \textbf{E} &
    \textbf{F} &
    \textbf{G} &
    \textbf{H} \\
         {\includegraphics[width=0.22\textwidth]{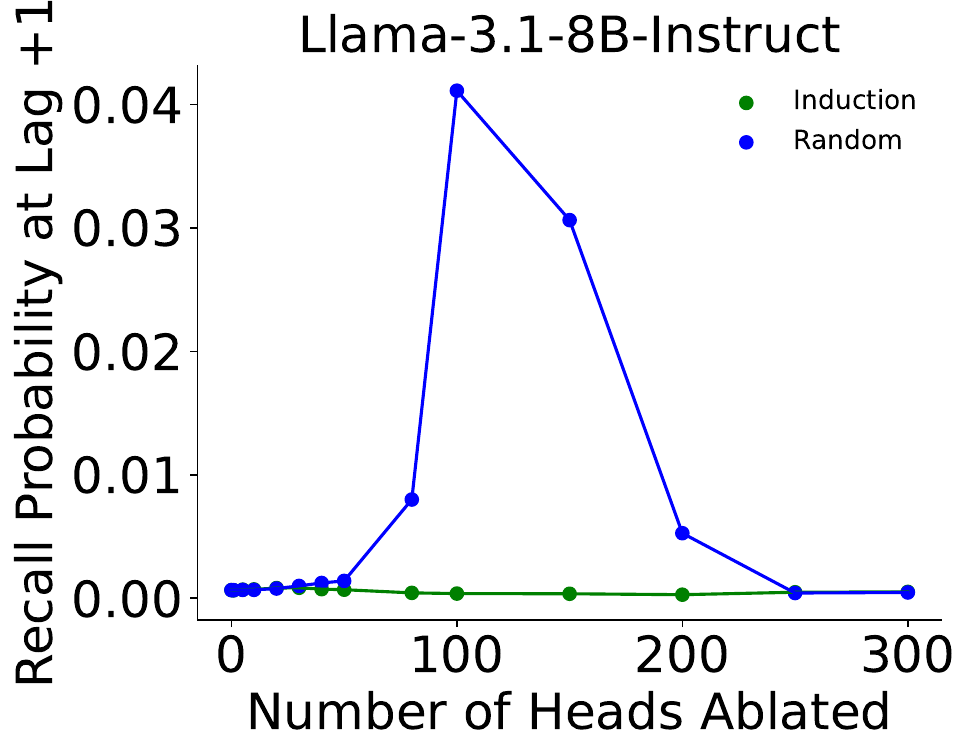
        }} &
        {\includegraphics[width=0.22\textwidth]{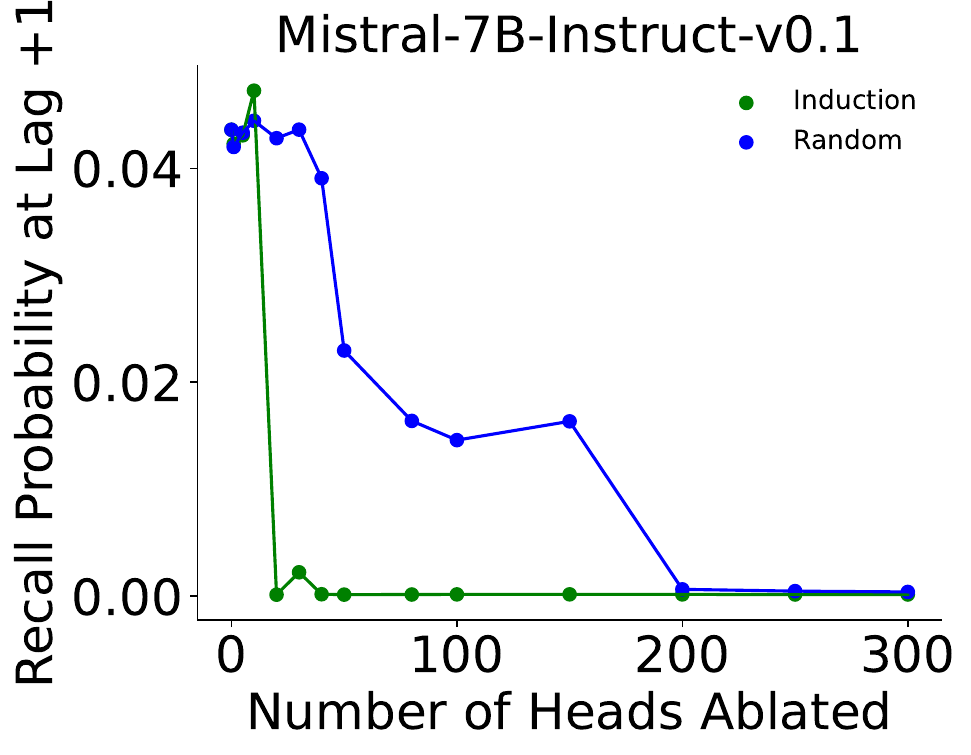}} &
        {\includegraphics[width=0.22\textwidth]{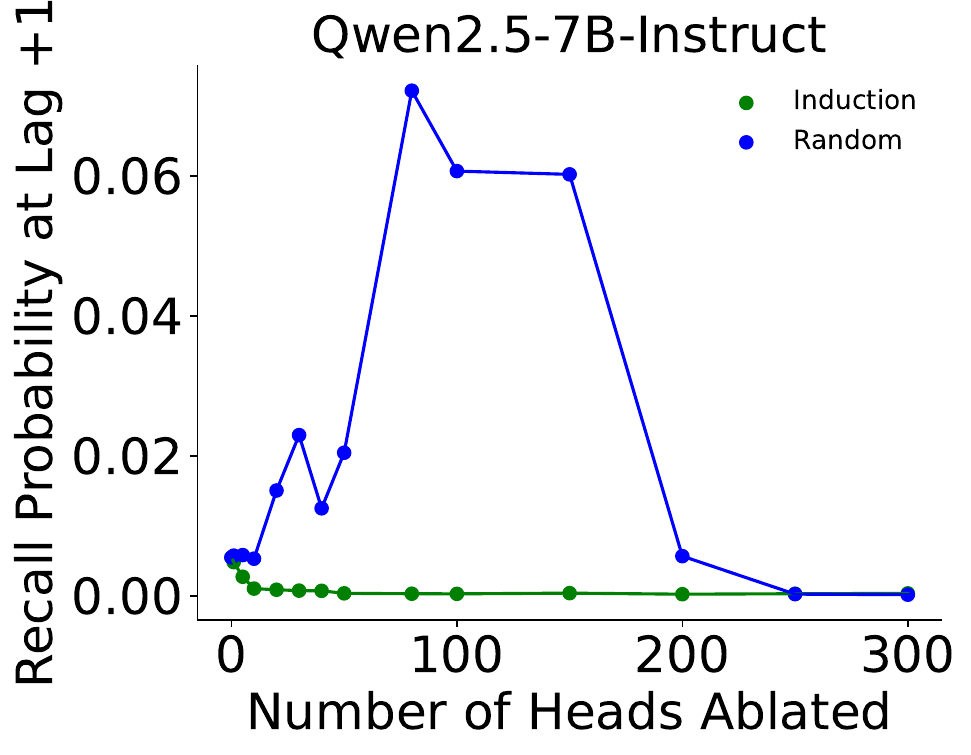}} &
        {\includegraphics[width=0.22\textwidth]{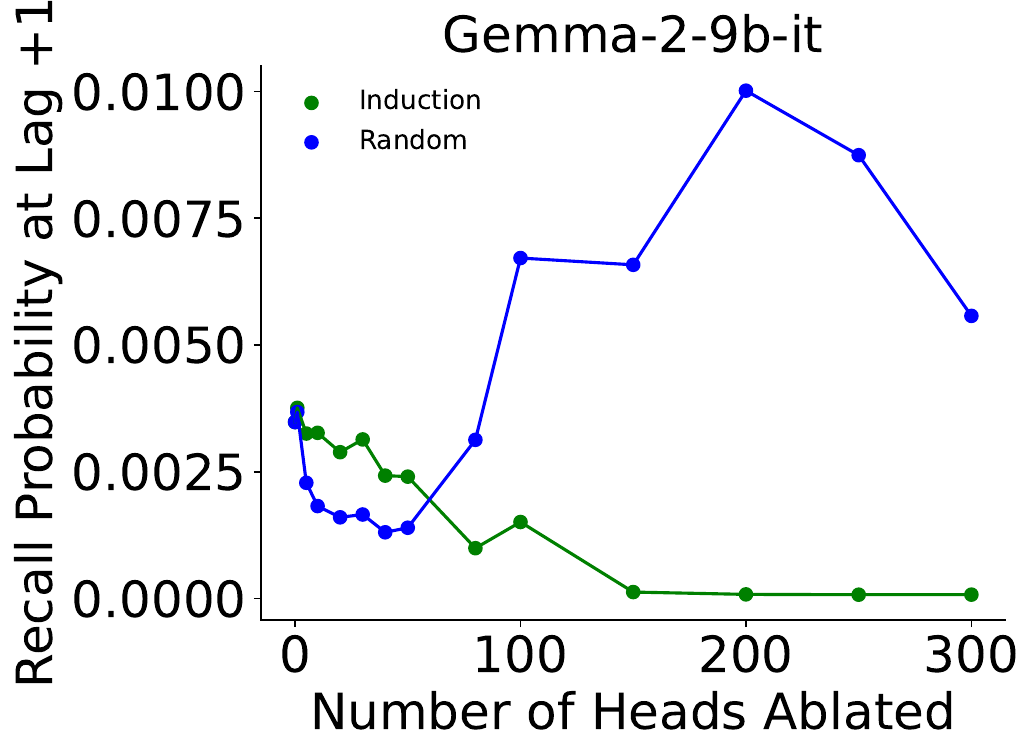}} \\
    \end{tabular}
    \caption{Impact of induction head mean ablation on the model output probability for the token at lag +1. The models were presented with a sequence of 501 tokens where the last token repeated the token at index 250 and the lag is defined relative to the repeated token, hence the probability at lag +1 is the probability that the model assigns to token 251 (see Methods for more details). The results show averages across 5000 runs with shuffled token sequences.  We ablated the following numbers of induction heads (sorted by the induction scores) and random heads (x-axis): 1, 5, 10, 20, 30, 40, 50, 80, 100, 150, 200, 250 and 300. Top row: Base models. Bottom row: Instruction-tuned models.
    \label{fig:lag1_probability_mean}}
\end{figure*}

\begin{figure*}[h!]
    \centering
    \begin{tabular}{lllll}
    \textbf{A} &
    \textbf{B} &
    \textbf{C} &
    \textbf{D} \\
        {\includegraphics[width=0.22\textwidth]{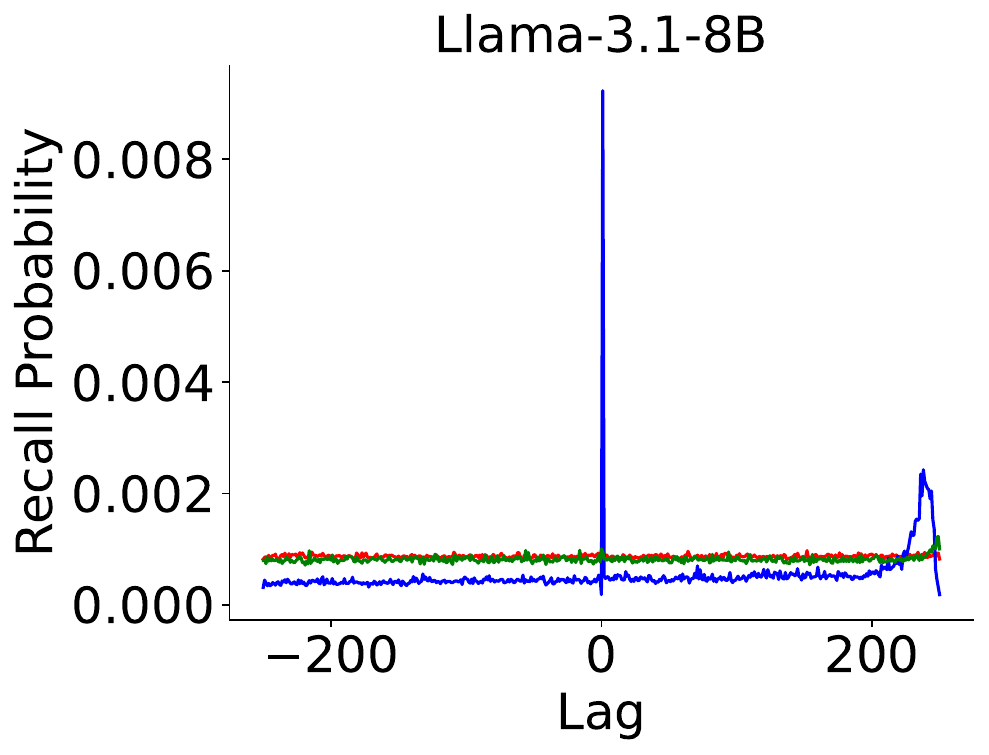
        }} &
        {\includegraphics[width=0.22\textwidth]{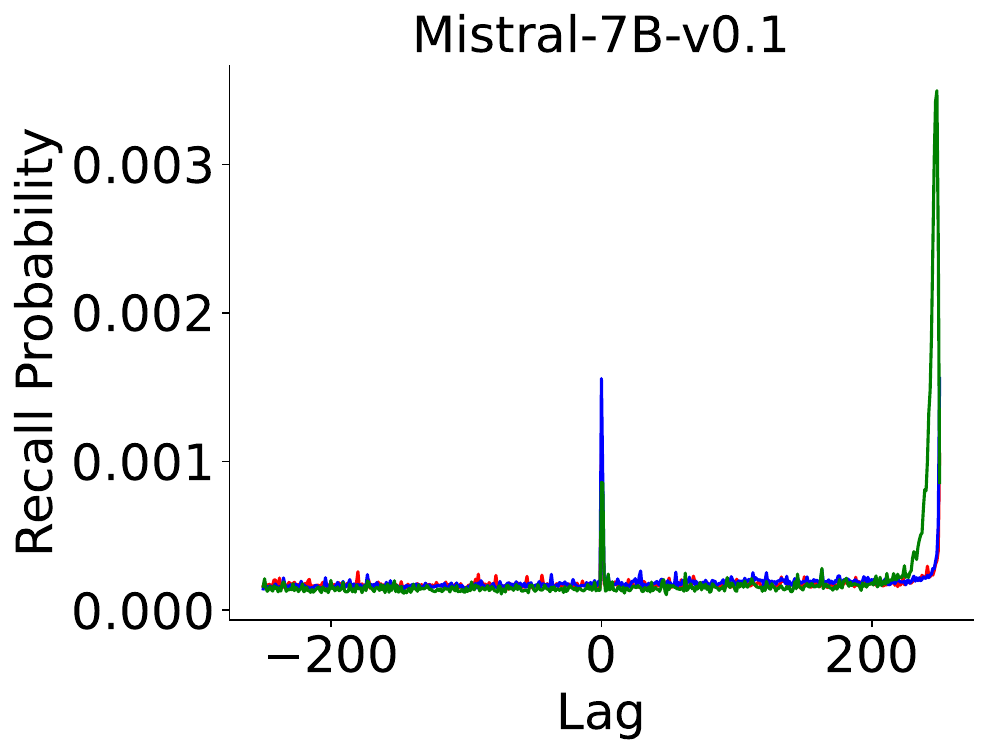}} &
        {\includegraphics[width=0.22\textwidth]{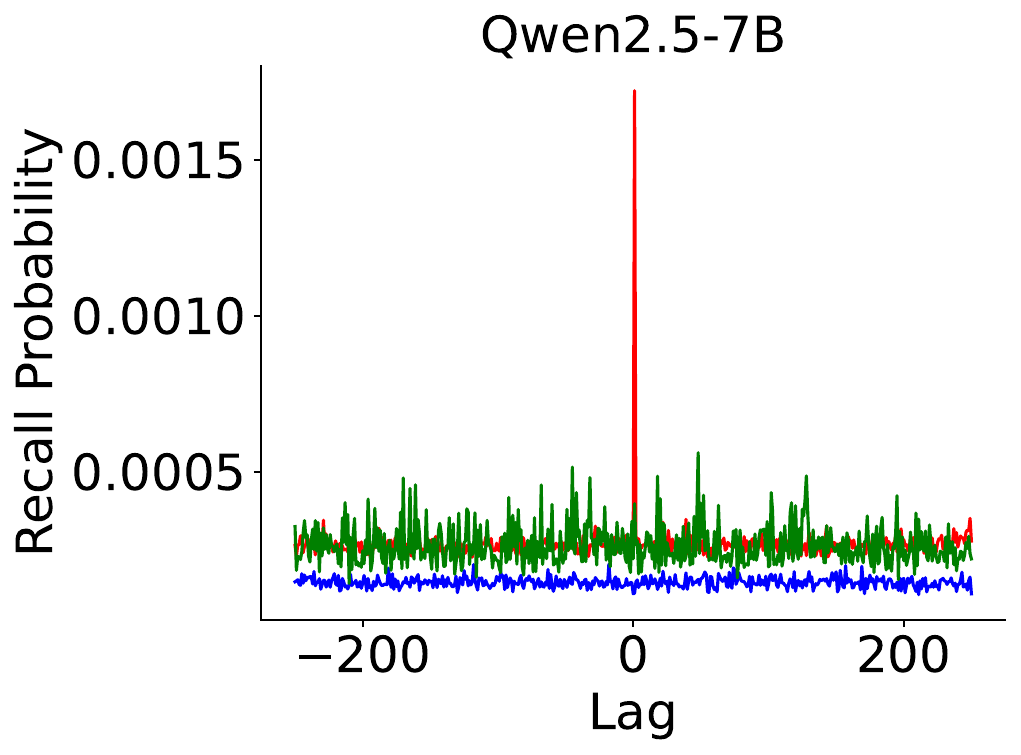}} &
        {\includegraphics[width=0.22\textwidth]{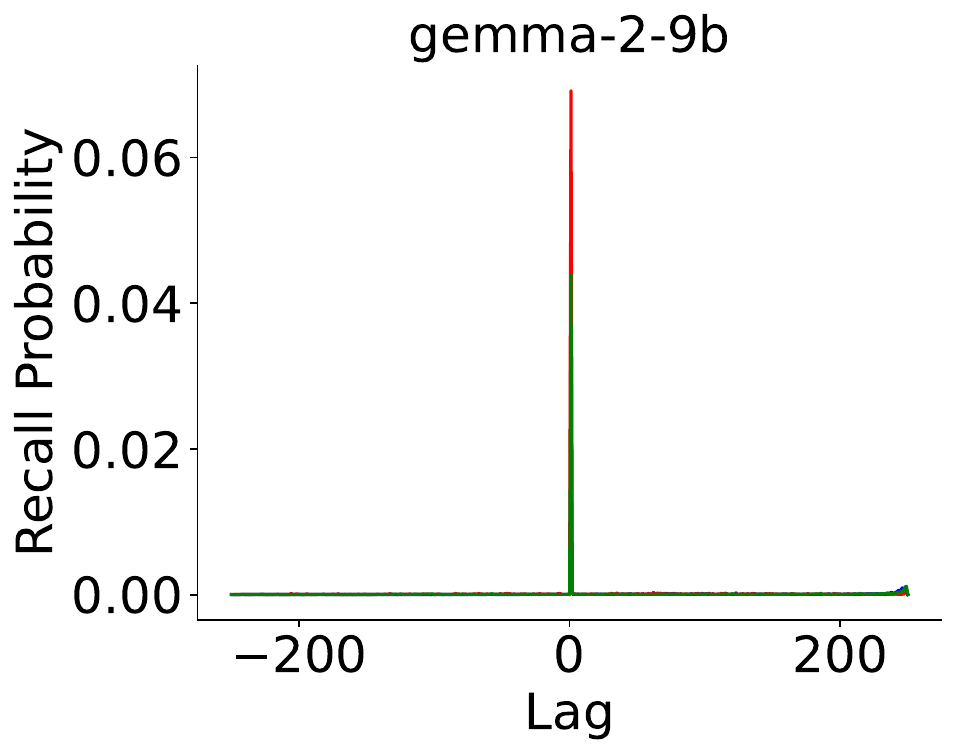}}
        \\
    \end{tabular}
    \begin{tabular}{lllll}
    \textbf{E} &
    \textbf{F} &
    \textbf{G} &
    \textbf{H} \\
        {\includegraphics[width=0.22\textwidth]{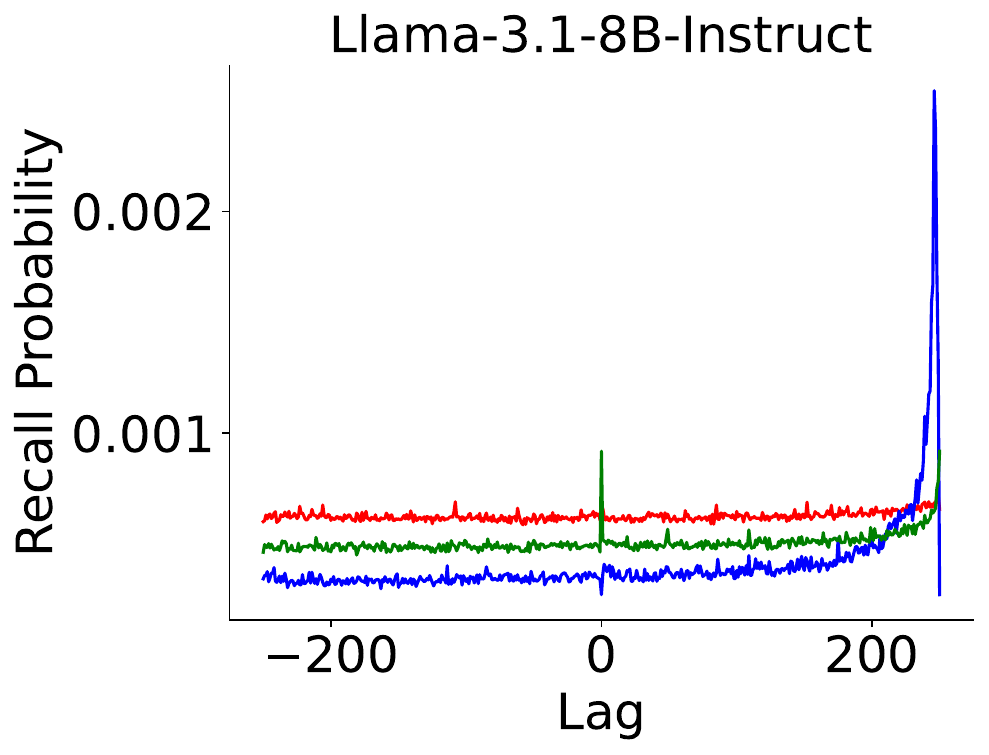
        }} &
        {\includegraphics[width=0.22\textwidth]{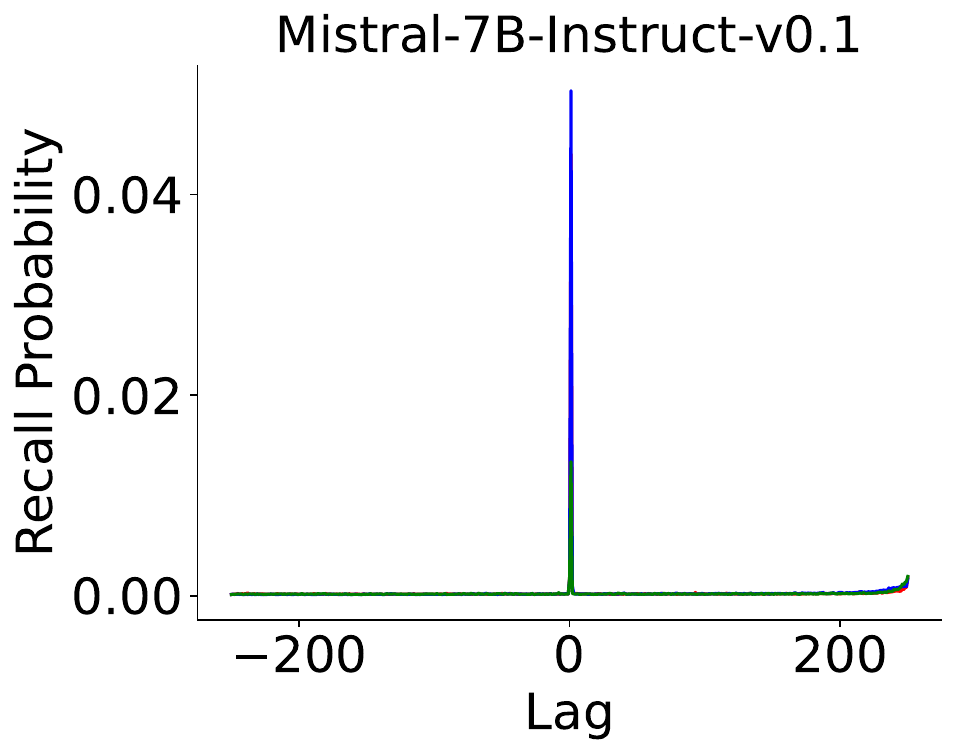}} &
        {\includegraphics[width=0.22\textwidth]{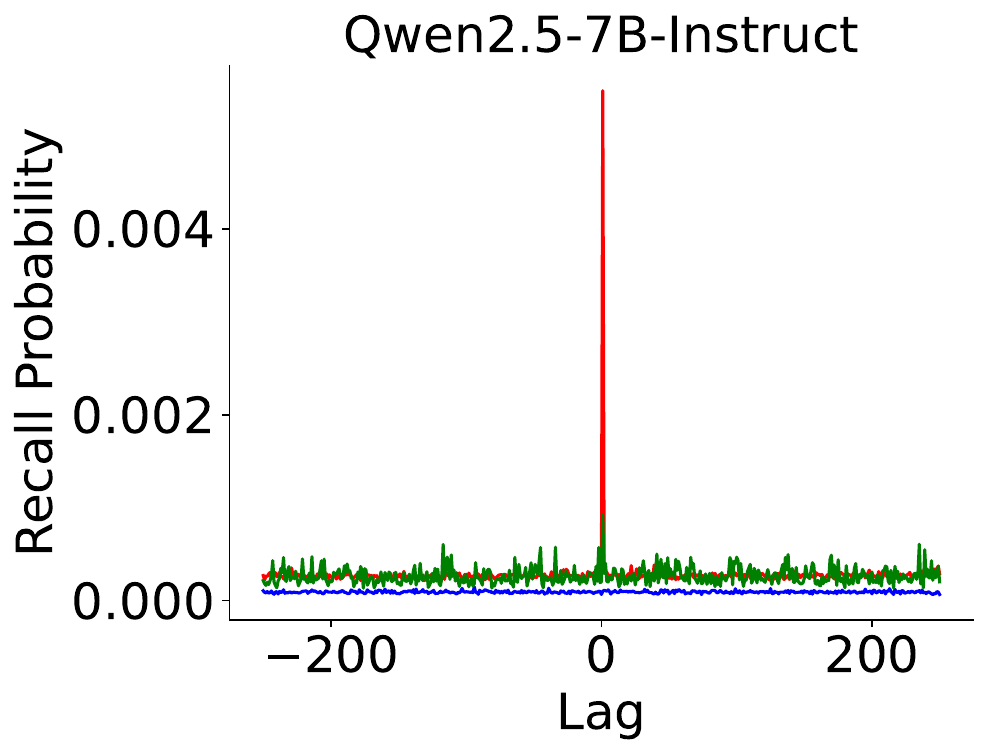}} &
        {\includegraphics[width=0.22\textwidth]{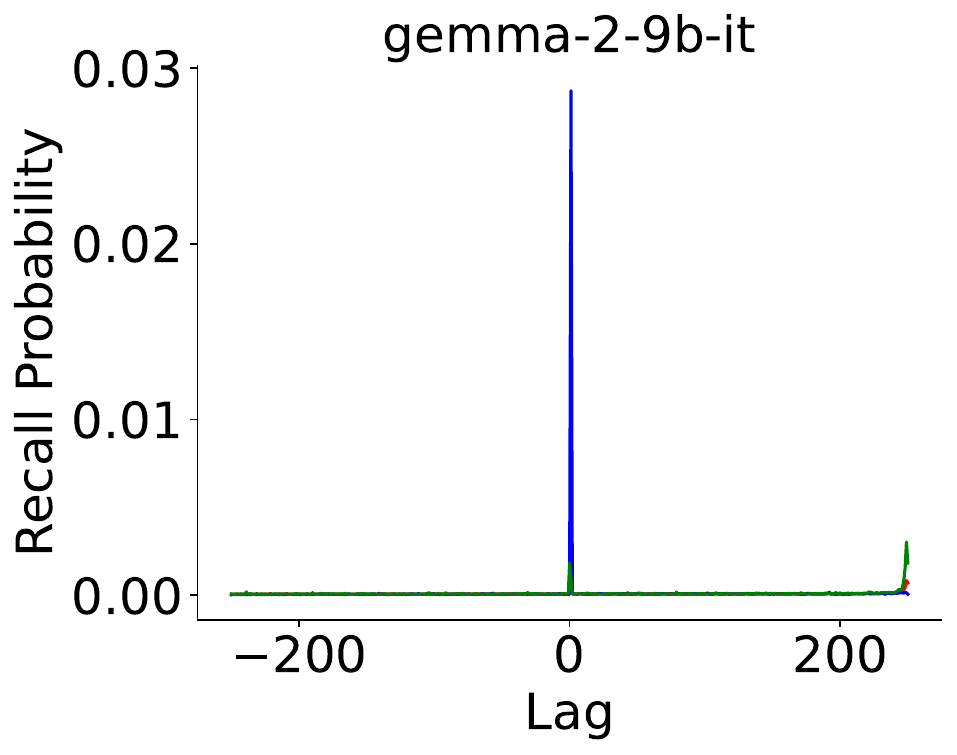}} \\
    \end{tabular}
    \includegraphics[scale=0.5]{Figures/legend.pdf}
    \caption{Impact of induction and random head top layers ablation (100 heads in each case) on the model output probability as a function of lag. The models were presented with a sequence of 501 tokens where the last token repeated the token at index 250 and the lag is defined relative to the repeated token (see Methods for more details and Fig.~\ref{fig:CRP_6_top} for a zoomed version showing lags -6 to 6). The results show averages across 5000 runs with shuffled token sequences.
    Top row: Base models. Bottom row: Instruction-tuned models.
    \label{fig:CRP_250_top}}
\end{figure*}
\begin{figure*}[h!]
    \centering
    \begin{tabular}{lllll}
    \textbf{A} &
    \textbf{B} &
    \textbf{C} &
    \textbf{D} \\
        {\includegraphics[width=0.22\textwidth]{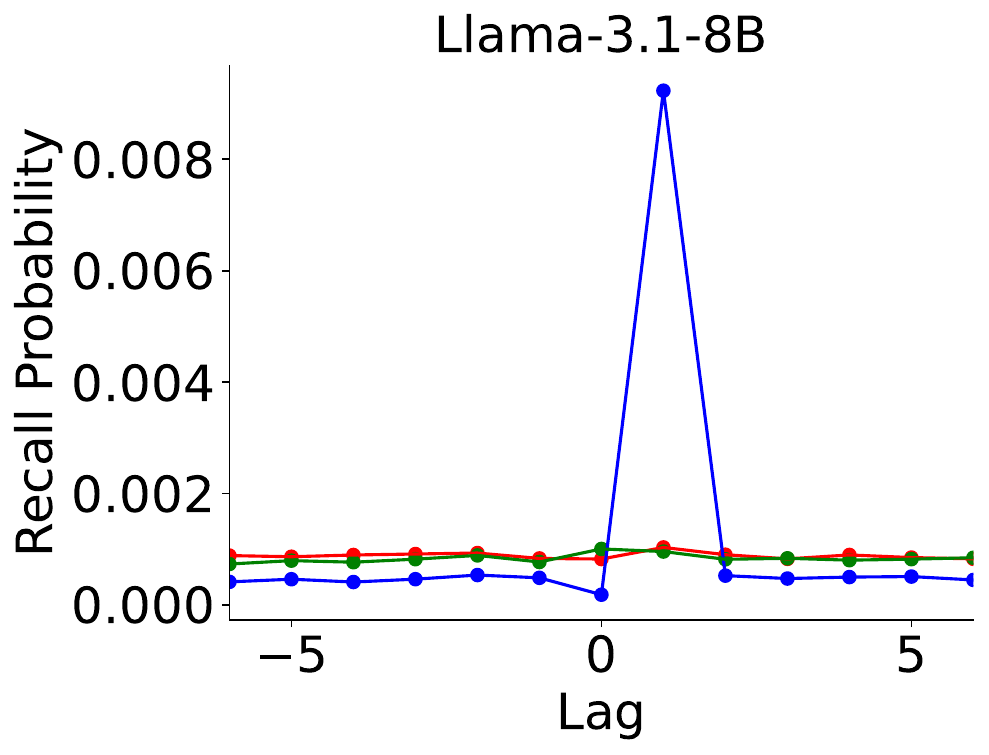
        }} &
        {\includegraphics[width=0.22\textwidth]{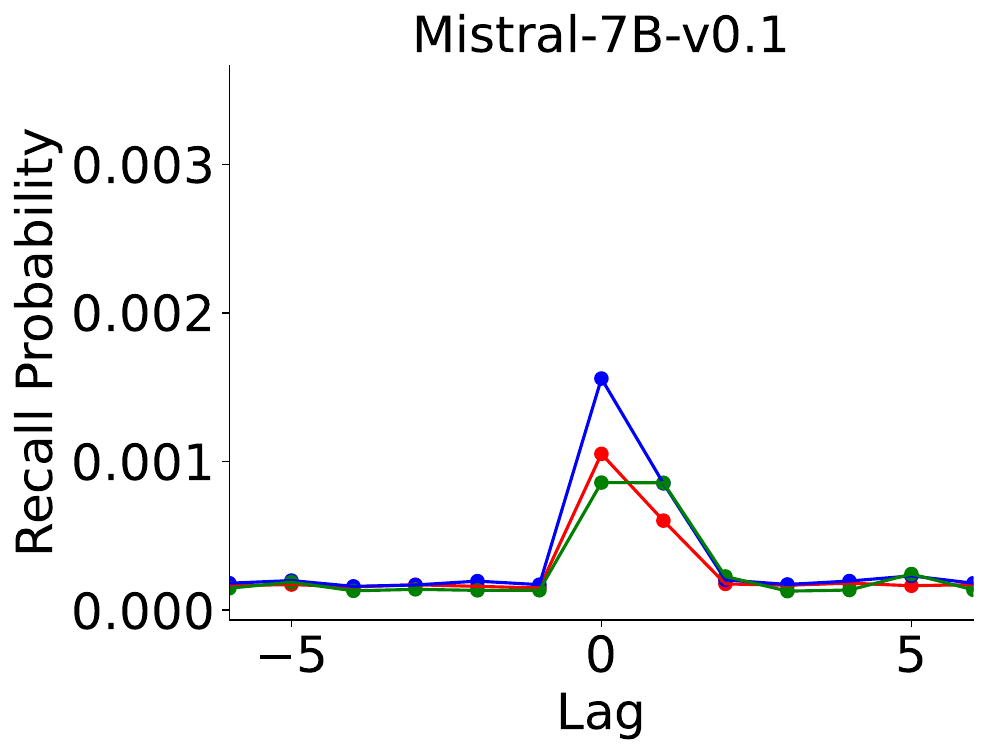}} &
        {\includegraphics[width=0.22\textwidth]{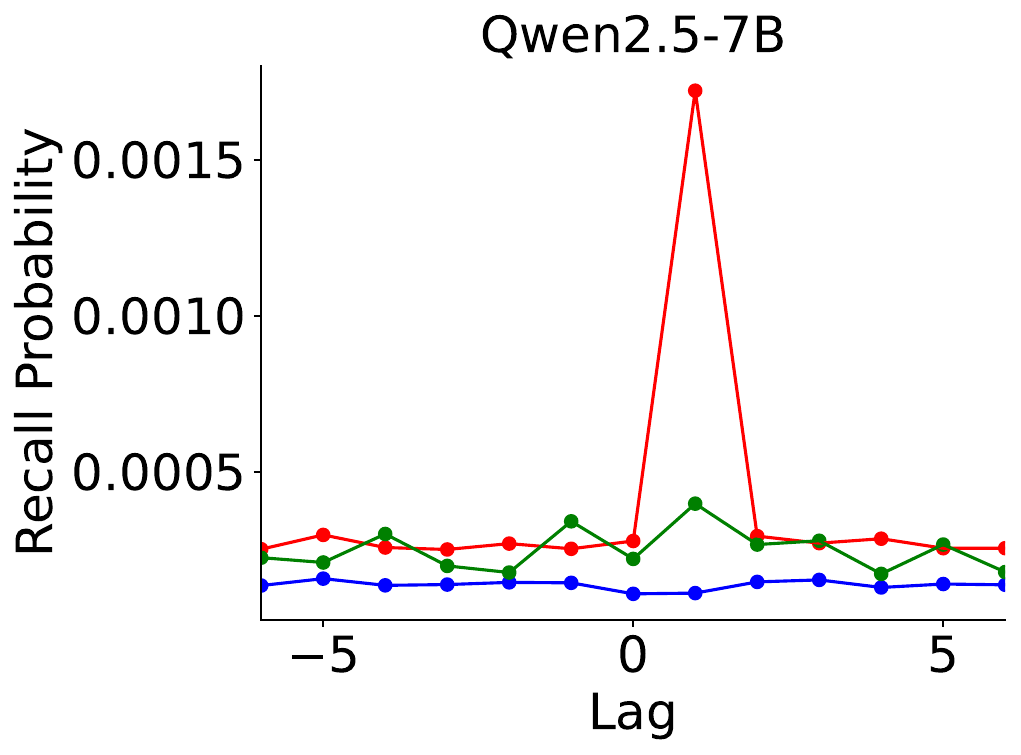}} &
        {\includegraphics[width=0.22\textwidth]{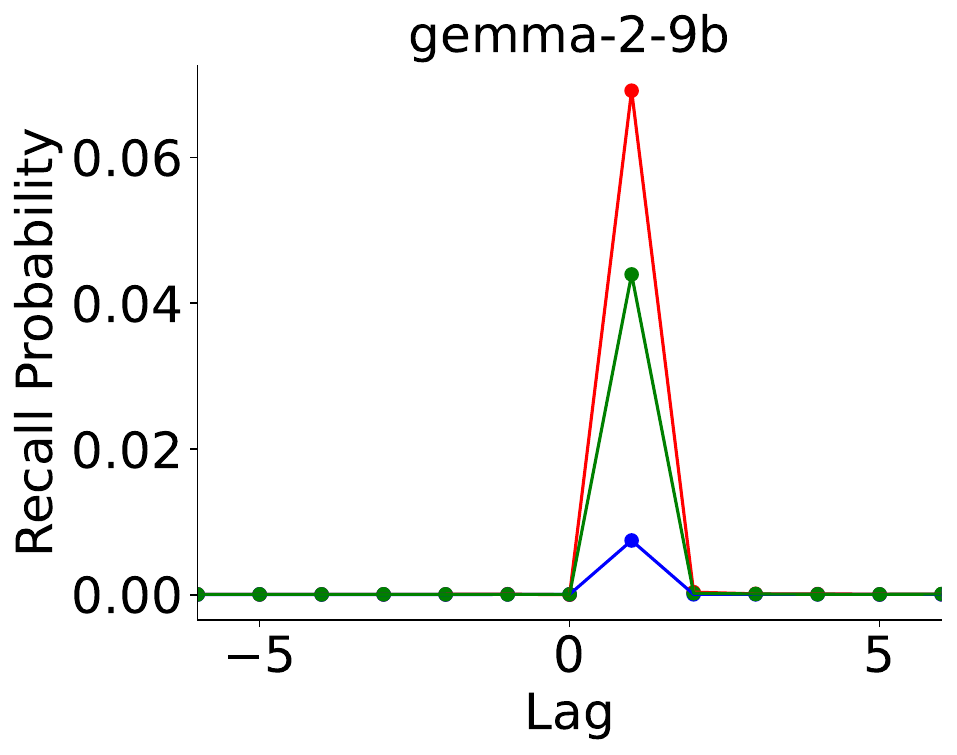}}
        \\
    \end{tabular}
    \begin{tabular}{lllll}
    \textbf{E} &
    \textbf{F} &
    \textbf{G} &
    \textbf{H} \\
        {\includegraphics[width=0.22\textwidth]{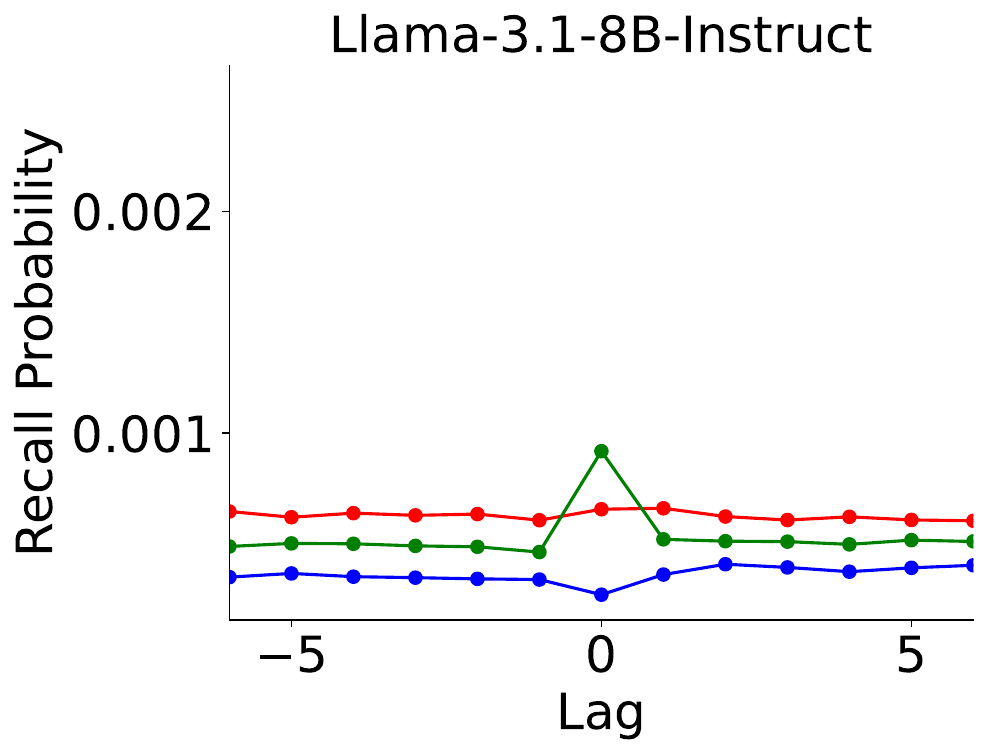
        }} &
        {\includegraphics[width=0.22\textwidth]{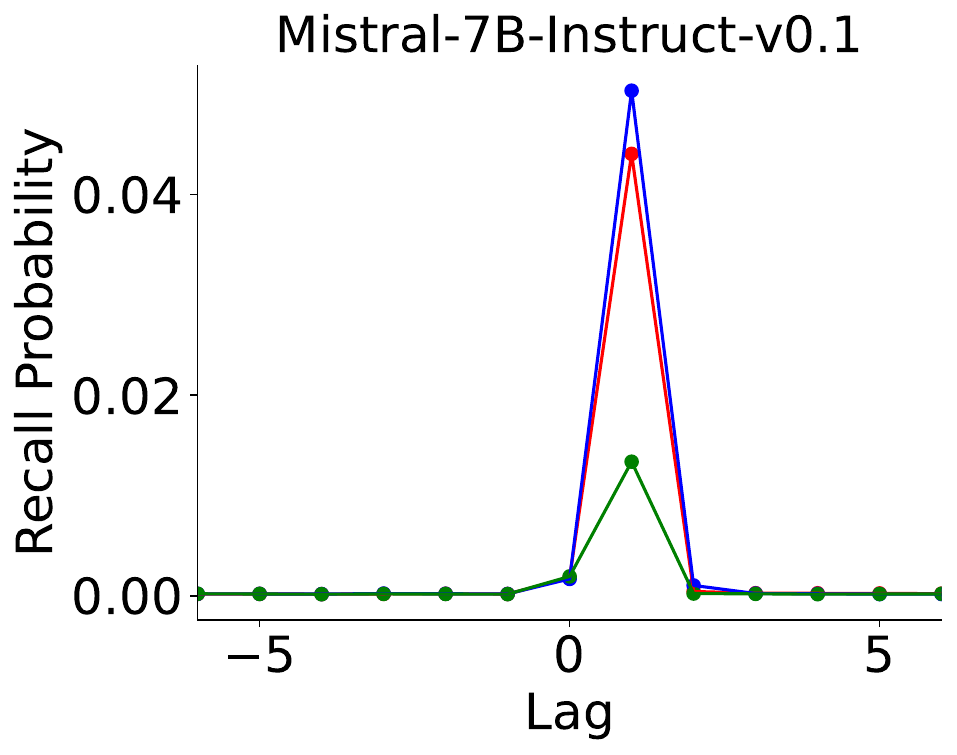}} &
        {\includegraphics[width=0.22\textwidth]{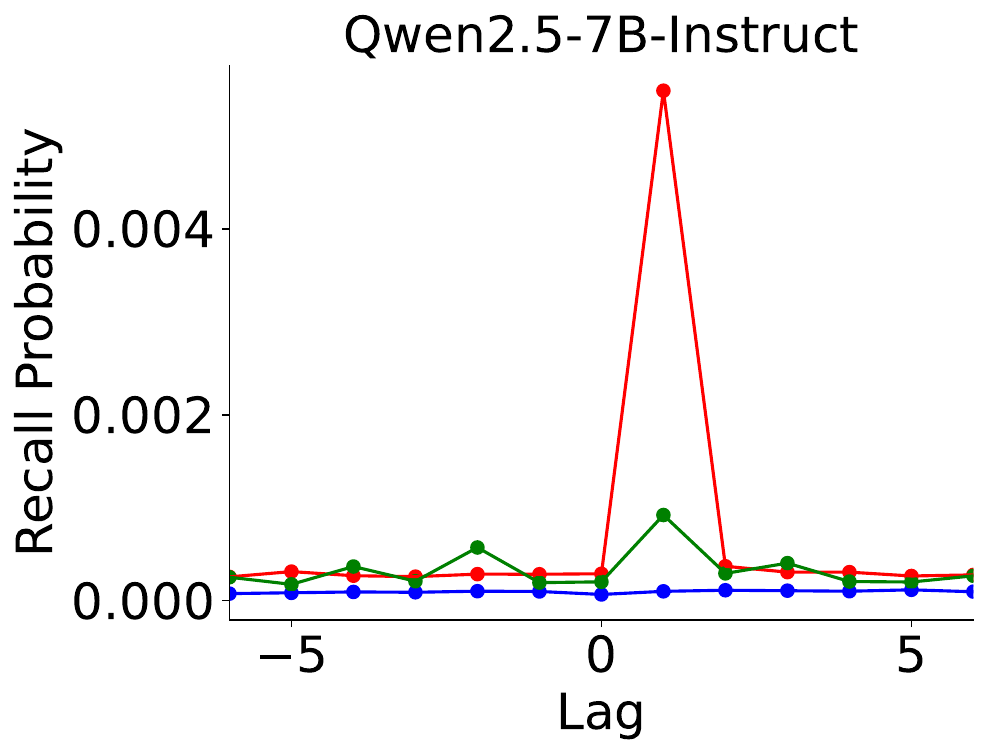}} &
        {\includegraphics[width=0.22\textwidth]{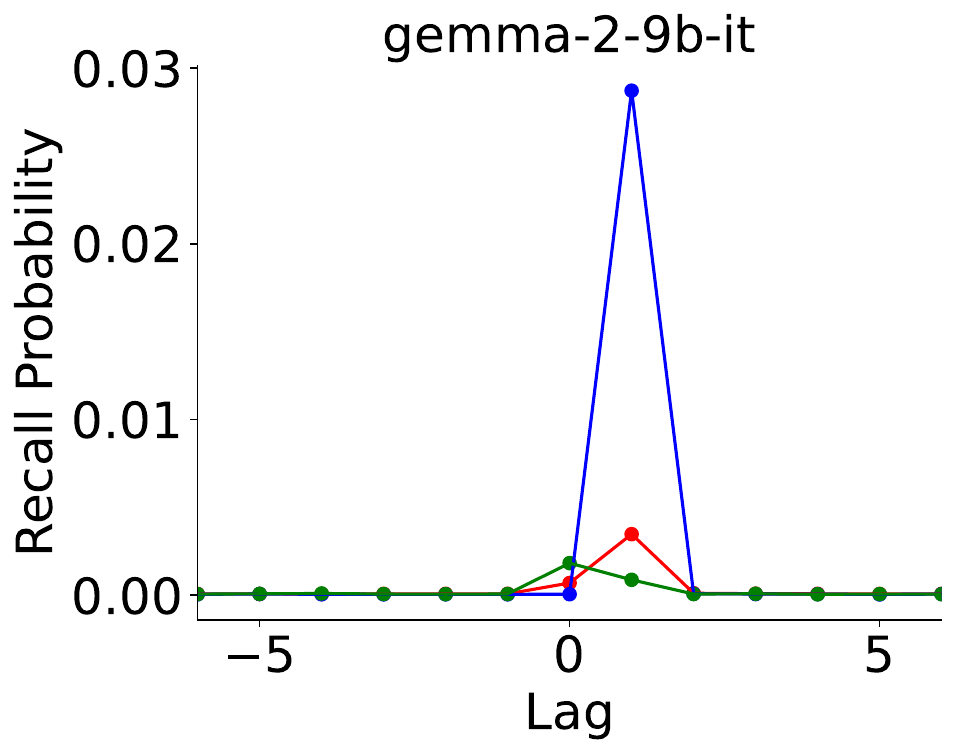}} \\
    \end{tabular}
    \includegraphics[scale=0.5]{Figures/legend.pdf}
    \caption{Impact of induction and random head top layers ablation (100 heads in each case) on the model output probability as a function of lag: same as Fig.~\ref{fig:CRP_250_top}, but with zoom on lags -6 to 6 to emphasize that the highest probabilities were at lags 1 or 0. Top row: Base models. Bottom row: Instruction-tuned models.
    \label{fig:CRP_6_top}}
\end{figure*}

\begin{figure*}[h!]
    \centering
    \begin{tabular}{lllll}
    \textbf{A} &
    \textbf{B} &
    \textbf{C} &
    \textbf{D} \\
        {\includegraphics[width=0.22\textwidth]{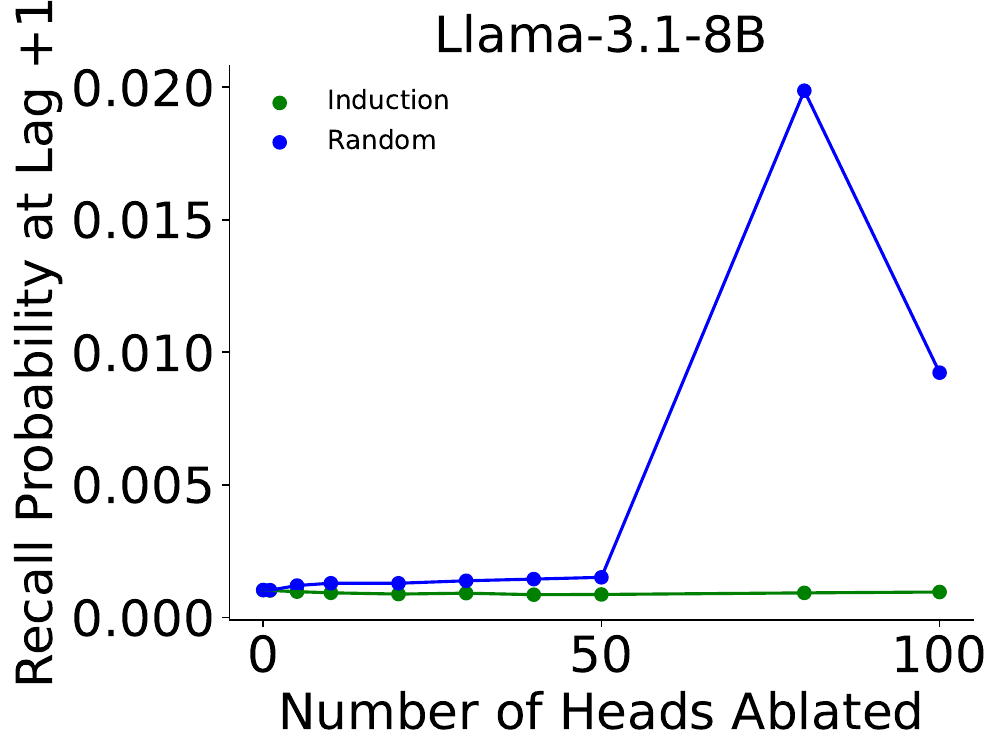
        }} &
        {\includegraphics[width=0.22\textwidth]{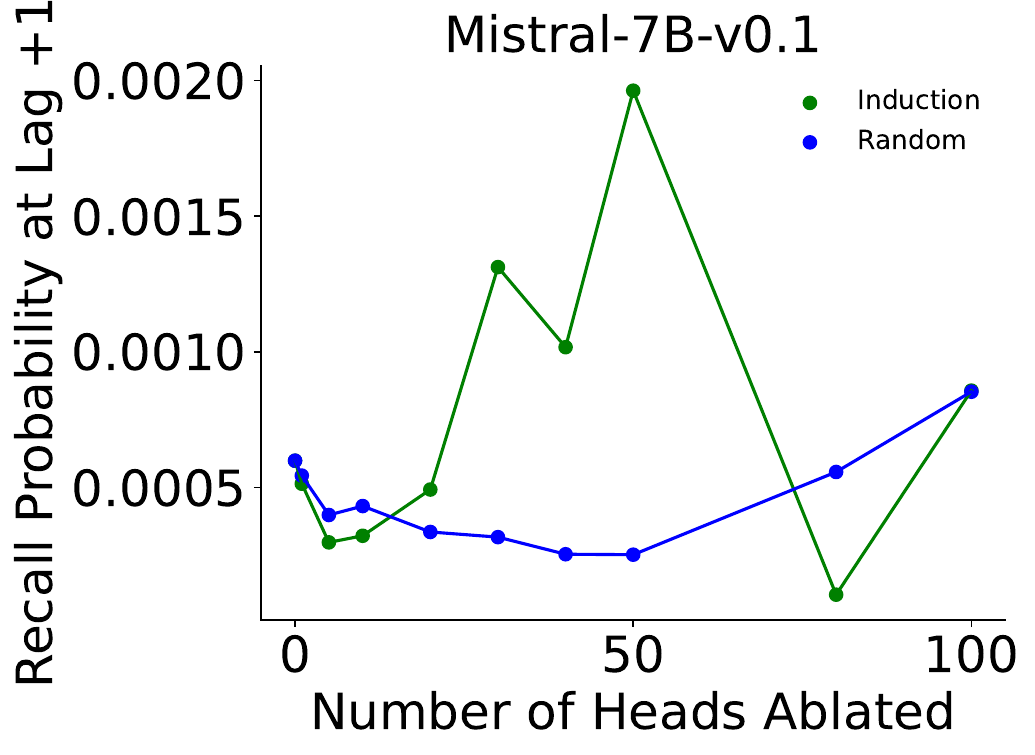}} &
        {\includegraphics[width=0.22\textwidth]{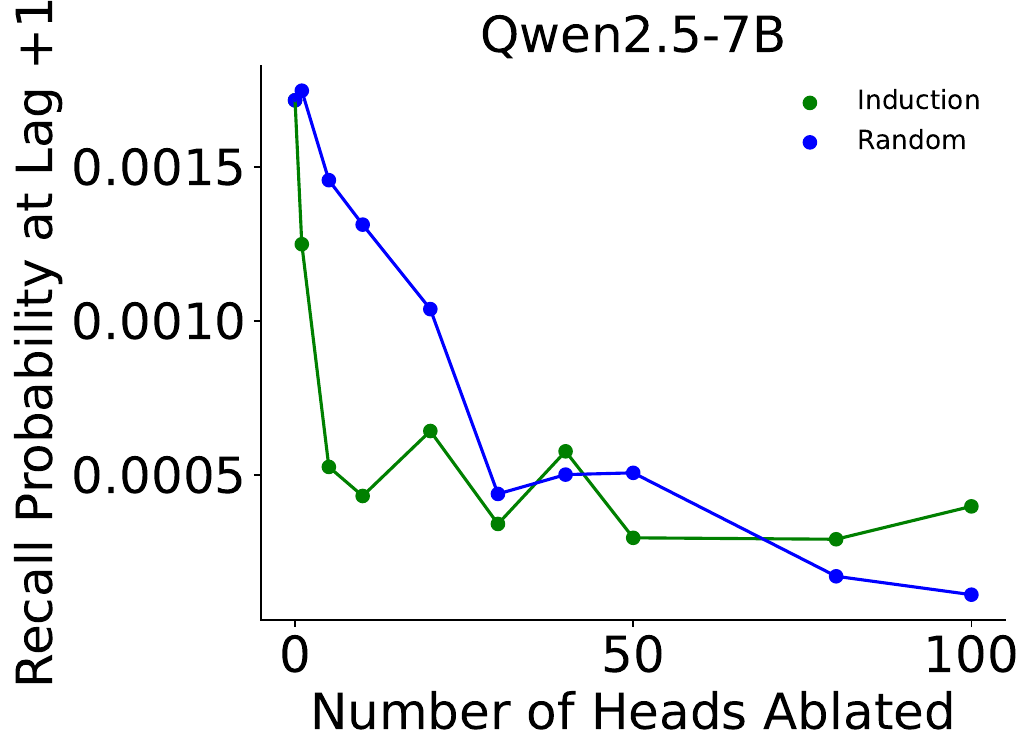}} &
        {\includegraphics[width=0.22\textwidth]{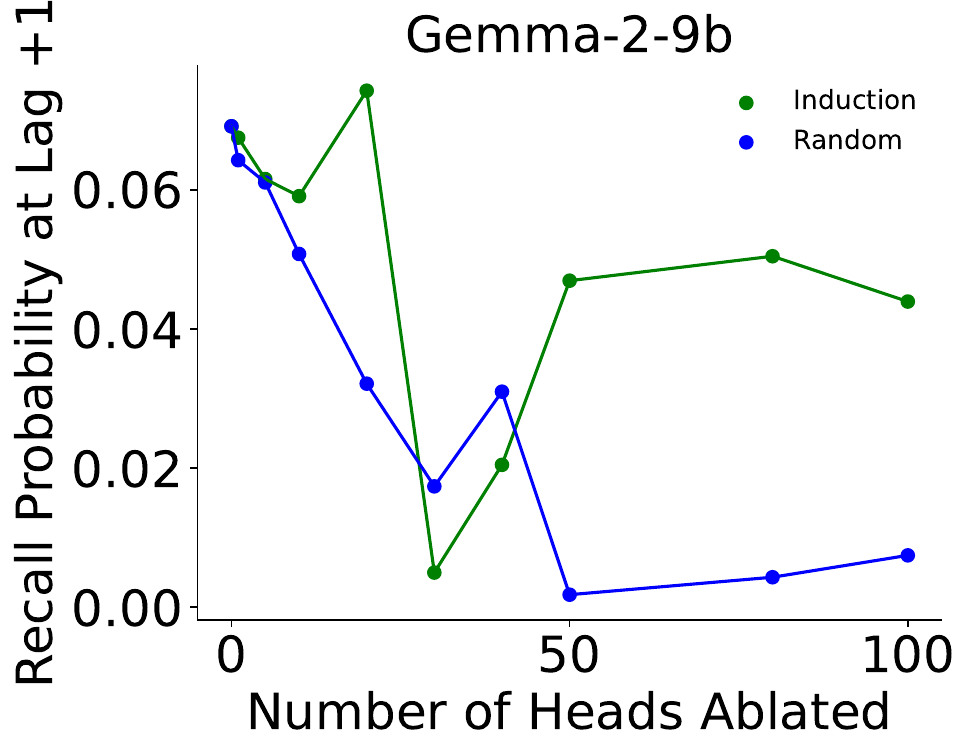}} \\
    \end{tabular}
    \begin{tabular}{lllll}
    \textbf{E} &
    \textbf{F} &
    \textbf{G} &
    \textbf{H} \\
         {\includegraphics[width=0.22\textwidth]{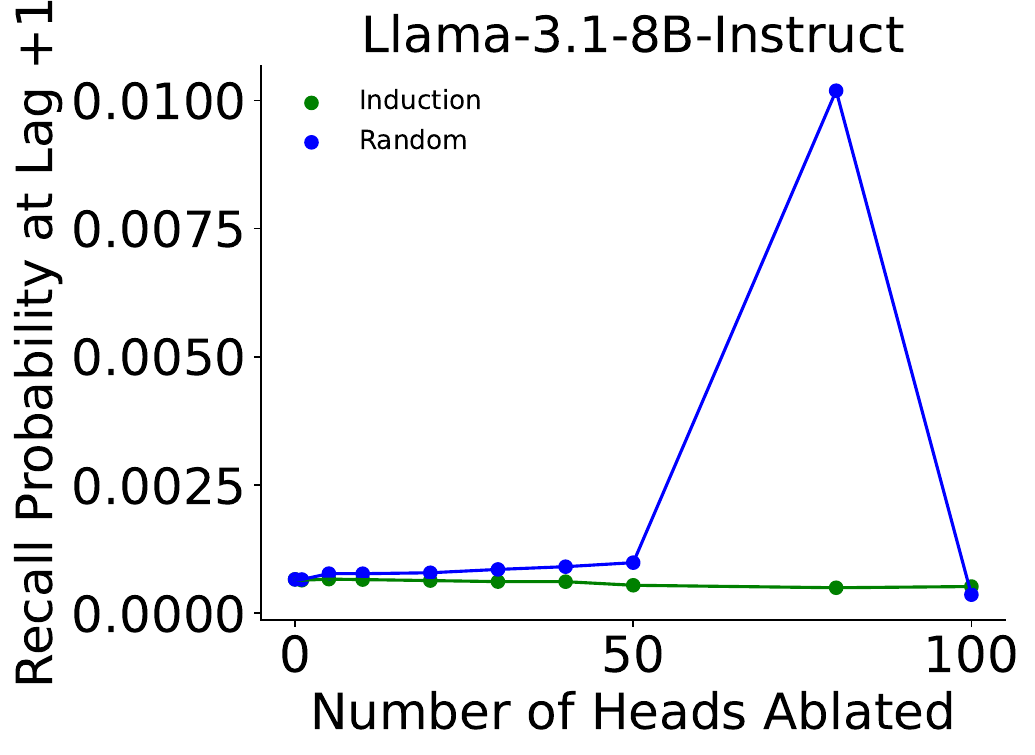
        }} &
        {\includegraphics[width=0.22\textwidth]{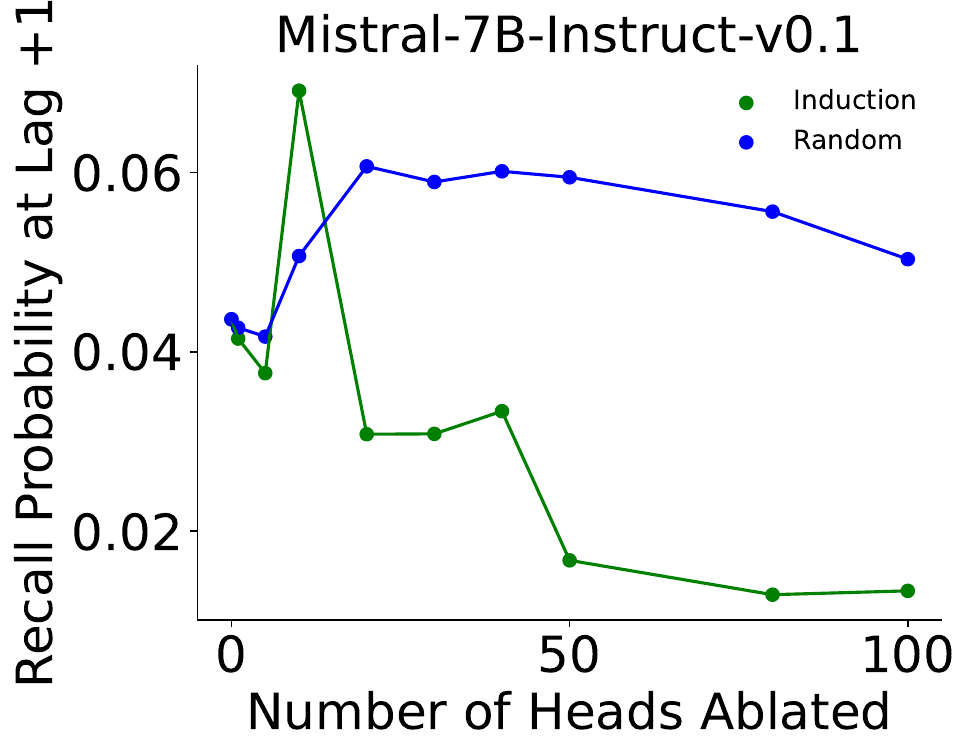}} &
        {\includegraphics[width=0.22\textwidth]{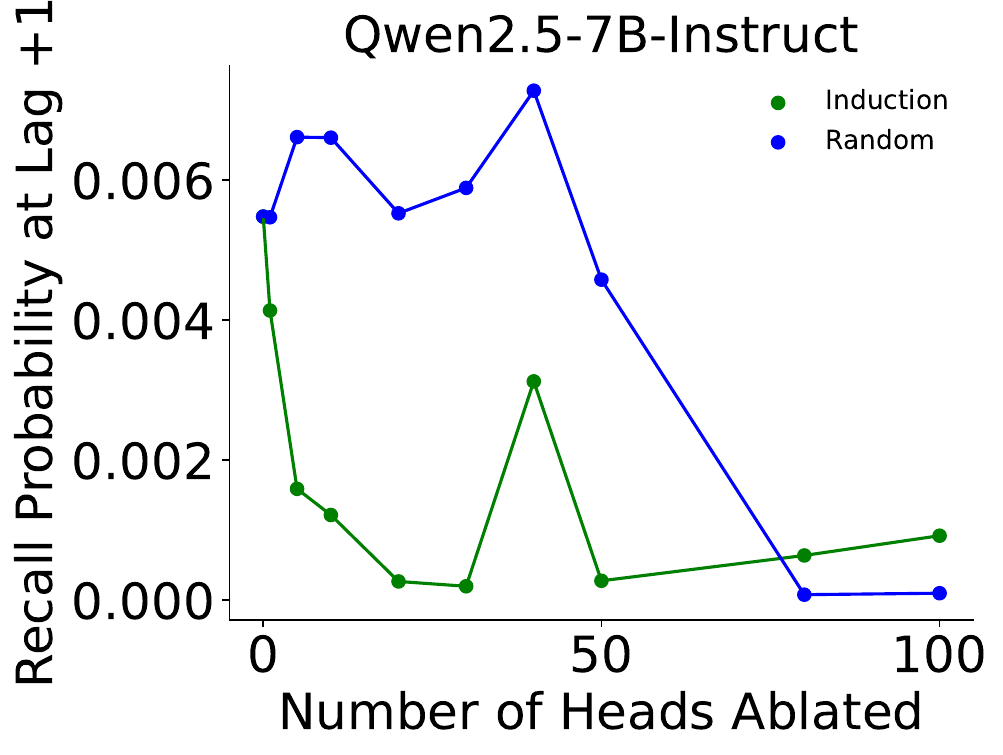}} &
        {\includegraphics[width=0.22\textwidth]{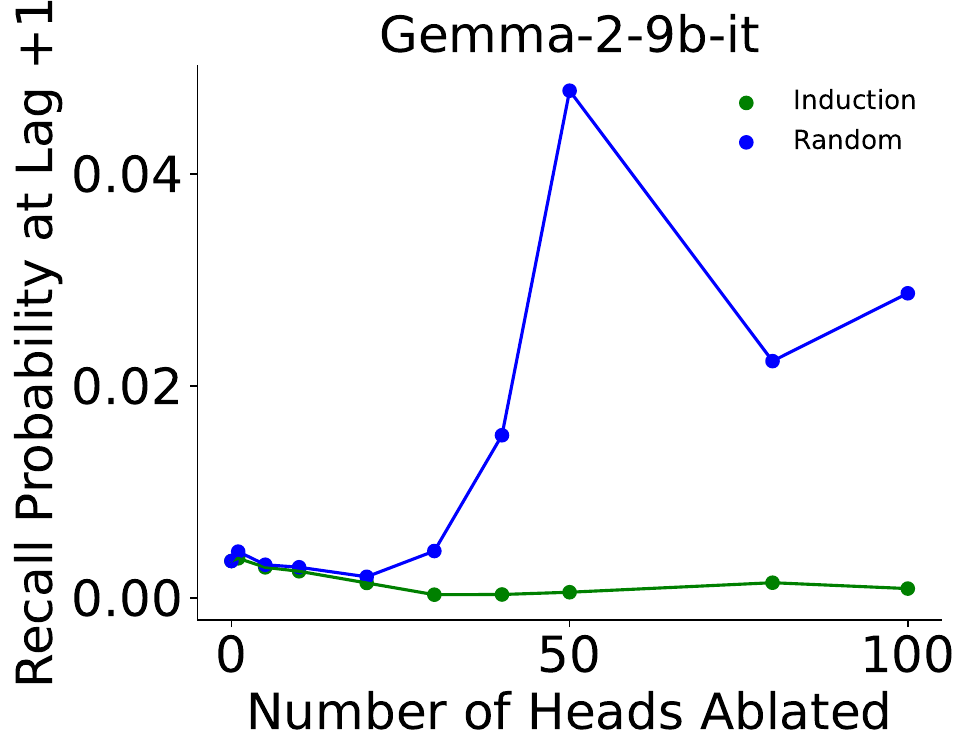}} \\
    \end{tabular}
    \caption{Impact of induction head top layers ablation on the model output probability for the token at lag +1. The models were presented with a sequence of 501 tokens where the last token repeated the token at index 250 and the lag is defined relative to the repeated token, hence the probability at lag +1 is the probability that the model assigns to token 251 (see Methods for more details). The results show averages across 5000 runs with shuffled token sequences.  We ablated the following numbers of induction heads (sorted by the induction scores) and random heads (x-axis): 1, 5, 10, 20, 30, 40, 50, 80, 100, 150, 200, 250 and 300. Top row: Base models. Bottom row: Instruction-tuned models.
    \label{fig:lag1_probability_top}}
\end{figure*}

\begin{figure*}[h!]
    \centering
    \begin{tabular}{lllll}
    \textbf{A} &
    \textbf{B} &
    \textbf{C} &
    \textbf{D} \\
        {\includegraphics[width=0.22\textwidth]{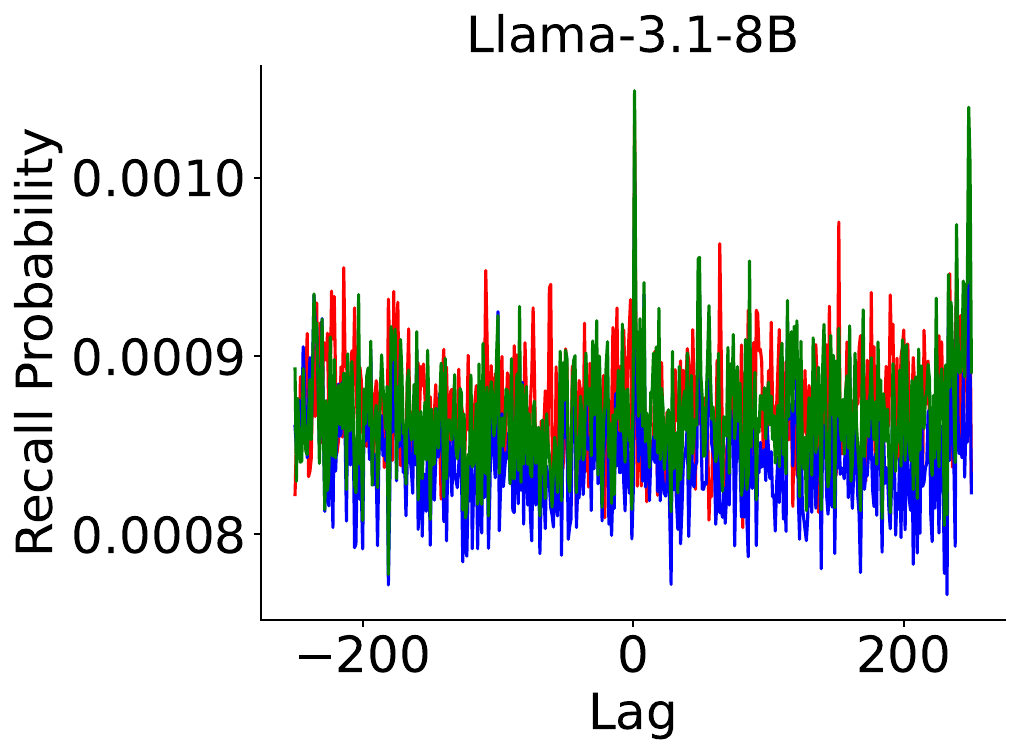
        }} &
        {\includegraphics[width=0.22\textwidth]{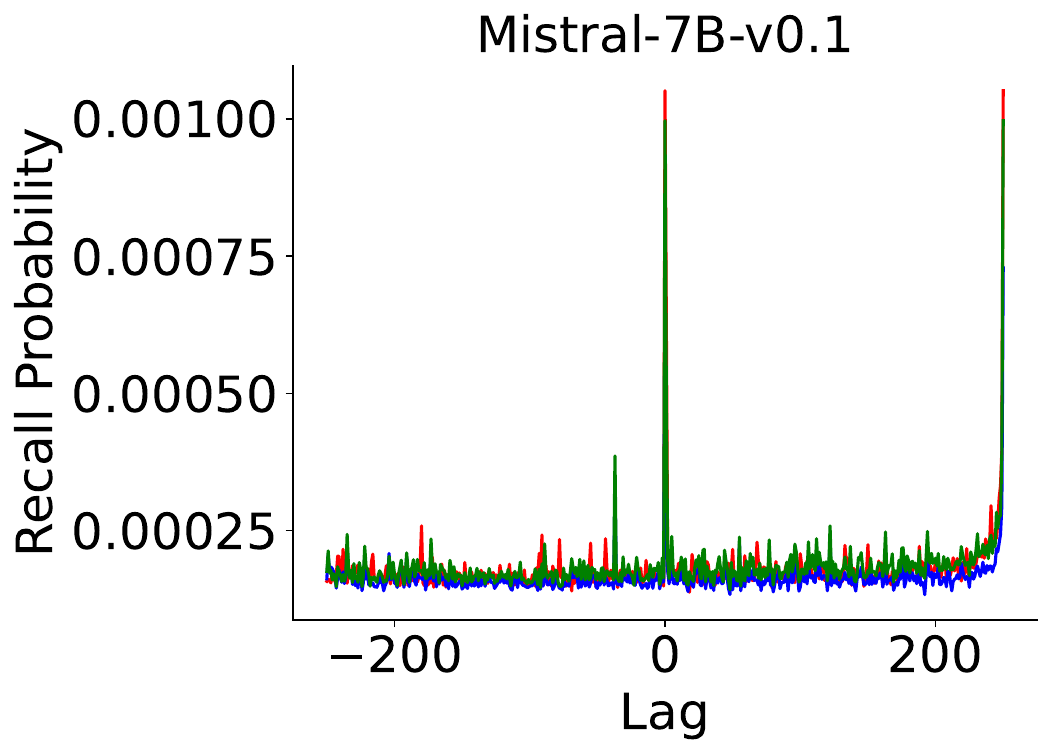}} &
        {\includegraphics[width=0.22\textwidth]{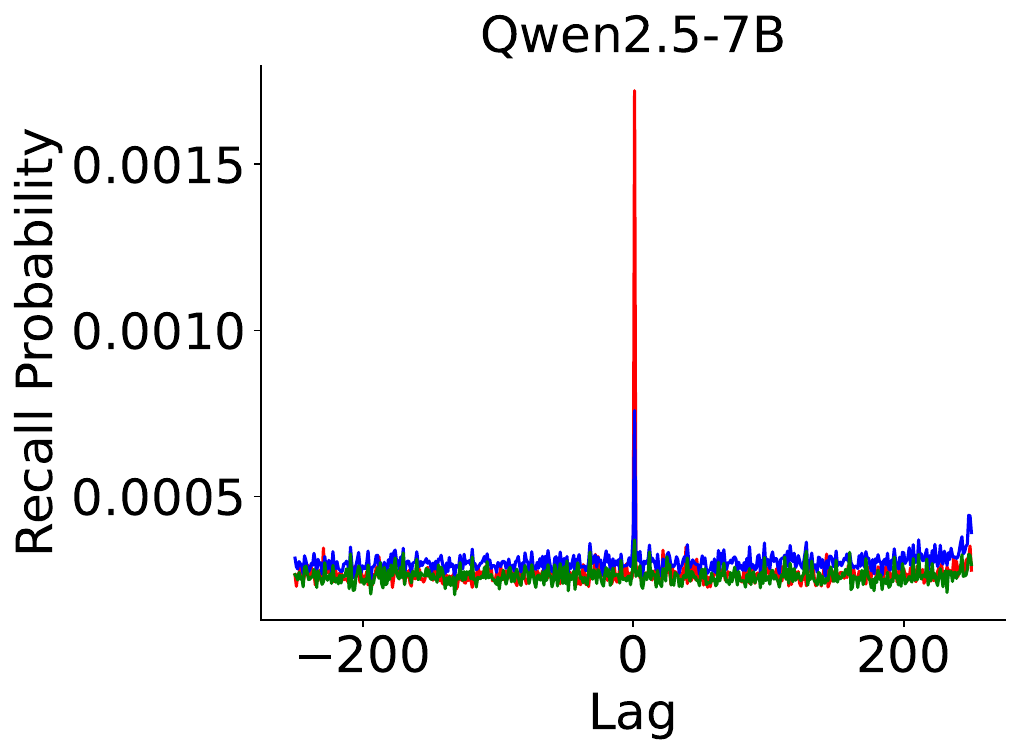}} &
        {\includegraphics[width=0.22\textwidth]{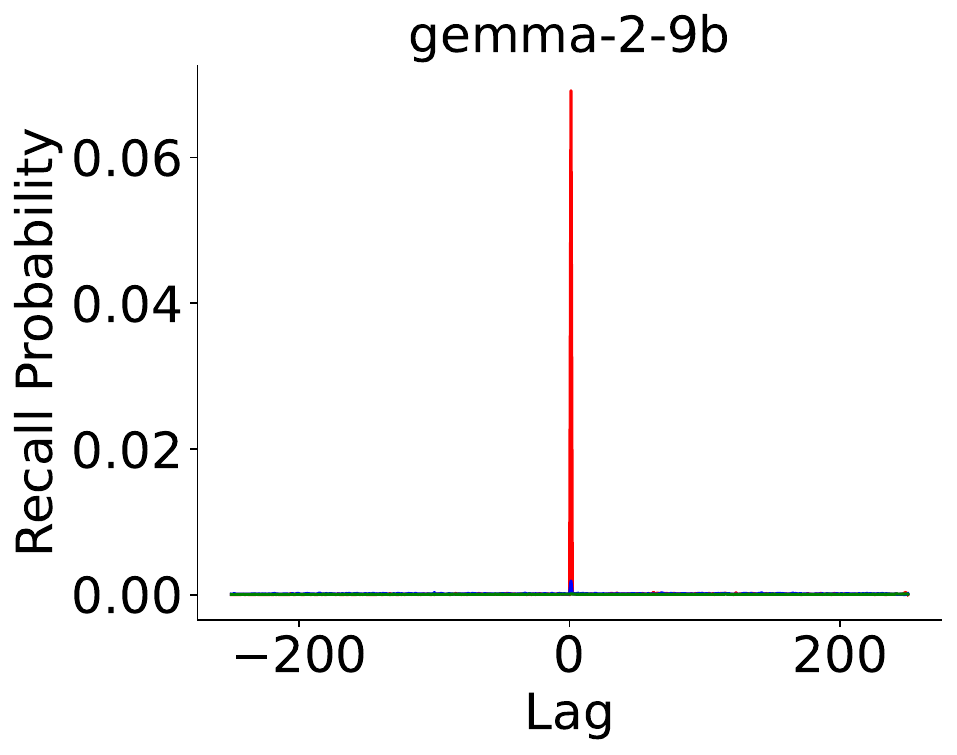}}
        \\
    \end{tabular}
    \begin{tabular}{lllll}
    \textbf{E} &
    \textbf{F} &
    \textbf{G} &
    \textbf{H} \\
        {\includegraphics[width=0.22\textwidth]{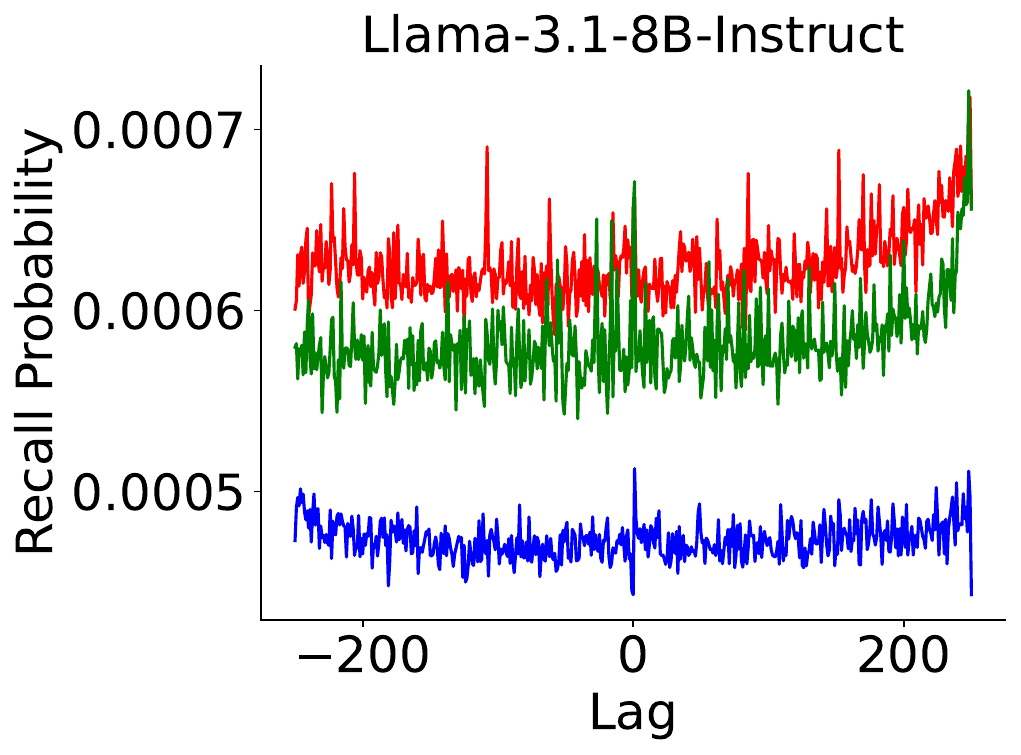
        }} &
        {\includegraphics[width=0.22\textwidth]{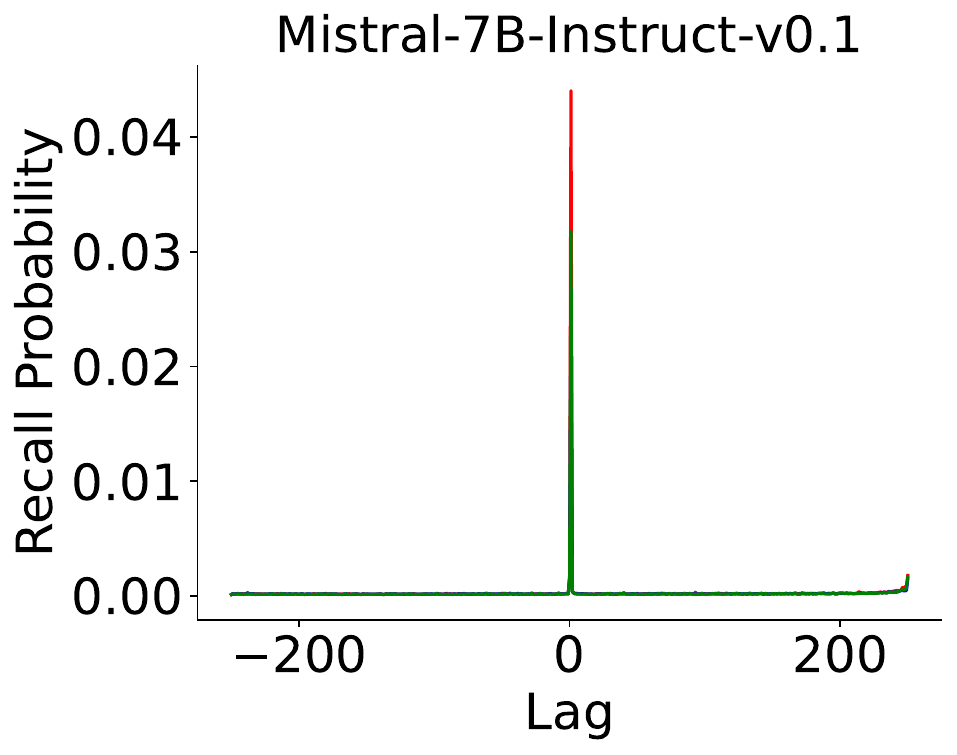}} &
        {\includegraphics[width=0.22\textwidth]{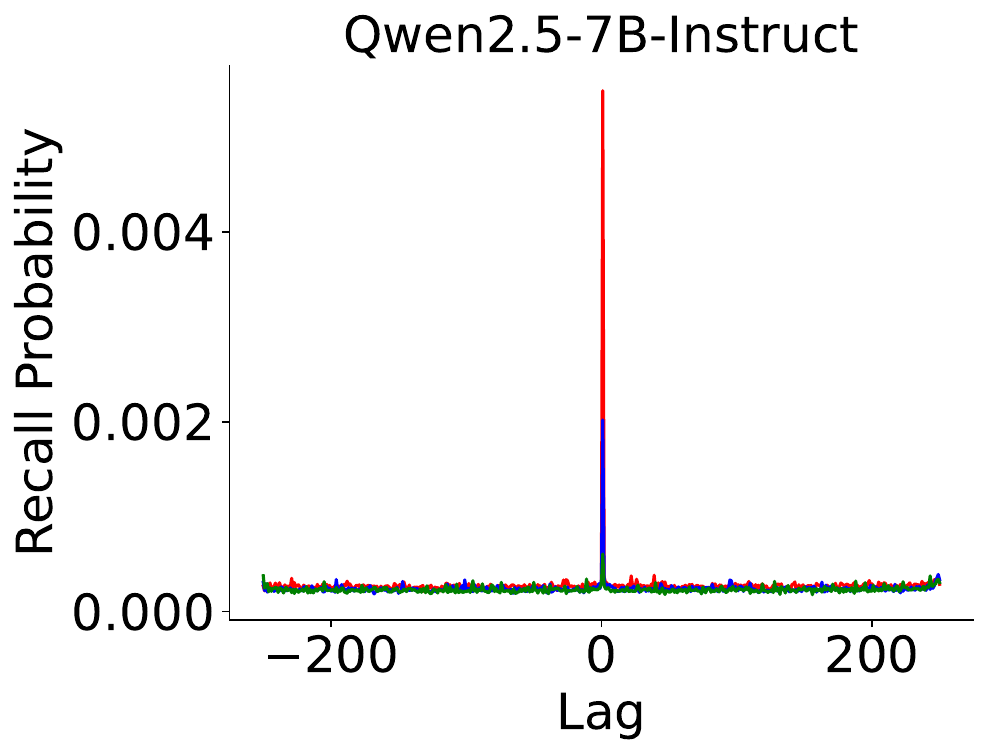}} &
        {\includegraphics[width=0.22\textwidth]{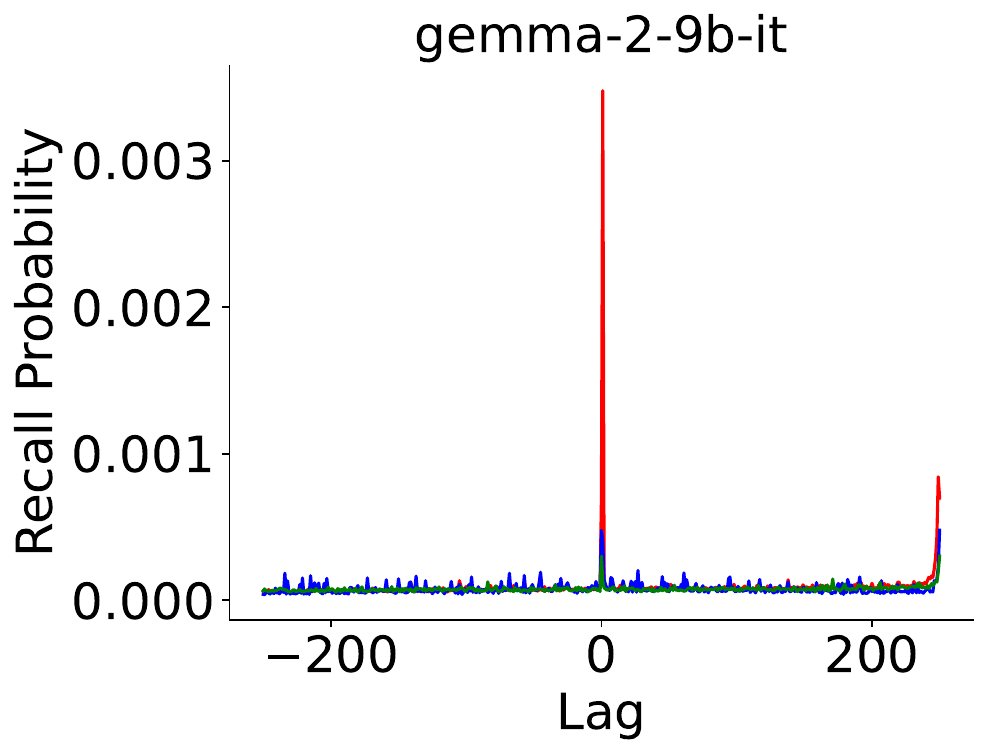}} \\
    \end{tabular}
    \includegraphics[scale=0.5]{Figures/legend.pdf}
    \caption{Impact of induction and random head bottom layers ablation (100 heads in each case) on the model output probability as a function of lag. The models were presented with a sequence of 501 tokens where the last token repeated the token at index 250 and the lag is defined relative to the repeated token (see Methods for more details and Fig.~\ref{fig:CRP_6_bottom} for a zoomed version showing lags -6 to 6). The results show averages across 5000 runs with shuffled token sequences.
    Top row: Base models. Bottom row: Instruction-tuned models.
    \label{fig:CRP_250_bottom}}
\end{figure*}
\begin{figure*}[h!]
    \centering
    \begin{tabular}{lllll}
    \textbf{A} &
    \textbf{B} &
    \textbf{C} &
    \textbf{D} \\
        {\includegraphics[width=0.22\textwidth]{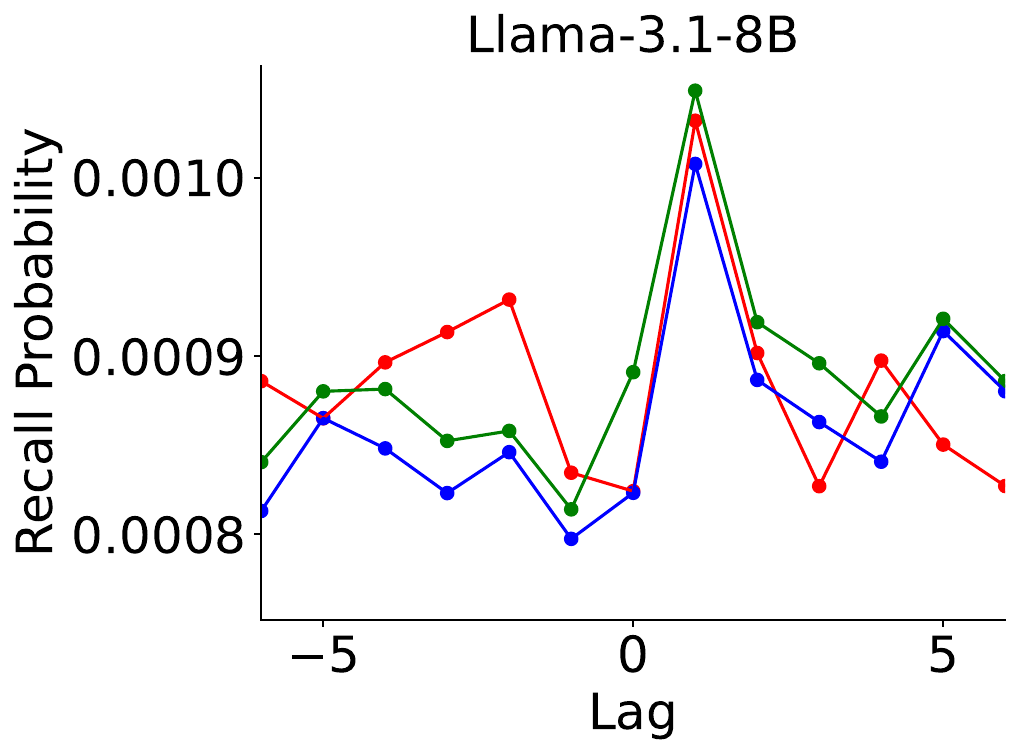
        }} &
        {\includegraphics[width=0.22\textwidth]{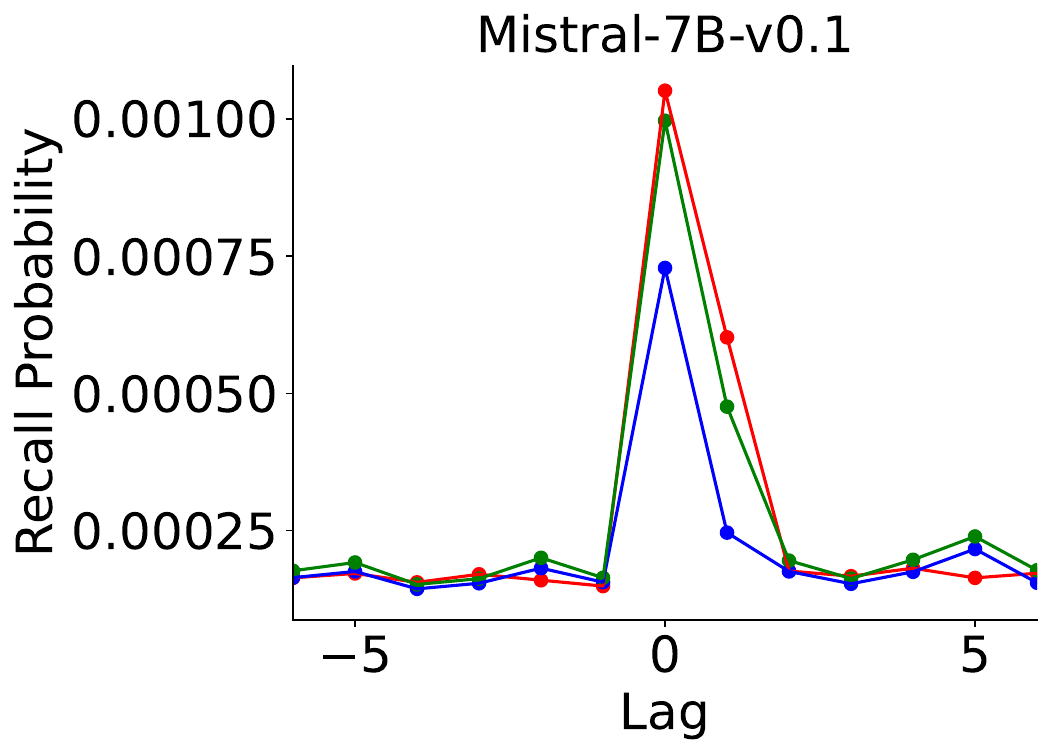}} &
        {\includegraphics[width=0.22\textwidth]{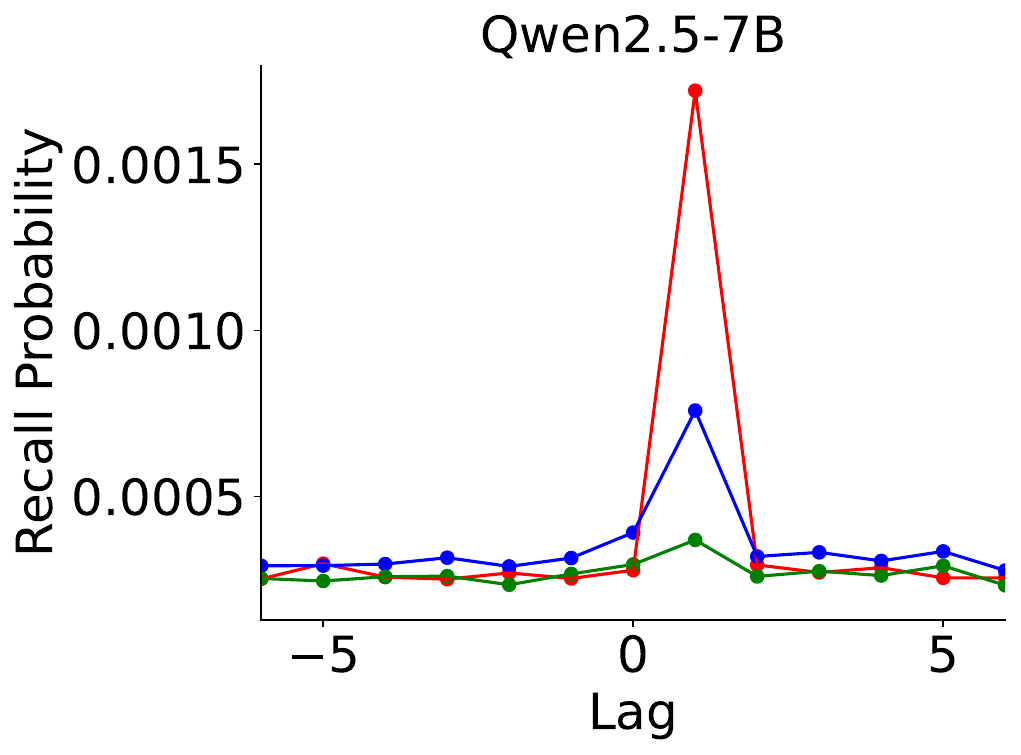}} &
        {\includegraphics[width=0.22\textwidth]{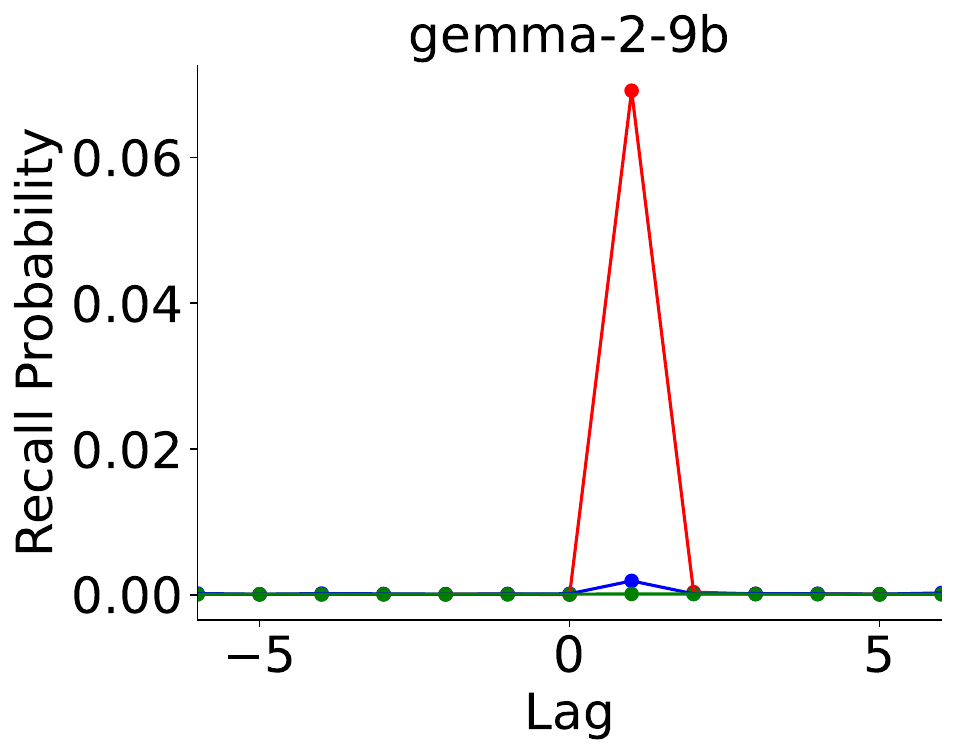}}
        \\
    \end{tabular}
    \begin{tabular}{lllll}
    \textbf{E} &
    \textbf{F} &
    \textbf{G} &
    \textbf{H} \\
        {\includegraphics[width=0.22\textwidth]{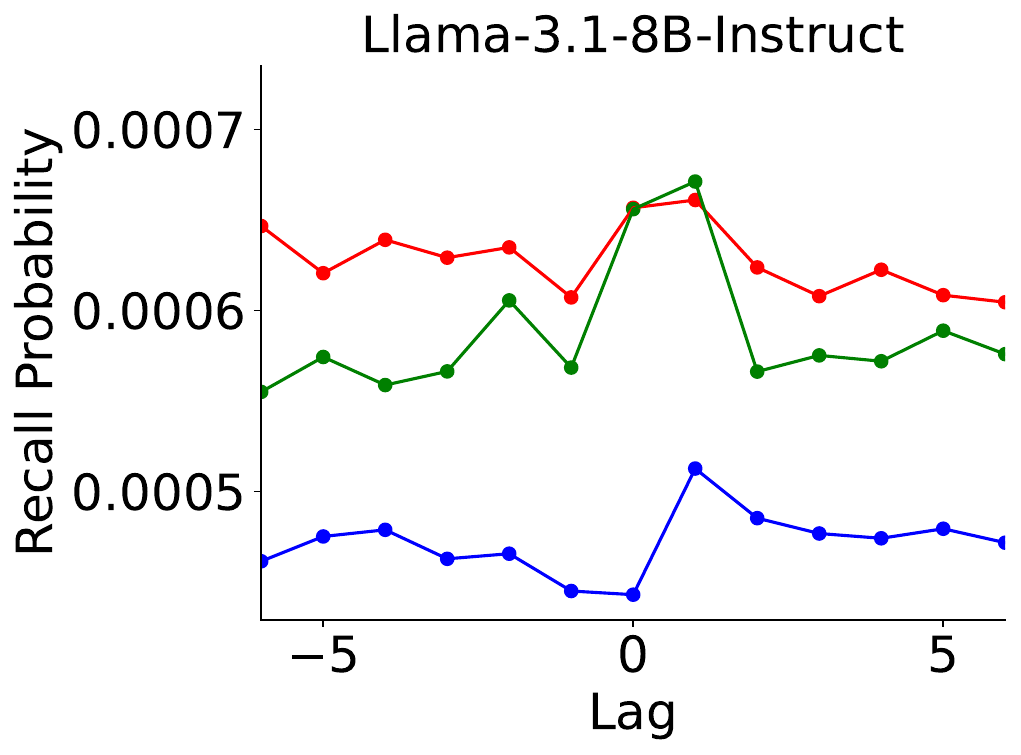
        }} &
        {\includegraphics[width=0.22\textwidth]{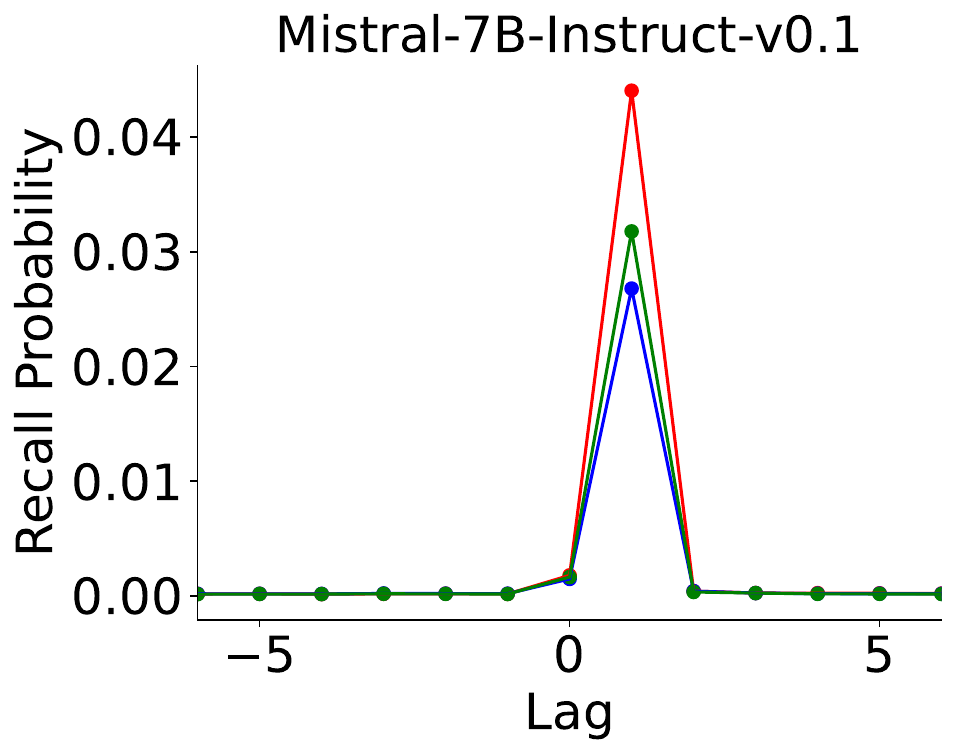}} &
        {\includegraphics[width=0.22\textwidth]{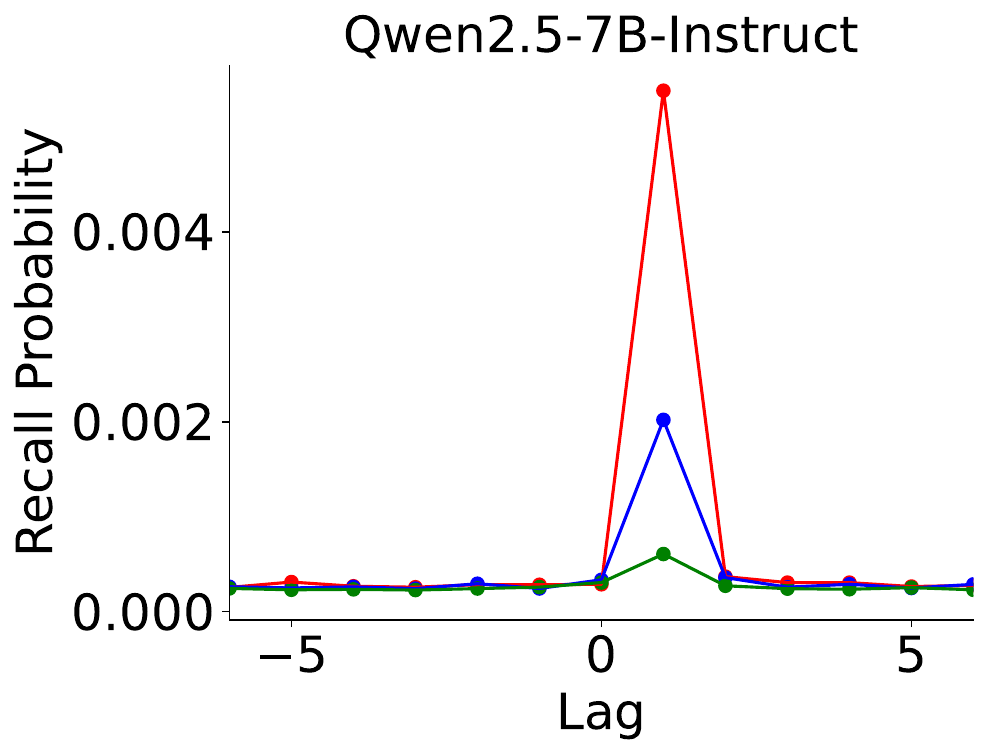}} &
        {\includegraphics[width=0.22\textwidth]{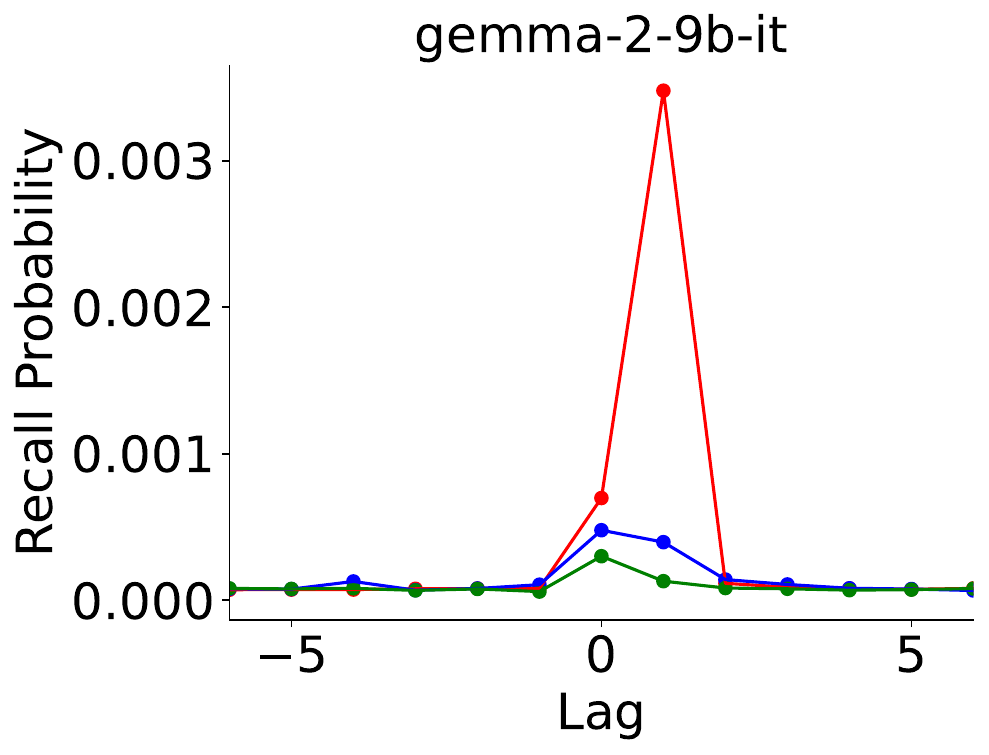}} \\
    \end{tabular}
    \includegraphics[scale=0.5]{Figures/legend.pdf}
    \caption{Impact of induction and random head bottom layers ablation (100 heads in each case) on the model output probability as a function of lag: same as Fig.~\ref{fig:CRP_250_bottom}, but with zoom on lags -6 to 6 to emphasize that the highest probabilities were at lags 1 or 0. Top row: Base models. Bottom row: Instruction-tuned models.
    \label{fig:CRP_6_bottom}}
\end{figure*}

\begin{figure*}[h!]
    \centering
    \begin{tabular}{lllll}
    \textbf{A} &
    \textbf{B} &
    \textbf{C} &
    \textbf{D} \\
        {\includegraphics[width=0.22\textwidth]{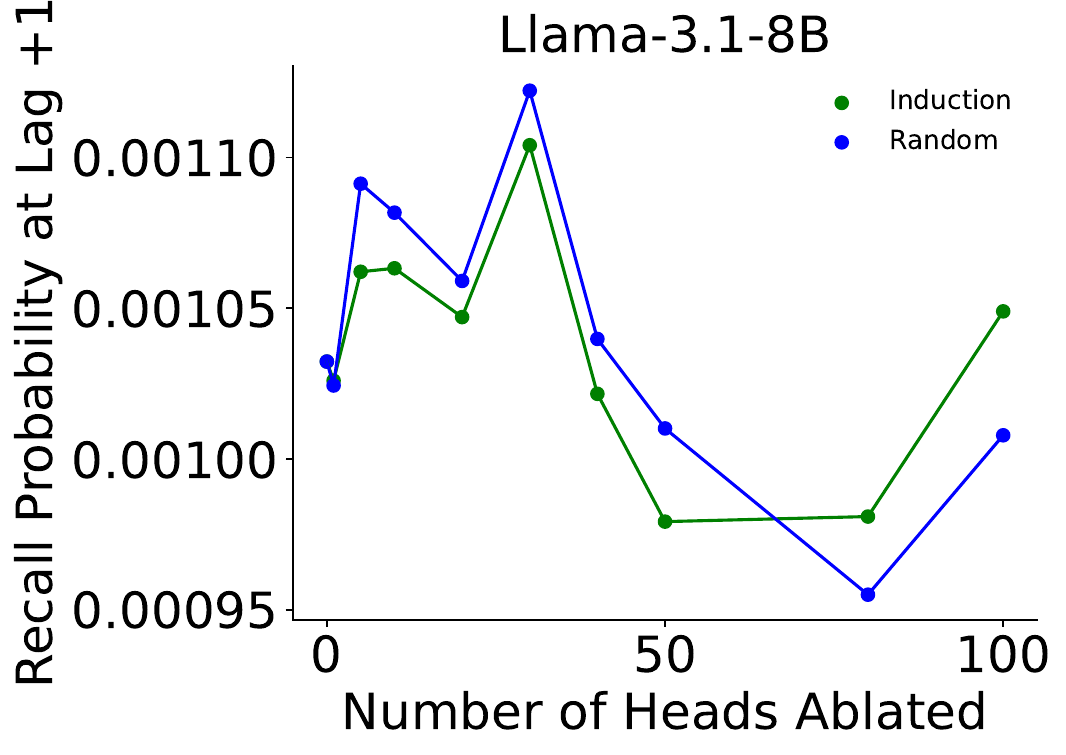
        }} &
        {\includegraphics[width=0.22\textwidth]{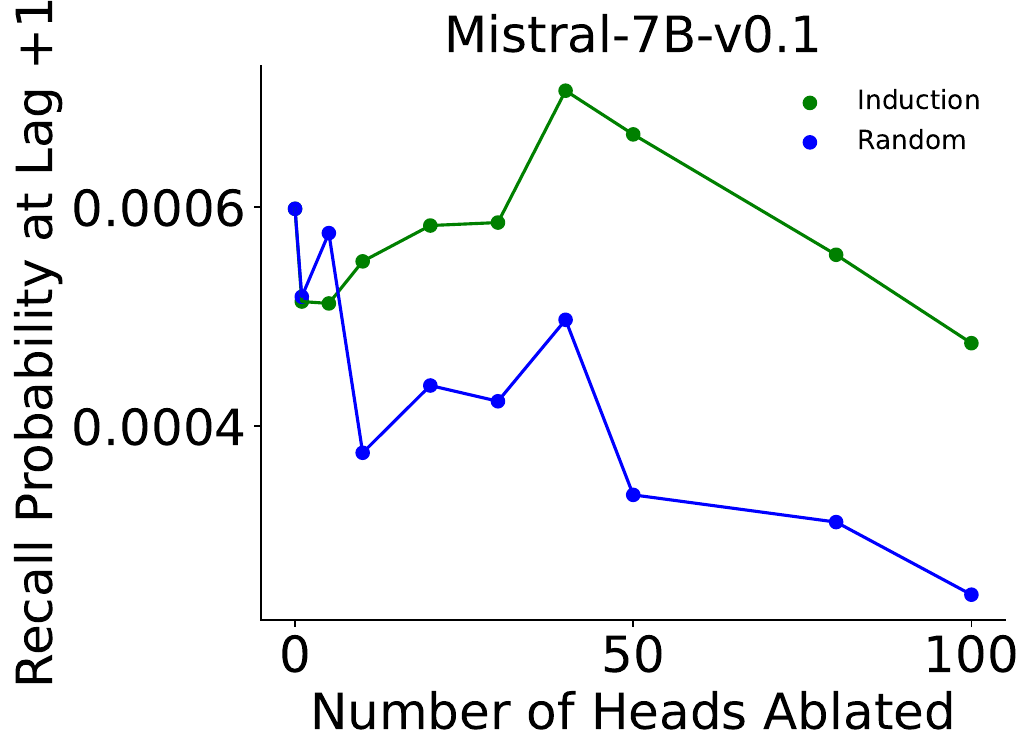}} &
        {\includegraphics[width=0.22\textwidth]{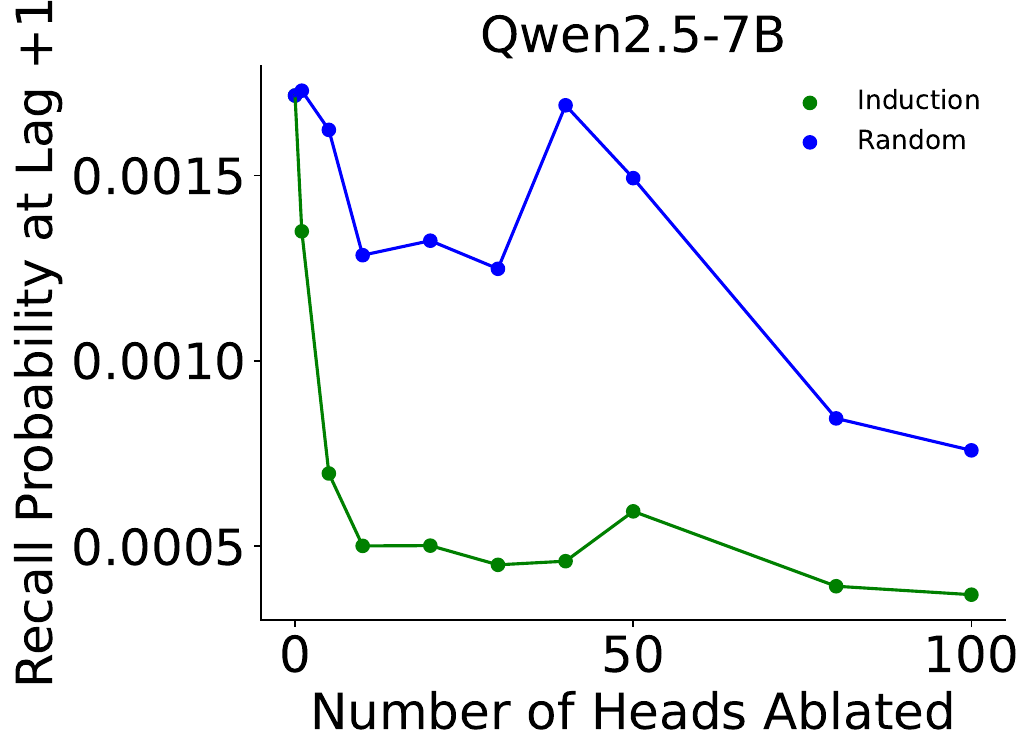}} &
        {\includegraphics[width=0.22\textwidth]{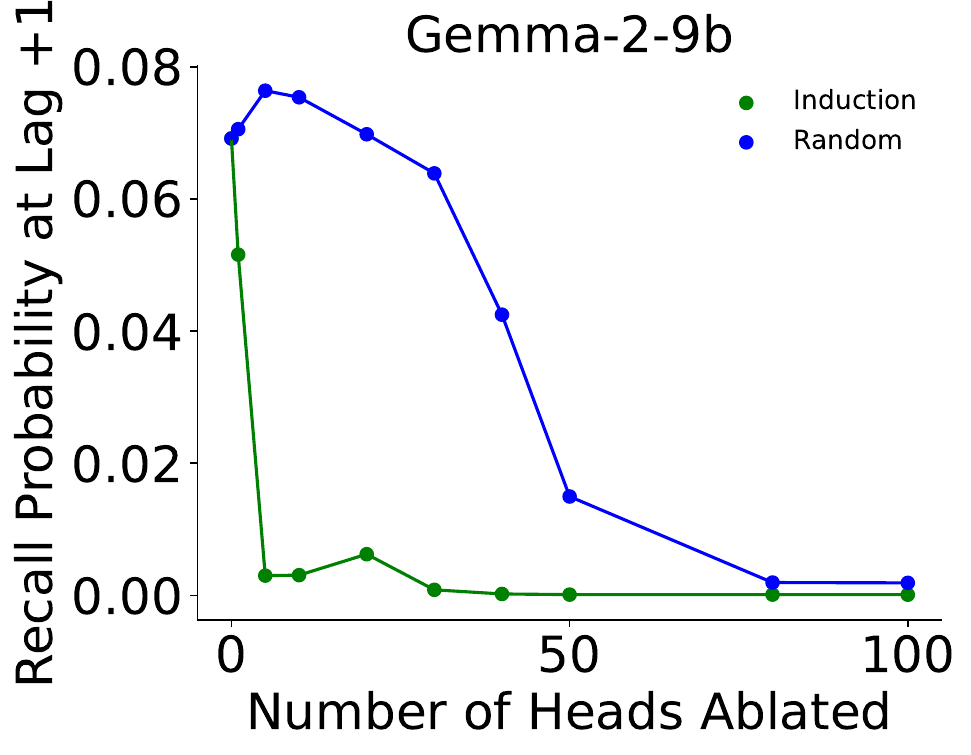}} \\
    \end{tabular}
    \begin{tabular}{lllll}
    \textbf{E} &
    \textbf{F} &
    \textbf{G} &
    \textbf{H} \\
         {\includegraphics[width=0.22\textwidth]{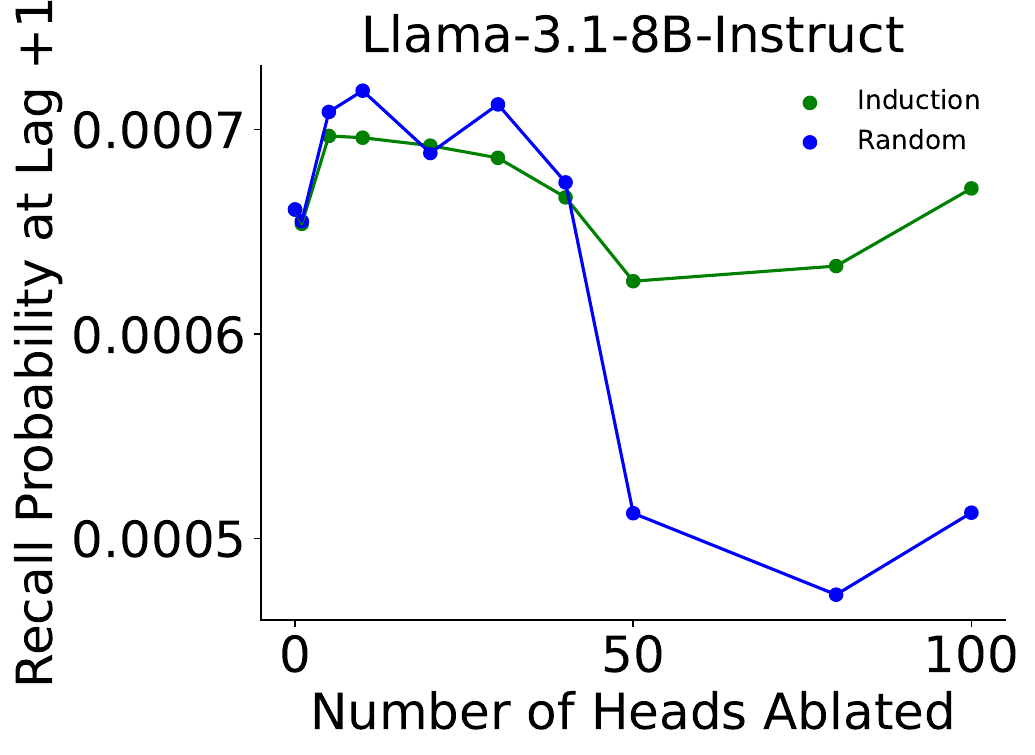
        }} &
        {\includegraphics[width=0.22\textwidth]{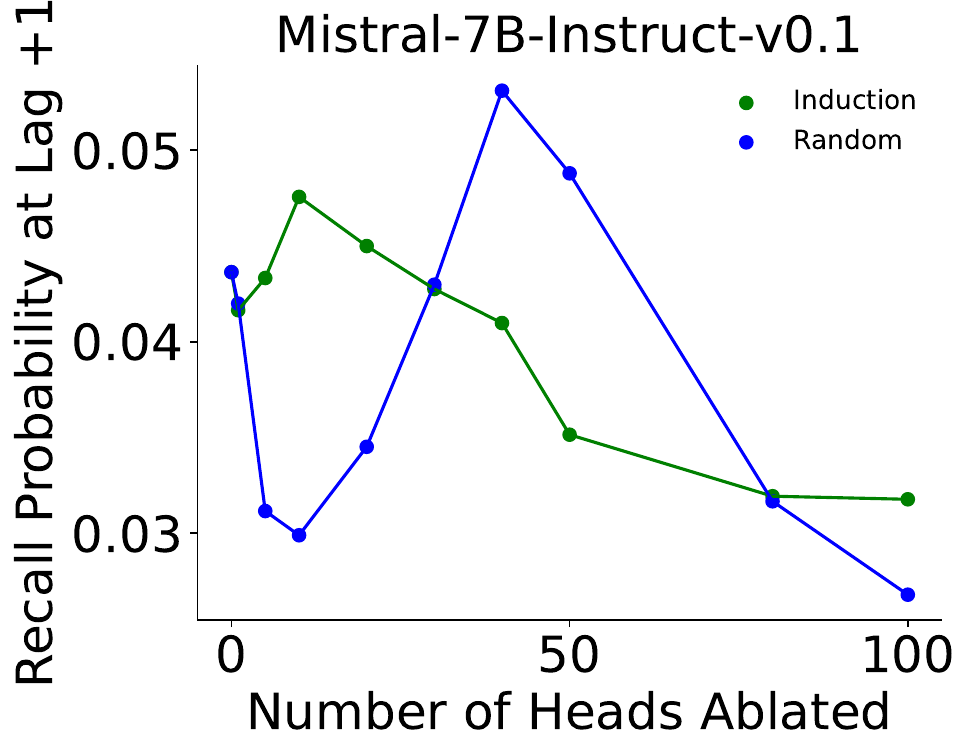}} &
        {\includegraphics[width=0.22\textwidth]{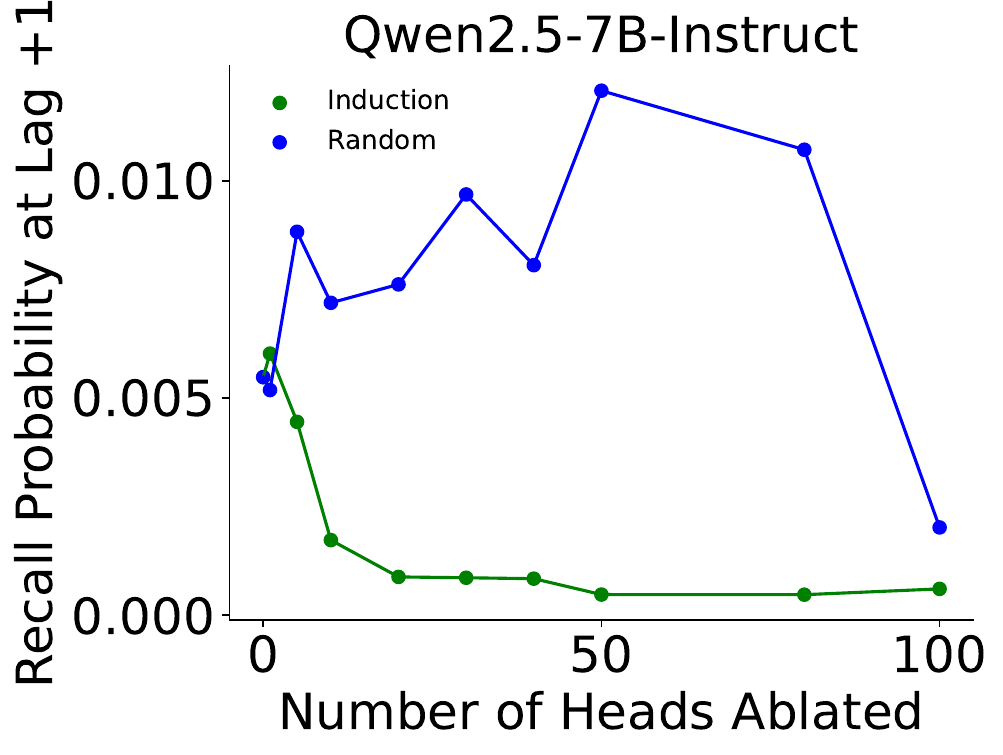}} &
        {\includegraphics[width=0.22\textwidth]{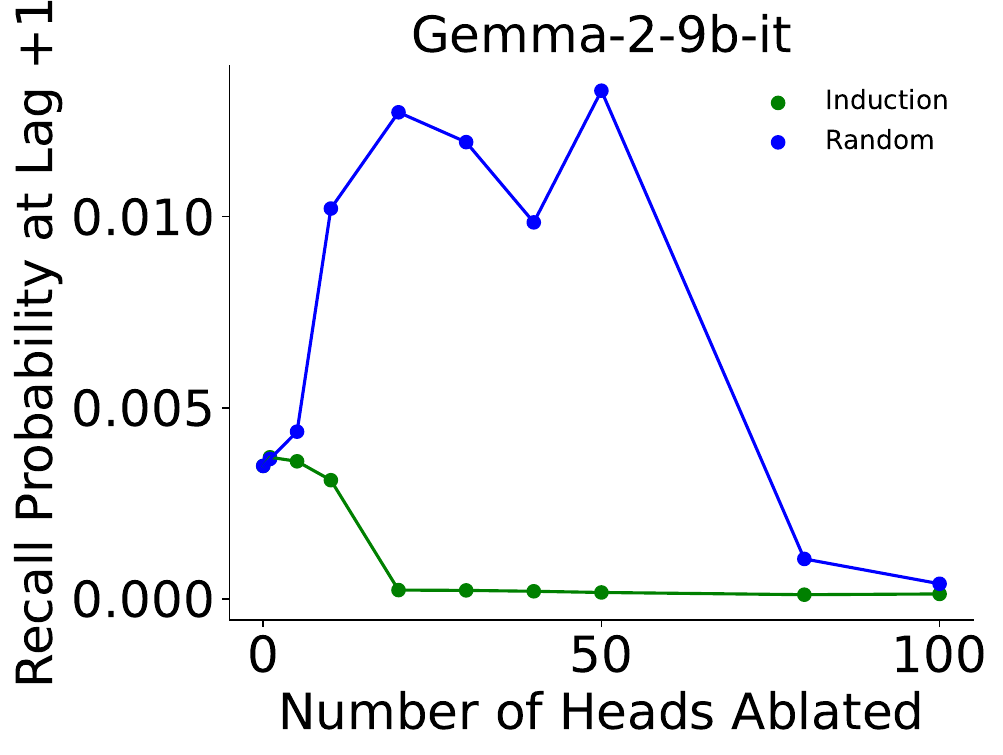}} \\
    \end{tabular}
    \caption{Impact of induction head bottom layers ablation on the model output probability for the token at lag +1. The models were presented with a sequence of 501 tokens where the last token repeated the token at index 250 and the lag is defined relative to the repeated token, hence the probability at lag +1 is the probability that the model assigns to token 251 (see Methods for more details). The results show averages across 5000 runs with shuffled token sequences.  We ablated the following numbers of induction heads (sorted by the induction scores) and random heads (x-axis): 1, 5, 10, 20, 30, 40, 50, 80, 100, 150, 200, 250 and 300. Top row: Base models. Bottom row: Instruction-tuned models.
    \label{fig:lag1_probability_bottom}}
\end{figure*}

\end{appendices}

\end{document}